\documentclass{article} 
\usepackage{iclr2024_conference,times}

\usepackage{amsmath,amsfonts,bm}

\def\lemref#1{Lemma\;\ref{#1}}
\def\corref#1{Corollary\;\ref{#1}}
\def\thmref#1{Theorem\;\ref{#1}}
\def\Appref#1{Appendix\;\ref{#1}}

\def\figref#1{figure\;\ref{#1}}
\def\Figref#1{Figure\;\ref{#1}}

\def\secref#1{section\;\ref{#1}}
\def\Secref#1{Section\;\ref{#1}}

\def\eqref#1{equation\;\ref{#1}}

\def\1{\bm{1}}

\def\vtheta{{\bm{\theta}}}

\def\vpi{{\bm{\pi}}}
\def\vsigma{{\bm{\sigma}}}
\def\va{{\bm{a}}}
\def\vb{{\bm{b}}}

\def\vr{{\bm{r}}}
\def\vs{{\bm{s}}}

\def\vu{{\bm{u}}}
\def\vv{{\bm{v}}}
\def\vw{{\bm{w}}}
\def\vx{{\bm{x}}}

\def\mA{{\bm{A}}}
\def\mB{{\bm{B}}}
\def\mC{{\bm{C}}}
\def\mD{{\bm{D}}}

\def\mH{{\bm{H}}}
\def\mI{{\bm{I}}}

\def\mM{{\bm{M}}}

\def\mP{{\bm{P}}}
\def\mQ{{\bm{Q}}}

\def\mS{{\bm{S}}}

\def\mU{{\bm{U}}}
\def\mV{{\bm{V}}}
\def\mW{{\bm{W}}}
\def\mX{{\bm{X}}}

\def\mZ{{\bm{Z}}}

\def\mPhi{{\bm{\Phi}}}
\def\mLambda{{\bm{\Lambda}}}
\def\mSigma{{\bm{\Sigma}}}
\def\mPsi{{\bm{\Psi}}}
\def\mTheta{{\bm{\Theta}}}
\def\mOmega{{\bm{\Omega}}}

\DeclareMathAlphabet{\mathsfit}{\encodingdefault}{\sfdefault}{m}{sl}
\SetMathAlphabet{\mathsfit}{bold}{\encodingdefault}{\sfdefault}{bx}{n}

\newcommand{\E}{\mathbb{E}}

\newcommand{\R}{\mathbb{R}}

\DeclareMathOperator*{\argmin}{arg\,min}

\DeclareMathOperator{\Tr}{Tr}

\usepackage[unicode=true, colorlinks=true]{hyperref}
\usepackage{url}
\usepackage{graphicx}
\usepackage{caption}
\usepackage{float}
\usepackage{subfig}
\usepackage{booktabs}
\usepackage{ulem}
\usepackage{tikz}
\usetikzlibrary{automata, positioning, arrows, chains, fit, shapes, calc, shadows}
\usepackage{pgfplots}

\title{On Double Descent in Reinforcement\\ Learning with LSTD and Random Features}

\iclrfinalcopy
\author{David Brellmann, Eloïse Berthier, David Filliat \& Goran Frehse  \\
U2IS, ENSTA Paris, Institut Polytechnique de Paris, Palaiseau, FRANCE \\
\texttt{\{first\_name.last\_name\}@ensta-paris.fr} 
}

\definecolor{Blue}{rgb}{0.16470588235,0.17647058823,0.45490196078}
\definecolor{Purple}{rgb}{0.50588235294,0.27843137254,0.52156862745}
\definecolor{Brown}{rgb}{0.3725490196,0.16862745098,0.10588235294}
\definecolor{PineGreen}{rgb}{0.07058823529,0.49019607843,0.40392156862}
\definecolor{RedOrange}{rgb}{0.81960784313,0.38823529411,0.26666666666}

\hypersetup{colorlinks,citecolor={Blue}}

\usepackage{amsthm}

\newtheorem{theorem}{Theorem}[section]
\newtheorem{corollary}{Corollary}[theorem]
\newtheorem{lemma}[theorem]{Lemma}
\newtheorem{assumption}{Assumption}

\newtheorem{remark}{Remark}

\newcommand{\N}{\mathcal{N}}

\newcommand{\numin}{\nu_\text{min}}
\newcommand{\numax}{\nu_\text{max}}
\newcommand{\State}{\mathcal{S}}
\newcommand{\SVisited}{\hat{\State}}

\newcommand{\rtrain}{\vr}
\newcommand{\Xtrain}{\mX_n}

\newcommand{\FState}{\mSigma_{\State}}
\newcommand{\FSVisited}{\mSigma_{\SVisited}}
\newcommand{\FXtrain}{\mSigma_{\mX_n}}

\newcommand{\FSW}{\mSigma_{\SVisited}(\mW)}
\newcommand{\FSWH}{\mSigma_{\SVisited}(\mW+\mH)}

\newcommand{\FXtrainN}{\mSigma_{\mX'_n}}

\newcommand{\MSBE}{\operatorname{MSBE}}

\newcommand{\MSVE}{\operatorname{MSVE}}
\newcommand{\DMSBE}{\overline{\operatorname{MSBE}}(\hat \vtheta)}
\newcommand{\DMSVE}{\overline{\operatorname{MSVE}}(\hat \vtheta)}
\newcommand{\Etrain}{\widehat{\operatorname{MSBE}}}
\newcommand{\EVtrain}{\widehat{\operatorname{MSVE}}}
\newcommand{\DEtrain}{\widehat{\overline{\operatorname{MSBE}}}(\hat \vtheta)}
\newcommand{\Dtrain}{\mathcal{D}_\text{train}}

\usepackage{epstopdf}
\DeclareGraphicsExtensions{.png,.eps}

\begin{document}

\maketitle

\begin{abstract}
   
    Temporal Difference (TD) algorithms are widely used in Deep Reinforcement Learning (RL). Their performance is heavily influenced by the size of the neural network. 
    While in supervised learning,  the regime of over-parameterization and its benefits are well understood, the situation in RL is much less clear. In this paper, we present a theoretical analysis of the influence of network size and $l_2$-regularization on performance. We identify the ratio between the number of parameters and the number of visited states as a crucial factor and define over-parameterization as the regime when it is larger than one. Furthermore, we observe a double descent phenomenon, i.e., a sudden drop in performance around the parameter/state ratio of one. Leveraging random features and the lazy training regime, we study the regularized Least-Squared Temporal Difference (LSTD) algorithm in an asymptotic regime, as both the number of parameters and states go to infinity, maintaining a constant ratio. We derive deterministic limits of both the empirical and the true Mean-Squared Bellman Error (MSBE) that feature correction terms responsible for the double descent. Correction terms vanish when the $l_2$-regularization is increased or the number of unvisited states goes to zero. Numerical experiments with synthetic and small real-world environments closely match the theoretical predictions.
\end{abstract}

\section{Introduction \label{sec:introduction}}
In recent years, neural networks have seen increased use in  Reinforcement Learning (RL)~\citep{mnih2015human, schulman2017proximal, haarnoja2018}.
While they can outperform traditional RL algorithms on challenging tasks, their theoretical understanding remains limited.
Even for supervised learning, which can be considered a special case of RL with discount factor equal to zero, deep neural networks are still far from being fully understood despite significant research efforts~\citep{arora2019fine, mei2018mean, rotskoff2018parameters, lee2019wide, bietti2019inductive, cao2019towards}. The difficulty is further exacerbated in RL by a myriad of new challenges that limit the scope of these works, such as the absence of true targets or the non-i.i.d nature of the collected samples~\citep{Kumar2020, Luo2020, lyle2021understanding, dong2020expressivity}.
Temporal-Difference (TD) methods are widely used RL algorithms that frequently use neural networks, are simple, and efficient in practice. 
We use the regularized Least-squares Temporal Difference (LSTD) algorithm~\citep{bradtke1996linear}, which is easier to analyze since it doesn't use gradient descent, and because it converges to the same solution as other TD algorithms~\citep{bradtke1996linear, boyan1999least, berthier2022non}. 

Theoretical studies of TD algorithms often explore asymptotic regimes where the number of samples $n \to \infty$ while the number of model parameters $N$ remains constant~\citep{tsitsiklis1996analysis, sutton1988learning}. When TD learning algorithms are applied to neural networks, it is commonly assumed that the number of parameters $N \to \infty$ with either a fixed or infinite number of samples without providing details on the relative magnitudes of these parameters~\citep{cai2019neural, agazzi2022temporal, berthier2022non, xiao2021understanding}.
Inspired by advancements in supervised learning~\citep{louart2018random, liao2020random}, we apply Random Matrix tools and propose a novel \emph{double asymptotic regime} where the number of parameters $N$ and the number of \emph{distinct visited} states $m$ go to infinity, maintaining a constant ratio, called \emph{model complexity}. 
We use a linear model and nonlinear random features (RF)~\citep{rahimi2007random} to approximate an overparameterized single-hidden-layer network in the lazy training regime~\citep{chizat2019lazy}. 
The results of our theoretical and empirical analyses are outlined below.
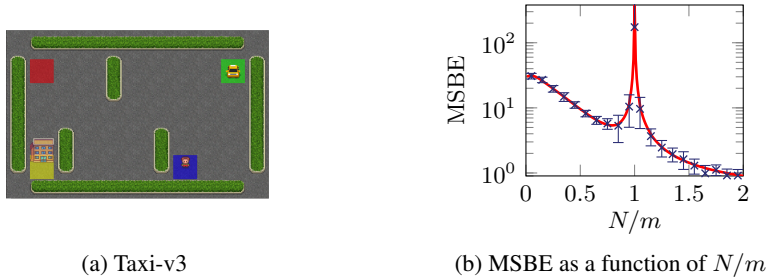
\begin{figure*}[t] 
    \begin{center} 
    \subfloat[Taxi-v3 \label{fig:taxi_gymnasium}]
    {
        \includegraphics[width=0.25\linewidth]{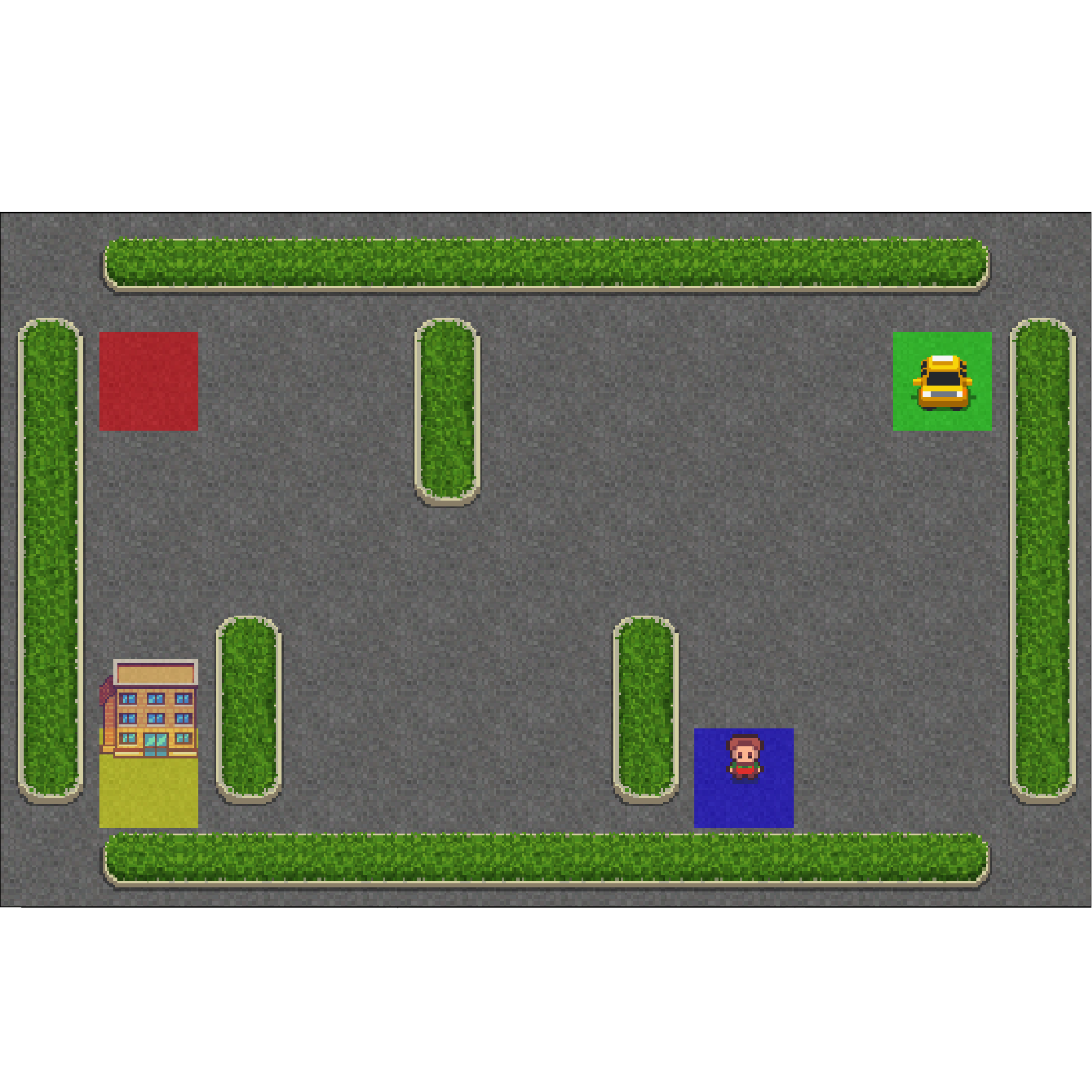}
    } \hspace{2cm}
        \subfloat[MSBE as a function of $N/m$]
        {
            \begin{minipage}[b]{0.32\linewidth}
    \begin{tikzpicture}[font=\small]
            \pgfplotsset{every major grid/.style={style=densely dashed}}
            \begin{axis}[
            width=\linewidth,
            xmin=0,
            xmax=2,
            ymin=0.900574544803151,
            ymax=380.13488806475823,
            ymode=log,
            ymajorgrids=false,
            scaled ticks=true,
            xlabel={ $N/m$ },
            ylabel={ $\MSBE$ },
            ylabel style={yshift=-0.3cm},
            xlabel style={yshift=0.2cm},
            legend style = {at={(0.98,0.98)}, anchor=north east, font=\small}
            ]\addplot[red, line width=1pt] coordinates{
            (0.01, 30.07202824655844)(0.02, 30.66787586692705)(0.03, 31.063319419990652)(0.04, 31.266155190319097)(0.05, 31.268102535758374)(0.060000000000000005, 31.096908736606817)(0.06999999999999999, 30.80314644856258)(0.08, 30.408539957384914)(0.09, 29.932883025317544)(0.09999999999999999, 29.393718340976605)(0.11, 28.806288171019897)(0.12, 28.183637865819072)(0.13, 27.536795303771942)(0.14, 26.874978277835663)(0.15000000000000002, 25.98221826459151)(0.16, 25.312669292888913)(0.17, 24.64835943256081)(0.18000000000000002, 23.99286969474573)(0.19, 23.348899292757444)(0.2, 22.71839411749262)(0.21000000000000002, 22.10266540833907)(0.22, 21.502499425056534)(0.23, 20.918258194603016)(0.24000000000000002, 20.349970986748545)(0.25, 19.616661732731913)(0.26, 19.084437187060402)(0.27, 18.566834480980752)(0.28, 18.06325516424636)(0.29000000000000004, 17.57308740548782)(0.3, 17.095735629015326)(0.31, 16.630642180129897)(0.32, 16.17730196579464)(0.33, 15.735271065790531)(0.34, 15.304170317001747)(0.35000000000000003, 14.745836400689127)(0.36000000000000004, 14.339118948406504)(0.37, 13.9425077960043)(0.38, 13.55586515164323)(0.39, 13.179089721449518)(0.4, 12.812108285972851)(0.41000000000000003, 12.454867606660493)(0.42000000000000004, 12.107326984101174)(0.43, 11.769451692575005)(0.44, 11.441207427793334)(0.45, 11.018462678651396)(0.46, 10.712528369460454)(0.47000000000000003, 10.416064217415325)(0.48000000000000004, 10.12899932672461)(0.49, 9.851250852888615)(0.5, 9.582724304536422)(0.51, 9.323314391998503)(0.52, 9.072906307459268)(0.53, 8.831377332461233)(0.54, 8.598598682486866)(0.55, 8.374437513627932)(0.56, 8.08872793832997)(0.5700000000000001, 7.88415072202028)(0.5800000000000001, 7.687746326826114)(0.59, 7.499389259933038)(0.6, 7.31896162270545)(0.61, 7.146355358100331)(0.62, 6.98147454023815)(0.63, 6.82423773644702)(0.64, 6.6745804818080074)(0.65, 6.486753527351076)(0.66, 6.35464671451597)(0.67, 6.230062814093439)(0.68, 6.113044854071536)(0.6900000000000001, 6.003670600556213)(0.7000000000000001, 5.902057516865336)(0.7100000000000001, 5.808368608767427)(0.72, 5.722819392037038)(0.73, 5.645686283043285)(0.74, 5.577316799719146)(0.75, 5.51814207692092)(0.76, 5.4544785044145145)(0.77, 5.419028427720382)(0.78, 5.3950443614078365)(0.79, 5.38355811332736)(0.8, 5.385831151567458)(0.81, 5.403411469132558)(0.8200000000000001, 5.438208762645997)(0.8300000000000001, 5.4925951869207985)(0.8400000000000001, 5.569542430554098)(0.85, 5.713879170677222)(0.86, 5.859972116031734)(0.87, 6.045959609092487)(0.88, 6.280949113232502)(0.89, 6.57729483465503)(0.9, 6.952166083616165)(0.91, 7.430123355909057)(0.92, 8.047547869660836)(0.93, 8.860691793445774)(0.9400000000000001, 9.9613452232077)(0.9500000000000001, 11.510097550379466)(0.9600000000000001, 14.849848746901015)(0.97, 19.367624529319222)(0.98, 28.46487721979123)(0.99, 55.577540564301756)(1.0, 380.13488806475823)(1.01, 55.179869842840006)(1.02, 27.990654191845426)(1.03, 18.80944506032523)(1.04, 14.205483083238038)(1.05, 10.748319089191607)(1.06, 9.109740392825206)(1.07, 7.917726613368355)(1.08, 7.011640359912709)(1.09, 6.299632347995141)(1.1, 5.725384609819017)(1.11, 5.2524458030648296)(1.12, 4.85618136809079)(1.1300000000000001, 4.519344442150543)(1.1400000000000001, 4.229499638816999)(1.1500000000000001, 3.9005892014901633)(1.1600000000000001, 3.6884373931439147)(1.17, 3.5003000030519917)(1.18, 3.332318087088206)(1.19, 3.1814175006932537)(1.2, 3.0451189354577735)(1.21, 2.921400820068438)(1.22, 2.8085982298252476)(1.23, 2.705327629898007)(1.24, 2.610429809757631)(1.25, 2.4952540473575295)(1.26, 2.416336954896507)(1.27, 2.3430564983008497)(1.28, 2.274829726759374)(1.29, 2.211151169674137)(1.3, 2.1515807412882086)(1.31, 2.095733374753244)(1.32, 2.043270538290887)(1.33, 1.9938937167511814)(1.34, 1.9473382542875632)(1.35, 1.903369139426993)(1.36, 1.8484076921723307)(1.37, 1.809695730205398)(1.3800000000000001, 1.77295203290693)(1.3900000000000001, 1.7380302043498164)(1.4000000000000001, 1.7047982620027262)(1.4100000000000001, 1.6731362756868657)(1.42, 1.642935567401801)(1.43, 1.6140971527494)(1.44, 1.58653111493554)(1.45, 1.5516140175393465)(1.46, 1.526708199642454)(1.47, 1.5028258057146542)(1.48, 1.4799050029384082)(1.49, 1.4578892426337602)(1.5, 1.4367254876565745)(1.51, 1.4163654547597333)(1.52, 1.3967641562347317)(1.53, 1.3778799184846982)(1.54, 1.3596742628297833)(1.55, 1.3421111383200768)(1.56, 1.3196354660552096)(1.57, 1.3034457540709736)(1.58, 1.2877953967044793)(1.59, 1.272658182972439)(1.6, 1.2580092104659213)(1.61, 1.243825265440198)(1.62, 1.230084651613909)(1.6300000000000001, 1.2167668550499866)(1.6400000000000001, 1.2038526576391444)(1.6500000000000001, 1.187230204154921)(1.6600000000000001, 1.1751891862288812)(1.6700000000000002, 1.1634954065746101)(1.68, 1.1521342959138419)(1.69, 1.1410915255858431)(1.7, 1.1303542057640894)(1.71, 1.1199097057538385)(1.72, 1.1097461456423252)(1.73, 1.0998524916286347)(1.74, 1.0902181071531647)(1.75, 1.0808326334320935)(1.76, 1.068690242236972)(1.77, 1.0598500944318343)(1.78, 1.0512291622236005)(1.79, 1.0428191048959192)(1.8, 1.0346127173856101)(1.81, 1.026602559053389)(1.82, 1.0187815054240392)(1.83, 1.01114311697402)(1.84, 1.0036809976092103)(1.85, 0.9939951192991915)(1.86, 0.9869213938337926)(1.87, 0.9800047604859191)(1.8800000000000001, 0.9732398559867684)(1.8900000000000001, 0.9666223545312389)(1.9000000000000001, 0.9601470750780845)(1.9100000000000001, 0.9538097132563804)(1.9200000000000002, 0.9476056368554842)(1.93, 0.9415307553014028)(1.94, 0.9355810998202687)(1.95, 0.9297528941784334)(1.96, 0.9221646713767339)(1.97, 0.9166056368156477)(1.98, 0.9111565040442111)(1.99, 0.9058137980445828)(2.0, 0.900574544803151)};
            \addplot+[only marks,mark=x,Blue,line width=0.5pt, mark size=2pt, error bars/.cd, y fixed, y dir=both, y explicit] table [x=x, y=y,y error=error, col sep=comma] {
            x, y, error
            0.05, 30.811101190301297, 3.0184934881688994
            0.15000000000000002, 26.966873144096358, 2.587066849687243
            0.25, 19.724086826504823, 2.2865813304706166
            0.35000000000000003, 14.81602919538332, 2.2635675430500886
            0.45, 11.126290026538022, 1.2708078399679017
            0.55, 8.218225536079656, 0.9289770287736886
            0.6500000000000001, 6.48742606950872, 0.9022472425799718
            0.7500000000000001, 5.750319660409338, 1.0635926618911342
            0.8500000000000001, 5.315410880209127, 2.383106923274224
            0.9500000000000001, 10.449459520789398, 5.417251860991149
            1.0, 173.76184713661792, 457.11041537363553
            1.05, 9.649511592220348, 4.8362271641517465
            1.1500000000000001, 3.680050690255368, 1.0714784161228845
            1.2500000000000002, 2.45680788537364, 0.6969537763212688
            1.35, 1.9463596517016541, 0.4677083058226156
            1.4500000000000002, 1.6261033147559694, 0.4923380862001745
            1.55, 1.2967695891852569, 0.33717456802221807
            1.6500000000000001, 0.9865853853253088, 0.31678716216178343
            1.7500000000000002, 1.1192876232784836, 0.20973862074724797
            1.85, 0.9157388089445979, 0.2324313770676407
            1.9500000000000002, 0.912404888524148, 0.21847872099142274
            };
        \end{axis}
    \end{tikzpicture}
\end{minipage}
        } 
        \caption{\textbf{As the model complexity $N/m$ (for $N$ parameters, $m$ distinct visited states) increases, the MSBE first shows a U-shaped curve, peaking around the interpolation threshold ($N=m$). Double descent refers to the phenomenon for $N/m>1$ where the MSBE drops once again.} Continuous lines (\textcolor{red}{red}) indicate the theoretical values from \thmref{theorem:asy-behavior-true-MSBE}, the crosses (\textcolor{Blue}{blue}) are numerical results averaged over $30$ instances with their standard deviations after the learning with regularized LSTD on Taxi-v3 for $\gamma=0.95, \lambda=10^{-9}, n=5000, m=310$.}
        \label{fig:double_descent_taxi_intro}
    \end{center}
\end{figure*}
\paragraph{Contributions.} We make the following contributions, taking a step towards a better theoretical understanding of the influence of model complexity $N/m$ and $l_2$-regularization on the performance of Temporal Difference algorithms:
\begin{enumerate}
    \item We propose a novel double asymptotic regime, where the number of parameters $N$ and distinct visited states $m$ go to infinity while maintaining a constant ratio. This leads to a precise assessment of the performance in both \emph{over-parameterized} ($N/m>1$) and \emph{under-parameterized} regimes ($N/m<1$). 
    This is a nontrivial extension of existing work in supervised learning since several properties essential to proofs, such as the positive definiteness of key matrices, are voided by a discount factor in RL. 
    \item In the phase transition around $N/m=1$, we observe a peak in the Mean-Squared Bellman Error (MSBE), as illustrated in \figref{fig:double_descent_taxi_intro}, i.e., a \emph{double descent phenomenon} similar to what has been reported in supervised learning~\citep{mei2022generalization, liao2020random}. 
    \item We identify the resolvent of a non-symmetric positive-definite matrix that emerges as a crucial factor in the performance analysis of TD learning algorithms in terms of the MSBE and we provide its deterministic limit form in the double asymptotic regime.
    \item We derive analytical equations for both the asymptotic empirical MSBE on the collected transitions and the asymptotic true MSBE. The deterministic forms expose correction terms that we experimentally associate with the double descent phenomenon. We show that the correction terms vanish as the $l_2$-regularization is increased or $N/m$ goes to infinity. We also show that the influence of the $l_2$-regularization parameter decreases 
    as $N/m$ increases.%
    \item Our theory closely matches empirical results on a range of both toy and small real-world Markov Reward Processes where $m$ and $N$ are fixed, but for which the asymptotic regime still gives accurate predictions. Notably, we observe a peak in the true MSBE around $N/m=1$ that is not observed in the empirical MSBE. Correction terms, and therefore the difference between true and empirical MSBE, empirically vanish when the number of unvisited states goes to zero. 
\end{enumerate}  
\section{Related Work \label{sec:related_work}}
We review three related approaches to study neural networks in supervised learning or RL. Further technical results from the literature are cited where relevant throughout the paper.
\paragraph{Neural Tangent Kernel (NTK) regime.} In the NTK regime, one considers that infinitely wide neural networks, with appropriate scaling and initial conditions, behave like the linearization of the neural network around its initialization~\citep{jacot2018neural}. However, as highlighted by \citet{chizat2019lazy}, this behavior is not specific to neural networks and is not so much due to over-parameterization than to an implicit choice of scaling. In such a scenario, neural networks can be modeled as a random feature model~\citep{rahimi2007random}; we adopt this technique in order to abstract from the learning dynamics. The NTK regime was also considered in RL, in both finite and infinite state space, to prove the convergence of infinite-width neural TD learning algorithms towards the global optimum of the MSBE~\citep{cai2019neural, agazzi2022temporal, liu2019neural}.
\paragraph{Mean-Field regime.} Under appropriate initial conditions and scaling, the mean-field analysis models the neural network and its induced feature representation with an empirical distribution, which, at the infinite-width limit, corresponds to a population distribution. The evolution of such a population distribution is characterized by a partial differential equation (PDE) known as the continuity equation and captures Stochastic Gradient Descent (SGD) dynamics as a Wasserstein gradient flow of the objective function~\citep{chizat2018global, rotskoff2018parameters, mei2018mean}. Although more challenging than the NTK regime, the mean-field regime is more realistic since the weights are not restricted to staying in their initial regions~\citep{chizat2019lazy}. The mean-field regime was studied in RL to prove the convergence of infinite-width neural TD learning algorithms towards the global optimum of the MSBE~\citep{zhang2021understanding, agazzi2022temporal}. 
\paragraph{Double Asymptotic regime.} In the above regimes, the number of data points $n$ is negligible compared to the number of parameters $N$ since $N$ grows to infinity. However, this is rarely the case in practice, which is why the double descent phenomenon can not be explained with the previous regimes~\citep{zhang2020can, belkin2018reconciling}. The double descent phenomenon is characterized by a peak near the interpolation threshold ($N=n$). For this reason, many studies in supervised learning consider a double asymptotic regime \citep{mei2022generalization, louart2018random, liao2020random, belkin2020two}, where both $n, N$ go to infinity while maintaining their ratio constant. In the above work, techniques from Random Matrix theory are used to derive a precise description of the phase transition between under-($N<n$) and over-($N>n$) parameterization and the double descent phenomenon. 
Our work extends this approach from supervised learning to RL. Since several key properties from supervised learning are voided by the discount factor, the analysis is substantially more involved in the case of RL. 
\citet{thomas2022role} investigated off-policy linear TD methods in the limit of large number of states and parameters on a  transition matrix of rank~$1$, and observed a peaking behavior in the MSBE. We consider a more general setting on the on-policy setting and a different ratio in the double asymptotic regime without making assumptions about the rank of the transition matrix.
\section{Preliminaries \label{sec:preliminaries}}
\paragraph{Notations.} We define $[n] = \{1, 2, \ldots, n \}$. For a real matrix $\mA$,  $[\mA]_{ij}$ denotes its $(i, j)^{th}$ entry. For $\mA$ with real eigenvalues, we denote with $\numin(\mA)$ the smallest and $\numax(\mA)$ the largest eigenvalue. The symmetric part of $\mA$ is $H(\mA)=\tfrac{\mA+\mA^T}{2}$. The operator norm of a $\mA$ is written $\lVert \mA \rVert$.  The norm induced by $\mA$ on a vector $\vv$ is $\lVert \vv \rVert_\mA=\vv^T \mA \vv$ and $\lVert \vv \rVert$ depicts the Euclidean norm of $\vv$.
$\N(0, 1)$ denotes the standard Gaussian distribution. 
\paragraph{Markov Reward Processes.} We consider a Markov Reward Process (MRP) $(\State, P, r, \gamma)$, where $\State \subseteq \R^d$ is the state space; $P: \State \times \mathcal{S} \to [0, 1]$ is the transition kernel (stochastic kernel) and $P(\vs, \vs')$ denotes the probability of transitioning to state $s'$ from state $\vs$; $r: \State \times \mathcal{S} \to \R$ is the reward function; and $\gamma \in [0, 1)$ is the discount factor. 
For notational convenience, the state space $\State$ is described by the state matrix $\mS \in \R^{d \times \lvert \State \rvert}$, where each column of $\mS$, written $\mS_i$, represents a state in~$\State$. The transition probability matrix associated with the stochastic kernel~$P$ is $\mP \in \R^{\lvert \State \rvert \times \lvert \State \rvert}$.
The goal is to learn the value function $V: \State \to \R$, which maps each state~$\vs$ to the expected discounted sum of rewards when starting from~$\vs$ and following the dynamics of the MRP defined by~$\mP$ as $V(\vs) := \E_\mP \bigl[{\sum}_{k=1}^\infty \gamma^{k-1} r(\vs_k, \vs_{k+1})~ \bigm|~\vs_1=\vs\bigr].$ 
It is well-known that the value function is the unique fixed-point of Bellman’s equation
\begin{equation}
   \mV=\bar \vr + \gamma \mP \mV,
   \label{def:bellman_eq}
\end{equation}
where $\mV \in \R^{\lvert \State \rvert}$ is a vector whose $i$-th element is the value function of the $i$-th state $\mS_i$; and $\bar \vr \in \R^{\lvert \State \rvert}$ is the vector containing the expected rewards, for which $\bar \vr_i = \E_\mP[r|\mS_i]$ for all $i \in [~\lvert \State \rvert~]$.
\paragraph{Linear Function Approximation.} In practice,  \eqref{def:bellman_eq} cannot be solved since $\mP$ is unknown and $\lvert \State \rvert$ is too large.
One common solution is to use Linear Function Approximation (LFA). Using a parameter vector $\vtheta \in \R^N$ and a feature matrix $\FState \in \R^{N \times \lvert \State \rvert }$, whose columns are the feature vectors for every state, $\mV$ is approximated by $\mV \approx \FState^T \vtheta$. 
In Deep RL, the neural network learns both the feature vectors and the parameter vector.
For a given feature matrix, the learning process based on \eqref{def:bellman_eq} amounts to finding a parameter vector $\vtheta$ that minimizes the Mean-Squared Bellman error (MSBE) 
\begin{equation}
    \operatorname{MSBE}(\vtheta) = \lVert \bar \vr + \gamma \mP \FState^T \vtheta -  \FState^T \vtheta \rVert^2_{\mD_\vpi},
    \label{def:MSBE}
\end{equation}
where $\vpi \in \R^{\lvert \State \rvert}$ is the stationary distribution induced by the MRP and $\mD_\pi \in \R^{\lvert \State \rvert \times \lvert \State \rvert}$ is its diagonal matrix. 
Since $\bar \vr + \gamma \mP \FState^T \vtheta$ may not lie in the span of the bases $\FState$, there may not be a parameter vector $\vtheta$ that brings the MSBE to zero.
\paragraph{Linear Temporal-Difference Methods.} Linear Temporal-Difference (TD) methods are LFA methods that try to minimize the MSBE in \eqref{def:MSBE} by replacing the second occurrence of $\vtheta$ in \eqref{def:MSBE} with an auxiliary vector $\vu$, minimizing on $\vu$ and then finding a $\vtheta$ close to $\vu$ \citep{dann2014policy}: 
\begin{align}
        \vu^* &= \argmin_{\vu \in \R^{N}}~ \lVert \bar \vr + \gamma \mP \FState^T \vtheta - \FState^T \vu \rVert^2_{\mD_\vpi}  &\text{(projection step)}, \label{eq:projection_step} \\
        \vtheta^*&=\argmin_{\vtheta \in \R^{N}}~ \lVert \FState^T \vu^* - \FState^T \vtheta \rVert^2_{\mD_\vpi}  &\text{(fixed-point step)}.
        \label{eq:fixed_point_step}
\end{align}
The projection step (\eqref{eq:projection_step}) implies that TD methods actually minimize the Mean-Squared Projected Bellman error (MSPBE) rather than the MSBE~\citep{dann2014policy}.
In our asymptotic regime, as the number of features $N \to \infty$, the class of representable value functions becomes richer, and the MSBPE converges to the MSBE~\citep{cai2019neural, agazzi2022temporal}. 
\section{System Model \label{sec:system_model}}
We now describe the key elements on which we base our asymptotic analysis of the MSBE in TD learning: Random features, the regularized LSTD algorithm, and the double asymptotic regime. 
\subsection{Regularized LSTD with Random Features}
\paragraph{Random Features.} We consider value function approximation using the Random Feature (RF) mapping $\operatorname{RF}: \State \to \R^N$ defined for all $\vs \in \State$ as 
\begin{equation}
    \operatorname{RF}(\vs) = \sigma(\mW\vs),
    \label{def:rf_map}
\end{equation}
where $\sigma: \R \to \R$ is $K_\sigma$-Lipschitz continuous and applied component-wise; $\mW = \varphi(\tilde \mW) \in \R^{N \times d}$ is a random weight matrix fixed throughout training for which $\tilde \mW \in \R^{N \times d}$ has independent and identically distributed $\N(0, 1)$ entries, and $\varphi: \R \to \R$ is $K_\varphi$-Lipschitz continuous and applied component-wise. 
From the perspective of neural networks, the~$N$ random features can be seen as~$N$ outputs from a single-hidden-layer neural network. 
In our asymptotic regime, this simplification becomes even more accurate as the number of features $N$ of the single layer grows towards infinity and we enter into the \emph{lazy training regime}, where weights barely deviate from their random initial values~\citep{chizat2019lazy}.
In the literature on Deep Learning and double descent, large-width neural networks are often modeled using asymptotic random features~\citep{louart2018random, liao2020random, mei2022generalization}, including in RL~\citep{cai2019neural, agazzi2022temporal, liu2019neural}.
In the following, we denote the random feature matrix of any state matrix $\mA \in \R^{d \times p}$ as~$\mSigma_\mA$ where $\operatorname{RF}$ is applied column-wise, i.e.,  $\mSigma_\mA=\sigma(\mW\mA)$.
\paragraph{Sample Matrices and Empirical MSBE.} We assume that the transition probability matrix~$\mP$ is unknown during the training phase. Instead, we have a dataset of $n$ transitions consisting of states, rewards, and next-states drawn from the MRP, i.e., ${\mathcal{D}_\text{train}:=\bigl\{( \vs_i, r_i, \vs_i')\bigr\}_{i=1}^n}$ where $\vs_i' \sim P(\vs_i, \cdot)$. We consider the \emph{on-policy setting}, where $\mathcal{D}_\text{train}$ is derived from a sample path of the MRP or its stationary distribution $\vpi$. We collect the states and rewards in the sample matrices
\begin{equation}
        \Xtrain = [\vs_1, \ldots, \vs_n] \in \R^{d \times n}, \quad \rtrain = [r_1, \ldots, r_n]^T \in \R^{n}, \quad \Xtrain' = [\vs_1', \ldots, \vs_n'] \in \R^{d \times n}.
    \label{def:datasets}
\end{equation}
Let $\SVisited \subseteq \State$ be the set of distinct states in $\mathcal{D}_\text{train}$, which we call  \emph{visited states}, and let $m=|\SVisited|$. 
Let $\hat \mS \in \R^{d \times m}$ be the state matrix of $\SVisited$, i.e., each column ${\hat \mS}_i$ of $\hat \mS$ describes a state in $\SVisited$. 
We denote by $\FSVisited \in \R^{N \times m}$, $\FXtrain \in \R^{N \times n}$, and $\FXtrainN \in \R^{N \times n}$ the random feature matrices of $\SVisited$, $\Xtrain$, and $\Xtrain'$, respectively.
For the proof of our results, it will be mathematically advantageous to express $\FXtrain$ and $\FXtrainN$ as the product of $\FSVisited$ with auxiliary matrices $\hat \mU_n \in \R^{m \times n}$ and $\hat \mV_n \in \R^{m \times n}$ as follows:
\begin{equation}
        \FXtrain = \sqrt{n}\FSVisited \hat \mU_n \quad \text{and} \quad \FXtrainN = \sqrt{n}\FSVisited \hat \mV_n \, .
    \label{def:FXtrain_with_U_and_V}
\end{equation}
Each column $i$ of $\sqrt{n}\hat \mU_n$ is a one-hot vector, where the $j$-th element equals $1$ if the $i$-th state $\vs_i$ of $\Xtrain$ is $\hat \mS_j$, and similarly for $\sqrt{n}\hat \mV_n$ and $\Xtrain'$. 
Since $\mP$ is unknown, we aim to find $\vtheta$ that minimizes the empirical version of the MSBE (\eqref{def:MSBE}) obtained with transitions collected in $\mathcal{D}_\text{train}$:
\begin{equation}
    \widehat{\operatorname{MSBE}}(\vtheta) = \tfrac{1}{n} \lVert \rtrain + \gamma \FXtrainN^T \vtheta - \FXtrain^T\vtheta \rVert^2,
    \label{def:MSBE_train}
\end{equation}
which uses the Euclidean norm since the distribution is reflected by the samples. Assumming globally stable MRP, a fixed number of features, and all states being visited, $\widehat{\operatorname{MSBE}}(\vtheta)$ converges to $\operatorname{MSBE}(\vtheta)$ with probability~$1$, as the number of collected transitions $n\to\infty$. This follows from the law of large numbers \citep{stachurski2009economic}.
In our analysis, we will also consider the case where $n \to \infty$ without visiting all states, i.e., $m<|\State|$, such that there can be a significant difference between $\widehat{\operatorname{MSBE}}(\vtheta)$ and $\operatorname{MSBE}(\vtheta)$.
\paragraph{Regularized Least-Square Temporal-Difference Methods.} Regularized Least-Square Temporal-Difference (LSTD) Methods~\citep{bradtke1996linear} are linear TD methods that solve an empirical regularized version of \eqref{eq:projection_step} and \ref{eq:fixed_point_step} with transitions collected in $\mathcal{D}_\text{train}$:
\begin{align}
    \vu^* &= \argmin_{\vu \in \R^{N}}~ \lVert \rtrain + \gamma \FXtrainN^T\vtheta -  \FXtrain^T\vu \rVert^2 + \lambda_{m, n} \lVert \vu \rVert^2, \label{eq:empirical_projection_step}\\
    \hat \vtheta &= \argmin_{\vtheta \in \R^{N}}~ \lVert \FXtrain^T \vu^{*} - \FXtrain^T \vtheta \rVert^2, \label{eq:empirical_fixed_point_step}
\end{align}
where $\lambda_{m, n}>0$ is the \emph{effective $l_2$-regularization parameter}, introduced to mitigate overfitting \citep{hoffman2011regularized, chen2013efficient}. It is well known that for $\lambda_{m, n}=0$ and with the number of samples $n \to \infty$, the fixed point of the approximation \eqref{eq:empirical_projection_step}~and~\ref{eq:empirical_fixed_point_step} equals the fixed point of \eqref{eq:projection_step}~and~\ref{eq:fixed_point_step} with probability one. Solving the fixed-point of the linear system approximation given by \eqref{eq:empirical_projection_step}~and~\ref{eq:empirical_fixed_point_step} gives 
\begin{equation}
    \hat \vtheta = \left[\FXtrain \left[\FXtrain - \gamma \FXtrainN \right]^T + \lambda_{m, n} \mI_{N} \right]^{-1}\FXtrain\rtrain.
    \label{def:theta_LSTD}
\end{equation}
Under appropriate learning rates, linear TD methods based on gradient-descent converge towards the same fixed-point $\hat \vtheta$~\citep{robbins1951stochastic, dann2014policy, sutton2018reinforcement}. Besides reducing overfitting, an appropriate $\lambda_{m, n}$ ensures that $\FXtrain[\FXtrain - \gamma \FXtrainN]^T + \lambda_{m, n} \mI_{N}$ is invertible. 
\subsection{Double Asymptotic Regime and Resolvent in LSTD}
We study the regularized LSTD in the following double asymptotic regime:
\begin{assumption}[Double Asymptotic Regime]
    As $N, m, d \to \infty$, we have:
    \begin{enumerate}
        \item $0 <  \lim\min \left\{\tfrac{N}{m}, \tfrac{d}{m}, \tfrac{m}{\lvert \State \rvert}\right\} < \lim \max \left\{\tfrac{N}{m}, \tfrac{d}{m}, \tfrac{m}{\lvert \State \rvert}\right\} < \infty$. 
        \item There exists $K_\State, K_r>0$ such that $\lim \sup_{\lvert \State \rvert} \lVert \mS \rVert < K_\State$ and $r(\cdot, \cdot)$ is bounded by $K_r$. 
    \end{enumerate}
    \label{assumption:growth_rate}
\end{assumption}
In order to use Random Matrix tools, we rewrite \eqref{def:theta_LSTD} as (see proof in \lemref{lem:push_through_identity})
\begin{equation}
        \hat \vtheta = \tfrac{1}{mn} \FXtrain \left[\tfrac{1}{mn}\left[\FXtrain - \gamma \FXtrainN \right]^T \FXtrain + \tfrac{\lambda_{m, n}}{mn} \mI_{n} \right]^{-1}\rtrain.
    \label{def:theta_LSTD_with_m}
\end{equation}
Instead of the effective $l_2$-regularization parameter $\lambda_{m, n}$, we will use its scaled version $\lambda = \tfrac{\lambda_{m, n}}{mn}$ in the remainder of this paper.
We observe that $\hat \vtheta = \tfrac{1}{mn} \FXtrain \mQ_m(\lambda) \rtrain$ depends on the \emph{resolvent}
\begin{equation}
    \mQ_m(\lambda)=\left[\tfrac{1}{mn}\left[\FXtrain - \gamma \FXtrainN \right]^T \FXtrain + \lambda \mI_{n} \right]^{-1}=\left[\tfrac{1}{m}(\hat \mU_n - \gamma \hat \mV_n)^T \FSVisited^T \FSVisited \hat \mU_n + \lambda \mI_{n} \right]^{-1}
    \label{def:resolvent}
\end{equation}
when $\tfrac{1}{m}(\hat \mU_n - \gamma \hat \mV_n)^T \FSVisited^T \FSVisited \hat \mU_n + \lambda\mI_{n}$ is invertible, which in general may not be the case. 
We can guarantee invertibility if the \emph{empirical transition model matrix} $\hat \mA_m \in \R^{m \times m}$
\begin{equation}
    \hat \mA_m = \hat \mU_n(\hat \mU_n - \gamma \hat \mV_n)^T
    \label{def:empirical_transition_model_matrix}
\end{equation}
has a positive-definite symmetric part (see \Appref{sec:existence_resolvent} for a formal proof).
For the remainder of the paper, we therefore make the following assumption on $\hat \mA_m$: 
\begin{assumption}[Bounded Eigenspectrum]
    There exist $0 < \xi_{\min} < \xi_{\max}$ such that for every $m$, all the eigenvalues of $H(\hat \mA_m)$ are in $[\xi_{\min}, \xi_{\max}]$.
    \label{assumption:regime_n}
\end{assumption}
Note that the above assumption is satisfied for regularized pathwise LSTD~\citep{lazaric2012finite}, and also for sufficiently large $n$ (see \Appref{sec:existence_resolvent}).
\section{Asymptotic Analysis of Regularized LSTD \label{sec:main_results}}
In this section, we present our main theoretical results, which characterize the true and empirical MSBE under Assumptions~\ref{assumption:growth_rate} and \ref{assumption:regime_n}.
\subsection{An Equivalent Deterministic Resolvent}
The resolvent $\mQ_m(\lambda)$ (in \eqref{def:resolvent}) plays a significant role in the performance of regularized LSTD since $\hat \vtheta=\tfrac{1}{\sqrt{n}}\tfrac{1}{m}\FSVisited \hat \mU_n \mQ_m(\lambda)\rtrain$. To assess the asymptotic $\Etrain(\hat \vtheta)$ and true $\MSBE(\hat \vtheta)$, we first find a deterministic equivalent for the resolvent $\mQ_m(\lambda)$.
A natural deterministic equivalent would be $\E_\mW[\mQ_m(\lambda)]$, but it involves integration without having a closed form expression (due to the matrix inverse) and is inconvenient for practical computation. Leveraging Random Matrix tools, the following \thmref{theorem:asy-behavior-E[Q]} proposes an asymptotic form that is $i.$ close to $\E_\mW[\mQ_m(\lambda)]$ under Assumptions~\ref{assumption:growth_rate}~and~\ref{assumption:regime_n}, and $ii.$ numerically more accessible (for the proof, see \Appref{sec:proof_theorem_1}).
\begin{theorem}[Asymptotic Deterministic Resolvent]
    Under Assumptions~\ref{assumption:growth_rate} (double asymptotic regime) and \ref{assumption:regime_n} (bounded spectrum), let $\lambda >0$ and let the \emph{deterministic resolvent} $\bar \mQ_m(\lambda) \in \R^{n \times n}$  be  
    \begin{equation}
      \bar \mQ_m(\lambda) = \left[ \tfrac{N}{m} \tfrac{1}{1+\delta} (\hat \mU_n - \gamma \hat \mV_n)^T\mPhi_{\SVisited}\hat \mU_n + \lambda \mI_n \right]^{-1},
       \label{def:Q_bar}
    \end{equation}
    where the \emph{deterministic Gram feature matrix} $\mPhi_{\SVisited} \in \R^{m \times m}$ is 
    \begin{equation}
        \mPhi_{\SVisited} = \E_{\vw \sim \N(\mathbf{0},\mI_d)}\left[\sigma(\vw^T \hat \mS)^T\sigma(\vw^T \hat \mS)\right],
        \label{def:Phi}
    \end{equation}
    and the \emph{correction factor}  $\delta$  is the unique, positive, solution to
    \begin{equation}
        \delta = \tfrac{1}{m} \Tr \left((\hat \mU_n - \gamma \hat \mV_n)^T\mPhi_{\SVisited} \hat \mU_n \left[\frac{N}{m} \frac{1}{1+\delta} (\hat \mU_n - \gamma \hat \mV_n)^T\mPhi_{\SVisited}\hat \mU_n + \lambda \mI_n \right]^{-1}\right).
        \label{def:delta}
    \end{equation}
    Then 
    $\lim_{m \to \infty} \left\lVert \E_\mW\left[\mQ_m(\lambda)\right] - \bar \mQ_m(\lambda) \right\rVert = 0.$
     \label{theorem:asy-behavior-E[Q]}
     The correction factor $\delta$ diminishes as $N$ or $\lambda$ grows (see  \lemref{lem:delta_decreasing_N} and~\ref{lem:delta_decreasing_lambda}).
\end{theorem}
\begin{remark}
Since $\delta \to 0$ when $N/m \to \infty$, the correction factor $\tfrac{1}{1+\delta}$ arises from our asymptotic regime, which keeps the ratio $N/m$ asymptotically constant (see  \lemref{lem:delta_positive_unique} for existence and uniqueness). Similar correction factors arise in related Random Matrix literature, which, however, mostly deals with positive semi-definite matrices \citep{couillet2011random, liu2019neural, liao2020random}. Our problem exceeds this frame, so we prove the result, including existence and uniqueness, with a somewhat more involved analysis based on the eigenspectrum of the products of matrices with positive-definite symmetric part and skew-symmetric matrices (see \Appref{sec:delta}).    
\end{remark}
\begin{remark}
    For the case of supervised learning, a comparable proposition is presented by \citet[Theorem 1]{louart2018random}. It constitutes a special case of \thmref{theorem:asy-behavior-E[Q]} with $\gamma=0$, which corresponds to the case where we learn the reward function.
\end{remark}
\begin{remark}
    Note that the eigenvalues of $\bar \mQ_m(\lambda)$ are not necessarily real, which renders many tools from the related Random Matrix literature not applicable, e.g., Stieltjes transforms 
    would provide information on the eigenspectrum density of matrices based on the trace of their resolvents.
\end{remark}
\subsection{Asymptotic Empirical Mean-Squared Bellman Error}
TD methods learn by minimizing the empirical MSBE (\eqref{def:MSBE_train}) and, under appropriate learning rates, converge towards the empirical MSBE of LSTD, as mentioned in \Secref{sec:system_model}. It is straightforward to show that this leads to an optimal
$\Etrain(\hat \vtheta) =\tfrac{\lambda^2}{n} \lVert \mQ_m(\lambda)\rtrain \rVert^2$ (see \Appref{sec:proof_theorem_2}).
Using concentration arguments for Gaussian distributions and Lipschitz applications, as well as \thmref{theorem:asy-behavior-E[Q]}, we derive the following deterministic form (see proof in \Appref{sec:proof_theorem_2}). 
\begin{theorem}[Asymptotic Empirical MSBE]
    Under the conditions of \thmref{theorem:asy-behavior-E[Q]}, the \emph{deterministic asymptotic empirical MSBE} is
    $\DEtrain = \tfrac{\lambda^2}{n}\lVert \bar \mQ_m(\lambda) \rtrain \rVert^2  +  \hat \Delta$,  with second order correction factor
    \begin{equation}
           \hat \Delta = \tfrac{\lambda^2}{n}\tfrac{\tfrac{1}{N} \Tr \left(\bar \mQ_m(\lambda)  \mPsi_2 \bar \mQ_m(\lambda)^T \right)}{1-\tfrac{1}{N} \Tr \left( \mPsi_2 \bar \mQ_m(\lambda)^T \mPsi_1 \bar \mQ_m(\lambda)  \right)} \lVert \bar \mQ_m(\lambda) \rtrain\rVert^2_{\mPsi_1}, \text{ where}
           \label{def:Delta_hat}
    \end{equation}
    \begin{equation}
        \mPsi_1 = \tfrac{N}{m}\tfrac{1}{1+\delta}\hat \mU_n^T \mPhi_{\SVisited} \hat \mU_n, \quad \text{and} \quad \mPsi_2 = \tfrac{N}{m}\tfrac{1}{1+\delta}(\hat \mU_n-\gamma \hat \mV_n)^T \mPhi_{\SVisited} (\hat \mU_n-\gamma \hat \mV_n).
        \label{def:Psi_1_and_Psi_2}
    \end{equation}
\label{theorem:asy-behavior-MSBE}
As $N, m, d \to \infty$ with asymptotic constant ratio $N/m$, $
        \Etrain(\hat \vtheta) - \DEtrain \xrightarrow{a.s} 0$.
\end{theorem}
\begin{remark}
   As $N/m \to \infty$, we find $\DMSBE \to 0$.
\end{remark}
\begin{remark}
   In supervised learning, a comparable proposition is presented by \citet[Theorem 3]{louart2018random}. 
    It is a special case of \thmref{theorem:asy-behavior-MSBE} with $\gamma=0$, where we learn the reward function $r$.
\end{remark}
\subsection{Asymptotic Mean-Squared Bellman Error}
While the empirical MSBE only takes states from the data set into account, the true MSBE (\eqref{def:MSBE}) involves all states in $\State$. To extend the convergence results from the previous section to this case, we require some further notations. 
Using a decomposition similar to \eqref{def:FXtrain_with_U_and_V}, we express $\FXtrain$ and $\FXtrainN$ as a product of the random feature matrix of the entire state space $\FState \in \R^{N \times \lvert\State \rvert}$ with $\mU_n \in \R^{\lvert\State \rvert \times n}$ and $\mV_n \in \R^{\lvert\State \rvert \times n}$ instead of $\FSVisited, \hat \mU_n, \hat \mV_n$.
We obtain a decomposition of the transition model matrix $\mA_n = \mU_n(\mU_n - \gamma \mV_n)^T$. $\mA_n$ was used by \cite{boyan1999least} to interpret LSTD as model-based RL. \cite{tsitsiklis1996analysis} and \cite{ nedic2003least} showed that $\E\left[\mA_n\right] \to \mD_\vpi \left[\mI_{\lvert \State \rvert} - \gamma \mP \right]$ as $n \to \infty$. 
The bound on the difference  $\lVert \mA_n - \mD_\vpi \left[\mI_{\lvert \State \rvert} - \gamma \mP \right] \rVert$ as a function of $n$ was studied by \cite{tagorti2015rate}. 
We make the following assumption on this norm:
\begin{assumption}
    As $n,m \to \infty$, we have $\lVert \mA_n - \mD_\vpi [\mI_{\lvert \State \rvert} - \gamma \mP ] \rVert = \mathcal{O}\left(\frac{1}{\sqrt{m}}\right).$
    \label{assumption:new_regime_n}
\end{assumption}
Using an approach similar to that of \thmref{theorem:asy-behavior-MSBE}, plus a detailed analysis of operator norms, we obtain the following deterministic form of the asymptotic MSBE
(proof in \Appref{sec:proof_theorem_3}):
\begin{theorem}[Asymptotic MSBE]
    Under Assumptions~\ref{assumption:growth_rate},~\ref{assumption:regime_n},~and~\ref{assumption:new_regime_n}, the \emph{deterministic asymptotic MSBE} is $\DMSBE = \left\lVert \bar \vr + \gamma \tfrac{1}{\sqrt{n}} \tfrac{N}{m}\tfrac{1}{1+\delta}\mP\mPhi_\State \mU_n \bar \mQ_m(\lambda) \rtrain - \tfrac{1}{\sqrt{n}} \tfrac{N}{m}\tfrac{1}{1+\delta}\mPhi_\State \mU_n \bar \mQ_m(\lambda) \rtrain \right\rVert_{\mD_\vpi}^2 + \Delta$, with second-order correction factor
    \begin{equation}
        \Delta = \tfrac{1}{n} \tfrac{\tfrac{1}{N}\Tr \left(\mLambda_\mP \left[ \mTheta_\State \mPsi_2 \mTheta_\State^T - 2\mTheta_\State (\mU_n - \gamma \mV_n)^T \mPsi_\State + \mPsi_\State \right] \right)}{1-\tfrac{1}{N} \Tr \left(\mPsi_2 \bar \mQ_m(\lambda)^T \mPsi_1 \bar \mQ_m(\lambda) \right)}\lVert \bar \mQ_m(\lambda) \rtrain \rVert_{\mPsi_1}^2, \text{ where}
        \label{def:Delta}
    \end{equation}
    \begin{align}
        \mPsi_\State &= \tfrac{N}{m}\tfrac{1}{1+\delta}\mPhi_\State, \quad \mLambda_\mP = [\mI_{\lvert \State \rvert}-\gamma \mP ]^T \mD_{\vpi} [\mI_{\lvert \State \rvert}-\gamma \mP ], \quad \text{and} \quad \mTheta_{\State} = \mPsi_\State \mU_n \bar \mQ_m(\lambda).
        \label{def:Psi_State_Lambda_P_Theta_State}
    \end{align}
    As $N, m, d \to \infty$ with asymptotic constant ratio $N/m$, $\MSBE(\hat \vtheta) - \DMSBE \xrightarrow{a.s} 0$.
    \label{theorem:asy-behavior-true-MSBE}
\end{theorem}
\begin{remark}
    Like the empirical $\DEtrain$ in \thmref{theorem:asy-behavior-MSBE}, the true $\DMSBE$ is also influenced by the correction terms $\delta$ and $\Delta$.
    Note that in asymptotic regimes where $N/m \to \infty$ or $\lambda \to \infty$, the correction terms vanish. When $N/m \to \infty$, $\DMSBE$ is independent of $\lambda$ as shown in details in \Appref{sec:reformulation_main_results}.
\end{remark}
\begin{remark}
    When all states have been visited, the common subexpressions in the second-order correction factors $\hat \Delta$ and $\Delta$ dominate so that $\hat \Delta$, $\Delta$ become similar (for a proof, see \lemref{lem:DMSBE_A_invertible}). 
    \label{remark:dmsbe_all_states_visited}
\end{remark}
\section{Numerical Experiments \label{sec:experiments}} 
 In this section, we provide an empirical evaluation, including a discussion of the behavior of the
correction factor $\delta$ from \thmref{theorem:asy-behavior-E[Q]}, and its impact on the empirical and true MSBE from \thmref{theorem:asy-behavior-MSBE} and \thmref{theorem:asy-behavior-true-MSBE}. Additional experiments can be found in \Appref{sec:additional_experiments}. 
\input{tikz_figures/delta_taxi.tex}
%
\paragraph{Experimental Setup.} We use the recursive regularized LSTD implementation of~\citet{dann2014policy} on three MRPs: a synthetic ergodic MRP ($500$ states); a gridworld MRP ($400$ states) obtained from a random policy in a $20 \times 20$ gridworld~\citep{Zafarali2018}; and a Taxi-v3 MRP ($356$ states) obtained from a learned policy in OpenAI gym Taxi-v3~\citep{towers_gymnasium_2023} (\Figref{fig:taxi_gymnasium}). In all cases, states are described by $d$-Gaussian vectors where $d=50$. 
For the random features, $\mW$ is drawn from a Gaussian distribution 
and $\sigma(\cdot)=\max(0, \cdot)$ is the ReLU function. 
For all experiments, $\mathcal{D}_\text{train}:=\{( \vs_i, r_i, \vs_i')\}_{i=1}^n$ is derived from a sample path of $n$ transitions with the same seed ($42$). For each instance $i$, we sample random features using the seed $i$. The following graphs  show averages over $30$ instances.
\paragraph{Correction Factor $\delta$ vs.\ Model Complexity.} 
The correction factor $\delta$ (\eqref{def:delta})
plays a key role in the asymptotic $\Etrain$ and $\MSBE$.  
\Figref{fig:delta_ratio_taxi} shows $\delta$ as a function of the model complexity $N/m$ and for different values of the regularization parameter $\lambda$. It confirms that, as stated in
\thmref{theorem:asy-behavior-E[Q]}, $\delta$ is a decreasing function of $N/m$.
For a small $\lambda$, we observe a sharp decrease at the interpolation threshold ($N=m$). E.g., for $\lambda=10^{-9}$, $\delta$ falls from an order of $10^{7}$ in under-parameterized models ($N<m$) to an order of $10^{1}$ in over-parameterized models ($N>m$). 
For larger values of $\lambda$, $\delta$ decreases more smoothly and has smaller values.
Further experiments on the behavior of $\delta$ and experiments for other environments are provided in \Appref{sec:additional_experiments}. 
\input{tikz_figures/msbe_taxi.tex}
%
\paragraph{Double Descent Behavior.}
As a consequence of the sharp transition of the correction factor $\delta$ for small $l_2$-regularization parameters, as discussed above, \thmref{theorem:asy-behavior-MSBE} and \thmref{theorem:asy-behavior-true-MSBE} predict a change in behavior of the empirical $\Etrain$ and true $\MSBE$ between the under- and overparameterized regimes. 
\Figref{fig:double_descent_taxi} shows both
$\Etrain$ and $\MSBE$ as a function of the model complexity $N/m$ with different $l_2$-regularization penalties $\lambda$ in Taxi-v3. 
Despite the fact that the equations for $\Etrain$ and $\MSBE$ were derived for the asymptotic case $N\to \infty$, we observe an almost perfect match with the numerically evaluated original definitions in \eqref{def:MSBE_train} and \eqref{def:MSBE}. 
For small $\lambda$, the true $\MSBE$ exhibits a peak around the interpolation threshold $N=m$, leading to a double descent phenomenon. In contrast, the empirical $\Etrain$ is close to its minimum at $N=m$ and almost constant for $N\geq m$, so no double descent is observed.
While for the Taxi-v3, the empirical $\Etrain$ is smaller than the true $\MSBE$, this is not necessarily the case in other environments, where the empirical $\Etrain$ can be larger overall than the true $\MSBE$ (see further experiments in \Appref{sec:additional_experiments}).
For larger $\lambda$, the double descent in the true $\MSBE$ disappears and the difference between the true $\MSBE$ and the empirical $\Etrain$ is less pronounced, although it may not vanish. \Appref{sec:MSVE} shows a similar double descent phenomenon for the Mean-Squared Value Error.
All the above observations are in accordance with established results in supervised learning~\citep{liao2020random}. 
\paragraph{Impact of the Number of Unvisited States and the Discount Factor $\gamma$.}
 Once all states have been visited,  $\MSBE$ and $\Etrain$ have similar behavior, with no peak at the interpolation threshold ($N=m$), see also Remark~\ref{remark:dmsbe_all_states_visited}. The experiments in \Figref{fig:double_descent_transitions_ergodic_mrp} depict this behavior. They also illustrate that the double descent phenomenon diminishes as the number of distinct unvisited states goes to zero.
 The experiments in \Figref{fig:double_descent_gamma_large_gridworld} illustrate that the discount factor $\gamma$ has little impact on the double descent phenomenon. 
\input{tikz_figures/msbe_visited_states_gamma.tex}
\section{Conclusion}
In this work, we have analyzed the performance of regularized LSTD with random features in a novel double asymptotic regime, where the number of parameters $N$ and distinct visited states $m$ go to infinity with a constant ratio.
We have established deterministic limit forms for both the empirical MSBE and true MSBE that feature correction terms. We have observed
that these correction terms are responsible for a double descent phenomenon in the true MSBE, similar to supervised learning, resulting in a sudden drop in performance for $N=m$. 
The correction terms vanish, and so does the double descent phenomenon when the $l_2$-regularization is increased or the number of unvisited states goes to zero. 
Directions for future work include a study of the off-policy setting, extending our results beyond one hidden layer to deep neural networks, and going beyond policy evaluation in order to investigate other RL algorithms, such as Q-Learning.

\section{Acknowledgements}
This work was supported by Ecole Polytechnique. We also thank Zhenyu Liao for his helpful dis- ´
cussions on this work.

\bibliography{./main.bib}

\begin{thebibliography}{51}
\providecommand{\natexlab}[1]{#1}
\providecommand{\url}[1]{\texttt{#1}}
\expandafter\ifx\csname urlstyle\endcsname\relax
  \providecommand{\doi}[1]{doi: #1}\else
  \providecommand{\doi}{doi: \begingroup \urlstyle{rm}\Url}\fi

\bibitem[Agazzi \& Lu(2022)Agazzi and Lu]{agazzi2022temporal}
Andrea Agazzi and Jianfeng Lu.
\newblock Temporal-difference learning with nonlinear function approximation: lazy training and mean field regimes.
\newblock In \emph{Mathematical and Scientific Machine Learning}, pp.\  37--74, 2022.

\bibitem[Ahmed(2018)]{Zafarali2018}
Zafarali Ahmed.
\newblock emdp.
\newblock \url{https://github.com/zafarali/emdp}, 2018.

\bibitem[Arora et~al.(2019)Arora, Du, Hu, Li, and Wang]{arora2019fine}
Sanjeev Arora, Simon Du, Wei Hu, Zhiyuan Li, and Ruosong Wang.
\newblock Fine-grained analysis of optimization and generalization for overparameterized two-layer neural networks.
\newblock In \emph{International Conference on Machine Learning}, pp.\  322--332, 2019.

\bibitem[Belkin et~al.(2018)Belkin, Hsu, Ma, and Mandal]{belkin2018reconciling}
Mikhail Belkin, Daniel Hsu, Siyuan Ma, and Soumik Mandal.
\newblock Reconciling modern machine learning and the bias-variance trade-of.
\newblock \emph{arXiv preprint arXiv:1812.11118}, 321, 2018.

\bibitem[Belkin et~al.(2020)Belkin, Hsu, and Xu]{belkin2020two}
Mikhail Belkin, Daniel Hsu, and Ji~Xu.
\newblock Two models of double descent for weak features.
\newblock \emph{SIAM Journal on Mathematics of Data Science}, 2\penalty0 (4):\penalty0 1167--1180, 2020.

\bibitem[Berthier et~al.(2022)Berthier, Kobeissi, and Bach]{berthier2022non}
Elo{\"\i}se Berthier, Ziad Kobeissi, and Francis Bach.
\newblock A non-asymptotic analysis of non-parametric temporal-difference learning.
\newblock \emph{Advances in Neural Information Processing Systems}, 35:\penalty0 7599--7613, 2022.

\bibitem[Bietti \& Mairal(2019)Bietti and Mairal]{bietti2019inductive}
Alberto Bietti and Julien Mairal.
\newblock On the inductive bias of neural tangent kernels.
\newblock \emph{Advances in Neural Information Processing Systems}, 32, 2019.

\bibitem[Boyan(1999)]{boyan1999least}
Justin~A. Boyan.
\newblock Least-squares temporal difference learning.
\newblock In \emph{International Conference on Machine Learning}, pp.\  49--56, 1999.

\bibitem[Bradtke \& Barto(1996)Bradtke and Barto]{bradtke1996linear}
Steven~J. Bradtke and Andrew~G. Barto.
\newblock Linear least-squares algorithms for temporal difference learning.
\newblock \emph{Machine Learning}, 22\penalty0 (1):\penalty0 33--57, 1996.

\bibitem[Cai et~al.(2019)Cai, Yang, Lee, and Wang]{cai2019neural}
Qi~Cai, Zhuoran Yang, Jason~D. Lee, and Zhaoran Wang.
\newblock Neural temporal-difference learning converges to global optima.
\newblock \emph{Advances in Neural Information Processing Systems}, 32, 2019.

\bibitem[Cao et~al.(2019)Cao, Fang, Wu, Zhou, and Gu]{cao2019towards}
Yuan Cao, Zhiying Fang, Yue Wu, Ding-Xuan Zhou, and Quanquan Gu.
\newblock Towards understanding the spectral bias of deep learning.
\newblock \emph{arXiv preprint arXiv:1912.01198}, 2019.

\bibitem[Chen et~al.(2013)Chen, Chen, and Gu]{chen2013efficient}
Shenglei Chen, Geng Chen, and Ruijun Gu.
\newblock An efficient {L}2-norm regularized least-squares temporal difference learning algorithm.
\newblock \emph{Knowledge-Based Systems}, 45:\penalty0 94--99, 2013.

\bibitem[Chizat \& Bach(2018)Chizat and Bach]{chizat2018global}
Lénaïc Chizat and Francis Bach.
\newblock On the global convergence of gradient descent for over-parameterized models using optimal transport.
\newblock \emph{Advances in Neural Information Processing Systems}, 31, 2018.

\bibitem[Chizat et~al.(2019)Chizat, Oyallon, and Bach]{chizat2019lazy}
Lénaïc Chizat, Edouard Oyallon, and Francis Bach.
\newblock On lazy training in differentiable programming.
\newblock \emph{Advances in Neural Information Processing Systems}, 32, 2019.

\bibitem[Couillet \& Debbah(2011)Couillet and Debbah]{couillet2011random}
Romain Couillet and Merouane Debbah.
\newblock \emph{Random Matrix Methods for Wireless Communications}.
\newblock Cambridge University Press, 2011.

\bibitem[Dann et~al.(2014)Dann, Neumann, Peters, et~al.]{dann2014policy}
Christoph Dann, Gerhard Neumann, Jan Peters, et~al.
\newblock Policy evaluation with temporal differences: A survey and comparison.
\newblock \emph{Journal of Machine Learning Research}, 15:\penalty0 809--883, 2014.

\bibitem[Dong et~al.(2020)Dong, Luo, Yu, Finn, and Ma]{dong2020expressivity}
Kefan Dong, Yuping Luo, Tianhe Yu, Chelsea Finn, and Tengyu Ma.
\newblock On the expressivity of neural networks for deep reinforcement learning.
\newblock In \emph{International Conference on Machine Learning}, pp.\  2627--2637, 2020.

\bibitem[Haarnoja et~al.(2018)Haarnoja, Zhou, Hartikainen, Tucker, Ha, Tan, Kumar, Zhu, Gupta, Abbeel, and Levine]{haarnoja2018}
Tuomas Haarnoja, Aurick Zhou, Kristian Hartikainen, George Tucker, Sehoon Ha, Jie Tan, Vikash Kumar, Henry Zhu, Abhishek Gupta, Pieter Abbeel, and Sergey Levine.
\newblock Soft actor-critic algorithms and applications.
\newblock \emph{CoRR}, 2018.

\bibitem[Hoffman et~al.(2011)Hoffman, Lazaric, Ghavamzadeh, and Munos]{hoffman2011regularized}
Matthew~W. Hoffman, Alessandro Lazaric, Mohammad Ghavamzadeh, and R{\'e}mi Munos.
\newblock Regularized least squares temporal difference learning with nested {L}2 and {L}1 penalization.
\newblock In \emph{European Workshop on Reinforcement Learning}, pp.\  102--114. Springer, 2011.

\bibitem[Horn \& Johnson(2012)Horn and Johnson]{horn2012matrix}
Roger~A. Horn and Charles~R. Johnson.
\newblock \emph{Matrix Analysis}.
\newblock Cambridge University Press, 2012.

\bibitem[Jacot et~al.(2018)Jacot, Gabriel, and Hongler]{jacot2018neural}
Arthur Jacot, Franck Gabriel, and Cl{\'e}ment Hongler.
\newblock Neural tangent kernel: Convergence and generalization in neural networks.
\newblock \emph{Advances in Neural Information Processing Systems}, 31, 2018.

\bibitem[Kumar et~al.(2020)Kumar, Agarwal, Ghosh, and Levine]{Kumar2020}
Aviral Kumar, Rishabh Agarwal, Dibya Ghosh, and Sergey Levine.
\newblock Implicit under-parameterization inhibits data-efficient deep reinforcement learning.
\newblock In \emph{International Conference on Learning Representations}, 2020.

\bibitem[Lazaric et~al.(2012)Lazaric, Ghavamzadeh, and Munos]{lazaric2012finite}
Alessandro Lazaric, Mohammad Ghavamzadeh, and R{\'e}mi Munos.
\newblock Finite-sample analysis of least-squares policy iteration.
\newblock \emph{Journal of Machine Learning Research}, 13:\penalty0 3041--3074, 2012.

\bibitem[Ledoux(2001)]{ledoux2001concentration}
Michel Ledoux.
\newblock \emph{The Concentration of Measure Phenomenon}.
\newblock American Mathematical Soc., 2001.

\bibitem[Lee et~al.(2019)Lee, Xiao, Schoenholz, Bahri, Novak, Sohl-Dickstein, and Pennington]{lee2019wide}
Jaehoon Lee, Lechao Xiao, Samuel Schoenholz, Yasaman Bahri, Roman Novak, Jascha Sohl-Dickstein, and Jeffrey Pennington.
\newblock Wide neural networks of any depth evolve as linear models under gradient descent.
\newblock \emph{Advances in Neural Information Processing Systems}, 32, 2019.

\bibitem[Liao et~al.(2020)Liao, Couillet, and Mahoney]{liao2020random}
Zhenyu Liao, Romain Couillet, and Michael~W. Mahoney.
\newblock A random matrix analysis of random fourier features: beyond the {G}aussian kernel, a precise phase transition, and the corresponding double descent.
\newblock \emph{Advances in Neural Information Processing Systems}, 33:\penalty0 13939--13950, 2020.

\bibitem[Liu et~al.(2019)Liu, Cai, Yang, and Wang]{liu2019neural}
Boyi Liu, Qi~Cai, Zhuoran Yang, and Zhaoran Wang.
\newblock Neural trust region/proximal policy optimization attains globally optimal policy.
\newblock \emph{Advances in Neural Information Processing Systems}, 32, 2019.

\bibitem[Louart et~al.(2018)Louart, Liao, and Couillet]{louart2018random}
Cosme Louart, Zhenyu Liao, and Romain Couillet.
\newblock A random matrix approach to neural networks.
\newblock \emph{The Annals of Applied Probability}, 28\penalty0 (2):\penalty0 1190--1248, 2018.

\bibitem[Luo et~al.(2020)Luo, Meng, He, Chen, and Wang]{Luo2020}
Xufang Luo, Qi~Meng, Di~He, Wei Chen, and Yunhong Wang.
\newblock I4r: Promoting deep reinforcement learning by the indicator for expressive representations.
\newblock In \emph{IJCAI}, 2020.

\bibitem[Lyle et~al.(2021)Lyle, Rowland, and Dabney]{lyle2021understanding}
Clare Lyle, Mark Rowland, and Will Dabney.
\newblock Understanding and preventing capacity loss in reinforcement learning.
\newblock In \emph{International Conference on Learning Representations}, 2021.

\bibitem[Mei \& Montanari(2022)Mei and Montanari]{mei2022generalization}
Song Mei and Andrea Montanari.
\newblock The generalization error of random features regression: Precise asymptotics and the double descent curve.
\newblock \emph{Communications on Pure and Applied Mathematics}, 75\penalty0 (4):\penalty0 667--766, 2022.

\bibitem[Mei et~al.(2018)Mei, Montanari, and Nguyen]{mei2018mean}
Song Mei, Andrea Montanari, and Phan-Minh Nguyen.
\newblock A mean field view of the landscape of two-layer neural networks.
\newblock \emph{Proceedings of the National Academy of Sciences}, 115\penalty0 (33):\penalty0 E7665--E7671, 2018.

\bibitem[Mnih et~al.(2015)Mnih, Kavukcuoglu, Silver, Rusu, Veness, Bellemare, Graves, Riedmiller, Fidjeland, Ostrovski, et~al.]{mnih2015human}
Volodymyr Mnih, Koray Kavukcuoglu, David Silver, Andrei~A. Rusu, Joel Veness, Marc~G. Bellemare, Alex Graves, Martin Riedmiller, Andreas~K. Fidjeland, Georg Ostrovski, et~al.
\newblock Human-level control through deep reinforcement learning.
\newblock \emph{Nature}, 518\penalty0 (7540):\penalty0 529--533, 2015.

\bibitem[Nedi{\'c} \& Bertsekas(2003)Nedi{\'c} and Bertsekas]{nedic2003least}
A.~Nedi{\'c} and Dimitri~P. Bertsekas.
\newblock Least squares policy evaluation algorithms with linear function approximation.
\newblock \emph{Discrete Event Dynamic Systems}, 13\penalty0 (1-2):\penalty0 79--110, 2003.

\bibitem[Rahimi \& Recht(2007)Rahimi and Recht]{rahimi2007random}
Ali Rahimi and Benjamin Recht.
\newblock Random features for large-scale kernel machines.
\newblock \emph{Advances in Neural Information Processing Systems}, 20, 2007.

\bibitem[Robbins \& Monro(1951)Robbins and Monro]{robbins1951stochastic}
Herbert Robbins and Sutton Monro.
\newblock A stochastic approximation method.
\newblock \emph{The Annals of Mathematical Statistics}, pp.\  400--407, 1951.

\bibitem[Rotskoff \& Vanden-Eijnden(2018)Rotskoff and Vanden-Eijnden]{rotskoff2018parameters}
Grant Rotskoff and Eric Vanden-Eijnden.
\newblock Parameters as interacting particles: long time convergence and asymptotic error scaling of neural networks.
\newblock \emph{Advances in Neural Information Processing Systems}, 31, 2018.

\bibitem[Sch{\"o}lkopf \& Smola(2002)Sch{\"o}lkopf and Smola]{scholkopf2002learning}
Bernhard Sch{\"o}lkopf and Alexander~J Smola.
\newblock \emph{Learning with kernels: support vector machines, regularization, optimization, and beyond}.
\newblock MIT press, 2002.

\bibitem[Schulman et~al.(2017)Schulman, Wolski, Dhariwal, Radford, and Klimov]{schulman2017proximal}
John Schulman, Filip Wolski, Prafulla Dhariwal, Alec Radford, and Oleg Klimov.
\newblock Proximal policy optimization algorithms.
\newblock \emph{arXiv preprint arXiv:1707.06347}, 2017.

\bibitem[Stachurski(2009)]{stachurski2009economic}
John Stachurski.
\newblock \emph{Economic dynamics: theory and computation}.
\newblock MIT Press, 2009.

\bibitem[Sutton(1988)]{sutton1988learning}
Richard~S. Sutton.
\newblock Learning to predict by the methods of temporal differences.
\newblock \emph{Machine Learning}, 3\penalty0 (1):\penalty0 9--44, 1988.

\bibitem[Sutton \& Barto(2018)Sutton and Barto]{sutton2018reinforcement}
Richard~S. Sutton and Andrew~G. Barto.
\newblock \emph{Reinforcement Learning: An Introduction}.
\newblock MIT Press, 2018.

\bibitem[Tagorti \& Scherrer(2015)Tagorti and Scherrer]{tagorti2015rate}
Manel Tagorti and Bruno Scherrer.
\newblock On the rate of convergence and error bounds for {LSTD}($\lambda$).
\newblock In \emph{International Conference on Machine Learning}, pp.\  1521--1529. PMLR, 2015.

\bibitem[Tao(2012)]{tao2012topics}
Terence Tao.
\newblock \emph{Topics in Random Matrix Theory}, volume 132.
\newblock American Mathematical Soc., 2012.

\bibitem[Thomas(2022)]{thomas2022role}
Valentin Thomas.
\newblock On the role of overparameterization in off-policy temporal difference learning with linear function approximation.
\newblock \emph{Advances in Neural Information Processing Systems}, 35:\penalty0 37228--37240, 2022.

\bibitem[Towers et~al.(2023)Towers, Terry, Kwiatkowski, Balis, Cola, Deleu, Goulão, Kallinteris, K.G., Krimmel, Perez-Vicente, Pierré, Schulhoff, Tai, Shen, and Younis]{towers_gymnasium_2023}
Mark Towers, Jordan~K. Terry, Ariel Kwiatkowski, John~U. Balis, Gianluca~de Cola, Tristan Deleu, Manuel Goulão, Andreas Kallinteris, Arjun K.G., Markus Krimmel, Rodrigo Perez-Vicente, Andrea Pierré, Sander Schulhoff, Jun~Jet Tai, Andrew Tan~Jin Shen, and Omar~G. Younis.
\newblock Gymnasium, March 2023.
\newblock URL \url{https://zenodo.org/record/8127025}.

\bibitem[Tsitsiklis \& Van~Roy(1996)Tsitsiklis and Van~Roy]{tsitsiklis1996analysis}
John Tsitsiklis and Benjamin Van~Roy.
\newblock Analysis of temporal-difference learning with function approximation.
\newblock \emph{Advances in Neural Information Processing Systems}, 9, 1996.

\bibitem[Xiao et~al.(2021)Xiao, Dai, Mei, Ramirez, Gummadi, Harris, and Schuurmans]{xiao2021understanding}
Chenjun Xiao, Bo~Dai, Jincheng Mei, Oscar~A Ramirez, Ramki Gummadi, Chris Harris, and Dale Schuurmans.
\newblock Understanding and leveraging overparameterization in recursive value estimation.
\newblock In \emph{International Conference on Learning Representations}, 2021.

\bibitem[Yates(1995)]{yates1995framework}
Roy~D. Yates.
\newblock A framework for uplink power control in cellular radio systems.
\newblock \emph{IEEE Journal on Selected Areas in Communications}, 13\penalty0 (7):\penalty0 1341--1347, 1995.

\bibitem[Zhang et~al.(2021)Zhang, Bengio, Hardt, Recht, and Vinyals]{zhang2021understanding}
Chiyuan Zhang, Samy Bengio, Moritz Hardt, Benjamin Recht, and Oriol Vinyals.
\newblock Understanding deep learning (still) requires rethinking generalization.
\newblock \emph{Communications of the ACM}, 64\penalty0 (3):\penalty0 107--115, 2021.

\bibitem[Zhang et~al.(2020)Zhang, Cai, Yang, Chen, and Wang]{zhang2020can}
Yufeng Zhang, Qi~Cai, Zhuoran Yang, Yongxin Chen, and Zhaoran Wang.
\newblock Can temporal-difference and {Q}-learning learn representation? {A} mean-field theory.
\newblock \emph{Advances in Neural Information Processing Systems}, 33:\penalty0 19680--19692, 2020.

\end{thebibliography}
\bibliographystyle{iclr2024_conference}

\appendix
\newpage
\section{Additional Experiments \label{sec:additional_experiments}}
This appendix shows additional empirical results which cannot be put in the main body due to space limitations.
\subsection{$\delta$ in the Double Asymptotic Regime}
Like \Figref{fig:delta_ratio_taxi} in \Secref{sec:experiments}, \Figref{fig:delta_ratio_appendix} depicts the correction factor $\delta$ (\eqref{def:delta}) as a function of the complexity model $N/m$ in synthetic ergodic and Gridworld MRPs. $\delta$ shows a similar behavior than for the one observed in Taxi-v3 in \Figref{fig:delta_ratio_taxi}. 
\input{tikz_figures/delta_complexity.tex}
%

\Figref{fig:delta_weight_decay_appendix} depicts $\delta$ as a function of the $l_2$-regularization parameter for different ratio $N/m$. It confirms $\delta$ decreases monotonically as the $l_2$-regularization parameter $\lambda$ increases, as stated by \lemref{lem:delta_decreasing_lambda}. Furthermore, we observe the impact of regularization parameter $\lambda$ becomes less significant as the model complexity $N/m$ increases. Indeed, as $N/m$ increases, we observe a larger initially flat region and smaller values of $\delta$.
\input{tikz_figures/delta_weight_decay.tex}
%
\newpage
\subsection{Double Descent Behavior}
\Figref{fig:double_descent_appendix} shows both $\Etrain$ and $\MSBE$ as a function of the model complexity $N/m$ with different $l_2$-regularization penalties $\lambda$, in synthetic ergodic and Gridworld MRPs. Both $\MSBE$ and $\Etrain$ depict a similar double descent behavior for small $\lambda$ than in \Figref{fig:double_descent_taxi} in \Secref{sec:experiments}. We observe the empirical $\Etrain$ is not necessarily lower for over-parameterized ($N>m$) models. 
\input{tikz_figures/msbe_complexity.tex}
%
\newpage
\subsection{Impact of the Number of Unvisited States and of the Discount Factor $\gamma$}
Like \Figref{fig:double_descent_transitions_ergodic_mrp}, \Figref{fig:double_descent_transitions_appendix} depicts the behavior of the true MSBE for different numbers of distinct visited states $m$ and shows that as the number of distinct unvisited states goes to zero, the double descent phenomenon diminishes.
\input{tikz_figures/msbe_distinct_visited_states.tex}

\Figref{fig:double_descent_gamma_appendix} describes the impact of the discount factor $\gamma$ on the double descent phenomenon, and shows it remains true for all $\gamma$. 
\input{tikz_figures/msbe_gamma.tex}
%
\newpage
\subsection{Impact of the $l_2$-regularization Parameter on the MSBE}
\Figref{fig:double_descent_weight_decay_train_appendix} and \Figref{fig:double_descent_weight_decay_appendix} depict the empirical $\Etrain$ and the true $\MSBE$ as a function of the $l_2$-regularization parameter.
In supervised learning, the training error is an increasing function of the $l_2$-regularization parameter $\lambda$~\citep{liao2020random}, whereas the discount factor induces a more intricated behavior in RL as described by \Figref{fig:double_descent_weight_decay_train_appendix}. 
\input{tikz_figures/msbe_emp_weight_decay.tex}

In \Figref{fig:double_descent_weight_decay_appendix} at the interpolation threshold ($N=m$), we observe that as $\lambda$ increases, $\MSBE$ decreases and so does the peak observed in the previous experiments. 
For other ratios, the true MSBE depicts complex behaviors that may differ between under- and over-parameterized models and depends on the environment. 
Yet, both empirical and true MSBE show similar and opposite trends for the same $\lambda=\frac{\lambda_{m,n}}{mn}$, regardless of the number of samples $n$ and distinct visited states $m$.  In practice, the effective $l_2$-regularization parameter $\lambda_{m,n }$ is tuned. This suggests that it depends on both the number of transitions collected $n$ and the number of distinct visited states $m$, and not just on the number of samples $n$ as it is commonly suggested~\citep{hoffman2011regularized, chen2013efficient}.
\input{tikz_figures/msbe_weight_decay.tex}
\newpage
\subsection{Impact of the Second-Order Correction Factor $\Delta$ in the True MSBE}
\Figref{fig:Delta_appendix} depicts $\Delta$ in \thmref{theorem:asy-behavior-true-MSBE} as a function of the model complexity $N/m$ for small $l_2$-regularization parameter ($\lambda=10^{-9}$). It shows the double descent phenomenon in the true $\MSBE$ is mainly due to the second-order correction term $\Delta$.
\input{tikz_figures/Delta_ratio.tex}
%
\section{Mean-Squared Value Error \label{sec:MSVE}}
In this section, we study the Mean-Squared Value Error (MSVE) and observes a similar double descent behavior than the one observed for the true $\MSBE$. The MSVE is defined as
\begin{equation}
    \MSVE(\hat \vtheta) = \bigl\lVert \mV - \FState^T \hat \vtheta \bigr\rVert_{\mD_\vpi}^2.
    \label{def:MSVE}
\end{equation}
Using a similar approach than for \thmref{theorem:asy-behavior-true-MSBE}, we obtain the following deterministic form of the asymptotic MSVE:
\begin{corollary}[Asymptotic MSVE]
    Under Assumptions~\ref{assumption:growth_rate},~\ref{assumption:regime_n},~and~\ref{assumption:new_regime_n}, the \emph{deterministic asymptotic MSVE} is $\DMSVE = \left\lVert \mV - \tfrac{1}{\sqrt{n}} \tfrac{N}{m}\tfrac{1}{1+\delta}\mPhi_\State \mU_n \bar \mQ_m(\lambda) \rtrain \right\rVert_{\mD_\vpi}^2 + \Delta'$, with second-order correction factor
    \begin{equation}
        \Delta' = \tfrac{1}{n} \tfrac{\tfrac{1}{N}\Tr \left(\mD_\vpi \left[ \mTheta_\State \mPsi_2 \mTheta_\State^T - 2\mTheta_\State (\mU_n - \gamma \mV_n)^T \mPsi_\State + \mPsi_\State \right] \right)}{1-\tfrac{1}{N} \Tr \left(\mPsi_2 \bar \mQ_m(\lambda)^T \mPsi_1 \bar \mQ_m(\lambda) \right)}\lVert \bar \mQ_m(\lambda) \rtrain \rVert_{\mPsi_1}^2.
        \label{def:Delta'}
    \end{equation}
    As $N, m, d \to \infty$ with asymptotic constant ratio $N/m$, $\MSVE(\hat \vtheta) - \DMSVE \xrightarrow{a.s} 0$.
    \label{corollary:asy-behavior-true-MSVE}
\end{corollary}
\begin{proof}
    Using $\mD_{\vpi} = [\mI_m - \gamma \mP]^T\mD_{\vpi}\mD_{\vpi}^{-1}[\mI_m - \gamma \mP]^{-1 T}\mD_{\vpi}[\mI_m - \gamma \mP]^{-1}\mD_{\vpi}^{-1}\mD_{\vpi}[\mI_m - \gamma \mP]$ and a with similar proof than for \thmref{theorem:asy-behavior-true-MSBE}, we find \corref{corollary:asy-behavior-true-MSVE}.
\end{proof}
\begin{remark}
    Like $\DMSBE$ in \thmref{theorem:asy-behavior-true-MSBE}, the $\DMSVE$ is also influenced by the correction terms $\delta$ and $\Delta'$.
    Note that in asymptotic regimes where $N/m \to \infty$ or $\lambda \to \infty$, the correction terms vanish. When $N/m \to \infty$, $\DMSVE$ is independent of $\lambda$ as shown in details in \Appref{sec:reformulation_main_results}.
\end{remark}
\Figref{fig:double_descent_msve_appendix} shows both
the empirical $\EVtrain$ and the true $\MSVE$ as a function of the model complexity $N/m$ with different $l_2$-regularization penalties $\lambda$ in the synthetic ergodic, Girdworld and Taxi MRPs. 
Like the true $\MSBE$, we observe an almost perfect match with the numerically evaluated original definition in \eqref{def:MSVE}.

For small $\lambda$, like for the true $\MSBE$, the true $\MSVE$ exhibits a peak around the interpolation threshold $N=m$, leading to a double descent phenomenon. In contrast, the empirical $\EVtrain$ is close to its minimum at $N=m$ and almost constant for $N\geq m$, so no double descent is observed.
Unlike than for the true $\MSBE$, we observe that the empirical $\EVtrain$ is always smaller than the true $\MSVE$. For larger $\lambda$, the double descent in the true $\MSVE$ disappears and the difference between the true $\MSVE$ and the empirical $\EVtrain$ is less pronounced, although it may not vanish.
\input{tikz_figures/msve.tex}
\section{Reformulation of the Main Results \label{sec:reformulation_main_results}}
Let $\mathcal{C}$ be a compact set such that $\State \subset \mathcal{C}$. We extend the stationary distribution $\pi$ on $\mathcal{C}$ by setting $\pi(\vs)=0$ for $\vs \in \State \backslash \mathcal{C}$. We denote by $L^2(\mathcal{C}, \pi)$ the set of squared integrable functions $f : \mathcal{C} \to \R$ with respect to the distribution $\pi$ on $\mathcal{C}$, and the norm on $L^2(\mathcal{C}, \pi)$ defined as $\lVert f \rVert^2_{L^2(\mathcal{C}, \pi)}= \langle f, f \rangle = \int f(x)^2 \pi(dx)$. The Frobenius norm of a matrix $\mA$ is denoted as $\lVert \mA \rVert_F = \sqrt{\langle \mA, \mA \rangle_F}=\sqrt{\Tr(\mA^T\mA)}$.

In this section, we aim to reformulate the results from Section \ref{sec:main_results} in a feature space derived from the eigendecomposition of $\mPhi_\State$. The section is organized into three subsections: in Section~\ref{sec:asymptotic_feature_space} we introduce the new asymptotic feature space, we reformulate the results of Section~\ref{sec:main_results} in the new feature space in Section \ref{sec:main_results_rkhs} and we provide the proofs for these results in Section \ref{sec:proof_main_results_rkhs}.

\subsection{Asymptotic Feature Space \label{sec:asymptotic_feature_space}}
\paragraph{Asymptotic Feature Space.} Both \thmref{theorem:asy-behavior-MSBE} and \thmref{theorem:asy-behavior-true-MSBE} use either the deterministic Gram feature matrix $\mPhi_{\SVisited}$ (defined in \eqref{def:Phi}) or the deterministic Gram feature matrix $\mPhi_{\State}$ that are given by the continuous kernel function $\Phi: \mathcal{C} \times \mathcal{C} \to \R$ defined as
\begin{equation*}
    \Phi(\vs, \vs') = \E_{\vw \sim \N(\mathbf{0},\mI_d)}\left[\sigma(\vw^T \vs)^T\sigma(\vw^T \vs')\right].
\end{equation*}
Since $\Phi$ is continuous and $\mathcal{C}$ is compact, the Mercer’s theorem states~\citep{scholkopf2002learning} that
\begin{equation}
    \Phi(\vs, \vs') = \sum_{i=1}^M \nu_i \varphi_i(\vs) \varphi_i(\vs') = \sum_{i=1}^M \omega_i(\vs) \omega_i(\vs') ;
    \label{eq:mercer_theorem_Phi}
\end{equation}
where  $\{\nu_i\}_{i=1}^M$ and $\{\varphi_i\}_{i=1}^M$ are the eigenvalues and eigenfunctions of the Hilbert-Schmidt integral operators $T_{\Phi}: L^2(\mathcal{C}, \pi) \to L^2(\mathcal{C}, \pi), f \mapsto T_{\Phi}(f)(\vs') = \int_{\R^d} \Phi(\vs, \vs')f(\vs) \pi(\vs) d\vs$ and $\{\omega_i(\cdot)=\sqrt{\nu_i}\varphi_i(\cdot)\}_{i=1}^M$ are the rescaled eigenfunction. $\{\omega_i(\cdot)\}_{i=1}^M$ forms an orthogonal basis in $L^2(\mathcal{C}, \pi)$. Usually, $M$ is infinite. 
In the following of this section, we will find convenient to define a vector representation of functions in the asymptotic feature space defined by the feature map $\{\omega_i\}_{i=1}^M$. For any state matrix $\mA \in \R^{d \times p}$, we denote by $\mOmega_\mA \in \R^{M \times p}$ the feature matrices of $\mA$ so that $[\mOmega_\mA]_{ij} = \omega_i(\mA_j)$ for $\mA_j$ the $j^{th}$ column of $\mA$. With those new notations, we can decompose $\mPhi_{\SVisited}$ and $\mPhi_{\State}$ as 
\begin{equation*}
    \mPhi_{\SVisited} = \mOmega_{\SVisited}^T \mOmega_{\SVisited} \qquad \text{and} \qquad \mPhi_{\State} = \mOmega_{\State}^T \mOmega_{\State}.
\end{equation*}
\paragraph{Regularized LSTD in the Asymptotic Feature Space.} Let $\vv_{\bar \theta} \in \R^M$ the weight vector returned by the regularized LSTD with the rescaled $l_2$-regularization parameter $\lambda \tfrac{m(1+\delta)}{N}$ for the asymptotic features $\{\omega_i\}_{i=1}^M$ on transitions collected in $\Dtrain$:
\begin{equation*}
    \vv_{\bar \theta} = \tfrac{1}{\sqrt{n}}\Bigl[\bar \mA + \lambda \tfrac{m(1+\delta)}{N} \mI_M \Bigr]^{-1}\mOmega_{\SVisited}^T \hat \mU_n \rtrain,
\end{equation*}
where
\begin{equation}
        \bar \mA = \mOmega_{\SVisited} \hat \mU_n (\hat \mU_n - \gamma \hat \mV_n)^T \mOmega_{\SVisited}^T.
\end{equation}
\paragraph{Second-Order Correction Factor in the Asymptotic Feature Space.} Let $f_\Delta(\cdot)$ be the second-order correction function defined as
\begin{equation*}
    f_\Delta(\mB) = \tfrac{\lambda^2}{n}\tfrac{m^2(1+\delta)^2}{N^2}\tfrac{\tfrac{1}{N} \lVert  \vv_{\bar \vtheta} \rVert^2}{1-\tfrac{1}{N} \left\lVert \Bigl[\bar \mA + \lambda \tfrac{m(1+\delta)}{N} \mI_M \Bigr]^{-1} \bar \mA \right\rVert_F^2} \left\lVert \mB \left[\bar\mA+ \lambda \tfrac{m(1+\delta)}{N} \mI_M\right]^{-1} \right\rVert_{F}^2.
\end{equation*}
From the definition of $f_\Delta(\mB)$, it directly follows that $f_\Delta(\mB) \to 0$ as $N/m \to \infty$ or as $\lambda \to \infty$.
\subsection{Reformulation of the Main Results \label{sec:main_results_rkhs}}
\subsubsection{Empirical Mean-Squared Bellman Error}
\begin{theorem}[Asymptotic Empirical MSBE]
    Under the conditions of \thmref{theorem:asy-behavior-E[Q]}, the \emph{deterministic asymptotic empirical MSBE} is
    $$
        \DEtrain = \tfrac{1}{n} \left\lVert \rtrain + \gamma  \mOmega_{\mX'}^T  \vv_{\bar \vtheta} - \mOmega_{\mX}^T  \vv_{\bar \vtheta} \right\rVert^2  +  f_\Delta\bigl((\hat \mU_n - \gamma \hat \mV_n)^T \mOmega_{\SVisited}^T\bigr).
    $$  
    As $N, m, d \to \infty$ with asymptotic constant ratio $N/m$, 
    $$
            \Etrain(\hat \vtheta) - \DEtrain \xrightarrow{a.s} 0.
    $$
\end{theorem}
\begin{remark}
    The left-hand term $\tfrac{1}{n}\left\lVert \rtrain + \gamma  \mOmega_{\mX'}^T  \vv_{\bar \vtheta} - \mOmega_{\mX}^T  \vv_{\bar \vtheta} \right\rVert^2$ in $\DMSBE$ depicts the empirical MSBE for the regularized LSTD in the asymptotic features $\{\omega_i\}_{i=1}^M$ with the rescaled $l_2$-regularization parameter $\lambda \tfrac{m(1+\delta)}{N}$ on transitions collected in $\Dtrain$. 
\end{remark} 
\begin{remark}
    As $N/m$ increases, the influence of the $l_2$-regularization parameter $\lambda$ decreases.
\end{remark}
\begin{remark}
    As $N/m \to \infty$, $\DEtrain \to 0$.
\end{remark}
\subsubsection{Mean-Squared Bellman Error}
\begin{theorem}[Asymptotic MSBE]
    Under Assumptions~\ref{assumption:growth_rate},~\ref{assumption:regime_n},~and~\ref{assumption:new_regime_n}, the \emph{deterministic asymptotic MSBE} is 
    $$
        \DMSBE = \left\lVert \rtrain + \gamma  \mP \mOmega_{\State}^T  \vv_{\bar \vtheta} - \mOmega_{\State}^T  \vv_{\bar \vtheta} \right\rVert^2_{\mD_\vpi} + f_\Delta\bigl(\mD_{\vpi}^{\tfrac{1}{2}} (\mI_{\lvert \State \rvert} - \gamma \mP) \mOmega_\State^T\bigr).
    $$
    As $N, m, d \to \infty$ with asymptotic constant ratio $N/m$, 
    $$
        \MSBE(\hat \vtheta) - \DMSBE \xrightarrow{a.s} 0.
    $$
\end{theorem}
\begin{remark}
    The left-hand term $\left\lVert \rtrain + \gamma \mP \mOmega_{\State}^T  \vv_{\bar \vtheta} - \mOmega_{\State}^T  \vv_{\bar \vtheta} \right\rVert_{\mD_\vpi}^2$ in $\DMSBE$ depicts the true MSBE for the regularized LSTD in the asymptotic features $\{\omega_i\}_{i=1}^M$ with the rescaled $l_2$-regularization parameter $\lambda \tfrac{m(1+\delta)}{N}$ on transitions collected in $\Dtrain$. 
\end{remark} 
\begin{remark}
    As $N/m$ increases, the influence of the $l_2$-regularization parameter $\lambda$ decreases.
\end{remark}
\begin{remark}
    As $N/m \to \infty$, $\DMSBE$ converges to the MSBE of $\vv_{\bar \vtheta}$ without the $l_2$ regularization parameter.
\end{remark}
\subsubsection{Mean-Squared Value Error}
\begin{corollary}[Asymptotic MSVE]
    Under Assumptions~\ref{assumption:growth_rate},~\ref{assumption:regime_n},~and~\ref{assumption:new_regime_n}, the \emph{deterministic asymptotic MSVE} is 
    $$
        \DMSVE = \left\lVert \mV - \mOmega_{\State}^T  \vv_{\bar \vtheta} \right\rVert^2_{\mD_\vpi} + f_\Delta\bigl(\mD_{\vpi}^{\tfrac{1}{2}} \mOmega_\State^T\bigr).
    $$
    As $N, m, d \to \infty$ with asymptotic constant ratio $N/m$, 
    $$
        \MSVE(\hat \vtheta) - \DMSVE \xrightarrow{a.s} 0.
    $$
\end{corollary}
\begin{remark}
    The left-hand term $\left\lVert \mV - \mOmega_{\State}^T  \vv_{\bar \vtheta} \right\rVert_{\mD_\vpi}^2$ in $\DMSVE$ depicts the MSVE for the regularized LSTD in the asymptotic features $\{\omega_i\}_{i=1}^M$ with the rescaled $l_2$-regularization parameter $\lambda \tfrac{m(1+\delta)}{N}$ on transitions collected in $\Dtrain$. 
\end{remark} 
\begin{remark}
    As $N/m$ increases, the influence of the $l_2$-regularization parameter $\lambda$ decreases.
\end{remark}
\begin{remark}
    As $N/m \to \infty$, $\DMSVE$ converges to the MSVE of $\vv_{\bar \vtheta}$ without the $l_2$ regularization parameter.
\end{remark}
\subsection{Proofs of Results found in \secref{sec:main_results_rkhs} \label{sec:proof_main_results_rkhs}}
\begin{theorem}[Asymptotic Empirical MSBE]
    Under the conditions of \thmref{theorem:asy-behavior-E[Q]}, the \emph{deterministic asymptotic empirical MSBE} is
    $$
        \DEtrain = \tfrac{1}{n} \left\lVert \rtrain + \gamma  \mOmega_{\mX'}^T  \vv_{\bar \vtheta} - \mOmega_{\mX}^T  \vv_{\bar \vtheta} \right\rVert^2  +  f_\Delta\bigl((\hat \mU_n - \gamma \hat \mV_n)^T \mOmega_{\SVisited}^T\bigr).
    $$  
    As $N, m, d \to \infty$ with asymptotic constant ratio $N/m$, 
    $$
            \Etrain(\hat \vtheta) - \DEtrain \xrightarrow{a.s} 0.
    $$
\end{theorem}
\begin{proof}
    We have 
    \begin{align*}
        \tfrac{\lambda^2}{n} \lVert \bar \mQ_m(\lambda) \rtrain \rVert^2 &= \tfrac{1}{n} \left\lVert \rtrain + \gamma \tfrac{N}{m}\tfrac{1}{1+\delta} \hat \mV_n^T \mOmega_{\SVisited}^T \mOmega_{\SVisited} \hat \mU_n \bar \mQ_m(\lambda) \rtrain - \tfrac{N}{m}\tfrac{1}{1+\delta} \hat \mU_n \mOmega_{\SVisited}^T \mOmega_{\SVisited} \hat \mU_n \bar \mQ_m(\lambda) \rtrain \right\rVert^2  \\
        &= \tfrac{1}{n} \left\lVert \rtrain + \gamma  \mOmega_{\mX'}^T  \vv_{\bar \vtheta} - \mOmega_{\mX}^T  \vv_{\bar \vtheta} \right\rVert^2.
    \end{align*}
    For $\hat \Delta$, we have
    \begin{align*}
        &\Tr \left(\bar \mQ_m(\lambda)  \mPsi_2 \bar \mQ_m(\lambda)^T\right)  \\
        &= \tfrac{N}{m}\tfrac{1}{1+\delta} \Tr \left(\bar \mQ_m(\lambda)  (\hat \mU_n - \gamma \hat \mV_n)^T \mPhi_{\SVisited} (\hat \mU_n - \gamma \hat \mV_n)  \bar \mQ_m(\lambda)^T\right) \\
        &= \tfrac{m(1+\delta)}{N} \Tr \left((\hat \mU_n - \gamma \hat \mV_n)^T \mOmega_{\SVisited}^T \Bigl[\bar \mA + \lambda \tfrac{m(1+\delta)}{N} \mI_M\Bigr]^{-1}  \Bigl[\bar \mA + \lambda \tfrac{m(1+\delta)}{N} \mI_M\Bigr]^{-1 T} \mOmega_{\SVisited} (\hat \mU_n - \gamma \hat \mV_n) \right) \\
        &= \tfrac{m(1+\delta)}{N} \left\lVert(\hat \mU_n - \gamma \hat \mV_n)^T \mOmega_{\SVisited}^T \Bigl[\bar \mA + \lambda \tfrac{m(1+\delta)}{N} \mI_M\Bigr]^{-1}  \right\rVert_F^2
    \end{align*}
    \begin{align*}
        &\Tr \left(\mPsi_2 \bar \mQ_m(\lambda)^T \mPsi_1 \bar \mQ_m(\lambda) \right) \\
        &= \tfrac{N^2}{m^2}\tfrac{1}{(1+\delta)^2}\Tr\left((\hat \mU_n - \gamma \hat \mV_n)^T \mOmega_{\SVisited}^{T} \mOmega_{\SVisited} (\hat \mU_n - \gamma \hat \mV_n) \bar \mQ_m(\lambda)^T \hat \mU_n^T \mOmega_{\SVisited}^{T} \mOmega_{\SVisited} \hat \mU_n \bar \mQ_m(\lambda) \right) \\
        &= \Tr\left(\bar \mA^T \Bigl[\bar \mA + \lambda \tfrac{m(1+\delta)}{N}\mI_M\Bigr]^{-1 T} \Bigl[\bar \mA + \lambda \tfrac{m(1+\delta)}{N} \mI_M \Bigr]^{-1} \bar \mA \right) \\
        &= \left\lVert \Bigl[\bar \mA + \lambda \tfrac{m(1+\delta)}{N}\mI_M \Bigr]^{-1} \bar \mA \right\rVert_F^2,
    \end{align*}
    and
    \begin{align*}
        \lVert \bar \mQ_m(\lambda) \rtrain \rVert_{\mPsi_1}^2 
          &= \tfrac{m(1+\delta)}{n} \lVert  \vv_{\bar \vtheta} \rVert^2
    \end{align*}
\end{proof}
\begin{theorem}[Asymptotic MSBE]
    Under Assumptions~\ref{assumption:growth_rate},~\ref{assumption:regime_n},~and~\ref{assumption:new_regime_n}, the \emph{deterministic asymptotic MSBE} is 
    $$
        \DMSBE = \left\lVert \rtrain + \gamma  \mP \mOmega_{\State}^T  \vv_{\bar \vtheta} - \mOmega_{\State}^T  \vv_{\bar \vtheta} \right\rVert^2_{\mD_\vpi} + f_\Delta\bigl(\mD_{\vpi}^{\tfrac{1}{2}} (\mI_{\lvert \State \rvert} - \gamma \mP) \mOmega_\State^T\bigr).
    $$
    As $N, m, d \to \infty$ with asymptotic constant ratio $N/m$, 
    $$
        \MSBE(\hat \vtheta) - \DMSBE \xrightarrow{a.s} 0.
    $$
\end{theorem}
\begin{proof}
    We can rewrite
    \begin{align*}
        &\Tr\left(\mLambda_\mP \mTheta_\State \mPsi_2 \mTheta_\State^T \right) \\
        &= \tfrac{N}{m}\tfrac{1}{1+\delta}\Tr\left(\bigl[\bar\mA+ \lambda \tfrac{m(1+\delta)}{N} \mI_M\bigr]^{-1T} \bar \mA^T  \mOmega_{\State} [\mI_{\lvert \State \rvert}-\gamma \mP ]^T \mD_{\vpi} [\mI_{\lvert \State \rvert}-\gamma \mP ] \mOmega_{\State}^T   \bar \mA\bigl[\bar\mA+ \lambda \tfrac{m(1+\delta)}{N} \mI_M\bigr]^{-1}  \right) \\
        &= \tfrac{N}{m}\tfrac{1}{1+\delta} \left\lVert  [\mI_{\lvert \State \rvert}-\gamma \mP ] \mOmega_{\State}^T  \bar \mA\bigl[\bar\mA+ \lambda \tfrac{m(1+\delta)}{N} \mI_M\bigr]^{-1}  \right\rVert_{F, \mD_{\vpi}}^2, \\
        &\tfrac{N}{m}\tfrac{1}{1+\delta}\Tr\left(\mLambda_\mP \mPhi_\State \right) = \tfrac{N}{m}\tfrac{1}{1+\delta}\left\lVert [\mI_{\lvert \State \rvert}-\gamma \mP ] \mOmega_{\State}^T  \right\rVert_{F, \mD_\vpi}^2, 
    \end{align*}
    and
    \begin{align*}
        \Tr\left(\mLambda_\mP \mTheta_\State (\mU_n - \gamma \mV_n)^T \mPsi_\State \right) 
        &= \tfrac{N}{m}\tfrac{1}{1+\delta}\Tr\left(\mLambda_\mP \mOmega_{\State}^T \bar \mA\bigl[\bar\mA+ \lambda \tfrac{m(1+\delta)}{N} \mI_M\bigr]^{-1}  \mOmega_{\State} \right) \\
        &=\tfrac{N}{m}\tfrac{1}{1+\delta}\Tr\left(\mOmega_{\State} [\mI_{\lvert \State \rvert}-\gamma \mP ]^T \mD_\vpi  [\mI_{\lvert \State \rvert}-\gamma \mP]\mOmega_{\State}^T  \bar \mA\bigl[\bar\mA+ \lambda \tfrac{m(1+\delta)}{N} \mI_M\bigr]^{-1} \right) \\
        &=\tfrac{N}{m}\tfrac{1}{1+\delta}\left\langle [\mI_{\lvert \State \rvert}-\gamma \mP ]\mOmega_{\State}^T ,  [\mI_{\lvert \State \rvert}-\gamma \mP] \mOmega_{\State}^T  \bar \mA\bigl[\bar\mA+ \lambda \tfrac{m(1+\delta)}{N} \mI_M\bigr]^{-1} \right\rangle_{F, \mD_\vpi}.
    \end{align*}
    Therefore,
    \begin{align*}
        &\Tr \left(\mLambda_\mP \left[ \mTheta_\State \mPsi_2 \mTheta_\State^T - 2\mTheta_\State (\mU_n - \gamma \mV_n)^T \mPsi_\State + \mPsi_\State \right] \right) \\
        &= \tfrac{N}{m}\tfrac{1}{1+\delta} \left\lVert [\mI_{\lvert \State \rvert}-\gamma \mP ]\mOmega_{\State}^T  -  [\mI_{\lvert \State \rvert}-\gamma \mP ] \mOmega_{\State}^T  \bar \mA\bigl[\bar\mA+ \lambda \tfrac{m(1+\delta)}{N} \mI_M\bigr]^{-1} \right\rVert_{F, \mD_{\vpi}}^2 \\
        &= \lambda^2\left\lVert  [\mI_{\lvert \State \rvert}-\gamma \mP ] \mOmega_{\State}^T  \bigl[\bar\mA+ \lambda \tfrac{m(1+\delta)}{N} \mI_M\bigr]^{-1} \right\rVert_{F, \mD_{\vpi}}^2.
    \end{align*}
\end{proof}
\section{Evaluation of \texorpdfstring{$\mPhi_{\State}$}{Phi_State} or \texorpdfstring{$\mPhi_{\SVisited}$}{Phi_SVisited} \label{sec:Phi}}
The evaluation of $\mPhi_{\SVisited} = \E_\vw\bigl[\sigma(\vw^T \hat \mS)^T\sigma(\vw^T \hat \mS)\bigr]$ or $\mPhi_{\State} = \E_\vw\bigl[\sigma(\vw^T \mS)^T\sigma(\vw^T \mS)\bigr]$ naturally boils down to the evaluation of its individual entries and thus to the calculus, for arbitrary vectors $\va, \vb \in \R^d$ of
\begin{equation}
    \mPhi_{ab} = \E\bigl[\sigma(\vw^T\va)\sigma(\vw^T \vb)\bigr] = (2\pi)^{-\frac{p}2} \int \sigma(\varphi(\tilde \vw)^T \va)\sigma( \varphi(\tilde \vw)^T \vb) e^{-\frac12 \lVert \tilde \vw \rVert^2}d\tilde \vw.
    \label{def:phi_a_b}
\end{equation}
The evaluation of \eqref{def:phi_a_b} can be obtained through various integration tricks for a wide family of mappings $\varphi(\cdot)$ and activation functions $\sigma(\cdot)$. We provide in Table~\ref{tab:Phi} (found in \citet{louart2018random}) the
values of $\mPhi_{\va\vb}$ for $\vw \sim \mathcal{N}(0,\mI_d)$ (i.e., for $\varphi(t) = t$) and for a set of activation functions $\sigma(\cdot)$ not necessarily satisfying the Lipschitz continuity. In experiments in \Secref{sec:experiments}, we focus only on the ReLU function, i.e., $\sigma(t)=\max(t, 0)$.
\begin{table}[htbp]
    \caption{Values of $\mPhi_{\va\vb}$ for $\vw \sim \mathcal{N}(0,\mI_d)$, $\angle(\va,\vb)\equiv \frac{\va^T \vb}{\lVert \va \rVert \lVert \vb\rVert}$~\citep{louart2018random}.}
    \label{tab:Phi}
	\renewcommand{\arraystretch}{1.2}
    \centering
    \begin{tabular}{cc}
    	\toprule
    	$\sigma(t)$ & $\mPhi_{\va\vb}$ \\ \hline
    	\noalign{\smallskip}
    	$t$ & $\va^T \vb$ \\
    	$\max(t,0)$ & $\frac1{2\pi}\lVert \va \rVert \lVert \vb \rVert \left( \angle(\va, \vb) \arccos (-\angle(\va,\vb)) + \sqrt{1-\angle(\va, \vb)^2} \right)$\\
    	$|t|$ & $\frac2{\pi}\lVert \va \rVert \lVert \vb \rVert \left( \angle(\va, \vb) \arcsin (\angle(\va, \vb)) + \sqrt{1-\angle(\va, \vb)^2} \right)$\\
    	${\rm erf}(t)$ & $\frac2{\pi}\arcsin  \left( \frac{2\va^T \vb}{\sqrt{ (1+2\lVert \va \rVert^2)(1+2\lVert \vb \rVert^2) }} \right)$ \\
    	$1_{\{t>0\}}$ & $\frac12-\frac1{2\pi} \arccos (\angle(\va, \vb))$ \\
    	${\rm sign}(t)$ & $\frac2\pi \arcsin (\angle(\va, \vb))$ \\
    	$\cos(t)$ & $\exp(-\frac12(\lVert \va \rVert^2 + \lVert \vb \rVert^2))\cosh(\va^T \vb)$ \\
    	$\sin(t)$ & $\exp(-\frac12(\lVert \va \rVert^2+\lVert \vb \rVert^2))\sinh(\va^T \vb)$. \\
    	\bottomrule
    \end{tabular}
\end{table}
\newpage
\section{Proof of Theorem~\ref{theorem:asy-behavior-E[Q]} \label{sec:proof_theorem_1}}
Under Assumptions~\ref{assumption:growth_rate}~and~\ref{assumption:regime_n}, this section is dedicated to prove the asymptotic equivalence between $\E[\mQ_m(\lambda)]$ and
 \begin{equation*}
    \bar \mQ_m(\lambda) = \biggl[ \frac{N}{m} \frac{1}{1+\delta} (\hat \mU_n - \gamma \hat \mV_n)^T\mPhi_{\SVisited}\hat \mU_n + \lambda \mI_n \biggr]^{-1}
\end{equation*}
defined in \thmref{theorem:asy-behavior-E[Q]}, when $N, m \to \infty$.
In order to prove \thmref{theorem:asy-behavior-E[Q]}, we shall proceed by introducing an intermediary resolvent $\tilde \mQ_m(\lambda)$ (defined in \eqref{def:Q_tilde}), and show subsequently under Assumptions~\ref{assumption:growth_rate}~and~\ref{assumption:regime_n} that
\begin{equation*}
    \lVert \E[\mQ_m(\lambda)] - \tilde \mQ_m(\lambda) \rVert \to 0 \quad \text{and} \quad \lVert \tilde \mQ_m(\lambda) - \bar \mQ_m(\lambda) \rVert \to 0,
\end{equation*}
as $N, m \to \infty$. 

In order to simplify the notations, we denote by $\mQ_m$ the resolvent $\mQ_m(\lambda)$. The first half of the proof is dedicated to \lemref{lem:first-asy-behavior-E[Q]}, which proposes a first characterization of $\E[\mQ_m]$ by $\tilde \mQ_m$ as $N, m \to \infty$ under Assumptions~\ref{assumption:growth_rate}~and~\ref{assumption:regime_n}. This preliminary step is classical in studying resolvents in Random Matrix literature~\citep{louart2018random, liao2020random} as the direct comparison of $\E[\mQ_m]$ to $\bar \mQ_m$ with the implicit $\delta$ (\eqref{def:delta}) may be cumbersome.
\begin{lemma}
    Under Assumptions~\ref{assumption:growth_rate}~and~\ref{assumption:regime_n}, let $\lambda >0$ and let $\tilde \mQ_m(\lambda) \in \R^{n \times n}$ be the resolvent defined as 
    \begin{equation}
      \tilde \mQ_m(\lambda) = \left[ \frac{N}{m} \frac{1}{1+\alpha} (\hat \mU_n - \gamma \hat \mV_n)^T\mPhi_{\SVisited}\hat \mU_n + \lambda \mI_n \right]^{-1};
      \label{def:Q_tilde}
    \end{equation}
    for the deterministic \emph{Gram feature matrix}  
    \begin{equation*}
        \mPhi_{\SVisited} = \E_{\vw \sim \N(\mathbf{0},\mI_d)}\bigl[\sigma(\vw^T \hat \mS)^T\sigma(\vw^T \hat \mS)\bigr],
    \end{equation*}
    and
    \begin{equation}
        \alpha = \frac1m \Tr \bigl((\hat \mU_n - \gamma \hat \mV_n)^T \mPhi_{\SVisited} \hat \mU_n \E[\mQ_-(\lambda)]\bigr),
        \label{def:def_alpha}
    \end{equation}
    where
    \begin{align}
        \mQ_{-}(\lambda)=\left[\frac{1}{m}(\hat \mU_n - \gamma \hat \mV_n)^T \sigma(\mW_{-}\hat \mS)^T \sigma(\mW_{-} \hat \mS) \hat \mU_n + \lambda \mI_{n} \right]^{-1},
        \label{def:Q_-}
    \end{align}
    for which $\mW_{-} \in \R^{(N-1) \times d}$ depicts the submatrix of the weight matrix $\mW$ (defined in \eqref{def:rf_map}) without the first row.
    Then, 
    \begin{equation*}
        \lim_{m \to \infty} \lVert \E_\mW[\mQ_m(\lambda)] - \tilde \mQ_m(\lambda) \rVert = 0.
    \end{equation*}
    \label{lem:first-asy-behavior-E[Q]}
\end{lemma}
\begin{remark}
    Firstly, we can note that $\alpha$ is uniformly bounded. Since $\frac{1}{m}\Tr(\mPhi_{\SVisited})=\E\Bigl[\frac{1}{m} \lVert \sigma(\vw^T\hat \mS) \rVert^2 \Bigr]$ and from \lemref{lem:normal_concentration_resolvent}, we have
    \begin{equation}
            \frac{1}{m}\Tr(\mPhi_{\SVisited})=\int_0^\infty \Pr\left(\frac{1}{m} \lVert \sigma(\vw^T\hat \mS) \rVert^2 > t \right) dt = \int_0^\infty 2t \Pr\left(\frac{1}{m} \lVert \sigma(\vw^T\hat \mS) \rVert > t \right) dt = \mathcal{O}(1).
        \label{eq:upper_bound_tr_phi}
    \end{equation}
    We deduce that
    \begin{equation}
        \alpha = \frac{1}{m} \Tr\bigl((\hat \mU_n - \gamma \hat \mV_n)^T \mPhi_{\SVisited} \hat \mU_n \E[\mQ_-]\bigr) \leq \lVert \hat \mU_n \E[\mQ_-] (\hat \mU_n - \gamma \hat \mV_n)^T \rVert \frac{1}{m}\Tr(\mPhi_{\SVisited}) = \mathcal{O}(1), 
        \label{eq:bound_alpha}
    \end{equation}
    where we used $\lvert \Tr(\mA\mB) \rvert \leq \lVert \mA \rVert \Tr(\mB)$ for non-negative definite matrix $\mB$ together with \lemref{lem:upper_bound_norm_resolvent} which asserts the operator norm of the resolvent $\mQ_-$ is uniformly bounded. Furthermore, both $\lVert \hat \mU_n \rVert$ and $\lVert \hat \mV_n \rVert$ are upper bounded by $1$.
\end{remark}
\begin{proof}
     We decompose the matrix $\FSVisited^T\FSVisited$ as 
    \begin{equation}
        \FSVisited^T\FSVisited = \sum_{i=1}^N \vsigma_i \vsigma_i^T,
        \label{eq:decompositionFState^TFstate}
    \end{equation}
    where $\vsigma_i = \sigma(\hat \mS^T\vw_i) \in \R^{m}$ for which $\vw_i \in \R^d$ denotes the i-th row of $\mW$ (defined in \eqref{def:rf_map}). Using the resolvent identity (\lemref{lem:resolvent_identity}), we write
    \begin{align*}
        &\E[\mQ_m] - \tilde \mQ_m \\
        &= \E \left[ \mQ_m \biggl[ \tilde \mQ_m^{-1} - \lambda \mI_n - \frac{1}{m}(\hat \mU_n - \gamma \hat \mV_n)^T \FSVisited^T \FSVisited \hat \mU_n \biggr] \tilde \mQ_m \right] \\
        &= \frac{N}{m} \frac{1}{1+\alpha} \E[\mQ_m]( \hat \mU_n - \gamma \hat \mV_n )^T \mPhi_{\SVisited} \hat \mU_n \tilde \mQ_m - \frac{1}{m} \sum_{i=1}^N \E\left[\mQ_m (\hat \mU_n - \gamma \hat \mV_n)^T \vsigma_i \vsigma_i^T \hat \mU_n\right] \tilde \mQ_m \\ 
        &=\frac{N}{m}\frac{1}{1+\alpha} \E[\mQ_m]( \hat \mU_n - \gamma \hat \mV_n )^T \mPhi_{\SVisited} \hat \mU_n \tilde \mQ_m - \frac{1}{m} \sum_{i=1}^N \E \left[\mQ_{-i}\frac{(\hat \mU_n - \gamma \hat \mV_n)^T \vsigma_i \vsigma_i^T \hat \mU_n}{1+\frac{1}{m}\vsigma_i^T\hat \mU_n \mQ_{-i}(\hat \mU_n-\gamma \hat \mV_n)^T \vsigma_i} \right]  \tilde \mQ_m,  
    \end{align*}
    where the last equality is obtained with the Sherman identity (\lemref{lem:sherman}) for
    \begin{equation}
        \mQ_{-i} = \biggl[\frac{1}{m}(\hat \mU_n - \gamma \hat \mV_n)^T \FSVisited^T \FSVisited \hat \mU_n - \frac{1}{m} (\hat \mU_n - \gamma \hat \mV_n)^T \vsigma_i\vsigma_i^T \hat \mU_n + \lambda \mI_{n}\biggr]^{-1}
        \label{def:Q_-i}
    \end{equation}
    independent of $\vsigma_i$ and thus $\vw_i$. Exploiting this independence, we decompose
    \begin{align}
            &\E[\mQ_m] - \tilde \mQ_m \\
            \begin{split}
                &=\frac{N}{m}\frac{1}{1+\alpha}\E[\mQ_m]( \hat \mU_n - \gamma \hat \mV_n )^T \mPhi_{\SVisited} \hat \mU_n \tilde \mQ_m - \frac{1}{1+\alpha} \frac{1}{m} \sum_{i=1}^N \E \left[\mQ_{-i}(\hat \mU_n - \gamma \hat \mV_n)^T \vsigma_i \vsigma_i^T \hat \mU_n \right]  \tilde \mQ_m \\
                &\quad \qquad + \frac{1}{m} \frac{1}{1+\alpha} \sum_{i=1}^N \E \left[\mQ_{-i}\frac{(\hat \mU_n - \gamma \hat \mV_n)^T \vsigma_i \vsigma_i^T \hat \mU_n(\frac{1}{m}\vsigma_i^T\hat \mU_n \mQ_{-i}(\hat \mU_n-\gamma \hat \mV_n)^T \vsigma_i - \alpha )}{1+\frac{1}{m}\vsigma_i^T\hat \mU_n \mQ_{-i}(\hat \mU_n-\gamma \hat \mV_n)^T \vsigma_i} \right]  \tilde \mQ_m 
            \end{split} \\
            \begin{split}
                &=\underbrace{\frac{1}{m}\frac{1}{1+\alpha}\sum_{i=1}^N \E[\mQ_m - \mQ_{-i}]( \hat \mU_n - \gamma \hat \mV_n )^T\mPhi_{\SVisited}\hat \mU_n \tilde \mQ_m}_{=\mZ_1} \\
                &\quad + \underbrace{\frac{1}{m}\frac{1}{1+\alpha} \sum_{i=1}^N \E \left[\mQ_m\left(\hat \mU_n - \gamma \hat \mV_n \right)^T \vsigma_i \vsigma_i^T \hat \mU_n \tilde \mQ_m \left(\frac{1}{m}\vsigma_i^T\hat \mU_n \mQ_{-i}(\hat \mU_n-\gamma \hat \mV_n)^T \vsigma_i - \alpha \right) \right]}_{=\mZ_2}.
            \end{split}
        \label{eq:decomposition_Q_-_Q-}
    \end{align}
    The last equality is obtained by exploiting the Sherman identity (\lemref{lem:sherman}) in reverse on the rightmost term, and from the independence of $\mQ_{-i}$ and $\vsigma_i\vsigma_i^T$ for the second right-hand term. We want to prove that both $\mZ_1$ and $\mZ_2$ have a vanishing spectral norm under Assumptions~\ref{assumption:growth_rate}~and~\ref{assumption:regime_n}. With both the resolvent identity (\lemref{lem:resolvent_identity}) and the Sherman identity (\lemref{lem:sherman}), we rewrite $\mZ_1$ as
    \begin{align*}
        \mZ_1 &= \frac{1}{m}\frac{1}{1+\alpha}\sum_{i=1}^N \E[\mQ_m - \mQ_{-i}]( \hat \mU_n - \gamma \hat \mV_n )^T\mPhi_{\SVisited}\hat \mU_n \tilde \mQ_m \\
        &=-\frac{1}{m^2}\frac{1}{1+\alpha} \sum_{i=1}^N \E\left[\mQ_m(\hat \mU_n-\gamma \hat \mV_n )^T\vsigma_i\vsigma_i^T\hat \mU_n \mQ_{-i}\right]( \hat \mU_n - \gamma \hat \mV_n )^T\mPhi_{\SVisited} \hat \mU_n \tilde \mQ_m \\
        &=-\frac{1}{m^2}\frac{1}{1+\alpha} \sum_{i=1}^N \E\biggl[\mQ_m(\hat \mU_n-\gamma \hat \mV_n )^T\vsigma_i \mD_i \vsigma_i^T\hat \mU_n \mQ_m \biggr](\hat \mU_n - \gamma \hat \mV_n )^T\mPhi_{\SVisited} \hat \mU_n \tilde \mQ_m  \\
        &=-\frac{1}{m^2}\frac{1}{1+\alpha}\E\left[\mQ_m (\hat \mU_n-\gamma \hat \mV_n)^T \FSVisited^T \mD \FSVisited \hat \mU_n \mQ_m \right]( \hat \mU_n - \gamma \hat \mV_n )^T\mPhi_{\SVisited} \hat \mU_n \tilde \mQ_m,
    \end{align*}
    where $\mD \in \R^{N \times N}$ is a diagonal matrix for which, for all $i \in [N]$, we have 
    \begin{equation}
        \mD_i=\left(1+\frac{1}{m}\vsigma_i^T\hat \mU_n\mQ_{-i}(\hat \mU_n-\gamma \hat \mV_n)^T\vsigma_i\right).
        \label{def:diagonal_matrix_lemma_theorem_1}
    \end{equation}
    With a similar proof than for \lemref{lem:upper_bound_norm_resolvent}, we can show there exists a $K_{\tilde \mQ_m}$ such that, for all $m$, we have $\lVert \tilde \mQ_m \rVert \leq K_{\tilde \mQ_m}$ and then
    \begin{equation}
        \left\lVert \frac{1}{1+\alpha} ( \hat \mU_n - \gamma \hat \mV_n )^T\mPhi_{\SVisited} \hat \mU_n \tilde \mQ_m \right\rVert = \left\lVert \frac{m}{N}(\mI_n - \lambda \tilde \mQ_m) \right\rVert \leq  \frac{m}{N}(1+ \lambda K_{\tilde \mQ_m}).
        \label{eq:first_part_left_hand_part_first_equivalence_Q}
    \end{equation}
    Furthermore, from \lemref{lem:bound_norm_E[Q_m(U_n-gammaV_n)^TFState^TDFStateU_nQ_M]}, we have 
    \begin{equation}
        \left\lVert \frac{1}{m^2}\E\left[\mQ_m (\hat \mU_n-\gamma \hat \mV_n)^T \FSVisited^T \mD \FSVisited \hat \mU_n \mQ_m \right] \right\rVert = \mathcal{O}\left(\frac{1}{m}\right).
        \label{eq:second_part_left_hand_part_first_equivalence_Q}
    \end{equation}
    Therefore, by combining both \eqref{eq:first_part_left_hand_part_first_equivalence_Q} and \eqref{eq:second_part_left_hand_part_first_equivalence_Q}, we conclude that $\mZ_1$ has a vanishing spectral norm, i.e.,
    \begin{equation}
        \lVert \mZ_1 \rVert = \left\lVert \frac{1}{m}\frac{1}{1+\alpha}\sum_{i=1}^N \E[\mQ_m - \mQ_{-i}]( \hat \mU_n - \gamma \hat \mV_n )^T\mPhi_{\SVisited}\hat \mU_n \tilde \mQ_m \right\rVert = \mathcal{O}\left(\frac{1}{m}\right).
        \label{eq:bound_left_hand_part_first_equivalence_Q}
    \end{equation}
    We want to show now that $\mZ_2$ has also a vanishing operator norm. For $i \in [N]$, by setting 
    \begin{equation*}
        \mB_i = m^{\frac{1}{4}}\mQ_m\left(\hat \mU_n - \gamma \hat \mV_n \right)^T \vsigma_i \left(\frac{1}{m}\vsigma_i^T\hat \mU_n \mQ_{-i}(\hat \mU_n-\gamma \hat \mV_n)^T \vsigma_i - \alpha \right)
    \end{equation*}
    and 
    \begin{equation*}
        \mC_i =m^{-\frac{1}{4}}\tilde \mQ_m^T \hat \mU_n^T \vsigma_i,
    \end{equation*}
     we decompose $\mZ_2$ with its symmetric and its skew-symmetric part as
     \begin{align*}
         \mZ_2&=\frac{1}{1+\alpha}\frac{1}{m} \sum_{i=1}^N \E \left[\mQ_m\left(\hat \mU_n - \gamma \hat \mV_n \right)^T \vsigma_i \vsigma_i^T \hat \mU_n \tilde \mQ_m \left(\frac{1}{m}\vsigma_i^T\hat \mU_n \mQ_{-i}(\hat \mU_n-\gamma \hat \mV_n)^T \vsigma_i - \alpha \right) \right] \\
         &= \frac{1}{1+\alpha}\frac{1}{m} \sum_{i=1}^N \E \left[\mB_i \mC_i^T\right] \\
         &= \frac{1}{1+\alpha}\frac{1}{m} \sum_{i=1}^N \E \left[\frac{\mB_i\mC_i^T + \mC_i\mB_i^T}{2} \right] +  \frac{1}{1+\alpha}\frac{1}{m} \sum_{i=1}^N \E \left[\frac{\mB_i\mC_i^T - \mC_i\mB_i^T}{2} \right].
     \end{align*}
     For the symmetric part, we use the relations $(\mB_i - \mC_i)(\mB_i - \mC_i)^T \succeq 0$ and $(\mB_i + \mC_i)(\mB_i + \mC_i)^T \succeq 0$ to deduce that
     \begin{equation*}
         -\mB_i\mB_i^T -\mC_i\mC_i^T \preceq \mB_i\mC_i^T + \mC_i\mB_i^T \preceq \mB_i\mB_i^T + \mC_i\mC_i^T, 
     \end{equation*}
     where $\preceq$ is the Loewner order for semi-positive-definite matrices. For the skew-symmetric part, we observe that  $\lVert \E \left[\mB_i\mC_i^T - \mC_i\mB_i^T \right] \rVert=\lVert i~\E \left[\mB_i\mC_i^T - \mC_i\mB_i^T \right] \rVert$ for $i^2=-1$. With a similar reasoning than above, using the relations $(\mB_i + i\mC_i)(\mB_i + i\mC_i)^* \succeq 0$ and $-(\mB_i - i\mC_i)(\mB_i - i\mC_i)^* \preceq 0$, we deduce the relation
     \begin{equation*}
         -\mB_i\mB_i^T -\mC_i\mC_i^T \preceq i(\mB_i\mC_i^T - \mC_i\mB_i^T) \preceq \mB_i\mB_i^T + \mC_i\mC_i^T.
     \end{equation*}
    From those relations, for both the symmetric and skew-symmetric parts, we have
     \begin{align}
        \begin{split}
            \lVert \mZ_2 \rVert&=\left\lVert \frac{1}{1+\alpha}\frac{1}{m} \sum_{i=1}^N \E \left[\mQ\left(\hat \mU_n - \gamma \hat \mV_n \right)^T \vsigma_i \vsigma_i^T \hat \mU_n \tilde \mQ_m \left(\frac{1}{m}\vsigma_i^T\hat \mU_n \mQ_{-i}(\hat \mU_n-\gamma \hat \mV_n)^T \vsigma_i - \alpha \right) \right] \right\rVert \\
            &\leq \frac{1}{1+\alpha} \left( \left\lVert \sum_{i=1}^N \E \left[\frac{1}{m}\mB_i\mB_i^T \right] \right\rVert + \left\lVert \sum_{i=1}^N \E \left[\frac{1}{m}\mC_i\mC_i^T \right] \right\rVert \right).
        \end{split}
        \label{eq:right_hand_part_bound_first_equivalence_Q}
     \end{align}
    From \lemref{lem:upper_bound_norm_resolvent_+_RF_matrix}, we know there exists a real $K_{\mQ_m}'>0$ such that, for all $m$, we have
    \begin{equation*}
        \left\lVert \frac{1}{\sqrt{m}}\mQ_m (\hat \mU_n - \gamma \hat \mV_n)^T \FSVisited^T \right\rVert \leq K_{\mQ_m}'.
    \end{equation*}
    
    At this point,
    \begin{align*}
        &\left\lVert \sum_{i=1}^N \E \left[\frac{1}{m}\mB_i\mB_i^T \right] \right\rVert \\
        &= \left\lVert \sum_{i=1}^N \E \left[ \frac{1}{\sqrt{m}} \mQ_m\left(\hat \mU_n - \gamma \hat \mV_n \right)^T \vsigma_i \vsigma_i^T \left(\hat \mU_n - \gamma \hat \mV_n \right)^T \mQ_m^T \left(\frac{1}{m}\vsigma_i^T\hat \mU_n \mQ_{-i}(\hat \mU_n-\gamma \hat \mV_n)^T \vsigma_i - \alpha \right)^2 \right] \right\rVert \\
        &= \left\lVert \sqrt{m} \, \E \left[ \frac{1}{m} \mQ_m\left(\hat \mU_n - \gamma \hat \mV_n \right)^T \FSVisited^T \mD^2_2 \FSVisited \left(\hat \mU_n - \gamma \hat \mV_n \right)^T \mQ_m^T \right] \right\rVert \\
        &\leq \sqrt{m}K_{\mQ_m}'^2\E[\lVert \mD^2_2\rVert],
    \end{align*}
    where $\mD_2 \in \R^{N \times N}$ is a diagonal matrix for which, for all $i \in [N]$, we have
    \begin{equation*}
        [\mD_2]_i=\left(\frac{1}{m}\vsigma_i^T\hat \mU_n\mQ_{-i}(\hat \mU_n-\gamma \hat \mV_n)^T\vsigma_i - \alpha \right).
    \end{equation*}
    From both \lemref{lem:concentration_coeff_D_i} and the union bound, we have
    \begin{equation*}
        \Pr\left(\lVert \mD_2 \rVert > t \right) = \Pr\left(\max_{1 \leq i \leq N} [\mD_2]_i > t \right) \leq CNe^{-cm\min(t, t^2)}
    \end{equation*}
    for some $c, C>0$ independent of $m$ and $N$. We have thus
    \begin{align*}
        \E\left(\lVert \mD_2 \rVert^2\right)=\E\left(\max_{1 \leq i \leq N} [\mD^2_2]_i\right) &= \int_0^\infty \Pr\left(\max_{1 \leq i \leq N} [\mD^2_2]_i > t\right)dt \\
        &= \int_0^\infty 2 t \Pr\left(\max_{1 \leq i \leq N} [\mD_2]_i > t\right)dt \\
        &\leq \int_0^\infty 2 t CNe^{-cm\min(t, t^2)}dt \\
        &= \int_0^1 2 t CNe^{-cmt^2}dt+\int_1^\infty 2 t CNe^{-cmt}dt \\
        &\leq \int_0^\infty 2 t CNe^{-cmt^2}dt+\int_0^\infty 2 t CNe^{-cmt}dt \\
        &=\frac{1}{m}\frac{2C}{c}\int_0^\infty tNe^{-t^2}dt+\frac{1}{m^2}\frac{2C}{c^2}\int_0^\infty tNe^{-t}dt\\
        &=\mathcal{O}\left(\frac{1}{m}\right).
    \end{align*}
    We deduce that
    \begin{equation*}
        \left\lVert \sum_{i=1}^N \E \left[\frac{1}{m}\mB_i\mB_i^T \right] \right\rVert = \mathcal{O}\left(\frac{1}{\sqrt{m}}\right).
    \end{equation*}
    In addition, with a similar proof than for \lemref{lem:upper_bound_norm_resolvent_+_RF_matrix}, we can show there exists a real $K_{\tilde \mQ_m}'>0$ such that, for all $m$, we have 
    \begin{equation*}
        \left\lVert \sqrt{\frac{N}{m}} \sqrt{\frac{1}{1+\alpha}} \bar \mZ^T \hat \mU_n \tilde \mQ_m \right\rVert \leq K_{\tilde \mQ_m}',
    \end{equation*}
    where $\bar \mZ \bar \mZ^T$ is the Cholesky decomposition of $\mPhi_{\SVisited}$. Therefore,
    \begin{align*}
        \left\lVert \sum_{i=1}^N \E \left[\frac{1}{m}\mC_i\mC_i^T \right] \right\rVert &= \left\lVert \sum_{i=1}^N \E \left[\frac{1}{m \sqrt{m}}\tilde \mQ_m^T \hat \mU_n^T \vsigma_i \vsigma_i^T \hat \mU_n \tilde \mQ_m \right] \right\rVert \\
        &= \left\lVert \frac{1}{\sqrt{m}} \frac{N}{m}\tilde \mQ_m^T \hat \mU_n^T \mPhi_{\SVisited} \hat \mU_n \tilde \mQ_m  \right\rVert \\
        &= \mathcal{O}\left(\frac{1}{\sqrt{m}}\right).
    \end{align*}
    From \eqref{eq:right_hand_part_bound_first_equivalence_Q} and above, we deduce that $\mZ_2$ vanishes under the operator nom, i.e.,
    \begin{equation}
        \left\lVert \mZ_2 \right\rVert = \mathcal{O}\left(\frac{1}{\sqrt{m}}\right).
        \label{eq:bound_right_hand_part_first_equivalence_Q}
    \end{equation}
    Using both \eqref{eq:bound_left_hand_part_first_equivalence_Q} and \eqref{eq:bound_right_hand_part_first_equivalence_Q} into \eqref{eq:decomposition_Q_-_Q-}, we conclude that  
    \begin{equation}
        \lVert \E[\mQ_m] - \tilde \mQ_m \rVert = \mathcal{O}\left(\frac{1}{\sqrt{m}}\right).
        \label{eq:first_equivalence_E_Q}
    \end{equation}
\end{proof}
To get \thmref{theorem:asy-behavior-E[Q]}, we start from \lemref{lem:first-asy-behavior-E[Q]} and we show that 
\begin{equation*}
    \lVert \bar \mQ_m(\lambda) - \tilde \mQ_m(\lambda) \rVert \to 0,
\end{equation*}
as $N, m \to \infty$.
\begin{theorem}[Asymptotic Deterministic Resolvent]
    Under Assumptions~\ref{assumption:growth_rate}~and~\ref{assumption:regime_n}, let $\lambda >0$ and let $\bar \mQ_m(\lambda) \in \R^{n \times n}$ be the resolvent defined as  
    \begin{equation*}
      \bar \mQ_m(\lambda) = \biggl[ \frac{N}{m} \frac{1}{1+\delta} (\hat \mU_n - \gamma \hat \mV_n)^T\mPhi_{\SVisited}\hat \mU_n + \lambda \mI_n \biggr]^{-1},
    \end{equation*}
    where $\delta$ is the \emph{correction factor} defined as the unique positive solution to
    \begin{equation*}
        \delta = \frac{1}{m} \Tr \Biggl((\hat \mU_n - \gamma \hat \mV_n)^T\mPhi_{\SVisited} \hat \mU_n \biggl[\frac{N}{m} \frac{1}{1+\delta} (\hat \mU_n - \gamma \hat \mV_n)^T\mPhi_{\SVisited}\hat \mU_n + \lambda \mI_n \biggr]^{-1}\Biggr).
    \end{equation*}
    Then, 
    \begin{equation*}
        \lim_{m \to \infty} \Bigl\lVert \E_\mW\bigl[\mQ_m(\lambda)\bigr] - \bar \mQ_m(\lambda) \Bigr\rVert = 0.
    \end{equation*}
\end{theorem}
\begin{proof}
     From \lemref{lem:delta_positive_unique} in \Appref{sec:delta}, we know that $\delta$ exists and is the unique positive solution of \eqref{def:delta} under Assumptions~\ref{assumption:growth_rate}~and~\ref{assumption:regime_n}. From \lemref{lem:first-asy-behavior-E[Q]} we have a first asymptotic equivalent of $\E_\mW[\mQ_m]$ given by
     \begin{equation*}
      \tilde \mQ_m = \left[\frac{N}{m} \frac{1}{1+\alpha} \left( \hat \mU_n - \gamma \hat \mV_n \right)^T \mPhi_{\SVisited} \hat \mU_n + \lambda \mI_n \right]^{-1},
    \end{equation*}
    where
    \begin{equation*}
        \alpha = \frac{1}{m} \Tr\bigl((\hat \mU_n - \gamma \hat \mV_n)^T \mPhi_{\SVisited} \hat \mU_n \E[\mQ_-]\bigr),
    \end{equation*}
    since
    \begin{equation*}
        \lim_{m \to \infty} \lVert \E_\mW[\mQ_m] - \tilde \mQ_m \rVert = 0.
    \end{equation*}
    In order to finish the proof of the Theorem, we want to show that 
    \begin{equation}
        \lim_{m \to \infty} \lVert \tilde \mQ_m - \bar \mQ_m \rVert = 0.
        \label{eq:objective_theorem_1}
    \end{equation}
    From the resolvent identity (\lemref{lem:resolvent_identity}), we have
    \begin{equation}
        \lVert \tilde \mQ_m - \bar \mQ_m \rVert  =  \frac{N}{m} \frac{\lvert \alpha-\delta \rvert}{(1+\delta)(1+\alpha)} \left\lVert \tilde \mQ_m \left( \hat \mU_n - \gamma \hat \mV_n \right)^T  \mPhi_{\SVisited} \hat \mU_n \bar \mQ_m \right\rVert.
        \label{eq:diff_Q_-_Q_bar}
    \end{equation}
    Let $\bar \mZ \bar \mZ^T$ be the Cholesky decomposition of $\mPhi_{\SVisited}$. With a similar proof than for \lemref{lem:upper_bound_norm_resolvent_+_RF_matrix}, we can show there exists a real $K_{\tilde \mQ}'>0$ such that, for all $m$, we have 
    \begin{equation*}
        \left\lVert \sqrt{\frac{1}{1+\alpha}}\sqrt{\frac{N}{m}} \tilde \mQ_m \left( \hat \mU_n - \gamma \hat \mV_n \right)^T \bar \mZ \right\rVert \leq K_{\tilde \mQ}'.
    \end{equation*}
    Similarly, we can show there exists a real $K_{\bar \mQ}'>0$ such that, for all $m$, we have  
    \begin{equation*}
        \left\lVert \sqrt{\frac{1}{1+\delta}}\sqrt{\frac{N}{m}} \bar \mZ^T \hat \mU_n \bar \mQ_m \right\rVert \leq K_{\bar \mQ}'.
    \end{equation*}
    Therefore,
    \begin{equation*}
        \left\lVert \tilde \mQ_m \left( \hat \mU_n - \gamma \hat \mV_n \right)^T  \mPhi_{\SVisited} \hat \mU_n \bar \mQ_m \right\rVert \leq \sqrt{(1+\delta)(1+\alpha)}\frac{m}{N}K_{\bar \mQ}'K_{\tilde \mQ}'.
    \end{equation*}
    As a consequence, in order to prove \eqref{eq:objective_theorem_1}, it remains to prove that
    \begin{equation*}
        \lim_{m \to \infty} \lvert \alpha - \delta \rvert = 0.
    \end{equation*}
    We decompose $\left\lvert \alpha - \delta \right\lvert$ as
    \begin{align}
        \left\lvert \alpha - \delta \right\lvert &= \left\lvert \frac{1}{m} \Tr\bigl((\hat \mU_n - \gamma \hat \mV_n)^T\mPhi_{\SVisited} \hat \mU_n \bigl[ \E[\mQ_-] - \bar \mQ_m\bigr]\bigr) \right\rvert \\
        &\leq \underbrace{\left\lvert \frac{1}{m} \Tr\bigl((\hat \mU_n - \gamma \hat \mV_n)^T\mPhi_{\SVisited} \hat \mU_n \bigl[\E[\mQ_-] - \tilde \mQ_m\bigr]\bigr) \right\rvert}_{=Z_1} + \underbrace{\left\lvert \frac{1}{m} \Tr\bigl((\hat \mU_n - \gamma \hat \mV_n)^T\mPhi_{\SVisited} \hat \mU_n\bigl[\tilde \mQ_m - \bar \mQ_m\bigr]\bigr) \right\rvert}_{=Z_2}.
        \label{eq:diff_alpha_delta}
    \end{align}
    To show $Z_1$ vanishes, we write $\alpha$ as 
    \begin{align*}
        \alpha &= \frac{1}{m} \Tr\bigl((\hat \mU_n - \gamma \hat \mV_n)^T\mPhi_{\SVisited} \hat \mU_n \E[\mQ_-]\bigr) \\ 
        &=  \frac{1}{m}\Tr\bigl((\hat \mU_n - \gamma \hat \mV_n)^T\mPhi_{\SVisited} \hat \mU_n \E[\mQ_m]\bigr) + \frac{1}{m} \Tr\bigl((\hat \mU_n - \gamma \hat \mV_n)^T\mPhi_{\SVisited}\hat \mU_n \bigl[\E[\mQ_-]-\E[\mQ_m]\bigr]\bigr).
    \end{align*}
    There exists a real $K>0$ such that
    \begin{equation*}
        \frac{1}{m} \Tr\bigl((\hat \mU_n - \gamma \hat \mV_n)^T\mPhi_{\SVisited}\hat \mU_n \bigl[\E[\mQ_-]-\E[\mQ_m]\bigr]\bigr) \leq K \left\lVert \bigl[\E[\mQ_-]-\E[\mQ_m]\bigr] \right\rVert;
    \end{equation*}
    since  both $\lVert \hat \mU_n \rVert$ and $\lVert \hat \mV_n \rVert$ are upper bounded by $1$, $\lvert \Tr(\mA\mB) \rvert \leq \lVert \mA \rVert \Tr(\mB)$ for non-negative definite matrix $\mB$, and from \eqref{eq:upper_bound_tr_phi} that uniformly bounds $\frac{1}{m}\Tr(\mPhi_{\SVisited})$. From \lemref{lem:bound_norm_E[Q_m(U_n-gammaV_n)^TFState^TDFStateU_nQ_M]}, we have
    \begin{align*}
    	\left\lVert \E\left[\mQ_m - \mQ_- \right] \right\rVert &= \left\lVert \frac{1}{m} \frac{m}N \sum_{i=1}^N \E\left[\mQ_m-\mQ_{-i}\right] \right\rVert \\
        &= \left\lVert \frac{1}{m^2} \frac{m}N \E\left[\mQ_m (\hat \mU_n-\gamma \hat \mV_n)^T \FSVisited^T \mD \FSVisited \hat \mU_n \mQ_m \right] \right\rVert \\
        &= \mathcal{O}\left(\frac{1}{m}\right),
    \end{align*}
    where $\mD \in \R^{N \times N}$ is a diagonal matrix for which, for all $i \in [N]$, we have
    \begin{equation*}
        \mD_i=\left(1+\frac{1}{m}\vsigma_i^T\hat \mU_n\mQ_{-i}(\hat \mU_n-\gamma \hat \mV_n)^T\vsigma_i\right).
    \end{equation*}
    As a consequence, by combining the results above and from \lemref{lem:first-asy-behavior-E[Q]}, we conclude for $Z_1$ that
    \begin{align*}
        \lvert Z_1 \rvert = \left\lvert \alpha - \frac{1}{m}\Tr\bigl((\hat \mU_n - \gamma \hat \mV_n)^T\mPhi_{\SVisited} \hat \mU_n \tilde \mQ_m\bigr) \right\rvert = \mathcal{O}\left(\frac{1}{\sqrt{m}}\right).
    \end{align*}
    Using the vanishing result of $Z_1$ into \eqref{eq:diff_alpha_delta} and applying the resolvent identity (\lemref{lem:resolvent_identity}) on $Z_2$, we get
    \begin{equation*}
        \left\lvert \alpha - \delta \right\lvert \leq \frac{N}{m} \frac{\lvert \alpha-\delta \rvert}{(1+\delta)(1+\alpha)} \left\lvert \frac{1}{m} \Tr\bigl((\hat \mU_n - \gamma \hat \mV_n)^T\mPhi_{\SVisited} \hat \mU_n \tilde \mQ_m (\hat \mU_n - \gamma \hat \mV_n)^T  \mPhi_{\SVisited} \hat \mU_n \bar \mQ_m\bigr) \right\rvert +  \mathcal{O}\left(\frac{1}{\sqrt m}\right),
    \end{equation*}
    which implies that 
    \begin{equation*}
        \left\lvert \alpha - \delta \right\lvert \left(1 - \frac{N}{m} \frac{1}{(1+\delta)(1+\alpha)}\frac{1}{m} \Tr\bigl((\hat \mU_n - \gamma \hat \mV_n)^T\mPhi_{\SVisited} \hat \mU_n\tilde \mQ_m (\hat \mU_n - \gamma \hat \mV_n)^T  \mPhi_{\SVisited} \hat \mU_n \bar \mQ_m \bigr) \right) =  \mathcal{O}\left(\frac{1}{\sqrt m}\right).
    \end{equation*}
    It remains to show
    \begin{equation*}
        \lim_{m \to \infty} \sup_m \frac{1}{m} \frac{N}{m} \frac{1}{(1+\delta)(1+\alpha)} \Tr\bigl((\hat \mU_n - \gamma \hat \mV_n)^T\mPhi_{\SVisited} \hat \mU_n \tilde \mQ_m \bigl( \hat \mU_n - \gamma \hat \mV_n \bigr)^T  \mPhi_{\SVisited} \hat \mU_n \bar \mQ_m\bigr)< 1.
        \label{eq:tr_bound_1}
    \end{equation*}
    Let the matrices $\mB_n = (\hat \mU_n - \gamma \hat \mV_n)^T\mPhi_{\SVisited} \hat \mU_n$, $\mB_n'=\bar \mZ^T \hat \mA_m \mZ$, $\bar \mQ_m'=\left[\frac{N}{m}\frac{1}{1+\delta}\mB_n' + \lambda \mI_m \right]^{-1}$, and $\tilde \mQ_m'=\left[\frac{N}{m}\frac{1}{1+\alpha} \mB_n' + \lambda \mI_m \right]^{-1}$; where $\hat \mA_m = \hat \mU_n (\hat \mU_n - \gamma \hat \mV_n)^T$ is the empirical transition model matrix defined in \eqref{def:empirical_transition_model_matrix}. Using the Cauchy–Schwarz inequality, we write
    \begin{align*}
        &\frac{1}{m} \frac{N}{m} \frac{1}{(1+\delta)(1+\alpha)} \Tr\bigl(\mB_n \tilde \mQ_m \mB_n \bar \mQ_m\bigr) \\
        &=\frac{1}{m} \frac{N}{m} \frac{1}{(1+\delta)(1+\alpha)} \Tr\bigl(\mB_n' \tilde \mQ_m' \mB_n' \bar \mQ_m'\bigr) \\
        &\leq \sqrt{\underbrace{\frac{N}{m}\frac{1}{m}\frac{1}{(1+\delta)^2}\Tr\bigl(\mB_n' \bar \mQ_m' \bar \mQ_m'^T \mB_n'^T\bigr)}_{Z_1'}
        \underbrace{\frac{N}{m}\frac{1}{m}\frac{1}{(1+\alpha)^2}\Tr\bigl(\mB_n' \tilde \mQ_m' \tilde \mQ_m'^T \mB_n'^T\bigr)}_{Z_2'}}.
    \end{align*}
    We observe that
    \begin{align*}
        \delta =\frac{1}{m}\Tr\bigl(\mB_n \bar \mQ_m \bigr)=\frac{1}{m}\Tr\bigl(\mB_n' \bar \mQ_m' \bigr)&=\frac{1}{m}\Tr\bigl(\mB_n' \bar \mQ_m'\mQ_m'^T\mQ_m^{-1T} \bigr) \\
        &=\frac{1}{m}\frac{N}{m}\frac{1+\delta}{(1+\delta)^2}\Tr\bigl(\mB_n' \bar \mQ_m' \bar \mQ_m'^T\mB_n'^T\bigr)+\frac{\lambda}{m}\Tr\bigl(\mB_n' \bar \mQ_m' \bar \mQ_m'^T\bigr).
    \end{align*}
    Since $H(\mB_n')$ is at least semi-positive-definite under Assumption~\ref{assumption:regime_n}, we have 
    \begin{equation*}
        \Tr\bigl(\mB_n' \bar \mQ_m' \bar \mQ_m'^T\bigr)=\Tr\bigl(\bar \mQ_m'^T \mB_n' \bar \mQ_m'\bigr) = \Tr\bigl(\bar \mQ_m'^T H(\mB_n') \bar \mQ_m'\bigr) \geq  0.
    \end{equation*}
    As a consequence we have
    \begin{align*}
        \frac{1}{m}\frac{N}{m}\frac{1}{(1+\delta)^2}\Tr\left(\mB_n' \bar \mQ_m' \bar \mQ_m'^T \mB_n'^T\right) \leq \frac{\delta-\frac{\lambda}{m}\Tr\bigl(\mB_n' \bar \mQ_m' \bar \mQ_m'^T\bigr)}{1+\delta} \leq \frac{\delta}{1+\delta}.
    \end{align*}
    To prove $\frac{\delta}{1+\delta} < 1$, it remains to show that $\delta < \infty$. With a similar proof than for \lemref{lem:upper_bound_norm_resolvent}, we can show there exists a real $K_{\bar \mQ}>0$ such that, for all $m$, we have $\lVert \bar \mQ_m \rVert \leq K_{\bar \mQ}$, and thus
    \begin{equation*}
        \delta = \frac{1}{m}\Tr\bigl(\mB_n \bar \mQ_m \bigr) = \frac{1}{m}\Tr\bigl((\hat \mU_n - \gamma \hat \mV_n)^T \mPhi_{\SVisited} \hat \mU_n \bar \mQ_m \bigr) \leq \frac{2}{m} \Tr(\mPhi_{\SVisited}) \lVert \bar \mQ_m(\delta) \rVert \leq \frac{2}{m} \Tr(\mPhi_{\SVisited}) K_{\bar \mQ} < \infty
    \end{equation*}
    where we used for the first inequality the relation $\lvert \Tr(\mA\mB) \rvert \leq \lVert \mA \rVert \Tr(\mB)$ for non-negative definite matrix $\mB$. Furthermore, from \eqref{eq:upper_bound_tr_phi}, $\frac{1}{m} \Tr(\mPhi_{\SVisited})$ is bounded under Assumptions~\ref{assumption:growth_rate}~and~\ref{assumption:regime_n}, and both $\lVert \hat \mU_n \rVert$ and $\lVert \hat \mV_n \rVert$ are upper bounded by $1$. We thus conclude for $Z_1'$ that
    \begin{equation}
        \lim \sup_m \frac{1}{m}\frac{N}{m}\frac{1}{(1+\delta)^2}\Tr\bigl(\mB_n' \bar \mQ_m' \bar \mQ_m'^T \mB_n'^T\bigr) < 1.
        \label{eq:lim_sup_Tr(BbarQBbarQ)}
    \end{equation}
    With similar arguments, we can show for $Z_2'$ that
    \begin{equation*}
        \lim \sup_m \frac{1}{m}\frac{N}{m}\frac{1}{(1+\alpha)^2}\Tr\bigl(\mB_n' \tilde \mQ_m' \tilde \mQ_m'^T \mB_n'^T\bigr) < 1,
    \end{equation*}
    which concludes the proof that 
    \begin{equation}
        \left\lvert \alpha - \delta \right\lvert = \mathcal{O}\left(\frac{1}{\sqrt m}\right).
        \label{eq:alpha_minus_delta_vanihes}
    \end{equation}
    Using the result above with \eqref{eq:diff_Q_-_Q_bar}, we get
    \begin{align*}
        \lVert \tilde \mQ_m - \bar \mQ_m \rVert  &= \lvert \alpha-\delta \rvert \left\lVert \frac{N}{m} \frac{1}{(1+\delta)(1+\alpha)}\tilde \mQ_m \left( \hat \mU_n - \gamma \hat \mV_n \right)^T  \mPhi_{\SVisited} \hat \mU_n \bar \mQ_m \right\Vert \\
        &=\mathcal{O}\left(\frac{1}{\sqrt m}\right),
    \end{align*}
    which concludes the proof.
\end{proof}
\begin{lemma}
    Under Assumptions~\ref{assumption:growth_rate}~and~\ref{assumption:regime_n}, let $\mD \in \R^{N \times N}$ be the diagonal matrix defined in \eqref{def:diagonal_matrix_lemma_theorem_1} for which, for all $i \in [N]$, we have 
    \begin{equation}
        \mD_i=1+\frac{1}{m}\vsigma_i^T \hat \mU_n \mQ_{-i}(\hat \mU_n - \gamma \hat \mV_n)^T \vsigma_i.
        \label{def:mD_lemma_theorem_2}
    \end{equation}
    Then 
    \begin{equation*}
        \E\left[\lVert \mD \rVert \right] = \mathcal{O}(1).
    \end{equation*}
    \label{lem:boundness_norm_E[D]}
\end{lemma}
\begin{proof}
    Let $\alpha = \frac1m \Tr \bigl((\hat \mU_n - \gamma \hat \mV_n)^T \mPhi_{\SVisited} \hat \mU_n \E[\mQ_-(\lambda)]\bigr)$ defined in \eqref{def:def_alpha}. From \eqref{eq:bound_alpha}, $\alpha$ is uniformly bounded, i.e., there exists a real $K_\alpha>0$ such that $\alpha \leq K_\alpha$. From both \lemref{lem:concentration_coeff_D_i} and the union bound, we have
    \begin{equation*}
        \Pr\left(\lVert \mD \rVert > 1+ \alpha + t\right) = \Pr\left(\max_{1 \leq i \leq N} \mD_i > 1+ \alpha + t\right) \leq CNe^{-cm\min(t, t^2)},
    \end{equation*}
    for some $c, C>0$ independent of $m$ and $N$. Therefore,
    \begin{align*}
        \begin{split}
            \E\left[\lVert \mD \rVert\right]=\E\left[\max_{1 \leq i \leq N} \mD_i\right] &= \int_0^\infty \Pr\left(\max_{1 \leq i \leq N} \mD_i > t\right)dt \\
            &= \int_0^{2(1+K_\alpha)} \Pr\left(\max_{1 \leq i \leq N} \mD_i > t\right)dt+\int_{2(1+K_\alpha)}^\infty \Pr\left(\max_{1 \leq i \leq N} \mD_i > t\right)dt \\
            &\leq 2(1+K_\alpha) +\int_{2(1+K_\alpha)}^\infty CN e^{-cm\min\bigl( (t-(1+K_\alpha))^2,t-(1+K_\alpha)\bigr)}dt \\
            &=2(1+K_\alpha) + \int_{1+K_\alpha}^\infty CN e^{-cmt}dt \\
            &=2(1+K_\alpha) + \frac{CN}{cm}e^{-Cm(1+K_\alpha)} \\
            &= \mathcal{O}(1).
        \end{split}
    \end{align*}
\end{proof}
\begin{lemma}
    Under Assumptions~\ref{assumption:growth_rate}~and~\ref{assumption:regime_n}, let $\mD \in \R^{N \times N}$ be the diagonal matrix defined in \eqref{def:diagonal_matrix_lemma_theorem_1} for which, for all $i \in [N]$, we have 
    \begin{equation*}
        \mD_i=1+\frac{1}{m}\vsigma_i^T\hat \mU_n\mQ_{-i}(\hat \mU_n-\gamma \hat \mV_n)^T\vsigma_i.
    \end{equation*}
    Then 
    \begin{equation*}
        \left\lVert \E\left[\frac{1}{m}\mQ_m (\hat \mU_n-\gamma \hat \mV_n)^T \FSVisited^T \mD \FSVisited \hat \mU_n \mQ_m \right] \right\rVert = \mathcal{O}\left(1\right).
    \end{equation*}
    \label{lem:bound_norm_E[Q_m(U_n-gammaV_n)^TFState^TDFStateU_nQ_M]}
\end{lemma}
\begin{proof}
     From \lemref{lem:upper_bound_norm_resolvent_+_RF_matrix}, there exists $K_{\mQ_m}'>0$ such that, for all $m$, we have $\Bigl\lVert \frac{1}{\sqrt{m}}\mQ_m(\hat \mU_n - \gamma \hat \mV_n)^T \FSVisited \Bigr\rVert \leq 2K_{\mQ_m}'$ and $\Bigl\lVert \frac{1}{\sqrt{m}}\FSVisited \hat \mU_n \mQ_m \Bigr\rVert \leq K_{\mQ_m}'$. Therefore,
    \begin{equation*}
        \left\lVert \E\left[\frac{1}{m}\mQ_m (\hat \mU_n-\gamma \hat \mV_n)^T \FSVisited^T \mD \FSVisited \hat \mU_n \mQ_m \right] \right\rVert \leq 2K_{\mQ_m}'^2\E\left[\lVert \mD \rVert\right].
    \end{equation*}
    From \lemref{lem:boundness_norm_E[D]}, we have 
    \begin{equation*}
            \E\left[\lVert \mD \rVert\right]=\mathcal{O}(1).
    \end{equation*}
    As a consequence, we deduce that
    \begin{equation}
        \left\lVert \E\left[\frac{1}{m}\mQ_m (\hat \mU_n-\gamma \hat \mV_n)^T \FSVisited^T \mD \FSVisited \hat \mU_n \mQ_m \right] \right\rVert = \mathcal{O}\left(1\right).
    \end{equation}
\end{proof}

\section{Proof of Theorem~\ref{theorem:asy-behavior-MSBE} \label{sec:proof_theorem_2}}
This section is dedicated to find an asymptotic deterministic limit of the empirical $\Etrain(\hat \vtheta)$ (\eqref{def:MSBE_train}) under Assumptions~\ref{assumption:growth_rate}~and~\ref{assumption:regime_n}. 
First, we observe that the empirical MSBE depends on the quadratic form $\rtrain^T \mQ_m(\lambda)^T \mQ_m(\lambda) \rtrain$ since
\begin{align}
    \Etrain(\hat \vtheta) &= \frac{1}{n}\lVert\rtrain + \gamma \FXtrainN^T\hat \vtheta - \FXtrain^T\hat \vtheta \rVert^2 \\
    &= \frac{1}{n}\biggl\lVert\rtrain - \frac{1}{mn}\bigl(\FXtrain-\gamma \FXtrainN\bigr)^T \FXtrain \mQ_m(\lambda)\rtrain \biggr\rVert^2 \\
    &= \frac{1}{n}\biggl\lVert\biggl[\frac{1}{mn}\bigl(\FXtrain-\gamma \FXtrainN\bigr)^T \FXtrain + \lambda \mI_n - \frac{1}{mn}\bigl(\FXtrain-\gamma \FXtrainN\bigr)^T \FXtrain\biggr] \mQ_m(\lambda)\rtrain \biggr\rVert^2 \\
    &=\frac{\lambda^2}{n} \left\lVert \mQ_m(\lambda)\rtrain \right\rVert^2.
    \label{eq:empirical_msbe_resolvent}
\end{align}
We determine in \thmref{theorem:asy-behavior-MSBE} a deterministic limit of $\Etrain(\hat \vtheta)$ by combining \thmref{theorem:asy-behavior-E[Q]}, which provides an asymptotically more tractable approximation of $\E_\mW\bigl[\mQ_m(\lambda)\bigr]$ under the form of a fixed-point equation, with concentration arguments. 
\thmref{theorem:asy-behavior-MSBE} is corollary of \lemref{lem:deterministic_Q^TMQ} and of the concentration result of \lemref{lem:normal_concentration_resolvent} in \Secref{sec:concentrations}. Both \lemref{lem:deterministic_E[Q_m^Psi_1Q_m]} and \lemref{lem:decomposition_Q^TPsi_1Q} are key Lemma used in the proof of \thmref{theorem:asy-behavior-MSBE} and \thmref{theorem:asy-behavior-true-MSBE}. 

To simplify the notations, we denote the matrix $\mQ_m$ as the resolvent $\mQ_m(\lambda)$ (defined in \eqref{def:resolvent}).
We define the matrix $\mPsi_{\SVisited} \in \R^{m \times m}$ as 
 \begin{equation*}
    \mPsi_{\SVisited} = \frac{N}{m}\frac{1}{1+\delta}\mPhi_{\SVisited}.
\end{equation*}
Furthermore, the notation $\mA = \mB + \mathcal{O_{\lVert \cdot \rVert}}\left(\frac{1}{\sqrt{m}}\right)$ means that $\lVert \mA - \mB \rVert = \mathcal{O}\left(\frac{1}{\sqrt{m}}\right)$.
\begin{theorem}[Asymptotic Empirical MSBE]
    Under the conditions of \thmref{theorem:asy-behavior-E[Q]}, the \emph{deterministic asymptotic empirical MSBE} is
    \begin{equation*}
        \DEtrain = \tfrac{\lambda^2}{n}\lVert \bar \mQ_m(\lambda) \rtrain \rVert^2  +  \hat \Delta,
    \end{equation*}
    with second-order correction factor
    \begin{equation*}
           \hat \Delta = \tfrac{\lambda^2}{n}\tfrac{\tfrac{1}{N} \Tr \left(\bar \mQ_m(\lambda)  \mPsi_2 \bar \mQ_m(\lambda)^T \right)}{1-\tfrac{1}{N} \Tr \left(\bar \mQ_m(\lambda) \mPsi_2 \bar \mQ_m(\lambda)^T \mPsi_1 \right)} \lVert \bar \mQ_m(\lambda) \rtrain\rVert^2_{\mPsi_1},
    \end{equation*}
    where
    \begin{equation*}
        \mPsi_1 = \hat \mU_n^T \mPsi_{\SVisited} \hat \mU_n, \quad \text{and} \quad \mPsi_2 = (\hat \mU_n-\gamma \hat \mV_n)^T \mPsi_{\SVisited} (\hat \mU_n-\gamma \hat \mV_n).
    \end{equation*}
   As $N, m, d \to \infty$ with asymptotic constant ratio $N/m$,
   \begin{equation*}
        \Etrain(\hat \vtheta) - \DEtrain \xrightarrow{a.s} 0.
   \end{equation*}
\end{theorem}
\begin{proof}
    From \eqref{eq:empirical_msbe_resolvent}, we have
    \begin{equation*}
        \Etrain(\hat \vtheta) =\frac{\lambda^2}{n} \left\lVert \mQ_m\ \rtrain \right\rVert^2=\frac{\lambda^2}{n}\rtrain^T \mQ_m^T \mQ_m \rtrain.
    \end{equation*}  
    From \lemref{lem:normal_concentration_msbe}, we have
    \begin{equation*}
            \Pr\left(~\left\lvert \frac{\lambda^2}{n}\rtrain^T \mQ_m^T \mQ_m \rtrain - \frac{\lambda^2}{n}\rtrain^T \E[\mQ_m^T \mQ_m]\rtrain  \right\rvert > t \right)
            \leq C e^{-cn^2mt^2},
    \end{equation*}
    for some $C, c >0$ independent of $m, n$ and $N$. Furthermore, from \lemref{lem:deterministic_Q^TMQ}, we have 
    \begin{align*}
        \biggl\lVert &\E\bigl[\mQ_m^T\mQ_m \bigr] - \bar \mQ_m^T\bar \mQ_m - \frac{\frac{1}{N} \Tr \left(\mPsi_2 \bar \mQ_m^T \bar \mQ_m \right)}{1-\frac{1}{N} \Tr \left(\mPsi_2 \bar \mQ_m^T  \mPsi_1 \bar \mQ_m  \right)}\bar \mQ_m^T \mPsi_1 \bar \mQ_m \biggr\rVert = \mathcal{O}\left(\frac{1}{\sqrt{m}}\right).
    \end{align*}
    As a consequence, we have
    \begin{equation*}
        \Etrain(\hat \vtheta) - \DEtrain \xrightarrow{a.s} 0,
    \end{equation*}
    as $m \to \infty$.
\end{proof}
\begin{lemma}
    Under Assumptions~\ref{assumption:growth_rate}~and~\ref{assumption:regime_n}, let $\mQ_m \in \R^{n \times n}$ be the resolvent defined in \eqref{def:resolvent}, let $\bar \mQ_m \in \R^{n \times n}$ be the deterministic resolvent defined in \eqref{def:Q_bar}, and let $\mM \in \R^{n \times n}$ be any matrix with a bounded operator norm. Then, 
    \begin{align*}
        \biggl\lVert \E\bigl[\mQ_m^T \mM \mQ_m \bigr] &- \bar \mQ_m^T \mM \bar \mQ_m  -\frac{\frac{1}{N} \Tr \left(\mPsi_2 \bar \mQ_m^T \mM \bar \mQ_m \right)}{1-\frac{1}{N} \Tr \left(\mPsi_2 \bar \mQ_m^T  \mPsi_1 \bar \mQ_m  \right)}\bar \mQ_m^T \mPsi_1 \bar \mQ_m \biggr\rVert = \mathcal{O}\left(\frac{1}{\sqrt{m}}\right),
    \end{align*}
    for $\mPsi_1, \mPsi_2 \in \R^{n \times n}$ defined in \eqref{def:Psi_1_and_Psi_2}.
    \label{lem:deterministic_Q^TMQ}
\end{lemma}
\begin{proof}
    From \lemref{lem:decomposition_Q^TMQ}, we have
    \begin{align*}
        \E\bigl[\mQ_m^T \mM \mQ_m \bigr] &= \bar \mQ_m^T \mM \bar \mQ_m  +  \E\left[\mQ_m^T \mM \bar \mQ_m(\hat \mU_n-\gamma \hat \mV_n)^T \mPsi_{\SVisited} \hat \mU_n \mQ_m \right] \\
        &\quad - \E\left[\mQ_{-}^T \mM \bar \mQ_m (\hat \mU_n-\gamma \hat \mV_n)^T \mPsi_{\SVisited} \hat \mU_n \mQ_{-}\right]  +\frac{1}{N} \Tr \left(\mPsi_2 \bar \mQ_m^T \mM \bar \mQ_m \right)\E\left[\mQ_{-}^T \mPsi_1 \mQ_{-} \right]   \\
        &\quad+\mathcal{O_{\lVert \cdot \rVert}}\left(\frac{1}{\sqrt{m}}\right).
    \end{align*}
    Let 
    \begin{equation*}
        \mM' = \mM \bar \mQ_m(\hat \mU_n-\gamma \hat \mV_n)^T \mPsi_{\SVisited} \hat \mU_n = \mM \bigl[\mI_n - \lambda \bar \mQ_m \bigr]
    \end{equation*}
    With a similar proof than for \lemref{lem:upper_bound_norm_resolvent}, we can show that there exists a real $K_{\bar \mQ}$ such that, for all $m$, we have $\lVert \bar \mQ_m \rVert \leq K_{\bar \mQ}$. We deduce thus that $\mM'$ is a matrix with a bounded operator norm since $\lVert \mM' \rVert \leq (1+\lambda K_{\bar \mQ}) \lVert \mM \rVert$. From \lemref{lem:deterministic_E[Q_m^TNQ_m]}, we have
    \begin{align*}
        \Bigl\lVert \E\bigl[ & \mQ_m^T \mM \bar \mQ_m(\hat \mU_n-\gamma \hat \mV_n)^T \mPsi_{\SVisited} \hat \mU_n \mQ_m \bigr] - \E\left[\mQ_{-}^T \mM \bar \mQ_m (\hat \mU_n-\gamma \hat \mV_n)^T \mPsi_{\SVisited} \hat \mU_n \mQ_{-}\right] \Bigr\rVert
        = \mathcal{O}\left(\frac{1}{m}\right).
    \end{align*}
    Therefore,
    \begin{equation*}
        \E\bigl[\mQ_m^T \mM \mQ_m \bigr] = \bar \mQ_m^T \mM \bar \mQ_m  + \frac{1}{N} \Tr \left(\mPsi_2 \bar \mQ_m^T \mM \bar \mQ_m \right)\E\left[\mQ_{-}^T \mPsi_1 \mQ_{-} \right] + \mathcal{O_{\lVert \cdot \rVert}}\left(\frac{1}{\sqrt{m}}\right).
    \end{equation*}
    Furthermore, from \lemref{lem:deterministic_E[Q_m^Psi_1Q_m]}, we have
    \begin{equation*}
        \left\lVert \E\left[\mQ_m^T \mPsi_1 \mQ_m \right]-\E\left[\mQ_{-}^T \mPsi_1 \mQ_{-}\right] \right\rVert = \mathcal{O}\left(\frac{1}{\sqrt{m}}\right),
    \end{equation*}
    and from \lemref{lem:decomposition_Q^TPsi_1Q} we have
    \begin{align*}
        \E\bigl[\mQ_m^T \mPsi_1 \mQ_m \bigr] &= \bar \mQ_m^T \mPsi_1 \bar \mQ_m  +\frac{1}{1-\frac{1}{N} \Tr \left(\mPsi_2 \bar \mQ_m^T \mPsi_1 \bar \mQ_m \right)}\bar \mQ_m^T \mPsi_1 \bar \mQ_m+\mathcal{O_{\lVert \cdot \rVert}}\left(\frac{1}{\sqrt{m}}\right).
    \end{align*}
    We conclude thus
    \begin{align*}
        \E\bigl[\mQ_m^T \mM \mQ_m \bigr] &= \bar \mQ_m^T \mM \bar \mQ_m+\frac{\frac{1}{N} \Tr \left(\mPsi_2 \bar \mQ_m^T \mM \bar \mQ_m \right)}{1-\frac{1}{N} \Tr \left(\mPsi_2 \bar \mQ_m^T  \mPsi_1 \bar \mQ_m  \right)}\bar \mQ_m^T \mPsi_1 \bar \mQ_m+ \mathcal{O_{\lVert \cdot \rVert}}\left(\frac{1}{\sqrt{m}}\right).
    \end{align*}
\end{proof}
\begin{lemma}
    Under Assumptions~\ref{assumption:growth_rate}~and~\ref{assumption:regime_n}, let $\mM \in \R^{n \times n}$ be any matrix with a bounded operator norm, let $\mQ_m \in \R^{n \times n}$ be the resolvent defined in \eqref{def:resolvent}, and let $\mQ_{-} \in \R^{n \times n}$ be the resolvent defined in \eqref{def:Q_-}. Then,
    \begin{equation*}
        \Bigl\lVert \E\left[\mQ_m^T \mM \mQ_m \right]-\E\left[\mQ_{-}^T \mM \mQ_{-}\right] \Bigr\rVert = \mathcal{O}\left(\frac{1}{m}\right).
    \end{equation*}
    \label{lem:deterministic_E[Q_m^TNQ_m]}
\end{lemma}
\begin{proof}
    We observe that
    \begin{align*}
        \Bigl\lVert\E\left[\mQ_m^T \mM \mQ_m \right]-\E\left[\mQ_{-}^T \mM \mQ_{-}\right] \Bigr\rVert \leq&~\Bigl\lVert\E\left[\mQ_m^T \mM \mQ_m \right]-\E\left[\mQ_{-}^T \mM \mQ_m\right] \Bigr\rVert \\
        &\quad + \Bigl\lVert\E\left[\mQ_{-}^T \mM \mQ_m \right]-\E\left[\mQ_{-}^T \mM \mQ_{-}\right] \Bigr\rVert.
    \end{align*}
    The objective is to show that both terms vanish.  By exchangeability arguments, we have
    \begin{align*}
        &\Bigl\lVert \E\left[\mQ_m^T \mM \mQ_m \right]-\E\left[\mQ_{-}^T \mM \mQ_m\right] \Bigr\rVert \\
        &=\left\lVert\frac{1}{N}\E\left[\sum_{i=1}^N[\mQ_m - \mQ_{-i}]^T \mM \mQ_m \right] \right\rVert\\
        &=\left\lVert\frac{1}{N}\E\left[\sum_{i=1}^N\frac{1}{m}\mQ_m^T \hat \mU_n^T \vsigma_i\vsigma_i^T (\hat \mU_n-\gamma \hat \mV_n) \mQ_{-i}^T \mM \mQ_m \right] \right\rVert \quad \text{(\lemref{lem:resolvent_identity})}\\
        &=\left\lVert\frac{1}{N}\E\left[\sum_{i=1}^N\frac{1}{m}\mQ_m^T \hat \mU_n^T \vsigma_i\vsigma_i^T (\hat \mU_n-\gamma \hat \mV_n) \mQ_m^T \mM \mQ_m \biggl(1+\frac{1}{m}\vsigma_i^T \hat \mU_n \mQ_{-i}(\hat \mU_n - \gamma \hat \mV_n)^T \vsigma_i\biggr)\right] \right\rVert\\
        &=\left\lVert\frac{1}{N}\E\left[\frac{1}{m}\mQ_m^T \hat \mU_n^T \FSVisited^T \mD \FSVisited (\hat \mU_n-\gamma \hat \mV_n) \mQ_m^T \mM \mQ_m\right] \right\rVert,
    \end{align*}
    where $\mD \in \R^{N \times N}$ is a diagonal matrix for which, for all $i \in [N]$, we have 
    \begin{equation*}
        \mD_i=1+\frac{1}{m}\vsigma_i^T \hat \mU_n \mQ_{-i}(\hat \mU_n - \gamma \hat \mV_n)^T \vsigma_i.
    \end{equation*}
    From \lemref{lem:boundness_norm_E[D]}, we know
    \begin{equation*}
        \E\left[\lVert \mD \rVert \right] = \mathcal{O}(1).
    \end{equation*}
    Furthermore, from \lemref{lem:upper_bound_norm_resolvent_+_RF_matrix}, we know there exists a real $K_\mQ' > 0$ such that, for all $m$, we have
    \begin{equation*}
        \left\lVert \frac{1}{\sqrt{m}} \FSVisited \hat \mU_n \mQ_m \right\rVert \leq K_\mQ'
    \end{equation*}
    and
    \begin{equation*}
        \left\lVert \frac{1}{\sqrt{m}} \mQ_m (\hat \mU_n - \gamma \hat \mV_n)^T \FSVisited^T \right\rVert \leq 2 K_\mQ'.
    \end{equation*}
    We deduce thus
    \begin{align*}
        \left\lVert \E\left[\mQ_m^T \mM \mQ_m \right]-\E\left[\mQ_{-}^T \mM \mQ_m\right] \right\rVert &= \left\lVert\frac{1}{m}\frac{m}{N}\E\left[\frac{1}{m}\mQ_m^T \hat \mU_n^T \FSVisited^T \mD \FSVisited (\hat \mU_n-\gamma \hat \mV_n) \mQ_m^T \mM \mQ_m\right] \right\rVert \\
        &= \mathcal{O}\left(\frac{1}{m}\right).
    \end{align*}
    With a similar reasoning, we can show that
    \begin{equation*}
        \left\lVert \E\left[\mQ_{-}^T \mM \mQ_m \right]-\E\left[\mQ_{-}^T \mM \mQ_{-}\right] \right\rVert = \mathcal{O}\left(\frac{1}{m}\right),
    \end{equation*}
    and we conclude thus
    \begin{equation*}
        \left\lVert\E\left[\mQ_m^T \mM \mQ_m \right]-\E\left[\mQ_{-}^T \mM \mQ_{-}\right] \right\rVert =  \mathcal{O}\left(\frac{1}{m}\right).
    \end{equation*}
\end{proof}
\begin{lemma}
    Under Assumptions~\ref{assumption:growth_rate}~and~\ref{assumption:regime_n}, let $\mQ_m \in \R^{n \times n}$ be the resolvent defined in \eqref{def:resolvent}, let $\mQ_{-} \in \R^{n \times n}$ be the resolvent defined in \eqref{def:Q_-}, and let $\mPsi_1 \in \R^{n \times n}$ be the matrix defined in \eqref{def:Psi_1_and_Psi_2}. Then,
    \begin{equation*}
        \Bigl\lVert \E\left[\mQ_m^T \mPsi_1 \mQ_m \right]-\E\left[\mQ_{-}^T \mPsi_1 \mQ_{-}\right] \Bigr\rVert = \mathcal{O}\left(\frac{1}{\sqrt{m}}\right).
    \end{equation*}
    \label{lem:deterministic_E[Q_m^Psi_1Q_m]}
\end{lemma}
\begin{proof}
    We observe that
    \begin{align*}
        \Bigl\lVert\E\left[\mQ_m^T \mPsi_1 \mQ_m \right]-\E\left[\mQ_{-}^T  \mPsi_1 \mQ_{-}\right] \Bigr\rVert \leq& \Bigl\lVert\E\left[\mQ_m^T \mPsi_1 \mQ_m \right]-\E\left[\mQ_{-}^T \mPsi_1 \mQ_m\right] \Bigr\rVert \\
        &+ \Bigl\lVert\E\left[\mQ_{-}^T \mPsi_1 \mQ_m \right]-\E\left[\mQ_{-}^T \mPsi_1 \mQ_{-}\right] \Bigr\rVert.
    \end{align*}
    The objective is to show that both terms vanish. By exchangeability arguments, we have
    \begin{align*}
        &\left\lVert \E\left[\mQ_m^T \mPsi_1 \mQ_m \right] - \E\left[\mQ_{-}^T \mPsi_1 \mQ_m \right] \right\rVert \\
        &=\left\lVert \frac{1}{N} \sum_{i=1}^N\E\left[\bigl[\mQ_m-\mQ_{-i}\bigr]^T \mPsi_1 \mQ_m \right] \right\rVert \\
        &=\left\lVert \frac{1}{N} \sum_{i=1}^N \E\left[\frac{1}{m}\mQ_m^T\hat \mU_n^T \vsigma_i \vsigma_i^T (\hat \mU_n-\gamma \hat \mV_n) \mQ_{-i}^T \mPsi_1 \mQ_m \right] \right\rVert \qquad \text{(\lemref{lem:resolvent_identity})} \\
        &=\left\lVert \frac{1}{N} \sum_{i=1}^N \E\left[\frac{1}{m}\mQ_m^T\hat \mU_n^T \vsigma_i \vsigma_i^T (\hat \mU_n-\gamma \hat \mV_n) \mQ_m^T \mPsi_1 \mQ_m \biggl(1+\frac{1}{m}\vsigma_i^T \hat \mU_n \mQ_{-i}(\hat \mU_n - \gamma \hat \mV_n)^T \vsigma_i\biggr) \right] \right\rVert \\
        &=\left\lVert \underbrace{\frac{1}{N}\E\left[\frac{1}{m}\mQ_m^T\hat \mU_n^T \FSVisited^T \mD \FSVisited (\hat \mU_n-\gamma \hat \mV_n) \mQ_m^T \mPsi_1 \mQ_m \right]}_{=\mZ} \right\rVert,
    \end{align*}   
    where $\mD \in \R^{N \times N}$ is a diagonal matrix for which, for all $i \in [N]$, we have 
    \begin{equation*}
        \mD_i=1+\frac{1}{m}\vsigma_i^T \hat \mU_n \mQ_{-i}(\hat \mU_n - \gamma \hat \mV_n)^T \vsigma_i.
    \end{equation*}
    Let the matrices
    \begin{equation*}
        \mB = \frac{1}{N}\frac{1}{m^{\frac{1}{4}}}\mQ_m^T\hat \mU_n^T \FSVisited^T \mD \FSVisited (\hat \mU_n-\gamma \hat \mV_n)\mQ_m^T
    \end{equation*}
    and 
    \begin{equation*}
        \mC^T = \frac{1}{m^{\frac{3}{4}}}  \mPsi_1 \mQ_m.
    \end{equation*}
    We decompose $\mZ$ with its symmetric and its skew-symmetric parts as 
    \begin{align*}
        \mZ = \E\bigl[\mB\mC^T \bigr] = \E\biggl[\frac{\mB \mC^T+\mC \mB^T}{2}\biggr] + \E\biggl[\frac{\mB \mC^T-\mC\mB^T}{2}\biggr].
    \end{align*}
    With the same reasoning on the symmetric part and the skew-symmetric part than for \eqref{eq:right_hand_part_bound_first_equivalence_Q}, we get for the operator norm
    \begin{equation*}
        \bigl\lVert \mZ \bigr\rVert \leq \bigl\lVert\E\left[\mB\mB^T\right]\bigr\rVert +  \bigl\lVert\E\left[\mC\mC^T\right]\bigr\rVert .
    \end{equation*}
    We want to show that both $\left\lVert\E\left[ \mB\mB^T\right]\right\rVert$ and $\left\lVert\E\left[\mC\mC^T\right]\right\rVert$ vanish. We have
    \begin{equation*}
        \E\left[\mB\mB^T\right] = \E\left[\frac{m^2}{N^2}\frac{1}{m^2\sqrt{m}}\mQ_m^T\hat \mU_n^T \FSVisited^T \mD \FSVisited (\hat \mU_n-\gamma \hat \mV_n) \mQ_m^T \mQ_m (\hat \mU_n-\gamma \hat \mV_n)^T \FSVisited^T \mD \FSVisited \hat \mU_n \mQ_m \right].
    \end{equation*}
    From \lemref{lem:boundness_norm_E[D]}, we know
    \begin{equation*}
        \E\left[\lVert \mD \rVert \right] = \mathcal{O}(1).
    \end{equation*}
    Furthermore, from \lemref{lem:upper_bound_norm_resolvent_+_RF_matrix}, we know there exists a real $K_\mQ' > 0$ such that, for all $m$, we have
    \begin{equation*}
        \left\lVert \frac{1}{\sqrt{m}} \FSVisited \hat \mU_n \mQ_m \right\rVert \leq K_\mQ'
    \end{equation*}
    and
    \begin{equation*}
        \left\lVert \frac{1}{\sqrt{m}} \mQ_m (\hat \mU_n - \gamma \hat \mV_n)^T \FSVisited^T \right\rVert \leq 2 K_\mQ'.
    \end{equation*}
    We have therefore
    \begin{equation*}
        \left\lVert \E\left[\mB\mB^T\right] \right\rVert = \mathcal{O}\left(\frac{1}{\sqrt{m}}\right).
    \end{equation*}
    For $\E\left[\mC\mC^T\right]$, we have
    \begin{equation*}
        \E\left[\mC\mC^T\right] = \E\biggl[ \frac{1}{m\sqrt{m}}\mQ_m^T \mPsi_1^2 \mQ_m \biggr].
    \end{equation*}
    Let $\vsigma_{N+1}$ and $\vsigma_{N+2}$ be independent vectors with the same law as $\vsigma_i$, we have
    \begin{align*}
        \E\biggl[\frac{1}{m\sqrt{m}}\mQ_m^T \mPsi_1 \mPsi_1 \mQ_m \biggr] = \E\biggl[\frac{1}{m\sqrt{m}} \frac{N^2}{m^2}\frac{1}{(1+\delta)^2}\mQ_m^T \hat \mU_n^T \vsigma_{N+1}\vsigma_{N+1}^T \hat \mU_n \hat \mU_n^T \vsigma_{N+2}\vsigma_{N+2}^T \hat \mU_n \mQ_m \biggr].
    \end{align*}
    Let
    \begin{equation*}
        \mB' = \frac{1}{m^{\frac{3}{4}}}\frac{N}{m}\frac{1}{1+\delta}\mQ_m^T \hat \mU_n^T \vsigma_{N+1}\vsigma_{N+1}^T \hat \mU_n
    \end{equation*}
    and 
    \begin{equation*}
        \mC'^T = \frac{1}{m^{\frac{3}{4}}}\frac{N}{m}\frac{1}{1+\delta}\hat \mU_n^T \vsigma_{N+2}\vsigma_{N+2}^T \hat \mU_n \mQ_m.
    \end{equation*}
    We decompose $\E\left[\mC\mC^T\right]$ with its symmetric and its skew-symmetric parts as 
    \begin{align*}
        \E\left[\mC\mC^T\right] = \E\bigl[\mB'\mC'^T \bigr] = \E\biggl[\frac{\mB' \mC'^T+\mC' \mB'^T}{2}\biggr] + \E\biggl[\frac{\mB' \mC'^T-\mC'\mB'^T}{2}\biggr],
    \end{align*}
    and we get for the operator norm 
    \begin{equation*}
        \bigl\lVert \E\left[\mC\mC^T\right] \bigr\rVert \leq \bigl\lVert\E\left[\mB'\mB'^T\right]\bigr\rVert +  \bigl\lVert\E\left[\mC'\mC'^T\right]\bigr\rVert .
    \end{equation*}
    To prove $\bigl\lVert \E\left[\mC\mC^T\right] \bigr\rVert$ vanish, we prove both $\bigl\lVert\E\left[\mB'\mB'^T\right]\bigr\rVert$ and $\bigl\lVert\E\left[\mC'\mC'^T\right]\bigr\rVert$ vanish. Let $K=\frac{1}{(1+\delta)^2}\frac{N}{N+1}\frac{N}{m}$, we write $\E\left[\mB'\mB'^T\right]$ as
    \begin{align*}
        \E\left[\mB'\mB'^T\right] &= \E\left[\frac{1}{m\sqrt{m}}\frac{N^2}{m^2}\frac{1}{(1+\delta)^2} \mQ_m^T \hat \mU_n^T \vsigma_{N+1}\vsigma_{N+1}^T \hat \mU_n \hat \mU_n^T \vsigma_{N+1}\vsigma_{N+1}^T \hat \mU_n \mQ_m \right] \\
        &= \E\left[\frac{1}{m\sqrt{m}}\frac{N^2}{m^2}\frac{1}{(1+\delta)^2} \mQ_{-N-1}^T \hat \mU_n^T \vsigma_{N+1}\vsigma_{N+1}^T \hat \mU_n \hat \mU_n^T \vsigma_{N+1}\vsigma_{N+1}^T \hat \mU_n \mQ_{-N-1} \right] \\
        &= \E\left[K\frac{1}{m\sqrt{m}}\sum_{i=1}^{N+1} \mQ_{-i}^T \hat \mU_n^T \vsigma_{i}\vsigma_{i}^T \hat \mU_n \mQ_{-i} \left(\frac{1}{m} \vsigma_{i}^T \hat \mU_n \hat \mU_n^T \vsigma_{i}\right) \right] \\
        &= \E\left[K\frac{1}{m\sqrt{m}}\sum_{i=1}^{N+1} \mQ_{-i}^T \hat \mU_n^T \vsigma_{i}\vsigma_{i}^T \hat \mU_n \mQ_{-i} \frac{1}{m}\Tr\bigl(\hat \mU_n \hat \mU_n^T \mPhi_{\SVisited}\bigr) \right] \\
        &+ \E\left[K\frac{1}{m\sqrt{m}}\sum_{i=1}^{N+1} \mQ_{-i}^T \hat \mU_n^T \vsigma_{i}\vsigma_{i}^T \hat \mU_n \mQ_{-i} \left(\frac{1}{m} \vsigma_{i}^T \hat \mU_n \hat \mU_n^T \vsigma_{i}-\frac{1}{m}\Tr\bigl(\hat \mU_n\hat \mU_n^T \mPhi_{\SVisited} \bigr)\right) \right] \\
        &= \underbrace{\E\left[K \frac{1}{m}\Tr\bigl(\hat \mU_n \hat \mU_n^T \mPhi_{\SVisited}\bigr)\frac{1}{m\sqrt{m}}\mQ_m^T \hat \mU_n^T \FSVisited^T \mD^2 \FSVisited \hat \mU_n \mQ_m \right]}_{=\mZ_1} \\
        &+ \underbrace{\E\left[K \frac{1}{m}\Tr\bigl(\hat \mU_n \hat \mU_n^T \mPhi_{\SVisited}\bigr)\frac{1}{m\sqrt{m}}\mQ_m^T \hat \mU_n^T \FSVisited^T \mD^2 \mD' \FSVisited \hat \mU_n \mQ_m \right]}_{=\mZ_2},
    \end{align*}
    where $\mD' \in \R^{N \times N}$ is a diagonal matrices for which, for all $i \in [N]$, we have
    \begin{equation*}
        \mD'_i = \frac{1}{m} \vsigma_{i}^T \hat \mU_n \hat \mU_n^T \vsigma_{i}-\frac{1}{m}\Tr\bigl(\hat \mU_n\hat \mU_n^T \mPhi_{\SVisited} \bigr).
    \end{equation*}
    From \lemref{lem:boundness_norm_E[D]}, from \lemref{lem:upper_bound_norm_resolvent_+_RF_matrix}, and from \eqref{eq:upper_bound_tr_phi}, we have 
    \begin{equation*}
        \lVert \mZ_1 \rVert = \mathcal{O}\left(\frac{1}{\sqrt{m}}\right).
    \end{equation*}
    From \lemref{lem:normal_concentration_resolvent}, we have 
    \begin{equation*}
        \E\bigl[\lVert \mD' \rVert \bigr] = \mathcal{O}\left(\frac{1}{\sqrt{m}}\right).
    \end{equation*}
    and thus
    \begin{equation*}
        \lVert \mZ_2 \rVert = \mathcal{O}\left(\frac{1}{m}\right).
    \end{equation*}
    We conclude that
    \begin{equation*}
        \left\lVert \E\left[\mB'\mB'^T\right] \right\rVert = \mathcal{O}\left(\frac{1}{\sqrt{m}}\right)
    \end{equation*}
    and 
    \begin{equation*}
        \left\lVert \E\left[\mC'\mC'^T\right] \right\rVert = \mathcal{O}\left(\frac{1}{\sqrt{m}}\right).
    \end{equation*}
    Therefore,
    \begin{equation*}
        \left\lVert \E\left[\mC\mC^T\right] \right\rVert = \mathcal{O}\left(\frac{1}{\sqrt{m}}\right)
    \end{equation*}
    and
    \begin{equation*}
        \left\lVert \E\left[\mQ_m^T\hat \mU_n^T \mPhi_{\SVisited} \hat \mU_n \mQ_m \right] - \E\left[\mQ_{-}^T\hat \mU_n^T \mPhi_{\SVisited} \hat \mU_n \mQ_m \right] \right\rVert =\mathcal{O}\left(\frac{1}{\sqrt{m}}\right). 
    \end{equation*}
    With a similar reasoning, we can show
    \begin{equation*}
        \left\lVert \E\left[\mQ_{-}^T \mPsi_1 \mQ_m \right]-\E\left[\mQ_{-}^T \mPsi_1 \mQ_{-}\right] \right\rVert = \mathcal{O}\left(\frac{1}{\sqrt{m}}\right).
    \end{equation*}
    We conclude thus
    \begin{equation*}
        \left\lVert\E\left[\mQ_m^T \mPsi_1 \mQ_m \right]-\E\left[\mQ_{-}^T \mPsi_1 \mQ_{-}\right] \right\rVert =  \mathcal{O}\left(\frac{1}{\sqrt{m}}\right).
    \end{equation*}
\end{proof}
\begin{lemma}
    Under Assumptions~\ref{assumption:growth_rate}~and~\ref{assumption:regime_n}, let $\mQ_m \in \R^{n \times n}$ be the resolvent defined in \eqref{def:resolvent}, let $\bar \mQ_m \in \R^{n \times n}$ be the deterministic resolvent defined in \eqref{def:Q_bar}, let $\mPsi_1, \mPsi_2 \in \R^{n \times n}$ be the matrices defined in \eqref{def:Psi_1_and_Psi_2}. Then,
    \begin{equation*}
        \Biggl\lVert \E\bigl[\mQ_m^T \mPsi_1 \mQ_m \bigr] -  \frac{1}{1-\frac{1}{N} \Tr \left(\mPsi_2 \bar \mQ_m^T \mPsi_1 \bar \mQ_m \right)}\bar \mQ_m^T \mPsi_1 \bar \mQ_m \Biggr\rVert =\mathcal{O}\left(\frac{1}{\sqrt{m}}\right).
    \end{equation*}
    \label{lem:decomposition_Q^TPsi_1Q}
\end{lemma}
\begin{proof}
    From \lemref{lem:decomposition_Q^TMQ}, we know that
    \begin{align*}
        \E\bigl[\mQ_m^T \mPsi_1 \mQ_m \bigr] &= \bar \mQ_m^T \mPsi_1 \bar \mQ_m  +  \E\left[\mQ_m^T \mPsi_1 \bar \mQ_m(\hat \mU_n-\gamma \hat \mV_n)^T \mPsi_{\SVisited} \hat \mU_n \mQ_m \right] \\
        &\quad - \E\left[\mQ_{-}^T \mPsi_1 \bar \mQ_m (\hat \mU_n-\gamma \hat \mV_n)^T \mPsi_{\SVisited} \hat \mU_n \mQ_{-}\right]  +\frac{1}{N} \Tr \left(\mPsi_2 \bar \mQ_m^T \mPsi_1 \bar \mQ_m \right)\E\left[\mQ_{-}^T \mPsi_1 \mQ_{-} \right]   \\
        &\quad +\mathcal{O_{\lVert \cdot \rVert}}\left(\frac{1}{\sqrt{m}}\right).
    \end{align*}
    Exploiting $\bar \mQ_m(\hat \mU_n-\gamma \hat \mV_n)^T \mPsi_{\SVisited} \hat \mU_n=\mI_n - \lambda \bar \mQ_n$ in the above equation, and from \lemref{lem:deterministic_E[Q_m^Psi_1Q_m]}, we obtain the simplification
    \begin{align*}
        \E\bigl[\mQ_m^T \mPsi_1 \mQ_m \bigr] &= \bar \mQ_m^T \mPsi_1 \bar \mQ_m  +\frac{1}{N} \Tr \left(\mPsi_2 \bar \mQ_m^T \mPsi_1 \bar \mQ_m \right)\E\left[\mQ_m^T \mPsi_1 \mQ_m \right]+\mathcal{O_{\lVert \cdot \rVert}}\left(\frac{1}{\sqrt{m}}\right).
    \end{align*}
    or equivalently 
    \begin{align*}
        &\E\bigl[\mQ_m^T \mPsi_1 \mQ_m \bigr]\left(1-\frac{1}{N} \Tr \left(\mPsi_2 \bar \mQ_m^T \mPsi_1 \bar \mQ_m \right)\E\left[\mQ_m^T \mPsi_1 \mQ_m \right]\right) = \bar \mQ_m^T \mPsi_1 \bar \mQ_m + \mathcal{O_{\lVert \cdot \rVert}}\left(\frac{1}{\sqrt{m}}\right).
    \end{align*}
    Let $\mB_n'=\bar \mZ^T \hat \mA_m \mZ$ and $\bar \mQ_m'=\left[\frac{N}{m}\frac{1}{1+\delta}\mB_n' + \lambda \mI_m \right]^{-1}$, for which $\hat \mA_m = \hat \mU_n (\hat \mU_n - \gamma \hat \mV_n)^T$ is the empirical transition model matrix (\eqref{def:empirical_transition_model_matrix}) and $\bar \mZ \bar \mZ^T=\mPhi_{\SVisited}$ is the Cholesky decompositon of $\mPhi_{\SVisited}$. We have from the cyclic properties of the trace
    \begin{align*}
        &\frac{1}{N} \Tr \left(\mPsi_2 \bar \mQ_m^T \mPsi_1 \bar \mQ_m \right) \\
        &=\frac{1}{m}\frac{N}{m}\frac{1}{(1+\delta)^2}\Tr \bigl((\hat \mU_n-\gamma \hat \mV_n)^T \mPhi_{\SVisited} (\hat \mU_n-\gamma \hat \mV_n) \bar \mQ_m^T \hat \mU_n^T \mPhi_{\SVisited} \hat \mU_n \bar \mQ_m \bigr) \\
        &= \frac{1}{m}\frac{N}{m}\frac{1}{(1+\delta)^2}\Tr \bigl(\mB_n' \mQ_m' \mQ_m'^T \mB_n'^T  \bigr).
    \end{align*}
    From \eqref{eq:lim_sup_Tr(BbarQBbarQ)}, we have
    \begin{equation*}
        \lim \sup_m \frac{1}{m}\frac{N}{m}\frac{1}{(1+\delta)^2}\Tr\bigl(\mB_n' \bar \mQ_m' \bar \mQ_m'^T \mB_n'^T\bigr) < 1.
    \end{equation*}
    Therefore, 
    \begin{align*}
        \E\bigl[\mQ_m^T \mPsi_1 \mQ_m \bigr] &= \bar \mQ_m^T \mPsi_1 \bar \mQ_m  +\frac{1}{1-\frac{1}{N} \Tr \left(\mPsi_2 \bar \mQ_m^T \mPsi_1 \bar \mQ_m \right)}\bar \mQ_m^T \mPsi_1 \bar \mQ_m+\mathcal{O_{\lVert \cdot \rVert}}\left(\frac{1}{\sqrt{m}}\right).
    \end{align*}
\end{proof}
\begin{lemma}
    Under Assumptions~\ref{assumption:growth_rate}~and~\ref{assumption:regime_n}, let $\mQ_m \in \R^{n \times n}$ be the resolvent defined in \eqref{def:resolvent}, let $\mQ_{-} \in \R^{n \times n}$ be the resolvent defined in \eqref{def:Q_-}, let $\bar \mQ_m \in \R^{n \times n}$ be the deterministic resolvent defined in \eqref{def:Q_bar}, let $\hat \mU_n, \hat \mV_n \in \R^{m \times n}$ be the shift matrices defined in \eqref{def:FXtrain_with_U_and_V}, and let $\mM$ be either any matrix with a bounded operator norm or $\mM=\mPsi_1$. Then, 
    \begin{align*}
        &\Biggl\lVert \E\bigl[\mQ_m^T \mM \mQ_m \bigr] - \bar \mQ_m^T \mM \bar \mQ_m  -  \E\left[\mQ_m^T \mM \bar \mQ_m(\hat \mU_n-\gamma \hat \mV_n)^T \mPsi_{\SVisited} \hat \mU_n \mQ_m \right] \\
        &\quad + \E\left[\mQ_{-}^T \mM \bar \mQ_m (\hat \mU_n-\gamma \hat \mV_n)^T \mPsi_{\SVisited} \hat \mU_n \mQ_{-}\right]  -\frac{1}{N} \Tr \left(\mPsi_2 \bar \mQ_m^T \mM \bar \mQ_m \right)\E\left[\mQ_{-}^T \mPsi_1 \mQ_{-} \right] \Biggr\rVert  \\
        &= \mathcal{O}\left(\frac{1}{\sqrt{m}}\right),
    \end{align*}
    for $\mPsi_1, \mPsi_2 \in \R^{n \times n}$ defined in \eqref{def:Psi_1_and_Psi_2}.
    \label{lem:decomposition_Q^TMQ}
\end{lemma}
\begin{proof}
    With the resolvent identity (\lemref{lem:resolvent_identity}), we decompose $\E\bigl[\mQ_m^T \mM \mQ_m \bigr]$  as 
    \begin{align}
        \E\bigl[\mQ_m^T \mM \mQ_m \bigr] &= \E\bigl[\mQ_m^T \mM \bar \mQ_m \bigr] - \E\bigl[\mQ_m^T \mM [\bar \mQ_m - \mQ_m] \bigr]  \\
        \begin{split}
            &= \E\bigl[\mQ_m^T \mM \bar \mQ_m \bigr] \\
            &\quad - \E\left[\mQ_m^T \mM \bar \mQ_m \biggl[\frac{1}{m} (\hat \mU_n-\gamma \hat \mV_n)^T \FSVisited^T \FSVisited \hat \mU_n -  (\hat \mU_n-\gamma \hat \mV_n)^T \mPsi_{\SVisited} \hat \mU_n \biggr] \mQ_m \right] \\
            &= \underbrace{\E\bigl[\mQ_m^T \mM \bar \mQ_m \bigr]}_{=\mZ_1} + \E\left[\mQ_m^T \mM \bar \mQ_m(\hat \mU_n-\gamma \hat \mV_n)^T \mPsi_{\SVisited} \hat \mU_n \mQ_m \right] \\
            &\qquad - \underbrace{\frac{1}{m} \sum_{i=1}^N \E\left[\mQ_m^T \mM \bar \mQ_m (\hat \mU_n-\gamma \hat \mV_n)^T \vsigma_i \vsigma_i^T \hat \mU_n \mQ_m \right]}_{=\mZ_2},
        \end{split} 
        \label{eq:decompositon_E[Q_m^TMQ_m]}
    \end{align}
    where $\FSVisited^T\FSVisited = \sum_{i=1}^N \vsigma_i \vsigma_i^T$ is the same decompositon of $\FSVisited^T\FSVisited$ than the one used in \eqref{eq:decompositionFState^TFstate}.  From \thmref{theorem:asy-behavior-E[Q]}, we have
    \begin{equation*}
        \left\lVert \E[\mQ_m] - \bar \mQ_m \right\rVert = \mathcal{O}\left(\frac{1}{\sqrt{m}}\right). 
    \end{equation*}
    Therefore, from above and from \lemref{lemma:upper_bound_norm_MQBar} which upper bounds $\lVert \mM \bar \mQ_m \rVert$, we deduce for $\mZ_1$ that 
    \begin{align*}
        \lVert \mZ_1 \rVert - \left\lVert \bar \mQ_m^T \mM \bar \mQ_m   \right\rVert&=\left\lVert \E\bigl[\mQ_m^T \mM \bar \mQ_m \bigr]  \right\rVert - \left\lVert \bar \mQ_m^T \mM \bar \mQ_m   \right\rVert \\
        &\leq \left\lVert \E[\mQ_m] - \bar \mQ_m \right\rVert \lVert \mM \bar \mQ_m \rVert \\
        &= \mathcal{O}\left(\frac{1}{\sqrt{m}}\right).
    \end{align*}
    We want to find now a deterministic approximation for $\mZ_2$ in \eqref{eq:decompositon_E[Q_m^TMQ_m]}. From the Sherman identity (\lemref{lem:sherman}) and with the resolvent $\mQ_{-i}$ defined in \eqref{def:Q_-i} as 
    \begin{equation*}
        \mQ_{-i} = \left[\frac{1}{m}(\hat \mU_n - \gamma \hat \mV_n)^T \FSVisited^T \FSVisited \hat \mU_n - \frac{1}{m} (\hat \mU_n - \gamma \hat \mV_n)^T \vsigma_i\vsigma_i^T \hat \mU_n + \lambda \mI_{n} \right]^{-1},
    \end{equation*}
    we obtain the following relation 
    \begin{equation*}
        \mQ_m = \mQ_{-i} - \frac{\frac{1}{m}\mQ_{-i}(\hat \mU_n - \gamma \hat \mV_n)^T \vsigma_i\vsigma_i^T \hat \mU_n\mQ_{-i}}{1+\frac{1}{m}\vsigma_i^T \hat \mU_n \mQ_{-i}(\hat \mU_n - \gamma \hat \mV_n)^T \vsigma_i}.
    \end{equation*}
    By remarking that for all $i \in [N]$, we have
    \begin{align*}
        &\mQ_m^T \mM \bar \mQ_m (\hat \mU_n-\gamma \hat \mV_n)^T \vsigma_i \vsigma_i^T \hat \mU_n \mQ_m \\
        &= \mQ_m^T \mM \bar \mQ_m (\hat \mU_n-\gamma \hat \mV_n)^T \vsigma_i \vsigma_i^T \hat \mU_n  \mQ_{-i} \frac{1}{1+\frac{1}{m}\vsigma_i^T \hat \mU_n \mQ_{-i}(\hat \mU_n - \gamma \hat \mV_n)^T \vsigma_i} \\
        &= \frac{1}{1+\delta}\mQ_m^T \mM \bar \mQ_m (\hat \mU_n-\gamma \hat \mV_n)^T \vsigma_i \vsigma_i^T \hat \mU_n \mQ_{-i} \\
        &+\frac{1}{1+\delta} \mQ_m^T \mM \bar \mQ_m (\hat \mU_n-\gamma \hat \mV_n)^T \vsigma_i \vsigma_i^T \hat \mU_n \mQ_{-i} \frac{\delta-\frac{1}{m}\vsigma_i^T \hat \mU_n \mQ_{-i}(\hat \mU_n - \gamma \hat \mV_n)^T \vsigma_i}{1+\frac{1}{m}\vsigma_i^T \hat \mU_n \mQ_{-i}(\hat \mU_n - \gamma \hat \mV_n)^T \vsigma_i}, 
    \end{align*}
    we decompose $\mZ_2$ as
    \begin{align}
        \mZ_2&=\frac{1}{m}\E\left[\sum_{i=1}^N\mQ_m^T \mM \bar \mQ_m (\hat \mU_n-\gamma \hat \mV_n)^T \vsigma_i \vsigma_i^T \hat \mU_n \mQ_m \right] \\
        &=\underbrace{\E\left[\mQ_{-}^T \mM \bar \mQ_m (\hat \mU_n-\gamma \hat \mV_n)^T \mPsi_{\SVisited} \hat \mU_n \mQ_{-}\right]}_{=\mZ_{21}} \\
        & -\underbrace{\frac{1}{m}\frac{1}{1+\delta}\sum_{i=1}^N\E\left[\mQ_{-i}^T\hat \mU_n^T \vsigma_i \vsigma_i^T \hat \mU_n \mQ_{-i} \frac{\frac{1}{m}\vsigma_i^T (\hat \mU_n - \gamma \hat \mV_n) \mQ_{-i}^T \mM \bar \mQ_m (\hat \mU_n-\gamma \hat \mV_n)^T \vsigma_i}{1+\frac{1}{m}\vsigma_i^T \hat \mU_n \mQ_{-i}(\hat \mU_n - \gamma \hat \mV_n)^T \vsigma_i} \right]}_{=\mZ_{22}} \label{def:Z_22_theorem_2} \\
        &+\underbrace{\frac{1}{m}\frac{1}{1+\delta}\sum_{i=1}^N\E\left[\mQ_{-i}^T \mM \bar \mQ_m (\hat \mU_n-\gamma \hat \mV_n)^T \vsigma_i \vsigma_i^T \hat \mU_n \mQ_{-i} \frac{\delta-\frac{1}{m}\vsigma_i^T \hat \mU_n \mQ_{-i}(\hat \mU_n - \gamma \hat \mV_n)^T \vsigma_i}{1+\frac{1}{m}\vsigma_i^T \hat \mU_n \mQ_{-i}(\hat \mU_n - \gamma \hat \mV_n)^T \vsigma_i}\right]}_{=\mZ_{23}} \label{def:Z_23_theorem_2} \\
        \begin{split}
            &-\frac{1}{m}\frac{1}{1+\delta}\sum_{i=1}^N\E\biggl[\mQ_{-i}^T\hat \mU_n^T \vsigma_i \vsigma_i^T \hat \mU_n \mQ_{-i} \\
            &\underbrace{\frac{\left(\frac{1}{m}\vsigma_i^T (\hat \mU_n - \gamma \hat \mV_n) \mQ_{-i}^T \mM \bar \mQ_m (\hat \mU_n-\gamma \hat \mV_n)^T \vsigma_i\right)\left(\delta-\frac{1}{m}\vsigma_i^T \hat \mU_n \mQ_{-i}(\hat \mU_n - \gamma \hat \mV_n)^T \vsigma_i\right)}{\left(1+\frac{1}{m}\vsigma_i^T \hat \mU_n \mQ_{-i}(\hat \mU_n - \gamma \hat \mV_n)^T \vsigma_i\right)^2} \biggr]}_{=\mZ_{24}}
        \end{split} \\
        &=\mZ_{21}-\mZ_{22}+\mZ_{23}-\mZ_{24}.
    \end{align}
    From \lemref{lem:deterministic_equivalent_Z22_theorem_2}, we have
    \begin{equation*}
        \Biggl\lVert \mZ_{22}-\frac{1}{N}\Tr \bigl(\mPsi_2 \bar \mQ_m^T \mM \bar \mQ_m \bigr)\E\left[\mQ_{-}^T\mPsi_1 \mQ_{-} \right]  \Biggr\rVert =\mathcal{O}\left(\frac{1}{\sqrt{m}}\right).
    \end{equation*}
    With a similar proof than for $\mZ_{22}$, we can show for $\mZ_{24}$ that
    \begin{equation*}
        \left\lVert \mZ_{24} \right\rVert = \mathcal{O}\left(\frac{1}{\sqrt{m}}\right).
    \end{equation*}
    From \lemref{lem:deterministic_equivalent_Z23_theorem_2}, we have 
    \begin{equation*}
        \lVert \mZ_{23} \rVert = \mathcal{O}\left(\frac{1}{\sqrt{m}}\right).
    \end{equation*}
    As a consequence, we conclude that
    \begin{align*}
            \E\bigl[\mQ_m^T \mM \mQ_m \bigr] &= \bar \mQ_m^T \mM \bar \mQ_m  +  \E\left[\mQ_m^T \mM \bar \mQ_m(\hat \mU_n-\gamma \hat \mV_n)^T \mPsi_{\SVisited} \hat \mU_n \mQ_m \right] \\
             &\quad - \E\left[\mQ_{-}^T \mM \bar \mQ_m (\hat \mU_n-\gamma \hat \mV_n)^T \mPsi_{\SVisited} \hat \mU_n \mQ_{-}\right] +\frac{1}{N} \Tr\left(\mPsi_2 \bar \mQ_m^T \mM \bar \mQ_m \right)\E\left[\mQ_{-}^T\mPsi_1 \mQ_{-} \right]  \\
             &\quad+ \mathcal{O_{\lVert \cdot \rVert}}\left(\frac{1}{\sqrt{m}}\right).
    \end{align*}
\end{proof}
\begin{lemma}
     Under Assumptions~\ref{assumption:growth_rate}~and~\ref{assumption:regime_n}, let $\mZ_{22} \in \R^{n \times n}$ be the matrix defined in \eqref{def:Z_22_theorem_2} as
    \begin{equation*}
        \mZ_{22} = \frac{1}{m}\frac{1}{1+\delta}\sum_{i=1}^N\E\left[\mQ_{-i}^T\hat \mU_n^T \vsigma_i \vsigma_i^T \hat \mU_n \mQ_{-i} \frac{\frac{1}{m}\vsigma_i^T (\hat \mU_n - \gamma \hat \mV_n) \mQ_{-i}^T \mM \bar \mQ_m (\hat \mU_n-\gamma \hat \mV_n)^T \vsigma_i}{1+\frac{1}{m}\vsigma_i^T \hat \mU_n \mQ_{-i}(\hat \mU_n - \gamma \hat \mV_n)^T \vsigma_i} \right].
    \end{equation*}
    Then, 
    \begin{equation*}
        \Biggl\lVert \mZ_{22}-\frac{1}{N}\Tr \bigl(\mPsi_2 \bar \mQ_m^T \mM \bar \mQ_m \bigr)\E\left[\mQ_{-}^T\mPsi_1 \mQ_{-} \right]  \Biggr\rVert =\mathcal{O}\left(\frac{1}{\sqrt{m}}\right),
    \end{equation*}
    where $\bar \mQ_m \in \R^{n \times n}$ is the deterministic resolvent defined in \eqref{def:Q_bar}, $\mQ_{-} \in \R^{n \times n}$ is the resolvent defined in \eqref{def:Q_-}, and $\mPsi_1, \mPsi_2 \in \R^{n \times n}$ defined in \eqref{def:Psi_1_and_Psi_2}.
    \label{lem:deterministic_equivalent_Z22_theorem_2}
\end{lemma}
\begin{proof}
    Let $\mD \in \R^{N \times N}$ be a diagonal matrix for which, for all $i \in [N]$, we have
    \begin{equation*}
        \mD_i=1+\frac{1}{m}\vsigma_i^T \hat \mU_n \mQ_{-i}(\hat \mU_n - \gamma \hat \mV_n)^T \vsigma_i,
    \end{equation*}
    and $\mD_2 \in \R^{N \times N}$ be another diagonal matrix for which, for all $i \in [N]$, we have
    \begin{align*}
       [\mD_2]_i&=\frac{\frac{1}{m}\vsigma_i^T (\hat \mU_n - \gamma \hat \mV_n) \mQ_{-i}^T \mM \bar \mQ_m (\hat \mU_n-\gamma \hat \mV_n)^T \vsigma_i}{1+\frac{1}{m}\vsigma_i^T \hat \mU_n \mQ_{-i}(\hat \mU_n - \gamma \hat \mV_n)^T \vsigma_i}  - \frac{1}{N} \Tr\left(\mPsi_2 \bar \mQ_m^T \mM \bar \mQ_m \right) \\
       &=\frac{\frac{1}{m}\vsigma_i^T (\hat \mU_n - \gamma \hat \mV_n) \mQ_{-i}^T \mM \bar \mQ_m (\hat \mU_n-\gamma \hat \mV_n)^T \vsigma_i}{1+\frac{1}{m}\vsigma_i^T \hat \mU_n \mQ_{-i}(\hat \mU_n - \gamma \hat \mV_n)^T \vsigma_i}  \\
       &\quad - \frac{\frac{1}{m}\Tr\left((\hat \mU_n-\gamma \hat \mV_n)^T \mPhi_{\SVisited} (\hat \mU_n-\gamma \hat \mV_n) \bar \mQ_m^T \mM \bar \mQ_m \right)}{1+\delta} \\
       &=\frac{\frac{1}{m}\vsigma_i^T (\hat \mU_n - \gamma \hat \mV_n) \mQ_{-i}^T \mM \bar \mQ_m (\hat \mU_n-\gamma \hat \mV_n)^T \vsigma_i}{1+\frac{1}{m}\vsigma_i^T \hat \mU_n \mQ_{-i}(\hat \mU_n - \gamma \hat \mV_n)^T \vsigma_i}  \\
       &\quad - \frac{\frac{1}{m}\Tr\left((\hat \mU_n-\gamma \hat \mV_n) \bar \mQ_m^T \mM \bar \mQ_m (\hat \mU_n-\gamma \hat \mV_n)^T \mPhi_{\SVisited} \right)}{1+\delta}.
    \end{align*}
    We have
    \begin{align*}
        &\Biggl\lVert \mZ_{22}-\frac{1}{N}\E\left[\mQ_{-}^T\mPsi_1 \mQ_{-} \right] \Tr \bigl(\mPsi_2 \bar \mQ_m^T \mM \bar \mQ_m \bigr) \Biggr\rVert \\
        &=\biggl\lVert \mZ_{22} -\frac{1}{m}\frac{1}{1+\delta}\E\left[\sum_{i=1}^N\mQ_{-i}^T\hat \mU_n^T \vsigma_i \vsigma_i^T \hat \mU_n \mQ_{-i} \right] \frac{1}{N} \Tr\left(\mPsi_2 \bar \mQ_m^T \mM \bar \mQ_m \right)\biggr\rVert \\
        &=\biggl\lVert \frac{1}{m} \frac{1}{1+\delta}\E\biggl[\sum_{i=1}^N \mQ_{-i}^T\hat \mU_n^T \vsigma_i \vsigma_i^T \hat \mU_n \mQ_{-i} \biggl(\frac{\frac{1}{m}\vsigma_i^T (\hat \mU_n - \gamma \hat \mV_n) \mQ_{-i}^T \mM \bar \mQ_m (\hat \mU_n-\gamma \hat \mV_n)^T \vsigma_i}{1+\frac{1}{m}\vsigma_i^T \hat \mU_n \mQ_{-i}(\hat \mU_n - \gamma \hat \mV_n)^T \vsigma_i} \\
        &\quad - \frac{1}{N} \Tr\left(\mPsi_2 \bar \mQ_m^T \mM \bar \mQ_m \right) \biggr) \biggr] \biggr\rVert \\
        &=\biggl\lVert \underbrace{\frac{1}{m} \frac{1}{1+\delta}\E\bigl[\mQ_m^T\hat \mU_n^T \FSVisited^T \mD^2 \mD_2 \FSVisited \hat \mU_n \mQ_m \bigr]}_{=\mZ_{221}} \biggr\rVert.
    \end{align*}
    Let the matrices
    \begin{equation*}
        \mB = m^{-\frac{1}{4}}\frac{1}{\sqrt{m}}\mQ_m^T\hat \mU_n^T \FSVisited^T \mD^2,
    \end{equation*}
    and
    \begin{equation*}
        \mC^T = m^\frac{1}{4}\frac{1}{\sqrt{m}} \mD_2 \FSVisited \hat \mU_n \mQ_m. 
    \end{equation*}
    Using the matrices above, we have
    \begin{equation*}
        \mZ_{221} = \frac{1}{m} \frac{1}{1+\delta}\E\bigl[\mQ_m^T\hat \mU_n^T \FSVisited^T \mD^2 \mD_2 \FSVisited \hat \mU_n \mQ_m \bigr] = \frac{1}{1+\delta}\E\bigl[\mB\mC^T \bigr] = \frac{1}{1+\delta}\E\biggl[\frac{\mB\mC^T + \mC\mB^T}{2} \biggr],
    \end{equation*}
    since $\mZ_{221}$ is symmetric. We use the relations $(\mB - \mC)(\mB - \mC)^T \succeq 0$ and $(\mB + \mC)(\mB+ \mC)^T \succeq 0$ to deduce the following relation
     \begin{equation*}
         -\mB\mB^T -\mC\mC^T \preceq \mB\mC^T + \mC\mB^T \preceq \mB\mB^T + \mC\mC^T.
     \end{equation*}
    From this relation, we obtain
    \begin{equation*}
        \lVert \mZ_{221} \rVert \leq \frac{1}{2(1+\delta)}\biggl( \E\bigl[\left\lVert \mB\mB^T \right\rVert\bigr] + \E\bigl[\left\lVert \mC\mC^T \right\rVert\bigr] \biggr),
    \end{equation*}
    where
    \begin{equation*}
        \mB\mB^T = \frac{1}{m\sqrt{m}}\mQ_m^T\hat \mU_n^T \FSVisited^T \mD^4 \FSVisited \hat \mU_n \mQ_m 
    \end{equation*}
    and
    \begin{equation*}
        \mC\mC^T = \frac{1}{\sqrt{m}}\mQ_m^T\hat \mU_n^T \FSVisited^T \mD_2^2 \FSVisited \hat \mU_n \mQ_m.
    \end{equation*}
    To get the Lemma, we prove that both $\E\bigl[\left\lVert \mB\mB^T \right\rVert\bigr]$ and $\E\bigl[\left\lVert \mC\mC^T \right\rVert\bigr]$ vanish. From \lemref{lem:upper_bound_norm_resolvent_+_RF_matrix}, we know there exists a real $K_{\mQ}'>0$ such that, for all $m$, we have
    \begin{equation*}
        \left\lVert \frac{1}{\sqrt{m}} \FSVisited \hat \mU_n \mQ_m \right\rVert \leq K_{\mQ}'
    \end{equation*}
    and
    \begin{equation*}
        \left\lVert \frac{1}{\sqrt{m}} \mQ_m (\hat \mU_n - \gamma \hat \mV_n)^T \FSVisited^T \right\rVert \leq 2K_{\mQ}'.
    \end{equation*}
    Furthermore, from \lemref{lem:boundness_norm_E[D]}, we know
    \begin{equation*}
        \E\left[\lVert \mD^4 \rVert \right] = \mathcal{O}(1).
    \end{equation*}
    We conclude that
    \begin{equation*}
        \E\bigl[\left\lVert \mB\mB^T \right\rVert\bigr] = \mathcal{O}\left(\frac{1}{\sqrt{m}}\right).
    \end{equation*}
    
    For $\E\bigl[\left\lVert \mC\mC^T \right\rVert\bigr]$, we remark that
    \begin{align*}
        \Pr\bigl([\mD_2]_i \geq t \bigr) &\leq \Pr\biggl(\frac{\frac{1}{m}\vsigma_i^T (\hat \mU_n - \gamma \hat \mV_n) \mQ_{-i}^T \mM \bar \mQ_m (\hat \mU_n-\gamma \hat \mV_n)^T \vsigma_i}{1+\frac{1}{m}\vsigma_i^T \hat \mU_n \mQ_{-i}(\hat \mU_n - \gamma \hat \mV_n)^T \vsigma_i}  \\
        &\quad - \frac{\frac{1}{m} \Tr\bigl(\mPhi_{\SVisited} (\hat \mU_n-\gamma \hat \mV_n) \bar \mQ_m^T \mM \bar \mQ_m (\hat \mU_n-\gamma \hat \mV_n)^T \bigr)}{1+\frac{1}{m}\vsigma_i^T \hat \mU_n \mQ_{-i}(\hat \mU_n - \gamma \hat \mV_n)^T \vsigma_i} \geq \frac{t}{2} \biggr) \\
        &+\Pr\biggl(\frac{\frac{1}{m} \Tr\bigl(\mPhi_{\SVisited} (\hat \mU_n-\gamma \hat \mV_n) \bar \mQ_m^T \mM \bar \mQ_m (\hat \mU_n-\gamma \hat \mV_n)^T \bigr)}{1+\frac{1}{m}\vsigma_i^T \hat \mU_n \mQ_{-i}(\hat \mU_n - \gamma \hat \mV_n)^T \vsigma_i} \\
        &\quad - \frac{\frac{1}{m} \Tr\bigl(\mPhi_{\SVisited} (\hat \mU_n-\gamma \hat \mV_n) \bar \mQ_m^T \mM \bar \mQ_m (\hat \mU_n-\gamma \hat \mV_n)^T \bigr)}{1+\delta} \geq \frac{t}{2}\biggr).
    \end{align*}
    Since $\lVert \mM \bar \mQ_m \rVert$ is bounded from \lemref{lemma:upper_bound_norm_MQBar}, with a similar proof than for \lemref{lem:concentration_coeff_D_i}, we can prove that
    \begin{align*}
        &\Pr \biggl(~\biggl\lvert\frac{1}{m}\vsigma^T (\hat \mU_n - \gamma \hat \mV_n) \mQ_{-i}^T \mM \bar \mQ_m (\hat \mU_n-\gamma \hat \mV_n)^T \vsigma \\
        &\quad - \frac{1}{m} \Tr\bigl((\hat \mU_n - \gamma \hat \mV_n) \E[\mQ_{-i}^T] \mM \bar \mQ_m (\hat \mU_n-\gamma \hat \mV_n)^T \mPhi_{\SVisited} \bigr) \biggr\rvert ~> t\biggr) \leq Ce^{-cm \max(t, t^2)},
    \end{align*}
    for some $C, c$ independent of $N, m$. Besides, from the proof of \thmref{theorem:asy-behavior-E[Q]}, we also have
    \begin{align*}
        &\biggl\lvert \frac{1}{m} \Tr\bigl((\hat \mU_n - \gamma \hat \mV_n) \E[\mQ_{-i}^T] \mM \bar \mQ_m (\hat \mU_n-\gamma \hat \mV_n)^T \mPhi_{\SVisited} \bigr)  \\
        &\qquad - \frac{1}{m} \Tr\bigl((\hat \mU_n - \gamma \hat \mV_n) \bar \mQ_m^T \mM \bar \mQ_m (\hat \mU_n-\gamma \hat \mV_n)^T \mPhi_{\SVisited} \bigr) \biggr\rvert = \mathcal{O}\left(\frac{1}{\sqrt{m}}\right),
    \end{align*}
    as both  $\lVert \hat \mU_n \rVert$ and $\lVert \hat \mV_n \rVert$ are upper bounded by $1$, $\lvert \Tr(\mA\mB) \rvert \leq \lVert \mA \rVert \Tr(\mB)$ for non-negative definite matrix $\mB$, and from \eqref{eq:upper_bound_tr_phi} that bounds $\frac{1}{m}\Tr(\mPhi_{\SVisited})$. From \lemref{lem:concentration_coeff_D_i}, we have 
    \begin{equation*}
        \Pr \biggl(\frac{1}{m}\vsigma_i^T \hat \mU_n \mQ_{-i}(\hat \mU_n - \gamma \hat \mV_n)^T \vsigma_i  - \alpha > t\biggr) \leq C'e^{-mc' \max(t, t^2)},
    \end{equation*}
    for some $C', c'$ independent of $N, m$. From \eqref{eq:alpha_minus_delta_vanihes} in the proof of \thmref{theorem:asy-behavior-E[Q]}, we have
    \begin{equation*}
        \left\lvert \alpha - \delta \right\lvert = \mathcal{O}\left(\frac{1}{\sqrt m}\right).
    \end{equation*}
    Combining all results above, we deduce that
    \begin{equation*}
        \E\left(\lVert \mD_2 \rVert^2\right)=\E\left(\max_{1 \leq i \leq N} [\mD^2_2]_i\right) = \int_0^\infty \Pr\left(\max_{1 \leq i \leq N} [\mD^2_2]_i > t\right)dt = \mathcal{O}\left(\frac{1}{m}\right),
    \end{equation*}
    and therefore
    \begin{equation*}
        \E\bigl[\left\lVert \mC\mC^T \right\rVert\bigr] = \mathcal{O}\left(\frac{1}{\sqrt{m}}\right).
    \end{equation*}
\end{proof}
\begin{lemma}
    Under Assumptions~\ref{assumption:growth_rate}~and~\ref{assumption:regime_n}, let $\mZ_{23} \in \R^{n \times n}$ be the matrix defined in \eqref{def:Z_23_theorem_2} as
    \begin{equation*}
        \mZ_{23} = \frac{1}{m}\frac{1}{1+\delta}\sum_{i=1}^N\E\left[\mQ_{-i}^T \mM \bar \mQ_m (\hat \mU_n-\gamma \hat \mV_n)^T \vsigma_i \vsigma_i^T \hat \mU_n \mQ_{-i} \frac{\delta-\frac{1}{m}\vsigma_i^T \hat \mU_n \mQ_{-i}(\hat \mU_n - \gamma \hat \mV_n)^T \vsigma_i}{1+\frac{1}{m}\vsigma_i^T \hat \mU_n \mQ_{-i}(\hat \mU_n - \gamma \hat \mV_n)^T \vsigma_i}\right].
    \end{equation*}
    Then,
    \begin{align*}
        \lVert \mZ_{23} \rVert=\mathcal{O}\left(\frac{1}{\sqrt{m}}\right).
    \end{align*}
    \label{lem:deterministic_equivalent_Z23_theorem_2}
\end{lemma}
\begin{proof}
    Let the matrices
    \begin{equation*}
        \mB_i = m^{-\frac{1}{4}}\frac{1}{\sqrt{m}}\mQ_{-i}^T \mM \bar \mQ_m (\hat \mU_n-\gamma \hat \mV_n)^T \vsigma_i
    \end{equation*}
    and 
    \begin{equation*}
        \mC_i^T = m^{\frac{1}{4}} \frac{1}{\sqrt{m}} \vsigma_i^T \hat \mU_n \mQ_{-i} \frac{\delta-\frac{1}{m}\vsigma_i^T \hat \mU_n \mQ_{-i}(\hat \mU_n - \gamma \hat \mV_n)^T \vsigma_i}{1+\frac{1}{m}\vsigma_i^T \hat \mU_n \mQ_{-i}(\hat \mU_n - \gamma \hat \mV_n)^T \vsigma_i}.
    \end{equation*}
    We decompose $\mZ_{23}$ with its symmetric and skew-symmetric parts as
    \begin{align*}
        \mZ_{23} &= \frac{1}{1+\delta} \sum_{i=1}^N \E\bigl[\mB_i \mC_i^T\bigr] \\
        &= \frac{1}{1+\delta} \sum_{i=1}^N \E\biggl[\frac{\mB_i \mC_i^T+\mC_i \mB_i^T}{2}\biggr] + \frac{1}{1+\delta} \E\biggl[\sum_{i=1}^N \frac{\mB_i \mC_i^T-\mC_i \mB_i^T}{2} \biggr].
    \end{align*}
    With the same reasoning on the symmetric part and on the skew-symmetric part than for \eqref{eq:right_hand_part_bound_first_equivalence_Q}, we get for the operator norm
    \begin{equation*}
        \lVert \mZ_{23} \rVert \leq \frac{1}{1+\delta}\left\lVert\E\left[\sum_{i=1}^N \mB_i\mB_i^T\right]\right\rVert + \frac{1}{1+\delta}\left\lVert\E\left[\sum_{i=1}^N \mC_i\mC_i^T\right]\right\rVert .
    \end{equation*}
    We want to show that both $\left\lVert\E\left[\sum_{i=1}^N \mB_i\mB_i^T\right]\right\rVert$ and $\left\lVert\E\left[\sum_{i=1}^N \mC_i\mC_i^T\right]\right\rVert$ vanish. We write $\E\left[\sum_{i=1}^N \mC_i\mC_i^T\right]$ as
    \begin{align*}
        \E\left[\sum_{i=1}^N \mC_i\mC_i^T\right] &= \E\left[\sum_{i=1}^N \frac{1}{\sqrt{m}} \mQ_{-i}^T \hat \mU_n^T \vsigma_i \vsigma_i^T \hat \mU_n \mQ_{-i} \frac{\bigl(\delta-\frac{1}{m}\vsigma_i^T \hat \mU_n \mQ_{-i}(\hat \mU_n - \gamma \hat \mV_n)^T \vsigma_i\bigr)^2}{\bigl(1+\frac{1}{m}\vsigma_i^T \hat \mU_n \mQ_{-i}(\hat \mU_n - \gamma \hat \mV_n)^T \vsigma_i\bigr)^2}\right] \\
        &= \E\left[\sum_{i=1}^N \frac{1}{\sqrt{m}} \mQ_m^T \hat \mU_n^T \vsigma_i \vsigma_i^T \hat \mU_n \mQ_m\biggl(\delta-\frac{1}{m}\vsigma_i^T \hat \mU_n \mQ_{-i}(\hat \mU_n - \gamma \hat \mV_n)^T \vsigma_i\biggr)^2\right] \\
        &= \E\left[\frac{1}{\sqrt{m}} \mQ_m^T \hat \mU_n^T \FSVisited^T \mD_3^2 \FSVisited \hat \mU_n \mQ_m\right],
    \end{align*}
    where $\mD_3 \in \R^{N \times N}$ is a diagonal matrix for which, for all $i \in [N]$, we have 
    \begin{equation*}
        [\mD_3]_i = \delta-\frac{1}{m}\vsigma_i^T \hat \mU_n \mQ_{-i}(\hat \mU_n - \gamma \hat \mV_n)^T \vsigma_i.
    \end{equation*}
    With a similar proof than for \lemref{lem:boundness_norm_E[D]}, and from \lemref{lem:first-asy-behavior-E[Q]} and \thmref{theorem:asy-behavior-E[Q]}, we find that
    \begin{equation*}
        \E\left(\lVert \mD_3 \rVert^2\right) = \mathcal{O}\left(\frac{1}{m}\right).
    \end{equation*}
    From \lemref{lem:upper_bound_norm_resolvent_+_RF_matrix}, we know there exists a real $K_\mQ'>0$ such that, for all $m$, we have 
    \begin{equation*}
        \left\lVert \frac{1}{\sqrt{m}} \FSVisited \hat \mU_n \mQ_m \right\rVert \leq K_\mQ'
    \end{equation*}
    and
    \begin{equation*}
        \left\lVert \frac{1}{\sqrt{m}} \mQ_m (\hat \mU_n - \gamma \hat \mV_n)^T \FSVisited^T \right\rVert \leq 2 K_\mQ'.
    \end{equation*}
    We deduce thus
    \begin{equation*}
        \left\lVert \E\left[\sum_{i=1}^N \mC_i\mC_i^T\right] \right\rVert = \mathcal{O}\left(\frac{1}{\sqrt{m}}\right).
    \end{equation*}
    We write $\E\left[\sum_{i=1}^N \mB_i\mB_i^T\right]$ as
    \begin{align*}
        \E\left[\sum_{i=1}^N \mB_i\mB_i^T\right] &= \E\left[\sum_{i=1}^N \frac{1}{m\sqrt{m}} \mQ_{-i}^T \mM \bar \mQ_m (\hat \mU_n-\gamma \hat \mV_n)^T \vsigma_i \vsigma_i^T (\hat \mU_n-\gamma \hat \mV_n) \bar \mQ_m^T \mM^T \mQ_{-i} \right] \\
        &= \frac{1}{\sqrt{m}} \frac{N}{m} \E\left[\mQ_-^T \mM \bar \mQ_m (\hat \mU_n-\gamma \hat \mV_n)^T \mPhi_{\SVisited} (\hat \mU_n-\gamma \hat \mV_n) \bar \mQ_m^T \mM^T \mQ_- \right],
    \end{align*}
    With a similar proof than for \lemref{lem:upper_bound_norm_resolvent}, we can show there exists a real $K_{\bar \mQ}>0$ such that, for all $m$, we have
    \begin{equation*}
        \lVert \bar \mQ_m \rVert \leq K_{\bar \mQ}.
    \end{equation*}
    Let $\hat \mA_m = \hat \mU_n (\hat \mU_n-\gamma \hat \mV_n)^T$ be the empirical transition model matrix defined in \eqref{def:empirical_transition_model_matrix}. Under Assumption~\ref{assumption:regime_n}, $\hat \mA_m$ is invertible. From \lemref{lem:upper_bound_norm_A}, we have
    \begin{align*}
        \left\lVert \bar \mQ_m (\hat \mU_n - \gamma \hat \mV_n)^T \mPhi_{\SVisited} \right\rVert &= \left\lVert \bar \mQ_m (\hat \mU_n - \gamma \hat \mV_n)^T \mPhi_{\SVisited} \hat \mU_n (\hat \mU_n - \gamma \hat \mV_n)^T \hat \mA_m^{-1} \right\rVert \\
        &=\left\lVert \frac{m}{N}(1+\delta)\bigl[\mI_n - \lambda \bar \mQ_m \bigr] (\hat \mU_n - \gamma \hat \mV_n)^T \hat \mA_m^{-1} \right\rVert \\
        &\leq 2\frac{m}{N}\frac{1+\delta}{\xi_{\min}} (1+K_{\bar \mQ}).
    \end{align*}
    From above and from \lemref{lemma:upper_bound_norm_MQBar} that upper bounds $\lVert \mM \bar \mQ_m \rVert$, we conclude that
    \begin{equation*}
        \left\lVert \E\left[\sum_{i=1}^N \mB_i\mB_i^T\right] \right\rVert = \mathcal{O}\left(\frac{1}{\sqrt{m}}\right).
    \end{equation*}
\end{proof}
\begin{lemma}
     Under Assumptions~\ref{assumption:growth_rate}~and~\ref{assumption:regime_n}, let $\bar \mQ_m \in \R^{n \times n}$ be the deterministic resolvent defined in \eqref{def:Q_bar}, and let $\mM$ be either any matrix with a bounded operator norm or $\mM=\hat \mU_n^T \mPsi_{\SVisited} \hat \mU_n$. Then there exists a real $K>0$ such that, for all $m$, we have
    \begin{equation*}
        \lVert \mM \bar \mQ_m \rVert \leq K.
    \end{equation*}
    \label{lemma:upper_bound_norm_MQBar}
\end{lemma}
\begin{proof}
    With a similar proof than for \lemref{lem:upper_bound_norm_resolvent}, we can show there exists a real $K_{\bar \mQ}>0$ such that, for all $m$, we have
    \begin{equation*}
        \lVert \bar \mQ_m \rVert \leq K_{\bar \mQ}.
    \end{equation*}
    In the case where $\mM$ is a matrix with a bounded operator norm, i.e., $\lVert \mM \rVert \leq K_{\mM}$ we have 
    \begin{equation*}
        \lVert \mM \bar \mQ_m \rVert \leq K_{\mM} K_{\bar \mQ}.
    \end{equation*}
    Otherwise, when $\mM = \hat \mU_n^T \mPsi_{\SVisited} \hat \mU_n$, we consider $\hat \mA_m = \hat \mU_n (\hat \mU_n-\gamma \hat \mV_n)^T$ the empirical transition model matrix defined in \eqref{def:empirical_transition_model_matrix}. Under Assumption~\ref{assumption:regime_n}, $\hat \mA_m$ is invertible. From \lemref{lem:upper_bound_norm_A}, we have
    \begin{align*}
        \lVert \mM \bar \mQ_m \rVert &= \left\lVert \frac{N}{m}\frac{1}{1+\delta} \hat \mU_n^T \mPhi_{\SVisited} \hat \mU_n \bar \mQ_m \right\rVert  \\
        &= \left\lVert \frac{N}{m}\frac{1}{1+\delta} \hat \mU_n^T \hat \mA_m^{-1} \hat \mU_n (\hat \mU_n-\gamma \hat \mV_n)^T \mPhi_{\SVisited} \hat \mU_n \bar \mQ_m \right\rVert \\
        &= \left\lVert \hat \mU_n^T \hat \mA_m^{-1} \hat \mU_n \bigl[\mI_n - \lambda \bar \mQ_m \bigr]  \right\rVert \\
        &\leq \frac{1}{\xi_{\min}} (1+\lambda K_{\bar \mQ}).
    \end{align*}
\end{proof}
\section{Proof of Theorem~\ref{theorem:asy-behavior-true-MSBE} \label{sec:proof_theorem_3}}
To simplify the notations, we denote the matrix $\mQ_m$ as the resolvent $\mQ_m(\lambda)$ (defined in \eqref{def:resolvent}), and we set $p=\lvert \State \rvert$. 
We define the matrices $\mPsi_{\SVisited} \in \R^{m \times m}$ and $\mPsi_{\State} \in \R^{p \times p}$ as 
 \begin{equation*}
    \mPsi_{\SVisited} = \frac{N}{m}\frac{1}{1+\delta}\mPhi_{\SVisited}~\text{ and }~\mPsi_{\State} = \frac{N}{m}\frac{1}{1+\delta}\mPhi_{\State}.
\end{equation*}
We also add the notation $\mA = \mB + \mathcal{O_{\lVert \cdot \rVert}}\left(\frac{1}{\sqrt{m}}\right)$ which means that $\lVert \mA - \mB \rVert = \mathcal{O}\left(\frac{1}{\sqrt{m}}\right)$.

Under Assumptions~\ref{assumption:growth_rate},~\ref{assumption:regime_n}~and~\ref{assumption:new_regime_n}, this section is dedicated to find an asymptotic deterministic version of the true $\MSBE(\hat \vtheta)$ defined in \eqref{def:MSBE} with a similar approach than the one used in \Appref{sec:proof_theorem_2} for $\Etrain(\hat \vtheta)$.  
In proofs of both \thmref{theorem:asy-behavior-E[Q]} and~\thmref{theorem:asy-behavior-MSBE}, we constantly use the fact that $\lVert \mQ_m \rVert$ is uniformly bounded by a constant $K_\mQ > 0$. We can also easily bound the operator norm of $\frac{1}{m}\FSVisited^T \FSVisited \hat \mU_n \mQ_m$ since the empirical transition model matrix $\hat \mA_m = \hat \mU_n(\hat \mU_n - \gamma \hat \mV_n)^T$ (\eqref{def:empirical_transition_model_matrix}) is invertible under Assumption~\ref{assumption:regime_n}. Indeed,
\begin{align*}
    \biggl\lVert \frac{1}{m}\FSVisited^T \FSVisited \hat \mU_n \mQ_m \biggr\rVert= \biggl\lVert \frac{1}{m} \hat \mA_m^{-1}\hat \mU_n(\hat \mU_n - \gamma \hat \mV_n)^T\FSVisited^T \FSVisited \hat \mU_n \mQ_m \biggr\rVert &= \bigl\lVert \hat \mA_m^{-1}\hat \mU_n\left[\mI_n - \lambda \mQ_m \right] \bigr\rVert \\
    &\leq \frac{1}{\xi_{\min}} \bigl(1 + \lambda K_\mQ \bigr) \\
\end{align*}
However, we do not have such simple control for $\frac{1}{m}\FState^T \FState \mU_n \mQ_m$ since $\mU_n(\mU_n - \gamma \mV_n)^T$ is not invertible until all states are visited. Furthermore, from \corref{corollary:operator_norm_FState}, only a $\mathcal{O}(\sqrt{m})$ upper bound can be derived for $\left\lVert \frac{1}{\sqrt{m}} \FSVisited \right\rVert$ or $\left\lVert \frac{1}{\sqrt{p}}\FState \right\rVert$. Nonetheless, with the additional Assumption~\ref{assumption:new_regime_n}, we can bound $\frac{1}{m}\mD_\vpi (\mI_p - \gamma \mP) \FState^T \FState \mU_n \mQ_m$  as stated by \lemref{lem:upper_bound_norm_transition_+_resolvent}. Fortunately for us, the proof of \thmref{theorem:asy-behavior-true-MSBE} indicates that controlling the operator norm of $\frac{1}{m}\mD_\vpi (\mI_p - \gamma \mP) \FState^T \FState \mU_n \mQ_m$ is sufficient.
\begin{lemma}
    Under Assumptions~\ref{assumption:growth_rate},~\ref{assumption:regime_n}~and~\ref{assumption:new_regime_n}, there exists $K>0$ such that, for all $m$, we have
    \begin{equation*}
        \biggl\lVert \frac{1}{m}\mD_\vpi (\mI_p - \gamma \mP) \FState^T \FState \mU_n \mQ_m \biggr\rVert \leq K.
    \end{equation*}
    \label{lem:upper_bound_norm_transition_+_resolvent}
\end{lemma}
\begin{proof}
    The main idea of the proof is to use a similar reasoning than for $\bigl\lVert \frac{1}{m}\hat \mA_m \FSVisited^T\FSVisited \mQ_m \bigr\rVert$. To this end, we use the triangular inequality, the decomposition $\mD_\vpi (\mI_p - \gamma \mP) = \mU_n(\mU_n - \gamma \mV_n)^T +  \mD_\vpi (\mI_p - \gamma \mP) - \mU_n(\mU_n - \gamma \mV_n)^T$, and Assumption~\ref{assumption:new_regime_n} as follows:
    \begin{align*}
        \biggl\lVert \frac{1}{m}\mD_\vpi (\mI_p - \gamma \mP) \FState^T \FState \mU_n \mQ_m \biggr\rVert
        &\leq \biggl\lVert \frac{1}{m} \mU_n(\mU_n - \gamma \mV_n)^T\FState^T \FState \mU_n \mQ_m \biggr\rVert \\
        & \qquad + \biggl\lVert \frac{1}{m} \bigl[\mD_\vpi (\mI_p - \gamma \mP) - \mU_n(\mU_n - \gamma \mV_n)^T \bigr]\FState^T \FState \mU_n \mQ_m \biggr\rVert \\
        &\leq \underbrace{\bigl\lVert \mU_n\bigl[\mI_n - \lambda \mQ_m \bigr] \bigr\rVert}_{=Z_1} \\
        & \qquad + \underbrace{\biggl\lVert \frac{1}{\sqrt{m}}\mU_n (\mU_n - \gamma \mV_n)^T \FState^T - \frac{1}{\sqrt{m}}\mD_\vpi \left[\mI_{\lvert \State \rvert} - \gamma \mP \right] \FState^T  \biggr\rVert \biggl\lVert \frac{1}{\sqrt{m}} \FState \mU_n \mQ_m \biggr\rVert}_{=Z_2}.
    \end{align*}
    From \lemref{lem:upper_bound_norm_resolvent}, we know there exists $K_\mQ > 0$ such that, for all $m$, we have $\lVert \mQ_m \rVert \leq K_\mQ$. Therefore, for the left-hand part $Z_1$, we have
    \begin{equation*}
        \bigl\lVert Z_1 \bigr\rVert = \bigl\lVert \mU_n\bigl[\mI_n - \lambda \mQ_m \bigr] \bigr\rVert \leq 1 + \lambda K_\mQ.
    \end{equation*}
    From Assumption~\ref{assumption:new_regime_n}, for the right-hand part $Z_2$, we have
    \begin{align*}
        &\biggl\lVert \frac{1}{\sqrt{m}}\mU_n (\mU_n - \gamma \mV_n)^T \FState^T - \frac{1}{\sqrt{m}}\mD_\vpi \left[\mI_{\lvert \State \rvert} - \gamma \mP \right] \FState^T \biggr\rVert \\
        &\leq \bigl\lVert \mU_n (\mU_n - \gamma \mV_n)^T - \mD_\vpi \bigl[\mI_{\lvert \State \rvert} - \gamma \mP \bigr] \bigr\rVert \biggl\lVert \frac{1}{\sqrt{m}} \FState \biggr\rVert \\
        &= \mathcal{O}\left(1\right),
    \end{align*}
    since $\lVert \FState \rVert = \mathcal{O}(m)$ from \corref{corollary:operator_norm_FState}.
    Furthermore, from \lemref{lem:upper_bound_norm_resolvent_+_RF_matrix}, we know there exists a real $K_\mQ'>0$ such that, for all $m$, we have
    \begin{equation*}
        \biggl\lVert \frac{1}{\sqrt{m}}\FState\mU_n\mQ_m \biggr\rVert = \biggl\lVert \frac{1}{\sqrt{m}}\FSVisited \hat \mU_n\mQ_m \biggr\rVert \leq K_\mQ'.
    \end{equation*}
    We have $\bigl\lVert Z_2 \bigr\rVert = \mathcal{O}(1)$ which concludes the proof.
\end{proof}
Assumption~\ref{assumption:new_regime_n} may hold for sufficiently large $n$ since $\E\left[\mU_n (\mU_n - \gamma \mV_n)^T\right] \to \mD_\vpi \left[\mI_p - \gamma \mP \right]$ as $n \to \infty$~\citep{tsitsiklis1996analysis, nedic2003least}. Indeed, \citet{tagorti2015rate} has established that we can control $\lVert \mU_n (\mU_n - \gamma \mV_n)^T - \mD_\vpi \left[\mI_{\lvert \State \rvert} - \gamma \mP \right] \rVert$ with a sufficiently large $n$.  

With the additional Assumption~\ref{assumption:new_regime_n}, we can now present the following result on the asymptotic Mean-Squared Bellman error.
\begin{theorem}[Asymptotic MSBE]
    Under Assumptions~\ref{assumption:growth_rate},~\ref{assumption:regime_n},~and~\ref{assumption:new_regime_n}, the \emph{determinsitic asymptotic MSBE} is 
    \begin{equation*}
        \DMSBE = \left\lVert \bar \vr + \gamma \tfrac{1}{\sqrt{n}} \mP \mPsi_\State \mU_n \bar \mQ_m(\lambda) \rtrain - \tfrac{1}{\sqrt{n}} \mPsi_\State \mU_n \bar \mQ_m(\lambda) \rtrain \right\rVert_{\mD_\vpi}^2 + \Delta,
    \end{equation*}
    with second-order correction factor
    \begin{equation*}
        \Delta = \tfrac{1}{n} \tfrac{\tfrac{1}{N}\Tr \left(\mLambda_\mP \left[ \mTheta_\State \mPsi_2 \mTheta_\State^T - 2\mTheta_\State (\mU_n - \gamma \mV_n)^T \mPsi_\State + \mPsi_\State \right] \right)}{1-\tfrac{1}{N} \Tr \left(\mPsi_2 \bar \mQ_m(\lambda)^T \mPsi_1 \bar \mQ_m(\lambda) \right)}\lVert \bar \mQ_m(\lambda) \rtrain \rVert_{\mPsi_1}^2,
    \end{equation*}
    where
    \begin{align*}
        \mLambda_\mP &= [\mI_{\lvert \State \rvert}-\gamma \mP ]^T \mD_{\vpi} [\mI_{\lvert \State \rvert}-\gamma \mP ], \\
        \mTheta_{\State} &= \mPsi_\State \mU_n \bar \mQ_m(\lambda).
    \end{align*}
    As $N, m, d \to \infty$ with asymptotic constant ratio $N/m$, $\MSBE(\hat \vtheta) - \DMSBE \xrightarrow{a.s} 0$.
\end{theorem}
\begin{proof}
    We decompose $\MSBE(\hat \vtheta)$ as
    \begin{align}
        \MSBE(\hat \vtheta) = \lVert \bar \vr + \gamma \mP \FState^T \hat \vtheta - \FState^T \hat \vtheta \rVert_{\mD_\vpi}^2 &= \lVert \bar \vr + \bigl[ \gamma \mP - \mI_p \bigr]\FState^T \vtheta \rVert_{\mD_\vpi}^2 \\
        &= \biggl\lVert \bar \vr - \frac{1}{m\sqrt{n}}\bigl[\mI_p-\gamma \mP \bigr]\FState^T \FState\mU_n \mQ_m \rtrain \biggr\rVert_{\mD_\vpi}^2 \\
        &= \lVert \bar \vr \rVert_{\mD_\vpi}^2 \\
        &\quad - \underbrace{\frac{2}{m\sqrt{n}}\bar \vr^T \mD_\vpi \bigl[\mI_p-\gamma \mP \bigr]\FState^T \FState\mU_n \mQ_m \rtrain}_{=Z_2} \label{def:Z_2_theorem_3} \\
        &\quad + \underbrace{\biggl\lVert \frac{1}{m\sqrt{n}}\bigl[\mI_p-\gamma \mP \bigr]\FState^T \FState\mU_n \mQ_m \rtrain \biggr\rVert_{\mD_\vpi}^2}_{=Z_3} \label{def:Z_3_theorem_3}.
    \end{align}
    We want to find an asymptotic equivalent for both $Z_2$ and $Z_3$.
    From \lemref{lem:Z2_theorem_3}, we have 
    \begin{equation*}
        \E\bigl[Z_{2}\bigr] = \frac{2}{\sqrt{n}}\bar \vr^T \mD_\vpi \bigl[\mI_p-\gamma \mP \bigr]\mPsi_\State \mU_n \bar \mQ_m \rtrain + \mathcal{O}\left(\frac{1}{\sqrt{m}}\right).
    \end{equation*}
    For $Z_3$, we have 
    \begin{equation*}
        Z_3 = \biggl\lVert \frac{1}{m\sqrt{n}}\bigl[\mI_p-\gamma \mP \bigr]\FState^T \FState\mU_n \mQ_m \rtrain \biggr\rVert_{\mD_\vpi}^2 = \frac{1}{nm^2} \rtrain^T \mQ_m^T \mU_n^T \FState^T \FState \Lambda_\mP \FState^T \FState \mU_n \mQ_m \rtrain.
    \end{equation*}
    From \lemref{lem:Z3_theorem_3}, we have 
    \begin{align*}
        \E\bigl[Z_{3}\bigr] &= \frac{1}{n} \rtrain^T \bar \mQ_m^T \mU_n^T \mPsi_\State \Lambda_\mP \mPsi_\State \mU_n \bar \mQ_m \rtrain  \\
        &+\frac{1}{n} \frac{\frac{1}{N} \Tr \left(\Lambda_\mP \bigl[\mTheta_\State \mPsi_2 \mTheta_\State^T - 2\mTheta_\State (\mU_n - \gamma \mV_n)^T \mPsi_\State + \mPsi_\State \bigr] \right)}{1-\frac{1}{N} \Tr \left(\mPsi_2 \bar \mQ_m^T  \mPsi_1 \bar \mQ_m  \right)} \lVert \bar \mQ_m \rtrain \rVert_{\mPsi_1}^2 \\
        &+ \mathcal{O}\left(\frac{1}{\sqrt{m}}\right).
    \end{align*}
    With a similar proof than for \lemref{lem:normal_concentration_msbe} we can 
    deduce that 
    \begin{equation*}
        \MSBE(\hat \vtheta) - \DMSBE \xrightarrow{a.s} 0,
    \end{equation*}
    as $m \to \infty$.
\end{proof}
\begin{lemma}
    Under Assumptions~\ref{assumption:growth_rate},~\ref{assumption:regime_n}~and~\ref{assumption:new_regime_n}, let $Z_2 \in \R$ defined in \eqref{def:Z_2_theorem_3} as 
    \begin{equation*}
        Z_2 = \frac{1}{m\sqrt{n}} \E\bigl[ \bar \vr^T \mD_\vpi \bigl[\mI_p-\gamma \mP \bigr]\FState^T \FState\mU_n \mQ_m \rtrain \bigr].
    \end{equation*}
    Then
    \begin{equation*}
        \biggl\lvert Z_2 - \frac{1}{\sqrt{n}}\bar \vr^T \mD_\vpi \bigl[\mI_p-\gamma \mP \bigr]\mPsi_\State \mU_n \bar \mQ_m \rtrain \biggr\rvert = \mathcal{O}\left(\frac{1}{\sqrt{m}}\right),
    \end{equation*}
    for $\bar \mQ_m$ the deterministic resolvent defined in \eqref{def:Q_bar}, and $\mPsi_\State \in \R^{p \times p}$ defined in \eqref{def:Psi_State_Lambda_P_Theta_State}.
    \label{lem:Z2_theorem_3}
\end{lemma}
\begin{proof}
    As in \eqref{eq:decompositionFState^TFstate}, we decompose the matrix $\FState^T\FState$ as 
    \begin{equation*}
        \FState^T\FState = \sum_{i=1}^N \vsigma_i \vsigma_i^T,
    \end{equation*}
    where $\vsigma_i = \sigma(\mS^T\vw_i) \in \R^{m}$ for which $\vw_i \in \R^d$ denotes the i-th row of $\mW$ defined in \eqref{def:rf_map}.
    Let $\mQ_{-i} \in \R^{n \times n}$ be the following resolvent 
    \begin{equation*}
        \mQ_{-i} = \biggl[\frac{1}{m}(\mU_n - \gamma \mV_n)^T \FState^T \FState \mU_n - \frac{1}{m} (\mU_n - \gamma \mV_n)^T \vsigma_i\vsigma_i^T \mU_n + \lambda \mI_{n} \biggr]^{-1},
    \end{equation*}
    independent of $\vsigma_i$ and thus $\vw_i$. From the Sherman identity (\lemref{lem:sherman}), we have
    \begin{align*}
        Z_2 &= \frac{1}{m\sqrt{n}} \E\left[\bar \vr^T \mD_\vpi \bigl[\mI_p-\gamma \mP \bigr]\FState^T \FState\mU_n \mQ_m \rtrain \right] \\
        &= \frac{1}{m\sqrt{n}}\E\left[\sum_{i=1}^N \bar \vr^T \mD_\vpi \bigl[\mI_p-\gamma \mP \bigr]\vsigma_i \vsigma_i^T \mU_n \mQ_m \rtrain\right] \\
        &= \frac{1}{m\sqrt{n}}\E\left[\sum_{i=1}^N \frac{\bar \vr^T \mD_\vpi \bigl[\mI_p-\gamma \mP \bigr]\vsigma_i \vsigma_i^T \mU_n \mQ_{-i} \rtrain}{1+\frac{1}{m}\vsigma_i^T \mU_n \mQ_{-i}(\mU_n - \gamma \mV_n)^T \vsigma_i}\right].
    \end{align*}
    Let $\mD \in \R^{N \times N}$ be a diagonal matrix for which, for all $i \in [N]$, we have
    \begin{equation*}
        \mD_i = \delta-\frac{1}{m}\vsigma_i^T \mU_n \mQ_{-i}(\mU_n - \gamma \mV_n)^T \vsigma_i.
    \end{equation*}
    We replace $1+\frac{1}{m}\vsigma_j^T \mU_n \mQ_{-j}(\mU_n - \gamma \mV_n)^T \vsigma_j$ by $1+\delta$ as following
    \begin{align*}
        Z_2 &=\underbrace{\frac{1}{m\sqrt{n}}\frac{1}{1+\delta}\E\left[\sum_{i=1}^N \bar \vr^T \mD_\vpi \bigl[\mI_p-\gamma \mP \bigr]\vsigma_i \vsigma_i^T \mU_n \mQ_{-i} \rtrain\right]}_{Z_{21}} \\
        &\quad +\underbrace{\frac{1}{m\sqrt{n}}\frac{1}{1+\delta}\E\left[\sum_{i=1}^N \frac{\bar \vr^T \mD_\vpi \bigl[\mI_p-\gamma \mP \bigr]\vsigma_i \vsigma_i^T \mU_n \mQ_{-i} \mD_i \rtrain}{1+\frac{1}{m}\vsigma_i^T \mU_n \mQ_{-i}(\mU_n - \gamma \mV_n)^T \vsigma_i}\right]}_{Z_{22}}.
    \end{align*}
    We have $Z_{22}$ vanishing since $\E\bigl[\lVert \mD \rVert \bigr]=\mathcal{O}\left(\frac{1}{\sqrt{m}}\right)$ and from \lemref{lem:upper_bound_norm_transition_+_resolvent}. From \thmref{theorem:asy-behavior-E[Q]}, we have thus 
    \begin{align*}
        Z_2 &= \frac{1}{m\sqrt{n}}\frac{1}{1+\delta}\E\left[\sum_{i=1}^N \bar \vr^T \mD_\vpi \bigl[\mI_p-\gamma \mP \bigr]\vsigma_i \vsigma_i^T \mU_n \mQ_{-i} \rtrain\right] + \mathcal{O}\left(\frac{1}{\sqrt{m}}\right) \\
        &= \frac{1}{\sqrt{n}}\frac{N}{m}\frac{1}{1+\delta}\bar \vr^T \mD_\vpi \bigl[\mI_p-\gamma \mP \bigr]\mPhi_\State \mU_n \E\bigl[\mQ_{-} \bigr] \rtrain + \mathcal{O}\left(\frac{1}{\sqrt{m}}\right) \\
        &= \frac{1}{\sqrt{n}}\bar \vr^T \mD_\vpi \bigl[\mI_p-\gamma \mP \bigr]\mPsi_\State \mU_n \bar \mQ_m \rtrain + \mathcal{O}\left(\frac{1}{\sqrt{m}}\right).
    \end{align*}
\end{proof}
\begin{lemma}
    Under Assumptions~\ref{assumption:growth_rate},~\ref{assumption:regime_n}~and~\ref{assumption:new_regime_n}, let $\mLambda_\mP \in \R^{p \times p}$ be the matrix defined in \eqref{def:Psi_State_Lambda_P_Theta_State}, and let $Z_3 \in \R$ be defined in \eqref{def:Z_3_theorem_3} as 
    \begin{equation*}
        Z_{3} = \E\biggl[\frac{1}{nm^2}\rtrain^T \mQ_m^T \mU_n^T \FState^T \FState \mLambda_\mP \FState^T \FState\mU_n \mQ_m \rtrain \biggr].
    \end{equation*}
    Then
    \begin{align*}
        \biggl\lvert Z_{3} &- \frac{1}{n} \rtrain^T \bar \mQ_m^T \mU_n^T \mPsi_\State \mLambda_\mP \mPsi_\State \mU_n \bar \mQ_m \rtrain  \\
        &-\frac{1}{n} \frac{\frac{1}{N}\Tr \Bigl(\mLambda_\mP \bigl[ \mTheta_\State \mPsi_2 \mTheta_\State^T - 2\mTheta_\State (\mU_n - \gamma \mV_n)^T \mPsi_\State + \mPsi_\State \bigl] \Bigr)}{1-\frac{1}{N} \Tr \left(\mPsi_2 \bar \mQ_m^T \mPsi_1 \bar \mQ_m \right)}\lVert \bar \mQ_m \rtrain \rVert_{\mPsi_1}^2 \biggr\rvert = \mathcal{O}\left(\frac{1}{\sqrt{m}}\right),
    \end{align*}
    where $\bar \mQ_m$ is the deterministic resolvent defined in \eqref{def:Q_bar},  $\mPsi_1, \mPsi_2 \in \R^{n \times n}$ are defined in \eqref{def:Psi_1_and_Psi_2}, $\mPsi_\State \in \R^{p \times p}$  and $\mTheta_\State \in \R^{p \times n}$ are defined in \eqref{def:Psi_State_Lambda_P_Theta_State}.
    \label{lem:Z3_theorem_3}
\end{lemma}
\begin{proof}
    As in \eqref{eq:decompositionFState^TFstate}, we decompose the matrix $\FState^T\FState$ as 
    \begin{equation*}
        \FState^T\FState = \sum_{i=1}^N \vsigma_i \vsigma_i^T,
    \end{equation*}
    where $\vsigma_i = \sigma(\mS^T\vw_i) \in \R^{m}$ for which $\vw_i \in \R^d$ denotes the i-th row of $\mW$ defined in \eqref{def:rf_map}.
    Let $\mQ_{-i} \in \R^{n \times n}$ be the following resolvent 
    \begin{equation*}
        \mQ_{-i} = \biggl[\frac{1}{m}(\mU_n - \gamma \mV_n)^T \FState^T \FState \mU_n - \frac{1}{m} (\mU_n - \gamma \mV_n)^T \vsigma_i\vsigma_i^T \mU_n + \lambda \mI_{n} \biggr]^{-1}
    \end{equation*}
    independent of $\vsigma_i$ and thus $\vw_i$. From the Sherman identity (\lemref{lem:sherman}), we decompose $Z_3$ as 
    \begin{align}
        Z_3 &= \E\biggl[\frac{1}{nm^2}\rtrain^T \mQ_m^T \mU_n^T \FState^T \FState \mLambda_\mP \FState^T \FState\mU_n \mQ_m \rtrain \biggr] \\
        &=\sum_{i, j=1}^N \E\biggl[\frac{1}{nm^2}\rtrain^T \mQ_m^T \mU_n^T \vsigma_i \vsigma_i^T \mLambda_\mP \vsigma_j \vsigma_j^T \mU_n \mQ_m \rtrain \biggr] \\
        &=\sum_{i, j=1}^N \E\Biggl[\frac{1}{nm^2}\rtrain^T \frac{\mQ_{-i}^T \mU_n^T \vsigma_i \vsigma_i^T}{1+\frac{1}{m}\vsigma_i^T \mU_n \mQ_{-i}(\mU_n - \gamma \mV_n)^T \vsigma_i} \mLambda_\mP \frac{\vsigma_j\vsigma_j^T \mU_n \mQ_{-j}}{1+\frac{1}{m}\vsigma_j^T \mU_n \mQ_{-j}(\mU_n - \gamma \mV_n)^T \vsigma_j}  \rtrain \Biggr] \\
        \begin{split}
            &=\underbrace{\sum_{i=1}^N \sum_{j \neq i} \E\Biggl[\frac{1}{nm^2}\rtrain^T \frac{\mQ_{-i}^T \mU_n^T \vsigma_i \vsigma_i^T}{1+\frac{1}{m}\vsigma_i^T \mU_n \mQ_{-i}(\mU_n - \gamma \mV_n)^T \vsigma_i} \mLambda_\mP \frac{\vsigma_j\vsigma_j^T \mU_n \mQ_{-j}}{1+\frac{1}{m}\vsigma_j^T \mU_n \mQ_{-j}(\mU_n - \gamma \mV_n)^T \vsigma_j}  \rtrain \Biggr]}_{=Z_{31}} \\
            &+\underbrace{\sum_{i=1}^N \E\Biggl[\frac{1}{nm^2}\rtrain^T \frac{\mQ_{-i}^T \mU_n^T \vsigma_i \vsigma_i^T \mLambda_\mP \vsigma_i\vsigma_i^T \mU_n \mQ_{-i}}{\left(1+\frac{1}{m}\vsigma_i^T \mU_n \mQ_{-i}(\mU_n - \gamma \mV_n)^T \vsigma_i \right)^2} \rtrain \Biggr]}_{=Z_{32}}.
            \label{eq:decomposition_Z_3_theorem_3}
        \end{split}
    \end{align}
    From \lemref{lem:Z31_theorem_3}, we have 
    \begin{align*}
        Z_{31} &= \frac{1}{n} \rtrain^T \bar \mQ_m^T \mU_n^T \mPsi_\State \mLambda_\mP \mPsi_\State \mU_n \bar \mQ_m \rtrain  \\
        &+\frac{1}{n} \frac{\frac{1}{N} \Tr \left(\mLambda_\mP \bigl[\mTheta_\State \mPsi_2 \mTheta_\State^T - 2\mTheta_\State (\mU_n - \gamma \mV_n)^T \mPsi_\State \bigr] \right)}{1-\frac{1}{N} \Tr \left(\mPsi_2 \bar \mQ_m^T  \mPsi_1 \bar \mQ_m  \right)} \lVert \bar \mQ_m \rtrain \rVert_{\mPsi_1}^2 \\
        &+ \mathcal{O}\left(\frac{1}{\sqrt{m}}\right).
    \end{align*}
    For the second term $Z_{32}$, we have from \lemref{lem:Z32_theorem_3},
     \begin{equation*}
        Z_{32}=\frac{1}{n}\frac{\frac{1}{N}\Tr(\mLambda_\mP \mPsi_\State )}{1-\frac{1}{N} \Tr \left(\mPsi_2 \bar \mQ_m^T \mPsi_1 \bar \mQ_m \right)}\lVert \bar \mQ_m \rtrain \rVert_{\mPsi_1}^2 + \mathcal{O}\left(\frac{1}{\sqrt{m}}\right).
    \end{equation*}
    We conclude that 
    \begin{align*}
        Z_{3} &= Z_{32} + Z_{32} \\
        &= \frac{1}{n} \rtrain^T \bar \mQ_m^T \mU_n^T \mPsi_\State \mLambda_\mP \mPsi_\State \mU_n \bar \mQ_m \rtrain  \\
        &+\frac{1}{n} \frac{\frac{1}{N} \Tr \left(\mLambda_\mP \bigl[\mTheta_\State \mPsi_2 \mTheta_\State^T - 2\mTheta_\State (\mU_n - \gamma \mV_n)^T \mPsi_\State + \mPsi_\State \bigr] \right)}{1-\frac{1}{N} \Tr \left(\mPsi_2 \bar \mQ_m^T  \mPsi_1 \bar \mQ_m  \right)} \lVert \bar \mQ_m \rtrain \rVert_{\mPsi_1}^2 \\
        &+ \mathcal{O}\left(\frac{1}{\sqrt{m}}\right).
    \end{align*}
\end{proof}
\begin{lemma}
    Under Assumptions~\ref{assumption:growth_rate},~\ref{assumption:regime_n}~and~\ref{assumption:new_regime_n}, let $Z_{31} \in \R$ defined in \eqref{eq:decomposition_Z_3_theorem_3} as
    \begin{equation*}
        Z_{31} = \frac{1}{nm^2} \sum_{i=1}^N \sum_{j \neq i} \E\Biggl[ \frac{\rtrain^T\mQ_{-i}^T \mU_n^T \vsigma_i \vsigma_i^T \mLambda_\mP \vsigma_j\vsigma_j^T \mU_n \mQ_{-j}\rtrain}{\left(1+\frac{1}{m}\vsigma_i^T \mU_n \mQ_{-i}(\mU_n - \gamma \mV_n)^T \vsigma_i\right)\left(1+\frac{1}{m}\vsigma_j^T \mU_n \mQ_{-j}(\mU_n - \gamma \mV_n)^T \vsigma_j\right)} \Biggr],
    \end{equation*}
    Then 
    \begin{align*}
        \Biggl\lvert Z_{31} &- \frac{1}{n} \rtrain^T \bar \mQ_m^T \mU_n^T \mPsi_\State \mLambda_\mP \mPsi_\State \mU_n \bar \mQ_m \rtrain  \\
        &-\frac{1}{n} \frac{\frac{1}{N}\Tr \left(\mLambda_\mP \bigl[ \mTheta_\State \mPsi_2 \mTheta_\State^T - 2\mTheta_\State (\mU_n - \gamma \mV_n)^T \mPsi_\State \bigr] \right)}{1-\frac{1}{N} \Tr \left(\mPsi_2 \bar \mQ_m^T  \mPsi_1 \bar \mQ_m \right)}  \lVert \bar \mQ_m \rtrain \rVert_{\mPsi_1}^2  \Biggr\rvert =\mathcal{O}\left(\frac{1}{\sqrt{m}}\right),
    \end{align*}
    where $\bar \mQ_m$ is the deterministic resolvent defined in \eqref{def:Q_bar},  $\mPsi_1, \mPsi_2 \in \R^{n \times n}$ are defined in \eqref{def:Psi_1_and_Psi_2}, $\mPsi_\State \in \R^{p \times p}$ and $\mTheta_\State \in \R^{p \times n}$ are defined in \eqref{def:Psi_State_Lambda_P_Theta_State}.
    \label{lem:Z31_theorem_3}
\end{lemma}
\begin{proof}
    Using the Sherman identity (\lemref{lem:sherman}), we decompose $Z_{31}$ as  
    \begin{align*}
        Z_{31} &= \frac{1}{nm^2} \sum_{i=1}^N \sum_{j \neq i} \E\Biggl[ \frac{\rtrain^T\mQ_{-i}^T \mU_n^T \vsigma_i \vsigma_i^T \mLambda_\mP \vsigma_j\vsigma_j^T \mU_n \mQ_{-j}\rtrain}{\left(1+\frac{1}{m}\vsigma_i^T \mU_n \mQ_{-i}(\mU_n - \gamma \mV_n)^T \vsigma_i\right)\left(1+\frac{1}{m}\vsigma_j^T \mU_n \mQ_{-j}(\mU_n - \gamma \mV_n)^T \vsigma_j\right)} \Biggr] \\
        &= \frac{1}{nm^2}\sum_{i=1}^N \sum_{j \neq i} \E\Biggl[\rtrain^T \frac{\mQ_m^T \mU_n^T \vsigma_i \vsigma_i^T \mLambda_\mP \vsigma_j\vsigma_j^T \mU_n \mQ_{-j}}{1+\frac{1}{m}\vsigma_j^T \mU_n \mQ_{-j}(\mU_n - \gamma \mV_n)^T \vsigma_j}  \rtrain \Biggr] \\
        &= \underbrace{\frac{1}{nm^2}\sum_{i=1}^N \sum_{j \neq i} \E\Biggl[\rtrain^T \frac{\mQ_{-j}^T \mU_n^T \vsigma_i \vsigma_i^T \mLambda_\mP \vsigma_j\vsigma_j^T \mU_n \mQ_{-j}}{1+\frac{1}{m}\vsigma_j^T \mU_n \mQ_{-j}(\mU_n - \gamma \mV_n)^T \vsigma_j}  \rtrain \Biggr]}_{=Z_{311}} \\
        &- \underbrace{\frac{1}{nm^3}\sum_{i=1}^N \sum_{j \neq i} \E\Biggl[\rtrain^T  \frac{\mQ_{-j}^T \mU_n^T \vsigma_j \vsigma_j^T (\mU_n - \gamma \mV_n) \mQ_{-j}^T \mU_n^T \vsigma_i \vsigma_i^T \mLambda_\mP \vsigma_j\vsigma_j^T \mU_n \mQ_{-j}}{\left(1+\frac{1}{m}\vsigma_j^T \mU_n \mQ_{-j}(\mU_n - \gamma \mV_n)^T \vsigma_j\right)^2}  \rtrain \Biggr]}_{=Z_{312}}.
    \end{align*}
    We want to find an asymptotic equivalent for both $Z_{311}$ and $Z_{312}$. For $Z_{312}$, we have 
    \begin{align*}
        Z_{312}&=\frac{1}{nm^3} \sum_{i=1}^N \sum_{j \neq i} \E\Biggl[\rtrain^T  \frac{\mQ_{-j}^T \mU_n^T \vsigma_j \vsigma_j^T (\mU_n - \gamma \mV_n) \mQ_{-j}^T \mU_n^T \vsigma_i \vsigma_i^T \mLambda_\mP \vsigma_j\vsigma_j^T \mU_n \mQ_{-j}}{\left(1+\frac{1}{m}\vsigma_j^T \mU_n \mQ_{-j}(\mU_n - \gamma \mV_n)^T \vsigma_j\right)^2}  \rtrain \Biggr] \\
        &=\frac{1}{nm} \sum_{j=1}^N \E\Biggl[\rtrain^T  \frac{\mQ_{-j}^T \mU_n^T \vsigma_j \vsigma_j^T \mU_n \mQ_{-j}\left(\frac{1}{m^2}\vsigma_j^T (\mU_n - \gamma \mV_n) \mQ_{-j}^T \mU_n^T \FState^{-jT}\FState^{-j} \mLambda_\mP \vsigma_j\right)}{\left(1+\frac{1}{m}\vsigma_j^T \mU_n \mQ_{-j}(\mU_n - \gamma \mV_n)^T \vsigma_j\right)^2}  \rtrain \Biggr],
    \end{align*}
    where $\FState^{-j}=\sigma(\mW^{-j}\mS) \in \R^{(N-1) \times n}$ for which $\mW^{-j} \in \R^{(N-1) \times d}$ depicts the same matrix than the weight matrix $\mW$ defined in \eqref{def:rf_map} without the $j^\text{th}$ row. Let $\mD_{312} \in \R^{N \times N}$ be a diagonal matrix for which, for all $j \in [N]$, we have
    \begin{equation*}
        [\mD_{312}]_j = \frac{1}{m^2}\vsigma_j^T (\mU_n - \gamma \mV_n) \mQ_{-j}^T \mU_n^T \FState^{-jT}\FState^{-j} \mLambda_\mP \vsigma_j - \frac{1}{m^2}\Tr\left((\mU_n - \gamma \mV_n) \mQ_{-j}^T \mU_n^T \FState^{-jT}\FState^{-j} \mLambda_\mP \mPhi_\State\right).
    \end{equation*}
    From \lemref{lem:upper_bound_norm_transition_+_resolvent}, we know there exists a real $K_1>0$ such that, for all $m$, we have $\lVert \mD_\vpi [\mI_p - \gamma \mP]\FState^T \FState \mU_n \mQ_m \rVert \leq K_1$. Therefore, we deduce that 
    \begin{equation*}
            \biggl\lVert \frac{1}{m}(\mU_n - \gamma \mV_n)\mQ_{-j}^T \mU_n^T \FState^{-jT}\FState^{-j} \mLambda_\mP \biggr\rVert =\mathcal{O}(1).
    \end{equation*}
    From \lemref{lem:normal_concentration_resolvent}, we deduce that $\E\bigl[\lVert \mD_{312} \rVert \bigr]=\mathcal{O}\left(\frac{1}{\sqrt{m}}\right)$. Therefore, we get
    \begin{align}
        \begin{split}
            Z_{312}&=\frac{1}{nm} \sum_{j=1}^N \E\Biggl[\rtrain^T  \frac{\mQ_{-j}^T \mU_n^T \vsigma_j \vsigma_j^T \mU_n \mQ_{-j}\frac{1}{m^2}\Tr\bigl((\mU_n - \gamma \mV_n) \mQ_{-j}^T \mU_n^T \FState^{-jT}\FState^{-j} \mLambda_\mP \mPhi_\State \bigr)}{\left(1+\frac{1}{m}\vsigma_j^T \mU_n \mQ_{-j}(\mU_n - \gamma \mV_n)^T \vsigma_j\right)^2}  \rtrain \Biggr] \\
            &+\frac{1}{nm} \sum_{j=1}^N \E\Biggl[\rtrain^T  \frac{\mQ_{-j}^T \mU_n^T \vsigma_j \vsigma_j^T \mU_n \mQ_{-j}[\mD_{312}]_j}{\left(1+\frac{1}{m}\vsigma_j^T \mU_n \mQ_{-j}(\mU_n - \gamma \mV_n)^T \vsigma_j\right)^2}  \rtrain \Biggr] \\
            &=\frac{1}{nm} \sum_{j=1}^N \E\Biggl[\rtrain^T  \frac{\mQ_{-j}^T \mU_n^T \vsigma_j \vsigma_j^T \mU_n \mQ_{-j}\frac{1}{m^2}\Tr\bigl((\mU_n - \gamma \mV_n) \mQ_{-j}^T \mU_n^T \FState^{-jT}\FState^{-j} \mLambda_\mP \mPhi_\State \bigr)}{\left(1+\frac{1}{m}\vsigma_j^T \mU_n \mQ_{-j}(\mU_n - \gamma \mV_n)^T \vsigma_j\right)^2}  \rtrain \Biggr] \\
            &+\frac{1}{nm} \E\Biggl[\rtrain^T  \mQ_m^T \mU_n^T \FState^T \mD_{312} \FState \mU_n \mQ_m  \rtrain \Biggr] \\
            &=\frac{1}{nm} \sum_{j=1}^N \E\Biggl[\rtrain^T  \frac{\mQ_{-j}^T \mU_n^T \vsigma_j \vsigma_j^T \mU_n \mQ_{-j}\frac{1}{m^2}\Tr\bigl((\mU_n - \gamma \mV_n) \mQ_{-j}^T \mU_n^T \FState^{-jT}\FState^{-j} \mLambda_\mP \mPhi_\State \bigr)}{\left(1+\frac{1}{m}\vsigma_j^T \mU_n \mQ_{-j}(\mU_n - \gamma \mV_n)^T \vsigma_j\right)^2}  \rtrain \Biggr] \\
            &+\mathcal{O}\left(\frac{1}{\sqrt{m}}\right),
            \label{eq:Z_312_trace}
        \end{split}
    \end{align}
    where the last equality is obtained since $\E\bigl[\lVert \mD_{312} \rVert \bigr]=\mathcal{O}\left(\frac{1}{\sqrt{m}}\right)$, and since we know there exists a real $K_{\mQ}'>0$ such that, for all $m$, we have
    \begin{equation*}
        \left\lVert \frac{1}{\sqrt{m}} \FState \mU_n \mQ_m \right\rVert \leq K_{\mQ}'
    \end{equation*}
    and
    \begin{equation*}
        \left\lVert \frac{1}{\sqrt{m}} \mQ_m (\mU_n - \gamma \mV_n)^T \FState^T \right\rVert \leq 2K_{\mQ}'.
    \end{equation*}
    from \lemref{lem:upper_bound_norm_resolvent_+_RF_matrix}. We replace $1+\frac{1}{m}\vsigma_j^T \mU_n \mQ_{-j}(\mU_n - \gamma \mV_n)^T \vsigma_j$ by $1+\delta$ in $Z_{312}$ as following
    \begin{align*}
        Z_{312}&=\frac{1}{nm^3} \frac{1}{(1+\delta)^2} \sum_{j=1}^N \E\Biggl[\rtrain^T  \mQ_{-j}^T \mU_n^T \vsigma_j \vsigma_j^T \mU_n \mQ_{-j}\Tr\bigl((\mU_n - \gamma \mV_n) \mQ_{-j}^T \mU_n^T \FState^{-jT}\FState^{-j} \mLambda_\mP \mPhi_\State \bigr)  \rtrain \Biggr] \\
        &+\frac{1}{nm^3} \frac{1}{(1+\delta)^2} \sum_{j=1}^N \E\Biggl[\rtrain^T  \frac{\mQ_{-j}^T \mU_n^T \vsigma_j \mD'_j \vsigma_j^T \mU_n \mQ_{-j}\Tr\bigl((\mU_n - \gamma \mV_n) \mQ_{-j}^T \mU_n^T \FState^{-jT}\FState^{-j} \mLambda_\mP \mPhi_\State \bigr)}{\left(1+\frac{1}{m}\vsigma_j^T \mU_n \mQ_{-j}(\mU_n - \gamma \mV_n)^T \vsigma_j\right)^2} \rtrain \Biggr] \\
        &+\mathcal{O}\left(\frac{1}{\sqrt{m}}\right) \\
        &=\underbrace{\frac{N}{nm^3} \frac{1}{(1+\delta)^2} \E\Biggl[\rtrain^T  \mQ_{-}^T \mU_n^T \mPhi_\State \mU_n \mQ_{-} \rtrain \Tr\bigl((\mU_n - \gamma \mV_n) \mQ_{-}^T \mU_n^T \FState^{-T}\FState^{-} \mLambda_\mP \mPhi_\State \bigr)  \Biggr]}_{=Z_{3121}} \\
        &+\underbrace{\frac{1}{nm^3} \frac{1}{(1+\delta)^2}\E\Biggl[\rtrain^T  \mQ_m^T \mU_n^T \FState^T \mD' \FState \mU_n \mQ_{m} \rtrain\Tr\bigl((\mU_n - \gamma \mV_n) \mQ_{-}^T \mU_n^T \FState^{-T}\FState^{-} \mLambda_\mP \mPhi_\State \bigr)  \Biggr]}_{=Z_{3122}} \\
        &+\mathcal{O}\left(\frac{1}{\sqrt{m}}\right),
    \end{align*}
    where $\mD' \in \R^{N \times N}$ is a diagonal matrix for which, for all $j \in [N]$, we have
    \begin{equation*}
        \mD'_j = (1+\delta)^2  - \left(1+\frac{1}{m}\vsigma_j^T \mU_n \mQ_{-j}(\mU_n - \gamma \mV_n)^T \vsigma_j\right)^2.
    \end{equation*}
    With a similar proof than for \eqref{eq:upper_bound_tr_phi}, we can show $\frac{1}{m}\Tr\bigl(\mPhi_\State\bigr)=\frac{p}{m}\frac{1}{p}\Tr\bigl(\mPhi_\State\bigr)$ is uniformly bounded under Assumption~\ref{assumption:new_regime_n}. Combining
    $\lvert \Tr(\mA\mB) \rvert \leq \lVert \mA \rVert \Tr(\mB)$ for non-negative definite matrix $\mB$ and  \lemref{lem:upper_bound_norm_transition_+_resolvent}, we have $\frac{1}{m^2}\Tr\bigl((\mU_n - \gamma \mV_n) \mQ_{-}^T \mU_n^T \FState^{-T}\FState^{-} \mLambda_\mP \mPhi_\State \bigr)=\mathcal{O}(1)$. From all these upper bounds, and since it can be shown that $\E\bigl[\lVert \mD' \rVert \bigr]=\mathcal{O}\left(\frac{1}{\sqrt{m}}\right)$, we deduce the second term, $Z_{3122}$, vanishes and thus 
    \begin{align*}
        Z_{312}&=\frac{1}{nm^2} \frac{1}{1+\delta} \E\Biggl[\rtrain^T  \mQ_{-}^T \mPsi_1 \mQ_{-} \rtrain \Tr\bigl((\mU_n - \gamma \mV_n) \mQ_{-}^T \mU_n^T \FState^{-T}\FState^{-} \mLambda_\mP \mPhi_\State \bigr)   \Biggr] \\
        &+\mathcal{O}\left(\frac{1}{\sqrt{m}}\right).
    \end{align*}
    Let $\mQ_{-ij} \in \R^{n \times n}$ be the resolvent defined as
    \begin{equation}
        \mQ_{-ij} = \biggl[\frac{1}{m}(\mU_n - \gamma \mV_n)^T \FState^{-ijT} \FState^{-ij} \mU_n + \lambda \mI_n \biggr]^{-1},
        \label{def:Q_-ij}
    \end{equation}
    where $\FState^{-ij}=\sigma(\mW^{-ij}\mS) \in \R^{(N-2) \times n}$ for which $\mW^{-ij} \in \R^{(N-2) \times d}$ depicts the same matrix than the weight matrix $\mW$ defined in \eqref{def:rf_map} without the $i^\text{th}$ and $j^\text{th}$ row.
    Using the Sherman identity (\lemref{lem:sherman}), the term $\frac{1}{m^2}\Tr\bigl((\mU_n - \gamma \mV_n) \mQ_{-}^T \mU_n^T \FState^{-T}\FState^{-} \mLambda_\mP \mPhi_\State \bigr)$ in $Z_{312}$ can be rewritten as
    \begin{align*}
        &\frac{1}{m^2}\Tr\bigl((\mU_n - \gamma \mV_n) \mQ_{-}^T \mU_n^T \FState^{-T}\FState^{-} \mLambda_\mP \mPhi_\State \bigr) \\
        &=\frac{1}{m^2}\Tr\left(\sum_{i \neq j}(\mU_n - \gamma \mV_n) \mQ_{-j}^T \mU_n^T \vsigma_i \vsigma_i^T \mLambda_\mP \mPhi_\State \right) \\
        &=\frac{1}{m^2}\Tr\left(\sum_{i \neq j}\frac{(\mU_n - \gamma \mV_n) \mQ_{-ij}^T \mU_n^T \vsigma_i \vsigma_i^T \mLambda_\mP \mPhi_\State}{1+\frac{1}{m}\vsigma_i^T \mU_n \mQ_{-ij}(\mU_n - \gamma \mV_n)^T \vsigma_i } \right) \\
        &=\frac{1}{m^2}\frac{1}{1+\delta}\Tr\left(\sum_{i \neq j}(\mU_n - \gamma \mV_n) \mQ_{-ij}^T \mU_n^T \vsigma_i \vsigma_i^T \mLambda_\mP \mPhi_\State \right) \\
        &+\frac{1}{m^2}\frac{1}{1+\delta}\Tr\left(\sum_{i \neq j}\frac{(\mU_n - \gamma \mV_n) \mQ_{-ij}^T \mU_n^T \vsigma_i \vsigma_i^T \mLambda_\mP \mPhi_\State \left(\delta -\frac{1}{m}\vsigma_i^T \mU_n \mQ_{-ij}(\mU_n - \gamma \mV_n)^T \vsigma_i\right)}{1+\frac{1}{m}\vsigma_i^T \mU_n \mQ_{-ij}(\mU_n - \gamma \mV_n)^T \vsigma_i } \right) \\
        &=\frac{N}{m^2}\frac{1}{1+\delta}\Tr\left((\mU_n - \gamma \mV_n) \mQ_{--}^T \mU_n^T \mPhi_\State \mLambda_\mP \mPhi_\State \right) \\
        &+\frac{1}{m^2}\frac{1}{1+\delta}\Tr\left((\mU_n - \gamma \mV_n) \mQ_{-j}^T \mU_n^T \FState^{-jT} \mD \FState^{-j} \mLambda_\mP \mPhi_\State \right),
    \end{align*}
    where $\mD \in \R^{N \times N}$ is a diagonal matrix for which, for all $i \in [N]$, we have
    \begin{equation*}
        \mD_i = \delta-\frac{1}{m}\vsigma_i^T \mU_n \mQ_{-ij}(\mU_n - \gamma \mV_n)^T \vsigma_i.
    \end{equation*}
    From the uniform boundness of $\frac{1}{m}\Tr\bigl(\mPhi_\State\bigr)=\frac{1}{K_r}\frac{1}{p}\Tr\bigl(\mPhi_\State\bigr)$, from \lemref{lem:upper_bound_norm_transition_+_resolvent}, we have $\frac{1}{m^2}\Tr\bigl((\mU_n - \gamma \mV_n) \mQ_{-}^T \mU_n^T \FState^{-T}\FState^{-} \mLambda_\mP \mPhi_\State \bigr)=\mathcal{O}(1)$. Since the operator norm of $\E\bigl[\lVert \mD \rVert\bigr]$ is of order $\mathcal{O}\left(\frac{1}{\sqrt{m}}\right)$, we deduce the second term vanishes, and thus 
    \begin{align*}
        &\frac{1}{m^2}\Tr\bigl((\mU_n - \gamma \mV_n) \mQ_{-}^T \mU_n^T \FState^{-T}\FState^{-} \mLambda_\mP \mPhi_\State \bigr) \\
        &=\frac{N}{m^2}\frac{1}{1+\delta}\Tr\left((\mU_n - \gamma \mV_n) \mQ_{--}^T \mU_n^T \mPhi_\State \mLambda_\mP \mPhi_\State \right) +\mathcal{O}\left(\frac{1}{\sqrt{m}}\right).
    \end{align*}
    Applying \lemref{lem:deterministic_E[Q_m^Psi_1Q_m]} and \lemref{lem:decomposition_Q^TPsi_1Q}, we deduce for $Z_{312}$ that
    \begin{align*}
        Z_{312}&=\frac{1}{n}\frac{\frac{1}{N}\Tr\bigl((\mU_n - \gamma \mV_n) \bar \mQ_m^T \mU_n^T \mPsi_\State \mLambda_\mP \mPsi_\State \bigr)}{1-\frac{1}{N} \Tr \left(\mPsi_2 \bar \mQ_m^T \mPsi_1 \bar \mQ_m \right)}\lVert \bar \mQ_m \rtrain \rVert_{\mPsi_1}^2 +\mathcal{O}\left(\frac{1}{\sqrt{m}}\right) \\
        &=\frac{1}{n}\frac{\frac{1}{N}\Tr\bigl(\mPsi_\State (\mU_n - \gamma \mV_n) \mTheta_{\State}^T \mLambda_\mP  \bigr)}{1-\frac{1}{N} \Tr \left(\mPsi_2 \bar \mQ_m^T \mPsi_1 \bar \mQ_m \right)}\lVert \bar \mQ_m \rtrain \rVert_{\mPsi_1}^2 +\mathcal{O}\left(\frac{1}{\sqrt{m}}\right).
    \end{align*}
    Now, we want to find an asymptotic equivalent for $Z_{311}$. We replace $1+\frac{1}{m}\vsigma_j^T \mU_n \mQ_{-j}(\mU_n - \gamma \mV_n)^T \vsigma_j$ by $1+\delta$ in $Z_{311}$ as following
    \begin{align*}
        Z_{311}&=\frac{1}{nm^2} \sum_{i=1}^N \sum_{j \neq i} \E\Biggl[\rtrain^T \frac{\mQ_{-j}^T \mU_n^T \vsigma_i \vsigma_i^T \mLambda_\mP \vsigma_j\vsigma_j^T \mU_n \mQ_{-j}}{1+\frac{1}{m}\vsigma_j^T \mU_n \mQ_{-j}(\mU_n - \gamma \mV_n)^T \vsigma_j}  \rtrain \Biggr] \\
        &=\frac{1}{nm^2} \sum_{j=1}^N \E\Biggl[\rtrain^T \frac{\mQ_{-j}^T \mU_n^T \FState^{-jT} \FState^{-j} \mLambda_\mP \vsigma_j\vsigma_j^T \mU_n \mQ_{-j}}{1+\frac{1}{m}\vsigma_j^T \mU_n \mQ_{-j}(\mU_n - \gamma \mV_n)^T \vsigma_j}  \rtrain \Biggr] \\
        &=\underbrace{\frac{1}{nm^2} \frac{1}{1+\delta} \sum_{j=1}^N \E\Biggl[\rtrain^T \mQ_{-j}^T \mU_n^T \FState^{-jT} \FState^{-j} \mLambda_\mP \vsigma_j\vsigma_j^T \mU_n \mQ_{-j}  \rtrain \Biggr]}_{=Z_{3111}} \\
        &+\underbrace{\frac{1}{nm^2} \frac{1}{1+\delta} \sum_{j=1}^N \E\Biggl[\rtrain^T \frac{\mQ_{-j}^T \mU_n^T \FState^{-jT} \FState^{-j} \mLambda_\mP \vsigma_j\vsigma_j^T \mU_n \mQ_{-j}\mD_j}{1+\frac{1}{m}\vsigma_j^T \mU_n \mQ_{-j}(\mU_n - \gamma \mV_n)^T \vsigma_j}  \rtrain \Biggr]}_{=Z_{3112}},
    \end{align*}
    where $\mD \in \R^{N \times N}$ is a diagonal matrix for which, for all $j \in [N]$, we have
    \begin{equation*}
        \mD_j = \delta-\frac{1}{m}\vsigma_j^T \mU_n \mQ_{-j}(\mU_n - \gamma \mV_n)^T \vsigma_j.
    \end{equation*}
    We observe that 
    \begin{align*}
        \mQ_{-j}^T\mU_n^T\frac{\FState^{-jT}\FState^{-j}}{m}&=\mQ_{-j}^T\mU_n^T\frac{\FState^T\FState}{m}-\mQ_{-j}^T \mU_n^T \frac{\vsigma_j \vsigma_j^T}{m} \\
        &=\mQ_m^T\mU_n^T\frac{\FState^T\FState}{m} + \frac{\frac{1}{m}\mQ_{-j}^T \mU_n^T \vsigma_j \vsigma_j^T (\mU_n-\gamma \mV_n) \mQ_{-j}^T}{1+\frac{1}{m}\vsigma_j^T \mU_n^T \mQ_{-j}(\mU_n - \gamma \mV_n)^T \vsigma_j}\mU_n^T\frac{\FState^T\FState}{m}-\mQ_{-j}^T\mU_n^T\frac{\vsigma_j \vsigma_j^T}{m} \\
        &=\mQ_m^T\mU_n^T\frac{\FState^T\FState}{m} \\
        &\quad + \mQ_m^T \mU_n^T \vsigma_j \vsigma_j^T (\mU_n-\gamma \mV_n) \mQ_m^T \left( 1+\frac{1}{m}\vsigma_j^T \mU_n^T \mQ_{-j}(\mU_n - \gamma \mV_n)^T \vsigma_j \right) \mU_n^T \frac{\FState^T\FState}{m} \\
        &\quad -\mQ_{-j}^T\mU_n^T\frac{\vsigma_j \vsigma_j^T}{m} \\
        &=\mQ_m^T\mU_n^T\frac{\FState^T\FState}{m} + \frac{\mQ_m^T \mU_n^T \vsigma_j \vsigma_j^T (\mU_n-\gamma \mV_n) \mQ_m^T}{1-\frac{1}{m}\vsigma_j^T \mU_n^T \mQ_m(\mU_n - \gamma \mV_n)^T \vsigma_j}\mU_n^T\frac{\FState^T\FState}{m}-\mQ_{-j}^T\mU_n^T\frac{\vsigma_j \vsigma_j^T}{m}.
    \end{align*}
    From above, we expand $Z_{3112}$ as
    \begin{align*}
        &Z_{3112}\\
        &=\frac{1}{nm^2}\frac{1}{1+\delta} \sum_{j=1}^N \E\Biggl[\rtrain^T \frac{\mQ_{-j}^T \mU_n^T \FState^{-jT} \FState^{-j} \mLambda_\mP \vsigma_j\vsigma_j^T \mU_n \mQ_{-j}\mD_j}{1+\frac{1}{m}\vsigma_j^T \mU_n \mQ_{-j}(\mU_n - \gamma \mV_n)^T \vsigma_j}  \rtrain \Biggr] \\
        &=\underbrace{\frac{1}{nm^2}\frac{1}{1+\delta} \E\Biggl[\rtrain^T \mQ_m^T \mU_n^T \FState^T \FState \mLambda_\mP \FState^T \mD \FState \mU_n \mQ_m  \rtrain \Biggr]}_{=Z_{31121}} \\
        &+\underbrace{\sum_{j=1}^N \E\Biggl[\rtrain^T \frac{\mQ_m^T \mU_n^T \vsigma_j \vsigma_j^T (\mU_n-\gamma \mV_n) \mQ_m^T\mU_n^T \FState^T\FState \mLambda_\mP \vsigma_j\vsigma_j^T \mU_n \mQ_m\mD_j}{nm^2(1+\delta)\left(1-\frac{1}{m}\vsigma_j^T \mU_n^T \mQ_m(\mU_n - \gamma \mV_n)^T \vsigma_j\right)}  \rtrain \Biggr]}_{=Z_{31122}} \\
        &- \underbrace{\frac{1}{nm^2}\frac{1}{1+\delta} \sum_{j=1}^N \E\Biggl[\rtrain^T \mQ_{-j}^T \mU_n^T \vsigma_j\vsigma_j^T \mLambda_\mP \vsigma_j\vsigma_j^T \mU_n \mQ_m\mD_j\left(1+\frac{1}{m}\vsigma_j^T \mU_n \mQ_{-j}(\mU_n - \gamma \mV_n)^T \vsigma_j\right)  \rtrain \Biggr]}_{=Z_{31123}}.
    \end{align*}
    We have $Z_{31121} = \mathcal{O}\left(\frac{1}{\sqrt{m}}\right)$ since $\E\left[\lVert \mD \rVert\right]=\mathcal{O}\left(\frac{1}{\sqrt{m}}\right)$ and from \lemref{lem:upper_bound_norm_transition_+_resolvent}. Subsequently, we rewrite $Z_{31122}$ as 
    \begin{align*}
        Z_{31122} = \frac{1}{nm}\frac{1}{1+\delta}\E\left[\rtrain^T \mQ_m^T \mU_n^T \FState^T \mD_{31122} \FState \mU_n \mQ_m \rtrain\right],
    \end{align*}
    with $\mD_{31122} \in \R^{N \times N}$ a diagonal matrix for which, for all $j \in [N]$, we have
    \begin{equation*}
        [\mD_{31122}]_j = \frac{1}{m}\frac{\mD_j\vsigma_j^T (\mU_n-\gamma \mV_n) \mQ_m^T\mU_n^T \FState^T\FState \mLambda_\mP \vsigma_j}{1-\frac{1}{m}\vsigma_j^T \mU_n^T \mQ_m(\mU_n - \gamma \mV_n)^T \vsigma_j}.
    \end{equation*}
    It can be shown that $\E\left[\lVert \mD_{31122} \rVert\right]=\mathcal{O}\left(\frac{1}{\sqrt{m}}\right)$, and we can deduce $Z_{31122} = \mathcal{O}\left(\frac{1}{\sqrt{m}}\right)$. Similarly, we have $Z_{31123} = \mathcal{O}\left(\frac{1}{\sqrt{m}}\right)$. $Z_{3112}$ vanishes, and thus
    \begin{align*}
        Z_{311} &= Z_{3111} + \mathcal{O}\left(\frac{1}{\sqrt{m}}\right) \\
        &=\frac{1}{nm^2} \frac{1}{1+\delta} \sum_{j=1}^N \E\Biggl[\rtrain^T \mQ_{-j}^T \mU_n^T \FState^{-jT} \FState^{-j} \mLambda_\mP \vsigma_j\vsigma_j^T \mU_n \mQ_{-j}  \rtrain \Biggr] + \mathcal{O}\left(\frac{1}{\sqrt{m}}\right).
    \end{align*}
    It remains to handle $Z_{3111}$ for which we have from the Sherman identity (\lemref{lem:sherman}),
    \begin{align*}
        Z_{3111}&=\frac{1}{nm^2} \frac{1}{1+\delta} \sum_{j=1}^N \E\Biggl[\rtrain^T \mQ_{-j}^T \mU_n^T \FState^{-jT} \FState^{-j} \mLambda_\mP \vsigma_j\vsigma_j^T \mU_n \mQ_{-j}  \rtrain \Biggr] \\
        &=\frac{1}{nm^2} \frac{1}{1+\delta} \sum_{j=1}^N \sum_{i \neq j} \E\Biggl[\rtrain^T \mQ_{-j}^T \mU_n^T \vsigma_i \vsigma_i^T \mLambda_\mP \mPhi_\State \mU_n \mQ_{-j}  \rtrain \Biggr] \\
        &=\underbrace{\frac{1}{nm^2} \frac{1}{1+\delta} \sum_{j=1}^N \sum_{i \neq j} \E\Biggl[\rtrain^T \frac{\mQ_{-ij}^T \mU_n^T \vsigma_i \vsigma_i^T\mLambda_\mP \mPhi_\State \mU_n \mQ_{-ij}}{1+\frac{1}{m}\vsigma_i^T \mU_n \mQ_{-ij}(\mU_n - \gamma \mV_n)^T \vsigma_i}  \rtrain \Biggr]}_{=Z_{31111}} \\ 
        &-\underbrace{\frac{1}{nm^3} \frac{1}{1+\delta} \sum_{j=1}^N \sum_{i \neq j} \E\Biggl[\rtrain^T \frac{\mQ_{-ij}^T \mU_n^T \vsigma_i \vsigma_i^T\mLambda_\mP \mPhi_\State \mU_n \mQ_{-ij} (\mU_n - \gamma \mV_n)^T \vsigma_i \vsigma_i^T \mU_n \mQ_{-ij}}{\left(1+\frac{1}{m}\vsigma_i^T \mU_n \mQ_{-ij}(\mU_n - \gamma \mV_n)^T \vsigma_i\right)^2} \rtrain \Biggr]}_{=Z_{31112}}.
    \end{align*}
    Again, we replace $1+\frac{1}{m}\vsigma_i^T \mU_n \mQ_{-ij}(\mU_n - \gamma \mV_n)^T \vsigma_i$ by $1+\delta$ in $Z_{31111}$ as following
    \begin{align*}
        Z_{31111} &= \frac{1}{nm^2} \frac{1}{1+\delta} \sum_{j=1}^N \sum_{i \neq j} \E\Biggl[\rtrain^T \frac{\mQ_{-ij}^T \mU_n^T \vsigma_i \vsigma_i^T\mLambda_\mP \mPhi_\State \mU_n \mQ_{-ij}}{1+\frac{1}{m}\vsigma_i^T \mU_n \mQ_{-ij}(\mU_n - \gamma \mV_n)^T \vsigma_i}  \rtrain \Biggr] \\
        &= \frac{1}{nm^2} \frac{1}{(1+\delta)^2} \sum_{j=1}^N \sum_{i \neq j} \E\Biggl[\rtrain^T \mQ_{-ij}^T \mU_n^T \vsigma_i \vsigma_i^T\mLambda_\mP \mPhi_\State \mU_n \mQ_{-ij} \rtrain \Biggr] \\
        &+\frac{1}{nm^2} \frac{1}{(1+\delta)^2} \sum_{j=1}^N \sum_{i \neq j} \E\Biggl[\rtrain^T \frac{\mQ_{-ij}^T \mU_n^T \vsigma_i \vsigma_i^T\mLambda_\mP \mPhi_\State \mU_n \mQ_{-ij}\mD_i}{1+\frac{1}{m}\vsigma_i^T \mU_n \mQ_{-ij}(\mU_n - \gamma \mV_n)^T \vsigma_i}  \rtrain \Biggr] \\
        &=\frac{1}{n}\frac{N^2}{m^2} \frac{1}{(1+\delta)^2} \E\Biggl[\rtrain^T \mQ_{--}^T \mU_n^T \mPhi_\State \mLambda_\mP \mPhi_\State \mU_n \mQ_{--} \rtrain \Biggr] \\
        &+\frac{1}{nm^2} \frac{1}{(1+\delta)^2} \sum_{j=1}^N \sum_{i \neq j} \E\Biggl[\rtrain^T \mQ_{-j}^T \mU_n^T \vsigma_i \vsigma_i^T\mLambda_\mP \mPhi_\State \mU_n \mQ_{-j}\mD_i \rtrain \Biggr] \\
        &+\frac{1}{nm^3} \frac{1}{(1+\delta)^2} \sum_{j=1}^N \sum_{i \neq j} \E\Biggl[\rtrain^T \frac{\mQ_{-j}^T \mU_n^T \vsigma_i \vsigma_i^T\mLambda_\mP \mPhi_\State \mU_n \mQ_{-j} (\mU_n - \gamma \mV_n)^T \vsigma_i \vsigma_i^T \mU_n \mQ_{-j}}{1-\frac{1}{m}\vsigma_i^T \mU_n \mQ_{-j}(\mU_n - \gamma \mV_n)^T \vsigma_i} \mD_i \rtrain \Biggr] \\
        &=\frac{1}{n} \E\Biggl[\rtrain^T \mQ_{--}^T \mU_n^T \mPsi_\State \mLambda_\mP \mPsi_\State \mU_n \mQ_{--} \rtrain \Biggr] \\
        &+\frac{1}{nm^2} \frac{1}{(1+\delta)^2} \sum_{j=1}^N\E\Biggl[\rtrain^T \mQ_{-j}^T \mU_n^T \FState^{-jT} \mD\FState^{-j}\mLambda_\mP \mPhi_\State \mU_n \mQ_{-j} \rtrain \Biggr] \\
        &+\frac{1}{nm} \frac{1}{(1+\delta)^2} \sum_{j=1}^N \E\Biggl[\rtrain^T \mQ_{-j}^T \mU_n^T \FState^{-jT} \mD_{31111} \FState^{-j} \mU_n \mQ_{-j}  \rtrain \Biggr],
    \end{align*}
    where $\mD_{31111} \in \R^{N \times N}$ is a diagonal matrix for which, for all $i \in [N]$, we have
    \begin{equation*}
        [\mD_{31111}]_i = \frac{1}{m} \frac{\vsigma_i^T\mLambda_\mP \mPhi_\State \mU_n \mQ_{-j} (\mU_n - \gamma \mV_n)^T \vsigma_i\left(\delta - \frac{1}{m}\vsigma_i^T \mU_n \mQ_{-ij}(\mU_n - \gamma \mV_n)^T \vsigma_i\right)}{1-\frac{1}{m}\vsigma_i^T \mU_n \mQ_{-j}(\mU_n - \gamma \mV_n)^T \vsigma_i},
    \end{equation*}
    and $\mQ_{--}$ is a resolvent with the same law than $\mQ_{-ij}$. With similar arguments that before, we can show that $\E\bigl[\lVert \mD \rVert\bigr]$ and $\E\bigl[\lVert \mD_{31111} \rVert\bigr]$ are of order $ \mathcal{O}\left(\frac{1}{\sqrt{m}}\right)$, and therefore
    \begin{equation*}
        Z_{31111} = \frac{1}{n} \E\Biggl[\rtrain^T \mQ_{--}^T \mU_n^T \mPsi_\State \mLambda_\mP \mPsi_\State \mU_n \mQ_{--} \rtrain \Biggr] + \mathcal{O}\left(\frac{1}{\sqrt{m}}\right).
    \end{equation*}
    By extending \lemref{lem:deterministic_Q^TMQ} to the matrix $\mLambda_\mP'=\mU_n^T \mPsi_\State \mLambda_\mP \mPsi_\State \mU_n$, and from \lemref{lem:deterministic_E[Q_m^Psi_1Q_m]} we obtain
    \begin{align*}
        Z_{31111} &= \frac{1}{n} \rtrain^T \bar \mQ_m^T \mU_n^T \mPsi_\State \mLambda_\mP \mPsi_\State \mU_n \bar \mQ_m \rtrain  \\
        &+\frac{1}{n} \frac{\frac{1}{N} \Tr \left(\mLambda_\mP \mTheta_\State \mPsi_2 \mTheta_\State^T \right)}{1-\frac{1}{N} \Tr \left(\mPsi_2 \bar \mQ_m^T  \mPsi_1 \bar \mQ_m  \right)} \lVert \bar \mQ_m \rtrain \rVert_{\mPsi_1}^2 \\
        &+ \mathcal{O}\left(\frac{1}{\sqrt{m}}\right).
    \end{align*}
    Following the same reasoning for $Z_{31112}$, and from \lemref{lem:normal_concentration_resolvent}, we have
    \begin{align*}
        Z_{31112}&=\frac{1}{nm^3} \frac{1}{1+\delta} \sum_{j=1}^N \sum_{i \neq j} \E\Biggl[\rtrain^T \frac{\mQ_{-ij}^T \mU_n^T \vsigma_i \vsigma_i^T\mLambda_\mP \mPhi_\State \mU_n \mQ_{-ij} (\mU_n - \gamma \mV_n)^T \vsigma_i \vsigma_i^T \mU_n \mQ_{-ij}}{\left(1+\frac{1}{m}\vsigma_i^T \mU_n \mQ_{-ij}(\mU_n - \gamma \mV_n)^T \vsigma_i\right)^2} \rtrain \Biggr] \\
        &=\frac{1}{nm^3} \frac{1}{(1+\delta)^3} \sum_{j=1}^N \sum_{i \neq j} \E\Biggl[\rtrain^T \mQ_{-ij}^T \mU_n^T \vsigma_i \vsigma_i^T\mLambda_\mP \mPhi_\State \mU_n \mQ_{-ij} (\mU_n - \gamma \mV_n)^T \vsigma_i \vsigma_i^T \mU_n \mQ_{-ij} \rtrain \Biggr] \\
        &\quad + \mathcal{O}\left(\frac{1}{\sqrt{m}}\right) \\
        &=\frac{1}{nm^2} \frac{1}{(1+\delta)^3} \sum_{j=1}^N \sum_{i \neq j} \E\Biggl[\rtrain^T \mQ_{-ij}^T \mU_n^T \vsigma_i \vsigma_i^T \mU_n \mQ_{-ij} \rtrain\left(\frac{1}{m}\vsigma_i^T\mLambda_\mP \mPhi_\State \mU_n \mQ_{-ij} (\mU_n - \gamma \mV_n)^T \vsigma_i\right) \Biggr] \\
        &\quad + \mathcal{O}\left(\frac{1}{\sqrt{m}}\right) \\
        &=\frac{1}{nm^2} \frac{1}{(1+\delta)^3} \sum_{j=1}^N \sum_{i \neq j} \E\Biggl[\rtrain^T \mQ_{-ij}^T \mU_n^T \vsigma_i \vsigma_i^T \mU_n \mQ_{-ij} \rtrain \frac{1}{m}\Tr(\mLambda_\mP \mPhi_\State \mU_n \mQ_{-ij} (\mU_n - \gamma \mV_n)^T \mPhi_\State) \Biggr] \\
        &\quad + \mathcal{O}\left(\frac{1}{\sqrt{m}}\right) \\
        &=\frac{1}{n}\frac{1}{N}  \E\Biggl[\rtrain^T \mQ_{--}^T \mPsi_1 \mQ_{--} \rtrain \Tr(\mLambda_\mP \mPsi_\State \mU_n \mQ_{--} (\mU_n - \gamma \mV_n)^T \mPsi_\State) \Biggr] \\
        &\quad + \mathcal{O}\left(\frac{1}{\sqrt{m}}\right),
    \end{align*}
    where the last equality is obtained with a similar reasoning than for \eqref{eq:Z_312_trace}. From \lemref{lem:deterministic_E[Q_m^Psi_1Q_m]} and \lemref{lem:decomposition_Q^TPsi_1Q}, we have
    \begin{align*}
        Z_{31112}=\frac{1}{n}\frac{\frac{1}{N}\Tr(\mLambda_\mP \mTheta_\State (\mU_n - \gamma \mV_n)^T \mPsi_\State)}{1-\frac{1}{N} \Tr \left(\mPsi_2 \bar \mQ_m^T \mPsi_1 \bar \mQ_m \right)}\lVert \bar \mQ_m \rtrain \rVert_{\mPsi_1}^2 + \mathcal{O}\left(\frac{1}{\sqrt{m}}\right).
    \end{align*}
    We conclude for $Z_{311}$ that 
    \begin{align*}
        Z_{311} &= \frac{1}{n} \rtrain^T \bar \mQ_m^T \mU_n^T \mPsi_\State \mLambda_\mP \mPsi_\State \mU_n \bar \mQ_m \rtrain  \\
        &+\frac{1}{n} \frac{\frac{1}{N} \Tr \left(\mLambda_\mP \bigl[\mTheta_\State \mPsi_2 \mTheta_\State^T - \mTheta_\State (\mU_n - \gamma \mV_n)^T \mPsi_\State \bigr] \right)}{1-\frac{1}{N} \Tr \left(\mPsi_2 \bar \mQ_m^T  \mPsi_1 \bar \mQ_m  \right)} \lVert \bar \mQ_m \rtrain \rVert_{\mPsi_1}^2 \\
        &+ \mathcal{O}\left(\frac{1}{\sqrt{m}}\right),
    \end{align*}
    and for $Z_{31}$ that 
    \begin{align*}
        Z_{31} &= \frac{1}{n} \rtrain^T \bar \mQ_m^T \mU_n^T \mPsi_\State \mLambda_\mP \mPsi_\State \mU_n \bar \mQ_m \rtrain  \\
        &+\frac{1}{n} \frac{\frac{1}{N} \Tr \left(\mLambda_\mP \bigl[\mTheta_\State \mPsi_2 \mTheta_\State^T - 2\mTheta_\State (\mU_n - \gamma \mV_n)^T \mPsi_\State \bigr] \right)}{1-\frac{1}{N} \Tr \left(\mPsi_2 \bar \mQ_m^T  \mPsi_1 \bar \mQ_m  \right)} \lVert \bar \mQ_m \rtrain \rVert_{\mPsi_1}^2 \\
        &+ \mathcal{O}\left(\frac{1}{\sqrt{m}}\right).
    \end{align*}
\end{proof}
\begin{lemma}
    Under Assumptions~\ref{assumption:growth_rate},~\ref{assumption:regime_n}~and~\ref{assumption:new_regime_n}, let $Z_{32} \in \R$ defined in \eqref{eq:decomposition_Z_3_theorem_3} as
    \begin{equation*}
        Z_{32} = \sum_{i=1}^N \E\Biggl[\frac{1}{nm^2}\rtrain^T \frac{\mQ_{-i}^T \mU_n^T \vsigma_i \vsigma_i^T \mLambda_\mP \vsigma_i\vsigma_i^T \mU_n \mQ_{-i}}{\left(1+\frac{1}{m}\vsigma_i^T \mU_n \mQ_{-i}(\mU_n - \gamma \mV_n)^T \vsigma_i \right)^2} \rtrain \Biggr].
    \end{equation*}
    Then
    \begin{equation*}
        \Biggl\lvert Z_{32}-\frac{1}{n}\frac{\frac{1}{N}\Tr(\mLambda_\mP \mPsi_\State)}{1-\frac{1}{N}\Tr \left(\mPsi_2 \bar \mQ_m^T \mPsi_1 \bar \mQ_m \right)} \lVert \bar \mQ_m \rtrain \rVert_{\mPsi_1}^2 \Biggr\rvert = \mathcal{O}\left(\frac{1}{\sqrt{m}}\right),
    \end{equation*}
    where $\bar \mQ_m$ is the deterministic resolent defined in \eqref{def:Q_bar}, $\mPsi_1, \mPsi_2 \in \R^{n \times n}$ are defined in \eqref{def:Psi_1_and_Psi_2}, and $\mPsi_\State \in \R^{p \times p}$ is defined in \eqref{def:Psi_State_Lambda_P_Theta_State}.
    \label{lem:Z32_theorem_3}
\end{lemma}
\begin{proof}
    We decompose $Z_{32}$ as 
    \begin{align*}
        Z_{32} &= \frac{1}{nm^2} \sum_{i=1}^N \E\Biggl[\rtrain^T \frac{\mQ_{-i}^T \mU_n^T \vsigma_i \vsigma_i^T \mLambda_\mP \vsigma_i\vsigma_i^T \mU_n \mQ_{-i}}{\left(1+\frac{1}{m}\vsigma_i^T \mU_n \mQ_{-i}(\mU_n - \gamma \mV_n)^T \vsigma_i \right)^2} \rtrain \Biggr] \\
        &= \frac{1}{nm} \sum_{i=1}^N \E\Biggl[\rtrain^T \frac{\mQ_{-i}^T \mU_n^T \vsigma_i \vsigma_i^T \mU_n \mQ_{-i}\frac{1}{m}\Tr\bigl(\mPhi_\State \mLambda_\mP \bigr)}{\left(1+\frac{1}{m}\vsigma_i^T \mU_n \mQ_{-i}(\mU_n - \gamma \mV_n)^T \vsigma_i \right)^2} \rtrain \Biggr] \\
        &\quad + \frac{1}{nm} \sum_{i=1}^N \E\Biggl[\rtrain^T \frac{\mQ_{-i}^T \mU_n^T \vsigma_i \vsigma_i^T \mU_n \mQ_{-i}\frac{1}{m} \bigl(\vsigma_i^T \mLambda_\mP \vsigma_i - \Tr\bigl(\mPhi_\State \mLambda_\mP \bigr) \bigr)}{\left(1+\frac{1}{m}\vsigma_i^T \mU_n \mQ_{-i}(\mU_n - \gamma \mV_n)^T \vsigma_i \right)^2} \rtrain  \Biggr] \\
        &= \frac{1}{n} \frac{\Tr\bigl(\mPhi_\State \mLambda_\mP \bigr)}{m} \rtrain^T \underbrace{\sum_{i=1}^N \E\Biggl[ \frac{1}{m} \frac{\mQ_{-i}^T \mU_n^T \vsigma_i \vsigma_i^T \mU_n \mQ_{-i}}{\left(1+\frac{1}{m}\vsigma_i^T \mU_n \mQ_{-i}(\mU_n - \gamma \mV_n)^T \vsigma_i \right)^2} \Biggr]}_{=\mZ_{321}}\rtrain \\
        &\quad + \frac{1}{n} \rtrain^T \underbrace{\sum_{i=1}^N\E\Biggl[\frac{1}{m}\mQ_m^T \mU_n^T \vsigma_i \vsigma_i^T \mU_n \mQ_m \frac{1}{m} \bigl(\vsigma_i^T \mLambda_\mP \vsigma_i - \Tr\bigl(\mPhi_\State \mLambda_\mP \bigr) \bigr) \Biggr]}_{\mZ_{322}} \rtrain.
    \end{align*}
    We want to show $\mZ_{322}$ vanishes and find an asymptotic equivalent for $\mZ_{321}$. Let $\mD_{322} \in \R^{N \times N}$ be a diagonal matrix for which, for all $i \in [N]$, we have 
    \begin{equation*}
        \bigl[\mD_{322}\bigr]_i = \frac{1}{m}\vsigma_i^T \mLambda_\mP \vsigma_i - \frac{1}{m}\Tr\bigl(\mPhi_\State \mLambda_\mP \bigr).
    \end{equation*}
    We rewrite $\mZ_{322}$ as 
    \begin{align*}
        \mZ_{322}&=\sum_{i=1}^N\E\Biggl[\frac{1}{m}\mQ_m^T \mU_n^T \vsigma_i \vsigma_i^T \mU_n \mQ_m \frac{1}{m} \bigl(\vsigma_i^T \mLambda_\mP \vsigma_i - \Tr\bigl(\mPhi_\State \mLambda_\mP \bigr) \bigr) \Biggr]\\
        &=\E\Biggl[\frac{1}{m}\mQ_m^T \mU_n^T \FState^T \mD_{322} \FState \mU_n \mQ_m  \Biggr]
    \end{align*}
    From \lemref{lem:upper_bound_norm_resolvent_+_RF_matrix}, we know there exists a real $K_\mQ'>0$ such that, for all $m$, we have 
    \begin{equation*}
        \left\lVert \frac{1}{\sqrt{m}} \FState \mU_n \mQ_m \right\rVert \leq K_\mQ'
    \end{equation*}
    and
    \begin{equation*}
        \left\lVert \frac{1}{\sqrt{m}} \mQ_m (\mU_n - \gamma \mV_n)^T \FState^T \right\rVert \leq 2 K_\mQ'.
    \end{equation*}
    Using \lemref{lem:normal_concentration_resolvent} we show that $\E\bigl[\lVert \mD_{322} \rVert \bigr]=\mathcal{O}\left(\frac{1}{\sqrt{m}}\right)$, and we deduce that 
    \begin{equation*}
        \lVert \mZ_{322} \rVert = \mathcal{O}\left(\frac{1}{\sqrt{m}}\right).
    \end{equation*}
    We want to find an asymptotic equivalent for $\mZ_{321}$. Let $\mD_{321} \in \R^{N \times N}$ be a diagonal matrix for which, for all $i \in [N]$, we have 
    \begin{equation*}
        \bigl[\mD_{321}\bigr]_i = (1+\delta)^2 - \left(1+\frac{1}{m}\vsigma_i^T \mU_n \mQ_{-i}(\mU_n - \gamma \mV_n)^T \vsigma_i \right)^2.
    \end{equation*}
    We replace $1+\frac{1}{m}\vsigma_i^T \mU_n \mQ_{-i}(\mU_n - \gamma \mV_n)^T \vsigma_i$ by $1+\delta$ as following
    \begin{align*}
        \mZ_{321} &= \sum_{i=1}^N \E\Biggl[ \frac{1}{m} \frac{\mQ_{-i}^T \mU_n^T \vsigma_i \vsigma_i^T \mU_n \mQ_{-i}}{\left(1+\frac{1}{m}\vsigma_i^T \mU_n \mQ_{-i}(\mU_n - \gamma \mV_n)^T \vsigma_i \right)^2} \Biggr] \\
        &= \frac{1}{(1+\delta)^2}\sum_{i=1}^N \E\Biggl[ \frac{1}{m} \mQ_{-i}^T \mU_n^T \vsigma_i \vsigma_i^T \mU_n \mQ_{-i} \Biggr] \\
        &+ \frac{1}{(1+\delta)^2}\sum_{i=1}^N \E\Biggl[ \frac{1}{m} \mQ_m^T \mU_n^T \vsigma_i \vsigma_i^T \mU_n \mQ_m \left((1+\delta)^2 - \left(1+\frac{1}{m}\vsigma_i^T \mU_n \mQ_{-i}(\mU_n - \gamma \mV_n)^T \vsigma_i \right)^2 \right) \Biggr] \\
        &=\frac{N}{m}\frac{1}{(1+\delta)^2} \E\bigl[\mQ_{-}^T \mU_n^T \mPhi_\State \mU_n \mQ_{-} \bigr] + \frac{1}{(1+\delta)^2}\E\biggl[\frac{1}{m}\mQ_m \mU_n^T \FState^T \mD_{321} \FState \mU_n \mQ_m \biggr] \\
        &=\frac{1}{1+\delta} \E\bigl[\mQ_{-}^T \mPsi_1 \mQ_{-} \bigr] + \mathcal{O_{\lVert \cdot \rVert}}\left(\frac{1}{\sqrt{m}}\right).
    \end{align*}
    The last equality is obtained since we can show that $\E\bigl[\lVert \mD_{321}\rVert\bigr]=\mathcal{O}\left(\frac{1}{\sqrt{m}}\right)$. We have from \lemref{lem:deterministic_E[Q_m^Psi_1Q_m]} 
    \begin{equation*}
        \left\lVert \E\left[\mQ_m^T \mPsi_1 \mQ_m \right]-\E\left[\mQ_{-}^T \mPsi_1 \mQ_{-}\right] \right\rVert = \mathcal{O}\left(\frac{1}{\sqrt{m}}\right),
    \end{equation*}
    and from \lemref{lem:decomposition_Q^TPsi_1Q}
    \begin{align*}
        \mZ_{321}=\frac{1}{1+\delta}\E\bigl[\mQ_m^T \mPsi_1 \mQ_m \bigr] = \frac{1}{1+\delta}\frac{1}{1-\frac{1}{N} \Tr \left(\mPsi_2 \bar \mQ_m^T \mPsi_1 \bar \mQ_m \right)}\bar \mQ_m^T \mPsi_1 \bar \mQ_m + \mathcal{O_{\lVert \cdot \rVert}}\left(\frac{1}{\sqrt{m}}\right).
    \end{align*}
    We conclude that 
    \begin{equation*}
        Z_{32}=\frac{1}{n}\frac{\frac{1}{N}\Tr(\mLambda_\mP \mPsi_\State )}{1-\frac{1}{N} \Tr \left(\mPsi_2 \bar \mQ_m^T \mPsi_1 \bar \mQ_m \right)}\lVert \bar \mQ_m \rtrain \rVert_{\mPsi_1}^2 + \mathcal{O}\left(\frac{1}{\sqrt{m}}\right).
    \end{equation*}
\end{proof}
\begin{lemma}
    When all states have been visited, the empirical transition model matrix $\hat \mA_m = \hat \mU_n(\hat \mU_n - \gamma \hat \mV_n)=\mU_n(\mU_n - \gamma \mV_n)$ defined in \eqref{def:empirical_transition_model_matrix} is invertible.
    \label{lem:A_invertible_all_states}
\end{lemma}
\begin{proof}
    Let $c: \State \to \mathbb{N}$ and $c': \State \to \mathbb{N}$ be defined such that, for all $i \in [p]$, $c(\mS_i)$ and $c'(\mS_i)$ represent the number of times $\mS_i$ occurs in $\Xtrain$ and $\Xtrain'$, respectively. 
    If all states have been visited ($m=p$), for all i in $[p]$, we have thus $c(\mS_i)>0$.
    The structure of $\sqrt{n}\mU_n \in \R^{m \times n}$ indicates each column $i$ of $\mU_n$ is a one-hot vector, where its $j$-th element is $1$ if the $i$-th state $\vs_i$ of $\Xtrain$ is $\mS_j$. 
    Conversely, each row $i$ of $\sqrt{n}\mU_n$ has a $j$-th element is one if the $j$-th state $\vs_j$ of $\Xtrain$ is $\mS_i$.
    A similar correspondence holds for $\sqrt{n}\mV_n$ and $\Xtrain'$. From interpretations of $\mU_n$ and $\mV_n$, we deduce $n\mU_n\mU_n^T \in \R^{m \times m}$ and $n\mV_n\mV_n^T \in \R^{m \times m}$ are diagonal matrices where the $i$-th element of its diagonal are $c(\mS_i)$ and $c'(\mS_i)$, respectively. 
    In the same way, $n\mU_n\mV_n^T \in \R^{m \times m}$ is matrix for which  $[n\mU_n\mV_n^T]_{ij}$ is $c(\mS_i \to \mS_j)$ which represents the number of times the state $\mS_i$ follows $\mS_j$ in $\mathcal{D}_\text{train}$. We are going to prove $\hat \mA_m$ is invertible by using the Gershgorin circle theorem to show $\hat \mA_m$ is strictly diagonally dominant, i.e., $\lvert [\hat \mA_m]_{ii} \rvert > \sum_{i \neq j}\lvert [\hat \mA_m]_{ij} \rvert$. From the interpretations of $\mU_n\mU_n^T$ and $\mU_n\mV_n^T$, we have
    \begin{equation*}
        [\hat \mA_m]_{ii}=[\mU_n\mU_n^T]_{ii} - \gamma [\mU_n\mV_n^T]_{ii} = \frac{c(\mS_i)-\gamma c(\mS_i \to \mS_i)}{n} > 0, \qquad \forall i \in [n],
    \end{equation*}
    and
    \begin{equation*}
        [\hat \mA_m]_{ij}=-\gamma [\mU_n\mV_n^T]_{ij} = \frac{-\gamma c(\mS_i \to \mS_j)}{n} < 0, \qquad \forall i \neq j.
    \end{equation*}
    To prove $\hat \mA_m$ is invertible it remains to show $\sum_j[\hat \mA_m]_{ij} = \sum_j [\mU_n(\mU_n - \gamma \mV_n)^T]_{ij} > 0$ for all $i \in [m]$.
    Let $i \in [m]$, we have 
    \begin{equation*}
        \sum_j [\mU_n(\mU_n - \gamma \mV_n)^T]_{ij} = \frac{c(\mS_i)}{n} - \gamma \sum_j \frac{c(\mS_i \to \mS_j)}{n} = (1-\gamma)\frac{c(\mS_i)}{n} > 0,
    \end{equation*}
    which concludes the proof.
\end{proof}
\begin{lemma}
    Let $\Delta$ be the second-order correction factor of $\DMSBE$ defined in \eqref{def:Delta}. If all states have been visited then
    \begin{equation*}
        \Delta = \frac{\lambda^2}{n} \frac{\frac{1}{N} \Tr \Bigl(\mU_n^T \hat \mA_m^{-1T} \mLambda_\mP \hat \mA_m^{-1} \mU_n \bar \mQ_m \mPsi_2 \bar \mQ_m^T \Bigr)}{1-\tfrac{1}{N} \Tr \left( \mPsi_2 \bar \mQ_m(\lambda)^T \mPsi_1 \bar \mQ_m(\lambda)  \right)}\lVert \bar \mQ_m(\lambda) \rtrain \rVert_{\mPsi_1}^2,
    \end{equation*}
    where $\hat \mA_m = \hat \mU_n(\hat \mU_n - \gamma \hat \mV_n)=\mU_n(\mU_n - \gamma \mV_n)$ is the empirical transition model matrix defined in \eqref{def:empirical_transition_model_matrix}.
    \label{lem:DMSBE_A_invertible}
\end{lemma}
\begin{proof}
    When all states have been visited, we have $\mU_n = \hat \mU_n$, $\mV_n = \hat \mV_n$ and $\FState = \FSVisited$. Furthermore, from \lemref{lem:A_invertible_all_states}, $\hat \mA_m = \hat \mU_n(\hat \mU_n - \gamma \hat \mV_n)=\mU_n(\mU_n - \gamma \mV_n)$ is invertible. We write
    \begin{align*}
        \mTheta_\State = \mPsi_\State \mU_n \bar \mQ_m(\lambda) &= \hat \mA_m^{-1} \mU_n (\mU_n - \gamma \mV_n)^T\mPsi_\State \mU_n \bar \mQ_m(\lambda) \\
        &= \hat \mA_m^{-1} \mU_n \bigl[\mI_n - \lambda \bar \mQ_m \bigr]
    \end{align*}
    Using the equality above and the cyclic properties of the trace we conclude that
    \begin{align*}
        &\Tr \Bigl(\mLambda_\mP \bigl[ \mTheta_\State \mPsi_2 \mTheta_\State^T - 2\mTheta_\State (\mU_n - \gamma \mV_n)^T \mPsi_\State + \mPsi_\State \bigl] \Bigr) \\
        &= \Tr \Bigl(\mLambda_\mP \bigl[\hat \mA_m^{-1} \mU_n \bigl[\mI_n - \lambda \bar \mQ_m \bigr] \mPsi_2 \bigl[\mI_n - \lambda \bar \mQ_m \bigr]^T \mU_n^T \hat \mA_m^{-1T} \\ 
        &\qquad - 2\hat \mA_m^{-1} \mU_n \bigl[\mI_n - \lambda \bar \mQ_m \bigr] (\mU_n - \gamma \mV_n)^T \mPsi_\State + \mPsi_\State \bigl] \Bigr) \\
        &= \lambda^2 \Tr \Bigl(\mLambda_\mP \hat \mA_m^{-1} \mU_n \bar \mQ_m \mPsi_2 \bar \mQ_m^T \mU_n^T \hat \mA_m^{-1T} \Bigr) -\lambda \Tr\Bigl(\mLambda_\mP \hat \mA_m^{-1} \mU_n \bar \mQ_m (\mU_n-\gamma \mV_n)^T \mPsi_\State \Bigr) \\
        & \qquad -\lambda \Tr \Bigl(\mLambda_\mP \mPsi_\State (\mU_n-\gamma \mV_n) \mQ_m^T \mU_n^T \hat \mA_m^{-1T} \Bigr) + 2 \lambda \Tr \Bigl( \mLambda_\mP \hat \mA_m^{-1} \mU_n  \bar \mQ_m (\mU_n - \gamma \mV_n)^T \mPsi_\State \Bigr) \\
        &= \lambda^2 \Tr \Bigl(\mLambda_\mP \hat \mA_m^{-1} \mU_n \bar \mQ_m \mPsi_2 \bar \mQ_m^T \mU_n^T \hat \mA_m^{-1T} \Bigr) \\
        &= \lambda^2 \Tr \Bigl(\mU_n^T \hat \mA_m^{-1T} \mLambda_\mP \hat \mA_m^{-1} \mU_n \bar \mQ_m \mPsi_2 \bar \mQ_m^T  \Bigr).
    \end{align*}
\end{proof}
\section{Existence of the Resolvent \texorpdfstring{$\mQ_m(\lambda)$}{Q_m(lambda)} \label{sec:existence_resolvent}}
In this section, we show that Assumption~\ref{assumption:regime_n} guarantees the existence of the resolvent $\mQ_m(\lambda)$ (\lemref{lem:existence_resolvent}), but also that Assumption~\ref{assumption:regime_n} may be true in practice under certain conditions (\lemref{lem:condition_H_A_positive_definite}). 
\begin{lemma}
     Under Assumption~\ref{assumption:regime_n}, for any $\lambda > 0$, the resolvent $\mQ_m(\lambda)$ defined in \eqref{def:resolvent} exists.
     \label{lem:existence_resolvent}
\end{lemma}
\begin{proof}
    From Assumption~\ref{assumption:regime_n}, we know that $\nu_{\min}\bigl(H(\hat \mA_m)\bigr)>\xi_{\min}>0$, and thus $H(\FSVisited \hat \mA_m \FSVisited^T)$ is at least semi-positive-definite. From the Min-Max theorem, we deduce that the eigenvalues of $\FSVisited \hat \mA_m \FSVisited^T$  have nonnegative real-parts. 
    Consequently, the eigenvalues of $\frac{1}{m}(\hat \mU_n - \gamma \hat \mV_n)^T \FSVisited^T \FSVisited \hat \mU_n$ have nonnegative real parts since both $\frac{1}{m}(\hat \mU_n - \gamma \hat \mV_n)^T \FSVisited^T \FSVisited \hat \mU_n$ and $\FSVisited \hat \mA_m \FSVisited^T$ share the same nonzero eigenvalues from the Weinstein–Aronszajn identity (\lemref{lem:weinstein}). 
\end{proof}
\begin{lemma}
    Let $c: \SVisited \to \mathbb{N}$ and $c': \SVisited \to \mathbb{N}$ be defined such that,
    for all $i \in [m]$, $c(\hat \mS_i)$ and $c'(\hat \mS_i)$ represent the number of times $\hat \mS_i$ occurs in $\Xtrain$ and $\Xtrain'$, respectively. If for all $i \in [m]$, $c(\hat \mS_i) \geq \gamma c'(\hat \mS_i)$ then the symmetric part of the empirical transition model matrix $\hat \mA_m$ (defined in \eqref{def:empirical_transition_model_matrix}) is positive-definite.
    \label{lem:condition_H_A_positive_definite}
\end{lemma}
\begin{proof}
    The structure of $\sqrt{n}\hat \mU_n \in \R^{m \times n}$ indicates each column $i$ of $\hat \mU_n$ is a one-hot vector, where its $j$-th element is $1$ if the $i$-th state $\vs_i$ of $\Xtrain$ is $\hat \mS_j$. 
    Conversely, each row $i$ of $\sqrt{n}\hat \mU_n$ has a $j$-th element equal to one if the $j$-th state $\vs_j$ of $\Xtrain$ is $\hat \mS_i$.
    A similar correspondence holds for $\sqrt{n}\hat \mV_n$ and $\Xtrain'$. 
    From interpretations of $\hat \mU_n$ and $\hat \mV_n$, we deduce that $n\hat \mU_n\hat \mU_n^T \in \R^{m \times m}$ and $n\hat \mV_n\hat \mV_n^T \in \R^{m \times m}$ are diagonal matrices where the $i$-th element of its diagonal is equal to $c(\hat \mS_i)$ and $c'(\hat \mS_i)$, respectively. 
    In the same way, $n\hat \mU_n\hat \mV_n^T \in \R^{m \times m}$ is matrix for which  $[n\hat \mU_n\hat \mV_n^T]_{ij}$ is $c(\hat \mS_i \to \hat \mS_j)$, i.e., the number of times the state $\hat \mS_i$ follows $\hat \mS_j$ in $\mathcal{D}_\text{train}$. 
    We want to prove $H(\hat \mA_m) = \frac{\hat \mA_m + \hat \mA_m^T}{2}$ is positive-definite by using the Gershgorin circle theorem and by showing $H(\hat \mA_m)$ is strictly diagonally dominant, i.e., $\lvert [H(\hat \mA_m)]_{ii} \rvert > \sum_{i \neq j}\lvert [H(\hat \mA_m)]_{ij} \rvert$. 
    From interpretations of $\hat \mU_n\hat \mU_n^T$ and $\hat \mU_n\hat \mV_n^T$, we have for all $i \in [n]$
    \begin{equation*}
        [H(\hat \mA_m)]_{ii}=[\hat \mU_n\hat \mU_n^T]_{ii} - \gamma [\hat \mU_n\hat \mV_n^T]_{ii} = \frac{c(\hat \mS_i)-\gamma c(\hat \mS_i \to \hat \mS_i)}{n} > 0, 
    \end{equation*}
    and for all $i \neq j$
    \begin{equation*}
        [H(\hat \mA_m)]_{ij}=\frac{-\gamma [\hat \mU_n\hat \mV_n^T]_{ij} -\gamma [\hat \mU_n\hat \mV_n^T]_{ji}}{2} = \frac{-\gamma c(\hat \mS_i \to \hat \mS_j)-\gamma c(\hat \mS_j \to \hat \mS_i)}{2n} < 0.
    \end{equation*}
    To prove that $H(\hat \mA_m)$ is positive-definite it remains to show that
    \begin{align*}
        \sum_j[ H(\hat \mA_m)]_{ij} &= \sum_{j \neq i}[ H(\hat \mA_m)]_{ij} + [H(\hat \mA_m)]_{ii} \\ &= \sum_j \left[\frac{\hat \mU_n(\hat \mU_n - \gamma \hat \mV_n)^T}{2}\right]_{ij} + \sum_j \left[\frac{\hat \mU_n(\hat \mU_n - \gamma \hat \mV_n)^T}{2}\right]_{ji} > 0
    \end{align*}
    for all $i \in [m]$.
    Let $i \in [m]$, we have 
    \begin{equation*}
        \sum_j [\hat \mU_n(\hat \mU_n - \gamma \hat \mV_n)^T]_{ij} = \frac{c(\hat \mS_i)}{n} - \gamma \sum_j \frac{c(\hat \mS_i \to \hat \mS_j)}{n} = (1-\gamma)\frac{c(\hat \mS_i)}{n} > 0
    \end{equation*}
    and
    \begin{equation*}
        \sum_j [\hat \mU_n(\hat \mU_n - \gamma \hat \mV_n)^T]_{ji} = \frac{c(\hat \mS_i)}{n} - \gamma \sum_j \frac{c(\hat \mS_j \to \hat \mS_i)}{n} = \frac{c(\hat \mS_i)-\gamma c'(\hat \mS_i)}{n} > 0,
    \end{equation*}
    since $c(\hat \mS_i) \geq \gamma c'(\hat \mS_i)$ for all $i \in [m]$.
    We deduce for all $i \in [m]$ that
    \begin{equation*}
        \sum_j[H(\hat \mA_m)]_{ij} = \sum_j \left[\frac{\hat \mU_n(\hat \mU_n - \gamma \hat \mV_n)^T}{2}\right]_{ij} + \sum_j \left[\frac{\hat \mU_n(\hat \mU_n - \gamma \hat \mV_n)^T}{2}\right]_{ji} > 0,
    \end{equation*}
    and thus $H(\hat \mA_m)$ is strictly diagonally dominant and positive-definite.
\end{proof}
\begin{remark}
    Conditions of \lemref{lem:condition_H_A_positive_definite} may hold in practice. If $\mathcal{D}_\text{train}$ is derived from a sample path of the MRP, where $\vs'_{i+1} = \vs_{i}$ for all $i \in [n-1]$, and if $\hat \mS_l$ depicts the distinct visited state corresponding to the last next state visited $\vs_n'$ in  $\mathcal{D}_\text{train}$, then we have $c(\hat \mS_i)=c'(\hat \mS_i)$ for all $i \neq l$ and $c(\hat \mS_l)=c'(\hat \mS_l)-1$. For sufficiently large $n$, we may have $c(\hat \mS_l) \geq \frac{\gamma}{1-\gamma}$ which satisfies conditions of \lemref{lem:condition_H_A_positive_definite}. Similarly, conditions of \lemref{lem:condition_H_A_positive_definite} are satisfied for the pathwise LSTD algorithm, where $\mathcal{D}_\text{train}$ is perturbed slightly by setting the feature of the next state of the last transition to zero~\citep{lazaric2012finite} to get $c(\hat \mS_l) \geq c'(\hat \mS_l)$.
\end{remark}
\section{Technical Details on the Resolvent \texorpdfstring{$\mQ_m(\lambda)$}{Q_m(lambda)} \label{sec:operator_norm_resolvent}}
The objective of this section is to prove that the operator norm of $\mQ_m(\lambda)$ is uniformly upper bounded under Assumption~\ref{assumption:regime_n}.
Indeed, controlling the operator norm of $\mQ_m(\lambda)$ is crucial for proving the theorems in \Secref{sec:main_results}. When $\gamma=0$, which corresponds to the supervised learning case on the reward function, the result is straightforward with \lemref{lem:resolvent_bound_semi_positive_definite} since $\frac{1}{m}\FXtrain^T \FXtrain$ is positive-definite~\citep{louart2018random, liao2020random}. In the RL setting, the conclusion is less straightforward as the resolvent is no longer that of a symmetric positive-definite matrix. This issue is further exacerbated by the lack of results in the literature concerning the upper bounds for operator norm of resolvents of non-positive-definite matrices. \lemref{lem:upper_bound_norm_resolvent} aims to propose a solution for the RL setting under Assumptions~\ref{assumption:growth_rate}~and~\ref{assumption:regime_n}. Proof of the widely used \lemref{lem:upper_bound_norm_resolvent_+_RF_matrix} is also presented at the end of this section.
\begin{lemma}
    Under Assumptions~\ref{assumption:growth_rate}~and~\ref{assumption:regime_n}, let $\lambda>0$ and let $\mQ_m(\lambda) \in \R^{n \times n}$ be the resolvent defined in \eqref{def:resolvent} as
    \begin{equation*}
        \mQ_m(\lambda)=\left[\frac{1}{m}(\hat \mU_n - \gamma \hat \mV_n)^T \FSVisited^T \FSVisited \hat \mU_n + \lambda \mI_{n} \right]^{-1}.
    \end{equation*}
    Then there exists a real $K > 0$ such that, for all $m$, we have
    \begin{equation*}
       \lVert \mQ_m(\lambda) \rVert \leq K.
    \end{equation*}
    \label{lem:upper_bound_norm_resolvent}
\end{lemma}
\begin{proof}
    Under Assumption~\ref{assumption:regime_n}, the empirical transition model matrix $\hat \mA_m=\hat \mU_n(\hat \mU_n - \gamma \hat \mV_n)^T$ (\eqref{def:empirical_transition_model_matrix}) is invertible since the symmetric part of $\hat \mA_m$ is positive-definite. Let $$0 < \epsilon < \lambda\min\left\{\frac{1}{\xi_{\max}}, \frac{\xi_{\min}}{4}\right\},$$ for $\xi_{\min}, \xi_{\max} > 0$ defined in Assumption~\ref{assumption:regime_n}. We rewrite \eqref{def:resolvent} as
    \begin{align*}
        \mQ_m(\lambda)&=\left[\frac{1}{m}(\hat \mU_n - \gamma \hat \mV_n)^T \FSVisited^T \FSVisited \hat \mU_n + \lambda \mI_{n} \right]^{-1} \\
        &=\left[(\hat \mU_n - \gamma \hat \mV_n)^T \biggl[ \frac{1}{m} \FSVisited^T \FSVisited + \epsilon \mI_{m}\biggr]\hat \mU_n + \underbrace{\lambda \mI_{n} - \epsilon (\hat \mU_n - \gamma \hat \mV_n)^T \hat \mU_n}_{=\mB_n} \right]^{-1}.
    \end{align*}
    To apply the Woodbury identity (\lemref{lem:woodbury}) on $\mQ_m(\lambda)$, we check that both 
    \begin{equation*}
        \mB_n=\lambda \mI_{n} - \epsilon(\hat \mU_n - \gamma \hat \mV_n)^T \hat \mU_n,
    \end{equation*}
    and
    \begin{align*}
        \mM_m &= \biggl[\frac{1}{m}\FSVisited^T\FSVisited + \epsilon \mI_{m}\biggr]^{-1} + \hat \mU_n \mB_n^{-1} (\hat \mU_n - \gamma \hat \mV_n)^T \\
        &=\biggl[\frac{1}{m}\FSVisited^T\FSVisited + \epsilon \mI_{m}\biggr]^{-1} + \left[\lambda \mI_{n} - \epsilon \hat \mA_m \right]^{-1}\hat \mA_m \\
        &= \biggl[\frac{1}{m}\FSVisited^T\FSVisited + \epsilon \mI_{m}\biggr]^{-1} + \left[\lambda \hat \mA_m^{-1} - \epsilon \mI_m \right]^{-1} 
    \end{align*}
    are non-singular, since $\frac{1}{m} \FSVisited^T \FSVisited + \epsilon \mI_{m}$ is non-singular. Given that $H(\hat \mA_m)$ is positive-definite, $\hat \mA_m$ has eigenvalues with positive real parts. Consequently, by the  Weinstein–Aronszajn identity (\lemref{lem:weinstein}), $(\hat \mU_n - \gamma \hat \mV_n)^T\hat \mU_n$ has non-zero eigenvalues with positive real parts.  As $\epsilon < \frac{\lambda}{\xi_{\max}} \leq \frac{\lambda}{\nu_{\max}(H(\hat \mA_m))} \leq \frac{\lambda}{\operatorname{Re}(\nu_{\max}(\hat \mA_m))}$, we deduce that the matrix $\mB_n=\lambda \mI_{n} - \epsilon (\hat \mU_n - \gamma \hat \mV_n)^T \hat \mU_n$ has eigenvalues with positive real parts and is non-singular. 
    In order to prove that the matrix $\mM_m$ is non-singular, we propose to show $\vx^T \mM_m \vx >0$ for all non-zero $\vx \in \R^m$. Since $\Bigl[\frac{1}{m}\FSVisited^T\FSVisited + \epsilon \mI_{m}\Bigr]^{-1}$ is at least positive-semi-definite, the statement $\vx^T \mM_m \vx > 0$ for all non-zero $\vx \in \R^m$ may be restated as
    \begin{align*}
        &\text{for all non-zero $\vx \in \R^m$,} \quad \vx^T \left[\lambda \hat \mA_m^{-1} - \epsilon \mI_m \right]^{-1} \vx > 0 \\
        \text{iff} \quad& \text{for all non-zero $\vx \in \R^m$,} \quad \vx^T \left[\lambda \hat \mA_m^{-1} - \epsilon \mI_m \right] \vx  > 0 \\
        \text{iff} \quad& \text{for all non-zero $\vx \in \R^m$,} \quad  \vx^T H\bigl(\hat \mA_m^{-1}\bigr) \vx - \frac{\epsilon}{\lambda} \vx^T \vx  > 0  \\
        \text{iff} \quad&  \nu_{\min}\bigl(H(\hat \mA_m^{-1})\bigr) > \frac{\epsilon}{\lambda}.
    \end{align*}
    By construction of $\hat \mU_n$ and $\hat \mV_n$, we have both $\lVert \hat \mU_n \rVert \leq 1$ and  $\lVert \hat \mV_n \rVert \leq 1$. We deduce thus
    \begin{equation*}
        \bigl\lVert \hat \mA_m \bigr\rVert= \bigl\lVert \hat \mU_n (\hat \mU_n - \gamma \hat \mV_n)^T \bigr\rVert  < 2.
    \end{equation*}
    Since $H(\hat \mA_m^{-1}) = [\hat \mA_m^{-1}]^T H(\hat \mA_m)\hat \mA_m^{-1}$, we deduce from Ostrowski’s Theorem (\lemref{lem:ostrowski}) that
    \begin{equation*}
        \nu_{\min}\bigl(H(\hat \mA_m^{-1})\bigr) \geq \frac{\nu_{\min}\bigl(H(\hat \mA_m)\bigr)}{\lVert \hat \mA_m \rVert^2} \geq \frac{\xi_{\min}}{4}.
    \end{equation*}
    Since $\epsilon < \frac{\lambda\xi_{\min}}{4}$, we have $\vx^T \mM_m \vx > 0$ for all non-zero $\vx \in \R^m$, and thus $\mM_m$ is non-singular. As a consequence, we apply the Woodbury identity (\lemref{lem:woodbury}) on the resolvent $\mQ_m(\lambda)$ to get
    \begin{align*}
        \mQ_m(\lambda)&=\left[(\hat \mU_n - \gamma \hat \mV_n)^T \biggl[ \frac{1}{m} \FSVisited^T \FSVisited + \epsilon \mI_{m}\biggr]\hat \mU_n + \mB_n \right]^{-1} \\
        &= \mB_n^{-1} - \mB_n^{-1} (\hat \mU_n - \gamma \hat \mV_n)^T \mM_m^{-1} \hat \mU_n \mB_n^{-1}.
    \end{align*}
    Multiplying the equation above by $\mB_n=\lambda \mI_{n} - \epsilon (\hat \mU_n - \gamma \hat \mV_n)^T \hat \mU_n$ on both sides, and after manipulating terms to isolate $\mQ_n$ on the left-hand side gives
    \begin{align*}
            \mQ_m(\lambda) = \frac{1}{\lambda^2}\Biggl[& \mB_n -  (\hat \mU_n - \gamma \hat \mV_n)^T \mM_m^{-1} \hat \mU_n \\
            &\quad + \lambda \epsilon \Bigl[(\hat \mU_n - \gamma \hat \mV_n)^T \hat \mU_n \mQ_m(\lambda) + \mQ_m(\lambda)(\hat \mU_n - \gamma \hat \mV_n)^T \hat \mU_n \Bigr] \\
            &\quad - \epsilon^2 (\hat \mU_n - \gamma \hat \mV_n)^T \hat \mU_n \mQ_m(\lambda) (\hat \mU_n - \gamma \hat \mV_n)^T \hat \mU_n \Biggr] \\
            = \frac{1}{\lambda^2}\Biggl[& \mB_n -  (\hat \mU_n - \gamma \hat \mV_n)^T \mM_m^{-1} \hat \mU_n \\
            &\quad + \lambda \epsilon (\hat \mU_n - \gamma \hat \mV_n)^T \left[\frac{1}{m}\hat \mA_m \FSVisited^T \FSVisited + \lambda \mI_m \right]^{-1} \hat \mU_n \\
            &\quad +\lambda \epsilon (\hat \mU_n - \gamma \hat \mV_n)^T \left[\frac{1}{m} \FSVisited^T \FSVisited \hat \mA_m + \lambda \mI_m \right]^{-1} \hat \mU_n  \\
            &\quad - \epsilon^2 (\hat \mU_n - \gamma \hat \mV_n)^T \left[\frac{1}{m}\hat \mA_m \FSVisited^T \FSVisited + \lambda \mI_m \right]^{-1} \hat \mU_n (\hat \mU_n - \gamma \hat \mV_n)^T \hat \mU_n \Biggr].
    \end{align*}
    Applying the operator nom on the equality above, we find
    \begin{equation}
        \begin{split}
            \lVert \mQ_m(\lambda) \rVert \leq \frac{1}{\lambda^2}\Biggl[& \lambda + 2 \epsilon +  2 \lVert \mM_m^{-1} \rVert \\
            &\quad + 2 \lambda \epsilon  \left\lVert \left[\frac{1}{m}\hat \mA_m \FSVisited^T \FSVisited + \lambda \mI_m \right]^{-1} \right\rVert + 2 \lambda \epsilon \left\lVert  \left[\frac{1}{m} \FSVisited^T \FSVisited \hat \mA_m + \lambda \mI_m \right]^{-1}\right\rVert \\
            &\quad + 4\epsilon^2 \left \lVert \left[\frac{1}{m}\hat \mA_m \FSVisited^T \FSVisited + \lambda \mI_m \right]^{-1} \right\rVert \Biggr],
        \end{split}
        \label{eq:eq_upper_bound_norm_resolvent_1}
    \end{equation}
    since 
    \begin{equation*}
        \lVert \mB_n \rVert = \bigl\lVert \lambda \mI_n - \epsilon (\hat \mU_n - \gamma \hat \mV_n)^T \hat \mU_n \bigr\rVert \leq \lambda + 2 \epsilon.
    \end{equation*}
    From \lemref{lem:upper_bound_norm_resolvent_dual}, we have
    \begin{equation*}
        \left\lVert \left[\frac{1}{m}\hat \mA_m \FSVisited^T \FSVisited + \lambda \mI_m \right]^{-1} \right\rVert  \leq \frac{1}{\lambda}\,\frac{4}{\xi_{\min}^2},
    \end{equation*}
    and 
    \begin{equation*}
        \left\lVert \left[\frac{1}{m}\FSVisited^T \FSVisited \hat \mA_m  + \lambda \mI_m \right]^{-1} \right\rVert  \leq \frac{1}{\lambda}\,\frac{4}{\xi_{\min}^2}.
    \end{equation*}
    To finish the proof, we find an upper bound for $\lVert \mM_m^{-1} \rVert$.
    By denoting by $\mZ^T\mZ$ the Cholesky decomposition of the positive-semi-definite matrix $\Bigl[\frac{1}{m}\FSVisited^T\FSVisited + \epsilon \mI_{m}\Bigr]^{-1}$,  we reuse the Woodbury identity (\lemref{lem:woodbury}) to rewrite $\mM_m^{-1}$ as
    \begin{align*}
        \mM_m^{-1} &= \left[~\biggl[\frac{1}{m}\FSVisited^T\FSVisited + \epsilon \mI_{m}\biggr]^{-1}+\bigl[\lambda \hat \mA_m^{-1} - \epsilon \mI_m \bigr]^{-1} \right]^{-1} \\
        &=\left[\mZ^T\mZ + \bigl[\lambda \hat \mA_m^{-1} - \epsilon \mI_m \bigr]^{-1} \right]^{-1} \\
        &=\bigl[\lambda \hat \mA_m^{-1} - \epsilon \mI_m \bigr] - \bigl[\lambda \hat \mA_m^{-1} - \epsilon \mI_m \bigr]\mZ^T\Bigl[\mZ\bigl[\lambda \hat \mA_m^{-1} - \epsilon \mI_m \bigr]\mZ^T + \mI_m \Bigr]^{-1}\mZ \bigl[\lambda \hat \mA_m^{-1} - \epsilon \mI_m \bigr].
    \end{align*}
    From \lemref{lem:resolvent_bound_semi_positive_definite},
    \begin{equation*}
        \biggl\lVert \Bigl[\mZ\bigl[\lambda \hat \mA_m^{-1} - \epsilon \mI_m \bigr]\mZ^T + \mI_m \Bigr]^{-1} \biggr\rVert \leq 1,
    \end{equation*} 
    since $H\bigl(\mZ\bigl[\lambda \hat \mA_m^{-1} - \epsilon \mI_m \bigr]\mZ^T\bigr)$ is positive-semi-definite, and from \lemref{lem:upper_bound_norm_A} we have
    \begin{equation*}
        \lVert \hat \mA_m^{-1} \rVert \leq \frac{1}{\xi_{\min}}.
    \end{equation*}
    Besides, 
    \begin{equation*}
        \lVert \mZ \rVert^2=\nu_{\max}\Bigl(\Bigl[\frac{1}{m}\FSVisited^T\FSVisited + \epsilon \mI_{m}\Bigr]^{-1}\Bigr)\leq \frac{1}{\epsilon}.
    \end{equation*}
    We deduce for the operator norm of $\mM_m^{-1}$ that
    \begin{equation*}
        \lVert \mM_m^{-1} \rVert \leq \biggl( \frac{\lambda}{\xi_{\min}}+ \epsilon \biggr) + \frac{1}{\epsilon}\biggl(  \frac{\lambda}{\xi_{\min}}+ \epsilon \biggr)^2.
    \end{equation*}
    Setting $\epsilon=\frac{\lambda}{2\epsilon'} < \lambda\min\left\{\frac{1}{\xi_{\max}}, \frac{\xi_{\min}}{4}\right\}$ for $\epsilon'>\frac{1}{2}\min\left\{\frac{1}{\xi_{\max}}, \frac{\xi_{\min}}{4}\right\}$ and putting upper bounds of $\lVert \mM_m^{-1} \rVert$, $\left\lVert \left[\frac{1}{m}\hat \mA_m \FSVisited^T \FSVisited + \lambda \mI_m \right]^{-1} \right\rVert$, $\left\lVert \left[\frac{1}{m} \FSVisited^T \FSVisited \hat \mA_m + \lambda \mI_m \right]^{-1} \right\rVert$ into \eqref{eq:eq_upper_bound_norm_resolvent_1} give
    \begin{align*}
            \lVert \mQ_m(\lambda) \rVert &\leq \frac{1}{\lambda^2}\Biggl[ \lambda + \frac{\lambda}{\epsilon'} + \lambda \biggl( \frac{2}{\xi_{\min}}+ \frac{1}{\epsilon'} \biggr) + \lambda \epsilon'\biggl(\frac{2}{\xi_{\min}}+ \frac{1}{\epsilon'} \biggr)^2 + \lambda \frac{8}{\xi_{\min}^2\epsilon'}+ \lambda \frac{4}{\xi_{\min}^2\epsilon'^2} \, \Biggr] \\
            &= \frac{1}{\lambda}\Biggl[1 + \frac{1}{\epsilon'} + \biggl( \frac{2}{\xi_{\min}}+ \frac{1}{\epsilon'} \biggr) +  \epsilon'\biggl(\frac{2}{\xi_{\min}}+ \frac{1}{\epsilon'} \biggr)^2 + \frac{8}{\xi_{\min}^2\epsilon'}+  \frac{4}{\xi_{\min}^2\epsilon'^2} \, \Biggr].
    \end{align*}
\end{proof}
\begin{remark}
    From the proof of \lemref{lem:upper_bound_norm_resolvent}, eigenspectrum constraints on the empirical transition model matrix $\hat \mA_m$ in Assumption~\ref{assumption:regime_n} ensure the resolvent $\mQ_m(\lambda)$ is uniformly bounded.
\end{remark}
\begin{lemma}
    Under Assumptions~\ref{assumption:growth_rate}~and~\ref{assumption:regime_n}, let $\lambda>0$ and let $\mQ_m'(\lambda), \mQ_m''(\lambda) \in \R^{m \times m}$ be the following resolvents
    \begin{align*}
        \mQ_m'(\lambda)&=\left[\frac{1}{m}\hat \mA_m \FSVisited^T \FSVisited + \lambda \mI_{m} \right]^{-1}
    \end{align*}
    and
    \begin{align*}
        \mQ_m''(\lambda)&=\left[\frac{1}{m} \FSVisited^T \FSVisited \hat \mA_m + \lambda \mI_{m} \right]^{-1},
    \end{align*}
     where $\hat \mA_m = \hat \mU_n(\hat \mU_n-\gamma \hat \mV_n)^T \in \R^{m \times m}$ is the empirical transition model matrix (\eqref{def:empirical_transition_model_matrix}). 
     Then, for all $m$, we have
    \begin{equation*}
       \lVert \mQ_m'(\lambda) \rVert \leq \frac{1}{\lambda}\,\frac{4}{\xi_{\min}^2}  \quad \text{and} \quad \lVert \mQ_m''(\lambda) \rVert \leq \frac{1}{\lambda}\,\frac{4}{\xi_{\min}^2}.
    \end{equation*}
    \label{lem:upper_bound_norm_resolvent_dual}
\end{lemma}
\begin{proof}
    Since the symmetric part of the empirical transition model matrix $\hat \mA_m$ is positive-definite under Assumption~\ref{assumption:regime_n}, the matrix $\hat \mA_m$ is non-singular. We write thus
    \begin{align*}
        \lVert \mQ_m'(\lambda)\rVert &= \left\lVert \left[\frac{1}{m}\hat \mA_m \FSVisited^T \FSVisited + \lambda \mI_m \right]^{-1} \right\rVert \\
        &= \left\lVert \left[\frac{1}{m}\FSVisited^T \FSVisited + \lambda \hat \mA_m^{-1} \right]^{-1} \hat \mA_m^{-1}\right\rVert \\
        &\leq \left\lVert \left[\frac{1}{m}\FSVisited^T \FSVisited + \lambda \hat \mA_m^{-1} - \lambda \nu_{\min}\bigl(H(\hat \mA_m^{-1})\bigr) \mI_m + \lambda \nu_{\min}\bigl(H(\hat \mA_m^{-1})\bigr) \mI_m \right]^{-1} \right\rVert \lVert \hat \mA_m^{-1} \rVert \\
        &= \frac{1}{\lambda}\,\frac{1}{\nu_{\min}\bigl(H(\hat \mA_m^{-1})\bigr)}\lVert \hat \mA_m^{-1} \rVert.
    \end{align*}
    The last inequality is obtained with \lemref{lem:resolvent_bound_semi_positive_definite} since $H\Bigl(\frac{1}{m}\FSVisited^T \FSVisited + \lambda \hat \mA_m^{-1} - \lambda \nu_{\min}\bigl(H(\hat \mA_m^{-1})\bigr) \mI_m\Bigr)$ is positive-semi-definite. 
    By construction of both $\hat \mU_n$ and $\hat \mV_n$, we have $\lVert \hat \mU_n \rVert \leq 1$ and  $\lVert \hat \mV_n \rVert \leq 1$. We deduce that
    \begin{equation*}
        \bigl\lVert \hat \mA_m \bigr\rVert= \bigl\lVert \hat \mU_n (\hat \mU_n - \gamma \hat \mV_n)^T \bigr\rVert  < 2.
    \end{equation*}
    Since $H(\hat \mA_m^{-1}) = [\hat \mA_m^{-1}]^T H(\hat \mA_m)\hat \mA_m^{-1}$, we deduce from  Ostrowski’s theorem (\lemref{lem:ostrowski}) that
    \begin{equation*}
        \nu_{\min}\bigl(H(\hat \mA_m^{-1})\bigr) \geq \frac{\nu_{\min}\bigl(H(\hat \mA_m)\bigr)}{\lVert \hat \mA_m \rVert^2} \geq \frac{\xi_{\min}}{4}.
    \end{equation*}
    Furthermore, from \lemref{lem:upper_bound_norm_A}, we have $\lVert \hat \mA_m^{-1} \rVert \leq \frac{1}{\xi_{\min}}$. We conclude that
    \begin{equation*}
        \lVert \mQ_m'(\lambda)\rVert \leq \frac{1}{\lambda}\,\frac{4}{\xi_{\min}^2}.
    \end{equation*}
    With a similar reasoning, we can find the same upper bound for $\lVert \mQ_m''(\lambda) \rVert$. 
\end{proof}
\begin{lemma}
    Let $\hat \mA_m = \hat \mU_n(\hat \mU_n - \gamma \hat \mV_n)^T$ be the empirical transition model matrix defined in \eqref{def:empirical_transition_model_matrix}. Under Assumption~\ref{assumption:regime_n}, for all $m$, we have
    \begin{equation*}
        \lVert \hat \mA_m^{-1} \rVert \leq \frac{1}{\xi_{\min}}.
    \end{equation*}
    \label{lem:upper_bound_norm_A}
\end{lemma}
\begin{proof}
    We rewrite $\hat \mA_m$ as 
    \begin{equation*}
        \hat \mA_m^{-1} = \Bigl[\,\bigl[\hat \mA_m - \nu_{\min}\bigl(H(\hat \mA_m)\bigr)\mI_m\bigr]  + \nu_{\min}\bigl(H(\hat \mA_m)\bigr)\mI_m  \Bigr]^{-1}.
    \end{equation*}
    Since the matrix $H\Bigl(\bigl[\hat \mA_m - \nu_{\min}\bigl(H(\hat \mA_m)\bigr)\mI_m\bigr]\Bigr)$ is positive-semi-definite, we apply \lemref{lem:resolvent_bound_semi_positive_definite} on $\hat \mA_m^{-1}$ to get 
    \begin{equation*}
        \lVert \hat \mA_m^{-1} \rVert \leq \frac{1}{\nu_{\min}\bigl(H(\hat \mA_m)\bigr)} \leq \frac{1}{\xi_{\min}}.
    \end{equation*}
\end{proof}
\begin{lemma}
    Under Assumption~\ref{assumption:growth_rate}~and~\ref{assumption:regime_n}, let $\lambda>0$ and let $\mQ_m(\lambda) \in \R^{n \times n}$ be the resolvent defined in \eqref{def:resolvent} as
    \begin{equation*}
        \mQ_m(\lambda)=\left[\frac{1}{m}(\hat \mU_n - \gamma \hat \mV_n)^T \FSVisited^T \FSVisited \hat \mU_n + \lambda \mI_{n} \right]^{-1}.
    \end{equation*}
     Then there exists a real $K > 0$ such that, for all $m$, we have
    \begin{equation*}
        \left\lVert \frac{1}{\sqrt{m}} \FSVisited \hat \mU_n \mQ_m(\lambda) \right\rVert \leq K
    \end{equation*} 
    and 
    \begin{equation*}
        \left\lVert \frac{1}{\sqrt{m}} \mQ_m(\lambda) (\hat \mU_n - \gamma \hat \mV_n)^T \FSVisited^T \right\rVert \leq 2K.
    \end{equation*} 
    \label{lem:upper_bound_norm_resolvent_+_RF_matrix}
\end{lemma}
\begin{proof}
    From \lemref{lem:upper_bound_norm_resolvent}, we know there exists a real $K>0$ such that, for all $m$, we have $\lVert \mQ_m(\lambda) \rVert \leq K$.
    Since the symmetric part of the empirical transition model matrix $\hat \mA_m=\hat \mU_n (\hat \mU_n - \gamma \hat \mV_n)^T$ (\eqref{def:empirical_transition_model_matrix}) is positive-definite under Assumption~\ref{assumption:regime_n}, the matrix $\hat \mA_m$ is non-singular.
     Furthermore, from \lemref{lem:upper_bound_norm_A} we have $\lVert \hat \mA_m^{-1} \rVert \leq \frac{1}{\xi_{\min}}$, and both $\lVert \hat \mU_n \rVert$ and $\lVert \hat \mV_n \rVert$  are upper bounded by $1$. We deduce that
    \begin{align*}
        \left\lVert \frac{1}{\sqrt{m}} \FSVisited \hat \mU_n \mQ_m(\lambda) \right\rVert &= \left\lVert \frac{1}{m} \mQ_m(\lambda)^T \hat \mU_n^T \FSVisited^T \FSVisited \hat \mU_n \mQ_m(\lambda) \right\rVert^{\frac{1}{2}} \\
        &= \left\lVert \frac{1}{m} \mQ_m(\lambda)^T \hat \mU_n^T \hat \mA_m^{-1} \hat \mU_n (\hat \mU_n-\gamma \hat \mV_n)^T \FSVisited^T \FSVisited \hat \mU_n \mQ_m(\lambda) \right\rVert^{\frac{1}{2}} \\
        &= \left\lVert \mQ_m(\lambda)^T \hat \mU_n^T \hat \mA_m^{-1} \hat \mU_n \bigl[\mI_n - \lambda \mQ_m(\lambda)\bigr] \right\rVert^{\frac{1}{2}} \\
        &\leq \sqrt{\frac{K(1+K)}{\xi_{\min}} }.
    \end{align*}
    Similarly, we have
    \begin{align*}
        &\left\lVert \frac{1}{\sqrt{m}}\mQ_m(\lambda) (\hat \mU_n-\gamma \hat \mV_n)^T \FSVisited^T \right\rVert \\ 
        &\quad = \left\lVert \frac{1}{m} \mQ_m(\lambda) (\hat \mU_n-\gamma \hat \mV_n)^T \FSVisited^T \FSVisited \hat \mU_n (\hat \mU_n-\gamma \hat \mV_n)^T \hat \mA_m^{-1} (\hat \mU_n-\gamma \hat \mV_n) \mQ_m(\lambda)^T \right\rVert^{\frac{1}{2}} \\
        &\quad = \left\lVert \bigl[\mI_n - \lambda \mQ_m(\lambda)\bigr] (\hat \mU_n-\gamma \hat \mV_n)^T \hat \mA_m^{-1} (\hat \mU_n-\gamma \hat \mV_n) \mQ_m(\lambda)^T \right\rVert^{\frac{1}{2}} \\
        &\quad \leq 2\sqrt{\frac{K(1+K)}{\xi_{\min}} }.
    \end{align*}
\end{proof}
\section{About the Existence, Positiveness, and Uniqueness of \texorpdfstring{$\delta$}{delta}  \label{sec:delta}}
This section is dedicated to prove that the fixed-point solution $\delta$ of \eqref{def:delta} is unique and positive under Assumptions~\ref{assumption:growth_rate}~and~\ref{assumption:regime_n}. This result is proven in the following Lemma.
\begin{lemma}
    Under Assumptions~\ref{assumption:growth_rate}~and~\ref{assumption:regime_n}, for all $m$, let $\delta$ be the solution to the fixed-point \eqref{def:delta} defined as
    \begin{equation*}
        \delta = \frac{1}{m} \Tr \Biggl((\hat \mU_n - \gamma \hat \mV_n)^T\mPhi_{\SVisited} \hat \mU_n \biggl[\frac{N}{m} \frac{1}{1+\delta} (\hat \mU_n - \gamma \hat \mV_n)^T\mPhi_{\SVisited}\hat \mU_n + \lambda \mI_n \biggr]^{-1}\Biggr).
    \end{equation*}
    Then $\delta$ exists, is positive, and is unique.
    \label{lem:delta_positive_unique}
\end{lemma}
\begin{proof}
    For ease of notations, we define the matrix $\mB_n=(\hat \mU_n - \gamma \hat \mV_n)^T\mPhi_{\SVisited} \hat \mU_n$. The proof is based on the use of \lemref{lem:fixed_point_standard_function} on the mapping $f: \delta \mapsto \frac{1}{m} \Tr\bigl(\mB_n \bar \mQ_m(\delta)\bigr)$. To apply \lemref{lem:fixed_point_standard_function}, we need to show $i.$ $f$ is positive on $[0, \infty)$, $ii.$ $f$ is monotonically increasing, $iii.$ $f$ is scalable, and $iv.$ $f$ admits $x_0 \in [0, \infty)$ such that $x_0 \geq f(x_0)$.
    Following this plan, we are going to show first $i.$, i.e., $f(\delta)>0$ for all $\delta>0$. 
    By denoting $\nu_j\bigl(\mB_n \bar \mQ_m(\delta)\bigr)$ the j-th eigenvalues of the matrix $\mB_n \bar \mQ_m(\delta)$, we have
    \begin{align}
        \begin{split}
            \nu_j\bigl(\mB_n \bar \mQ_m(\delta)\bigr) &= \nu_j\left(\mB_n \left[\frac{N}{m}\frac{1}{1+\delta}\mB_n+\lambda \mI_n\right]^{-1}\right) \\
            &=\nu_j(\mB_n)\nu_j\left(\left[\frac{N}{m}\frac{1}{1+\delta}\mB_n+\lambda \mI_n\right]^{-1}\right) \qquad \text{(from the Schur decomposition of $\mB_n$)} \\
            &=\frac{\nu_j(\mB_n)}{\frac{N}{m}\frac{1}{1+\delta}\nu_j\left(\mB_n\right) + \lambda} \\
            &=\frac{1}{\left\lvert \frac{N}{m}\frac{1}{1+\delta}\nu_j\left(\mB_n\right) + \lambda \right\rvert^2} \left(\frac{N}{m}\frac{1}{1+\delta}\lvert\nu_j(\mB_n)\rvert^2 + \lambda \nu_j(\mB_n) \right).
        \end{split}
        \label{eq:eig_A_Q_bar}
    \end{align}
    Let $\hat \mA_m=\hat \mU_n(\hat \mU_n - \gamma \hat \mV_n)^T$ be the transition model matrix defined in \eqref{def:empirical_transition_model_matrix}, and $\bar \mZ \bar \mZ^T$ be the Cholesky decompositon of $\mPhi_{\SVisited}$. From the Weinstein–Aronszajn identity (\lemref{lem:weinstein}), the matrices $\mB_n = (\hat \mU_n - \gamma \hat \mV_n)^T\mPhi_{\SVisited} \hat \mU_n$ and $\bar \mZ^T\hat \mA_m \bar \mZ$ share the same non-zero eigenvalues. Under Assumption~\ref{assumption:regime_n}, the matrix $H(\bar \mZ^T\hat \mA_m \bar \mZ)$ is at least semi-positive-definite, which implies that non-zero real parts of eigenvalues of $\bar \mZ^T\hat \mA_m \bar \mZ$ are positive. We deduce that $\operatorname{Re}\bigl(\nu_j(\mB_n)\bigr) \geq 0$, for all $j \in [m]$. As a consequence, 
    \begin{align}
        \begin{split}
            f(\delta)&=\frac{1}{m} \Tr\bigl(\mB_n \bar \mQ_m(\delta)\bigr) \\
            &=\frac{1}{m} \sum_{j=1}^n \frac{1}{\left\lvert \frac{N}{m}\frac{1}{1+\delta}\nu_j\left(\mB_n\right) + \lambda \right\rvert^2} \left(\frac{N}{m}\frac{1}{1+\delta}\lvert\nu_j\left(\mB_n\right)\rvert^2 + \lambda \nu_j(\mB_n) \right) \\
            &=\frac{1}{m} \sum_{j=1}^n \frac{1}{\left\lvert \frac{N}{m}\frac{1}{1+\delta}\nu_j\left(\mB_n\right) + \lambda \right\rvert^2} \left(\frac{N}{m}\frac{1}{1+\delta}\lvert\nu_j\left(\mB_n\right)\rvert^2 + \lambda \operatorname{Re}\bigl(\nu_j(\mB_n)\bigr) \right) \\
            &>0.
        \end{split}
        \label{eq:f_delta_sum_eigenvalues}
    \end{align}
    To prove $ii.$, i.e., $f$ is monotonically increasing 
    on $[0, \infty )$, we show the derivative 
    $f'$ of $f$ is positive on $[0, \infty)$. Let $\delta>0$,
    \begin{align*}
            f'(\delta) &= \frac{1}{m}\biggl( \sum_{j=1}^n \frac{\nu_j(\mB_n)}{\frac{N}{m}\frac{1}{1+\delta}\nu_j\left(\mB_n\right) + \lambda} \biggr)' \\
            &= \frac{1}{m}\sum_{j=1}^n \frac{\frac{N}{m}\frac{1}{(1+\delta)^2}}{\left(\frac{N}{m}\frac{1}{1+\delta}\nu_j\left(\mB_n\right) + \lambda\right)^2}\nu_j(\mB_n)^2 \\
            &= \frac{1}{m}\sum_{j=1}^n \frac{\frac{N}{m}\frac{1}{(1+\delta)^2}}{\left\lvert\frac{N}{m}\frac{1}{1+\delta}\nu_j\left(\mB_n\right) + \lambda\right\lvert^4}\biggl(\frac{N}{m}\frac{1}{(1+\delta)} \lvert\nu_j(\mB_n)\rvert^2  + \lambda \nu_j(\mB_n) \biggr)^2 \\
            &= \frac{1}{m}\sum_{j=1}^n \frac{\frac{N}{m}\frac{1}{(1+\delta)^2}}{\left\lvert\frac{N}{m}\frac{1}{1+\delta}\nu_j\left(\mB_n\right) + \lambda\right\lvert^4}\Bigl(\frac{N^2}{m^2}\frac{1}{(1+\delta)^2} \lvert\nu_j(\mB_n)\rvert^4 + 2\lambda \frac{N}{m}\frac{1}{1+\delta}\lvert\nu_j(\mB_n)\rvert^2 \nu_j(\mB_n) + \lambda^2 \nu_j(\mB_n)^2 \Bigr) \\
            &=\frac{1}{m}\underbrace{\sum_{j=1}^n \frac{\frac{N}{m}\frac{1}{(1+\delta)^2}}{\left\lvert\frac{N}{m}\frac{1}{1+\delta}\nu_j\left(\mB_n\right) + \lambda\right\lvert^4}\biggl(\frac{N^2}{m^2}\frac{1}{(1+\delta)^2} \lvert\nu_j(\mB_n)\rvert^4 + 2\lambda \frac{N}{m}\frac{1}{1+\delta}\lvert\nu_j(\mB_n)\rvert^2 \operatorname{Re}\bigl(\nu_j(\mB_n)\bigr) \biggr)}_{(1)}\\
            &\qquad +\underbrace{\sum_{j=1}^n \lambda^2 \frac{\frac{N}{m}\frac{1}{(1+\delta)^2}}{\left\lvert\frac{N}{m}\frac{1}{1+\delta}\nu_j\left(\mB_n\right) + \lambda\right\lvert^4} \operatorname{Re}\bigl(\nu_j(\mB_n^2)\bigr)}_{(2)}
    \end{align*}
    Since real parts of eigenvalues of $\mB_n$ are positive, $(1)$ is clearly positive. Since $\Tr(\mB_n^2)>0$ (\lemref{lem:trace_B_n_square}) and thus $(2)$ is positive, we can conclude $ii.$.
    We can use a similar proof for the scalability in $iii.$, i.e., $\alpha f(\delta) > f(\alpha \delta),~\forall \alpha > 1$. Let $\alpha>1$ and $\delta>0$,
    \begin{align}
        \alpha f(\delta)-f(\alpha \delta) &= \alpha\frac{1}{m} \Tr\bigl(\mB_n \bar \mQ_m(\delta)\bigr)-\frac{1}{m} \Tr\bigl(\mB_n \bar \mQ_m(\alpha \delta)\bigr) \\
        &= \frac{1}{m} \Tr\bigl(\mB_n \bigl[\alpha \bar \mQ_m(\delta) - \bar \mQ_m(\alpha \delta)\bigr]\bigr) \\
        &= \frac{1}{m} \Tr \Biggl(\alpha \mB_n \bar \mQ_m(\delta) \biggl[\frac{N}{m} \left(\frac{1}{1+\alpha \delta} - \frac{1}{\alpha(1+\delta)} \right) \mB_n + \left(\lambda - \frac{\lambda}{\alpha} \right) \mI_n \biggr] \bar \mQ_m(\alpha \delta)\Biggr) \\
        \begin{split}
            &= \underbrace{\alpha \frac{1}{m} \frac{N}{m} \left(\frac{1}{1+\alpha \delta} - \frac{1}{\alpha(1+\delta)} \right)}_{>0}\underbrace{\Tr\bigl( \mB_n \bar \mQ_m(\delta) \mB_n \bar \mQ_m(\alpha \delta)\bigr)}_{(1)} \\
            &\qquad + \underbrace{\alpha \frac{1}{m} \left(\lambda - \frac{\lambda}{\alpha} \right)}_{>0} \underbrace{\Tr\bigr(\mB_n \bar \mQ_m(\delta) \bar \mQ_m(\alpha \delta)\bigl)}_{(2)}.
            \label{eq:decomposition_scalable_f_delta}
        \end{split}
    \end{align}
    To prove $iii.$, we can show that both $(1)$ and $(2)$ in \eqref{eq:decomposition_scalable_f_delta} are positive. We prove in $ii.$ that $\Tr\bigl(\mB_n\bar \mQ_m(\delta')\mB_n\bar \mQ_m(\delta)\bigr)>0$ for any $\delta' > \delta$. Since $\alpha \delta > \delta$, we also deduce $(1)$ is positive.
    For $(2)$, we can write
    \begin{align*}
        &\Tr\bigl( \mB_n \bar \mQ_m(\delta) \bar \mQ_m(\alpha \delta)\bigr) \\
        &= \sum_{j=1}^n \nu_j\bigl(\mB_n \bar \mQ_m(\delta) \bar \mQ_m(\alpha \delta) \bigr) \\
        &=\sum_{j=1}^n \frac{\nu_j(\mB_n)}{\Bigl(\frac{N}{m}\frac{1}{1+\delta}\nu_j(\mB_n) + \lambda \Bigr)\Bigl(\frac{N}{m}\frac{1}{1+\alpha\delta}\nu_j(\mB_n) + \lambda \Bigr)} \\
        &=\sum_{j=1}^n c_j \Biggl(\biggl(\frac{N^2}{m^2}\frac{\lvert \nu_j(\mB_n) \rvert^2}{(1+\delta)(1+\alpha \delta)} +\lambda^2\biggr)\operatorname{Re}\bigl(\nu_j(\mB_n)\bigr) + \frac{N}{m} \left(\frac{\lambda}{1+\delta}+\frac{\lambda}{1+\alpha\delta}\right) \lvert \nu_j(\mB_n) \rvert^2 \Biggr) \\
        &>0,
    \end{align*}
    where 
    \begin{equation*}
        c_j=\frac{1}{\left\lvert\Bigl(\frac{N}{m}\frac{1}{1+\delta}\nu_j(\mB_n) + \lambda \Bigr)\Bigl(\frac{N}{m}\frac{1}{1+\alpha\delta}\nu_j(\mB_n) + \lambda \Bigr)\right\rvert^2}.
    \end{equation*}
    In order to apply \lemref{lem:fixed_point_standard_function}, we still need to demonstrate $iv.$, i.e., $f$ admits $x_0 \in [0, \infty)$ such that $x_0 \geq f(x_0)$. To prove $iv.$, it is sufficient to notice that if $f$ is bounded, i.e., $\forall \delta, f(\delta) \leq C$. Let $\delta>0$, we have
    \begin{align*}
        f(\delta)=\frac{1}{m} \Tr\bigl(\mB_n \bar \mQ_m(\delta)\bigr) &= \frac{1}{m} \Tr \bigl((\hat \mU_n - \gamma \hat \mV_n)^T\mPhi_{\SVisited} \hat \mU_n\bar \mQ_m(\delta)\bigr) \\
        &= \frac{1}{m} \Tr \bigl(\mPhi_{\SVisited} \hat \mU_n\bar \mQ_m(\delta) (\hat \mU_n - \gamma \hat \mV_n)^T \bigr) \\
        &\leq \frac{1}{m} \Tr(\mPhi_{\SVisited}) \lVert \hat \mU_n\bar \mQ_m(\delta) (\hat \mU_n - \gamma \hat \mV_n)^T \rVert \\
        &\leq \frac2n \Tr(\mPhi_{\SVisited}) \lVert \bar \mQ_m(\delta) \rVert \\
        &= \mathcal{O}(1),
    \end{align*}
    where we used for the first inequality $\lvert \Tr(\mA\mB) \rvert \leq \lVert \mB \rVert \Tr(\mA)$ for non-negative definite matrix $\mA$. The last inequality is obtained since $\frac{1}{m} \Tr(\mPhi_{\SVisited})$ is uniformly bounded under Assumptions~\ref{assumption:growth_rate}~and~\ref{assumption:regime_n} (see \eqref{eq:upper_bound_tr_phi}). Furthermore, both $\lVert \hat \mU_n \rVert$ and $\lVert \hat \mV_n \rVert$ are upper bounded by $1$ and, with a similar proof than for \lemref{lem:upper_bound_norm_resolvent}, we can show there exists a real $K_{\bar \mQ}>0$  such that, for all $m$ and for all $\delta \in [0, \infty)$, we have $\lVert \bar \mQ_m(\delta) \rVert \leq K_{\bar \mQ}$. Since all hypotheses required on $f$ to apply \lemref{lem:fixed_point_standard_function} are satisfied, we can apply this Lemma which concludes the proof.
\end{proof}
\begin{lemma}
    We have
    \begin{equation*}
        \Tr\bigl((\hat \mU_n - \gamma \hat \mV_n)^T\mPhi_{\SVisited} \hat \mU_n(\hat \mU_n - \gamma \hat \mV_n)^T\mPhi_{\SVisited} \hat \mU_n\bigr) > 0.
    \end{equation*}
    \label{lem:trace_B_n_square}
\end{lemma}
\begin{proof}
    Let $\mA=\hat \mU_n (\hat \mU_n - \gamma \hat \mV_n)^T$. We denote by $S(\mA)=\tfrac{\mA-\mA^T}{2}$ the skew-symmetric part of $\mA$. We have
    \begin{align*}
        &\Tr\bigl((\hat \mU_n - \gamma \hat \mV_n)^T\mPhi_{\SVisited} \hat \mU_n(\hat \mU_n - \gamma \hat \mV_n)^T\mPhi_{\SVisited} \hat \mU_n\bigr) \\
        &=\Tr\bigl(\mPhi_{\SVisited} \mA \mPhi_{\SVisited} \mA\bigr) \\
        &=\Tr\bigl(\mPhi_{\SVisited} \mA \mPhi_{\SVisited} H(\mA)\bigr)+\Tr\bigl(\mPhi_{\SVisited} \mA \mPhi_{\SVisited} S(\mA)\bigr) \\
        &=\Tr\bigl(\mPhi_{\SVisited} H(\mA) \mPhi_{\SVisited} H(\mA)\bigr)+\Tr\bigl(\mPhi_{\SVisited} S(\mA) \mPhi_{\SVisited} H(\mA)\bigr)+\Tr\bigl(\mPhi_{\SVisited} H(\mA) \mPhi_{\SVisited} S(\mA)\bigr)+\Tr\bigl(\mPhi_{\SVisited} S(\mA) \mPhi_{\SVisited} S(\mA)\bigr) \\
         &=\Tr\bigl(\mPhi_{\SVisited} H(\mA) \mPhi_{\SVisited} H(\mA)\bigr)+\Tr\bigl(\mPhi_{\SVisited} S(\mA) \mPhi_{\SVisited} S(\mA)\bigr) > 0.
    \end{align*}
\end{proof}
%
\begin{lemma}
    Under Assumptions~\ref{assumption:growth_rate}~and~\ref{assumption:regime_n}, let $\delta$ be the correction factor defined in \eqref{def:delta}. $\delta$ is a decreasing function with respect to $N$.
    \label{lem:delta_decreasing_N}
\end{lemma}
\begin{proof}
    For ease of notations, we define the matrix $\mB_n=(\hat \mU_n - \gamma \hat \mV_n)^T\mPhi_{\SVisited} \hat \mU_n$ and we denote by $\bar \mQ_m$ the resolvent $\bar \mQ_m(\lambda)$.
    The derivative of $\delta$ as function of $N$ is denoted as $\delta'(N)$ and defined as
    \begin{equation*}
        \delta'(N) = -\frac{1}{m}\frac{\frac{\frac{1}{m}\Tr(\mB_n \bar \mQ_m \mB_n \bar \mQ_m)}{(1+\delta)}}{1-\frac{N}{m}\frac{\frac{1}{m}\Tr(\mB_n \bar \mQ_m \mB_n \bar \mQ_m)}{(1+\delta)^2}}
    \end{equation*}
    For all $N$, we have $\delta'(N) \leq 0$ since $\frac{\frac{1}{m}\Tr(\mB_n \bar \mQ_m \mB_n \bar \mQ_m)}{(1+\delta)}>0$ and $\frac{N}{m}\frac{\frac{1}{m}\Tr(\mB_n \bar \mQ_m \mB_n \bar \mQ_m)}{(1+\delta)^2} < 1$ using a similar reasoning than for \eqref{eq:lim_sup_Tr(BbarQBbarQ)}.
\end{proof}
\begin{lemma}
    Under Assumptions~\ref{assumption:growth_rate}~and~\ref{assumption:regime_n}, let $\delta$ be the correction factor defined in \eqref{def:delta}. $\delta$ is a decreasing function with respect to $\lambda$.
    \label{lem:delta_decreasing_lambda}
\end{lemma}
\begin{proof}
    For ease of notations, we define the matrix $\mB_n=(\hat \mU_n - \gamma \hat \mV_n)^T\mPhi_{\SVisited} \hat \mU_n$ and we denote by $\bar \mQ_m$ the resolvent $\bar \mQ_m(\lambda)$.
    The derivative of $\delta$ as function of $\lambda$ is denoted as $\delta'(\lambda)$ and defined as
    \begin{equation*}
        \delta'(\lambda) = -\frac{1}{m}\Tr(\bar \mQ_m \mB_n \bar \mQ_m)
    \end{equation*}
    For all $\lambda$, we have $\delta'(\lambda) \leq 0$ using a similar reasoning than for $iii.$ in Lemma~\ref{lem:delta_positive_unique}.
\end{proof}
\section{Concentration Results \label{sec:concentrations}}
The following section is dedicated to a set of concentration results used for the proofs of Theorems. 
Preliminary results yield a concentration of measure properties for the random feature matrix $\FSVisited \in \R^{N \times m}$, which stem from the concentration inequality of \lemref{lem:normal_concentration} for Lipschitz applications of a Gaussian vector. 
Essentially, the guideline of the proofs involves the following steps; given $\mW_{ij}=\varphi(\tilde \mW_{ij})$, for which $\tilde \mW_{ij} \sim \N(0,1)$ and $\varphi$ a Lipschitz function, the normal concentration of $\tilde \mW$ is transferred to $\mW$. This process induces a normal concentration of the random vector $\sigma(\vw^T \hat \mS)$, for $\vw = \varphi(\tilde \vw)$ and $\vw \sim \N(\mathbf{0},\mI_d)$, and of the matrix $\FSVisited$. This implies that Lipschitz functionals of $\sigma(\vw^T \hat \mS)$ or $\FSVisited$ also concentrate. As highlighted earlier, these concentration results have multiple consequences on convergence of random variables, and are traditionally employed in Random Matrix theory and in \thmref{theorem:asy-behavior-E[Q]}. We start by revisiting \lemref{lem:sigma_bound} and \lemref{lem:normal_concentration_resolvent}, which are derived from \lemref{lem:normal_concentration} and that were previously introduced in~\citet{louart2018random}. Subsequently, we provide intermediary \lemref{lem:f_frobenius_lipschitz_concentration} and \lemref{lem:normal_concentration_matrix} to reach the principal results of this section articulated by \lemref{lem:concentration_coeff_D_i} and \lemref{lem:normal_concentration_msbe}, which are employed in proof of Theorems. 
In the remainder of this section, we denote by $\lVert \cdot \rVert_F$ the Frobenius norm of a matrix.
\begin{lemma}
    Let $\vsigma: \R \to \R$ be a $K_\vsigma$-Lipschitz continuous function, let $\mX \in \R^{d\times m}$ be a matrix, and let $\vw = \varphi(\tilde \vw)$ be a vector for which $\varphi: \R \to \R$ is a $K_\varphi$-Lipschitz continuous function and $\tilde \vw \sim \N(\mathbf{0},\mI_d)$. Let 
    \begin{equation*}
        t_0 = |\sigma(0)| + K_\vsigma K_\varphi \lVert \mX \rVert \sqrt{\frac{d}{m}}.
    \end{equation*}
    Then, for all $t\geq 4t_0$, we have
    \begin{equation*}
        \Pr\left(\:\left\lVert \frac1{\sqrt{m}} \sigma(\vw^T \mX) \right\rVert \geq t \right) \leq C e^{-\frac{cmt^2}{2 K_\vsigma^2 K_\varphi^2 \lVert \mX \rVert^2}},
    \end{equation*}
    for some $C,c>0$ are independent of all other parameters. 
    \label{lem:sigma_bound}
\end{lemma}
\begin{proof}
     The proof of this Lemma can be found in the first half of proof of~\citet[Lemma~2]{louart2018random}, and is based on \lemref{lem:normal_concentration}.
\end{proof}
\begin{corollary}
   \citep[Remark~2]{louart2018random} Let $\mX \in \R^{d \times m}$ and let $\mSigma_\mX=\sigma(\mW \mX) \in \R^{N \times m}$ be its random features matrix defined as in~\eqref{def:rf_map}. For all $t\geq 4t_0$, we have
    \begin{equation*}
        \Pr\biggl(~\biggl\lVert \frac{1}{m} \mSigma_\mX \biggr\rVert \geq t~\biggr) \leq C N e^{-\frac{c m^2 t^2}{2N\lVert \mX\rVert^2}},
    \end{equation*}
    where $t_0 = |\sigma(0)| + \lVert \mX\rVert \sqrt{\frac{d}{m}}$.
    \label{corollary:operator_norm_FState}
\end{corollary}
From the previous Lemma, we deduce the following key concentration result.
\begin{lemma}
     \citep[Lemma~2]{louart2018random} Let $\vsigma: \R \to \R$ be a $K_\vsigma$-Lipschitz continuous function, let $\mX \in \R^{d\times m}$ be a matrix, and let $\vw = \varphi(\tilde \vw)$ be a vector for which $\varphi: \R \to \R$ is a $K_\varphi$-Lipschitz continuous function and $\tilde \vw \sim \N(\mathbf{0},\mI_d)$. Let $\mA \in \R^{m \times m}$ be a matrix independent of $\vw$ such that $\Vert \mA \rVert \leq K_\mA$. Then, we have
    \begin{align*}
        &\Pr\left(~\left| \frac{1}{m} \sigma(\vw^T \mX)^T \mA \sigma(\vw^T \mX) - \frac{1}{m} \Tr\left(\mA \E\left[\sigma(\vw^T \mX) \sigma(\vw^T \mX)^T \right] \right) \right| > t \right) \\
        &\leq C e^{-\frac{cm}{2 K_\vsigma^2 K_\varphi^2 \lVert \mX \rVert^2} \min\left(\frac{t^2}{2^6t_0^2K_\mA^2}, \frac{t}{K_\mA}\right)},
    \end{align*}
    for $t_0 = |\sigma(0)| + \sqrt{\frac{d}{m}} K_\vsigma K_\varphi \lVert \mX \rVert$, and $c, C \in \R$ independent of all other parameters.
    \label{lem:normal_concentration_resolvent}
\end{lemma}
\begin{lemma}
    Let $f: \R^{N\times d}\to \R$, $\mW \mapsto f(\mW)$ be a $K_f$-Lipschitz function with respect to the Frobenius norm for which $\mW=\varphi(\tilde \mW)$ is the matrix defined in \eqref{def:rf_map}. Then, we have
    \begin{align*}
        \Pr\bigl(~\bigl| f(\mW) - \E \bigl[ f(\mW) \bigr] \bigr|~> t \bigr) \leq Ce^{-\frac{ct^2}{K_f^2 K_\varphi^2}},
    \end{align*}
    for some $C,c>0$.
    \label{lem:f_frobenius_lipschitz_concentration}
\end{lemma}
\begin{proof}
    The vectorization of $\tilde \mW$, $\operatorname{vec}(\tilde \mW)=\bigl[\tilde \mW_{11},\cdots,\tilde \mW_{nd}\bigr] \in \R^{N \times d}$ is a Gaussian vector. A $K_f$-Lipschitz function $f$ of $\mW$ with respect to the Frobenius norm is also a $K_f$-Lipschitz function of $\operatorname{vec}(\mW)$ with respect to the Euclidean norm. Applying \lemref{lem:normal_concentration} gives
    \begin{equation*}
        \Pr\bigl(~\bigl| f(\mW) - \E \bigl[ f(\mW) \bigr] \bigr|~> t \bigr) = \Pr\left( ~\left\lvert f\bigl(\varphi(\tilde \mW)\bigr) - \E\bigl[f\bigl(\varphi(\tilde \mW)\bigr)\bigr] \right\rvert~ > t \right) \leq Ce^{-\frac{ct^2}{K_\varphi^2 K_f^2}},
    \end{equation*}
    for some $C,c>0$.
\end{proof}
\begin{lemma}
    Under Assumptions~\ref{assumption:growth_rate}~and~\ref{assumption:regime_n}, let $\lambda>0$, let $\mW \in \R^{N \times d}$, and let the resolvent
    \begin{equation*}
        \mQ_m(\mW)=\left[\frac{1}{m}(\hat \mU_n - \gamma \hat \mV_n)^T \FSW^T \FSW \hat \mU_n + \lambda \mI_{n} \right]^{-1}
    \end{equation*}
    defined as in \eqref{def:resolvent}. Let $\vsigma \in \R^{m}$ independent of $\mW$ such that $\frac{1}{\sqrt{m}} \lVert \vsigma \rVert \leq \sqrt{K_v}$ for $K_v>0$. Then
    \begin{equation*}
            \Pr\left(~\left\lvert \frac{1}{m}\vsigma^T \hat \mU_n \mQ_m(\mW) (\hat \mU_n - \gamma \hat \mV_n)^T \vsigma - \frac{1}{m}\vsigma^T \hat \mU_n\E[\mQ_m(\mW)](\hat \mU_n - \gamma \hat \mV_n)^T \vsigma  \right\rvert > t \right) 
            \leq C e^{-cmt^2},
    \end{equation*}
    for some $C, c >0$ independent of $m$ and $N$.
    \label{lem:normal_concentration_matrix}
\end{lemma}
\begin{proof}
    Let the function $f: \mW \mapsto \frac{1}{m} \vsigma^T \hat \mU_n \mQ_m(\mW) (\hat \mU_n-\gamma \hat \mV_n)^T \vsigma$. We want to show $f$ is Lipschitz in order to apply \lemref{lem:f_frobenius_lipschitz_concentration}. From \lemref{lem:upper_bound_norm_resolvent}, we know there exists a real $K>0$ such that, for all $m$ and for any $\mW$, we have 
    \begin{equation*}
        \lVert \mQ_m(\mW) \rVert \leq K.
    \end{equation*}
    Furthemore, both $\lVert \hat \mU_n \rVert$ and $\lVert \hat \mV_n \rVert$ are upper bounded by $1$. Let $\mH \in \R^{N \times d}$, we have
    \begin{align*}
        &\left\lvert f(\mW+\mH) - f(\mW) \right\rvert \\
        &=\biggl\lvert \frac{1}{m} \vsigma^T \hat \mU_n \Bigl[\mQ_m(\mW+\mH) - \mQ_m(\mW) \Bigr] (\hat \mU_n-\gamma \hat \mV_n)^T \vsigma \biggr\rvert \\
        &=\biggl\lvert \frac{1}{m^2} \vsigma^T \hat \mU_n \mQ_m(\mW+\mH)
        (\hat \mU_n - \gamma \hat \mV_n )^T \Bigl[\FSWH^T\FSWH \\
        &\qquad - \FSW^T\FSW \Bigr] \hat \mU_n \mQ_m(\mW)(\hat \mU_n-\gamma \hat \mV_n)^T \vsigma \biggr\rvert \\
        &=\biggl\lvert \frac{1}{m^2} \vsigma^T \hat \mU_n \mQ_m(\mW+\mH)
        (\hat \mU_n - \gamma \hat \mV_n )^T \Bigl[\FSWH^T \bigl[\FSWH-\FSW\bigr] \\
        &\qquad + \bigl[\FSWH - \FSW\bigr]^T\FSW \Bigr] \hat \mU_n \mQ_m(\mW)(\hat \mU_n-\gamma \hat \mV_n)^T \vsigma \biggr\rvert \\
        &\leq\biggl\lvert \frac{1}{m^2} \vsigma^T \hat \mU_n \mQ_m(\mW+\mH)
        (\hat \mU_n - \gamma \hat \mV_n )^T \FSWH^T \bigl[\FSWH \\ 
        &\qquad -\FSW\bigr]\hat \mU_n \mQ_m(\mW)(\hat \mU_n-\gamma \hat \mV_n)^T \vsigma \biggr\rvert \\
        &\qquad + \biggl\lvert \frac{1}{m^2} \vsigma^T \hat \mU_n \mQ_m(\mW+\mH)(\hat \mU_n - \gamma \hat \mV_n )^T\bigl[\FSWH \\
        &\qquad - \FSW\bigr]^T\FSW  \hat \mU_n \mQ_m(\mW)(\hat \mU_n-\gamma \hat \mV_n)^T \vsigma \biggr\rvert \\
        &\leq 2 K_v K\biggl\lVert \frac{1}{\sqrt{m}} \mQ_m(\mW+\mH)
        (\hat \mU_n - \gamma \hat \mV_n )^T \FSWH^T \biggr\rVert~\biggl\lVert \frac{1}{\sqrt{m}} \Bigl[\FSWH-\FSW \Bigr] \biggr\rVert \\
        &\qquad +4 K_v K \biggl\lVert \frac{1}{\sqrt{m}} \Bigl[\FSWH-\FSW \Bigr] \biggr\rVert~\biggl\lVert \frac{1}{\sqrt{m}} \FSW \hat \mU_n \mQ_m(\mW)  \biggr\rVert.
    \end{align*}
    From \lemref{lem:upper_bound_norm_resolvent_+_RF_matrix}, we know there exists a real $K'>0$ such that, for all $m$, we have 
    \begin{equation*}
        \biggl\lVert \frac{1}{\sqrt{m}} \FSW \hat \mU_n \mQ_m(\mW) \biggr\rVert \leq K'
    \end{equation*}
    and
    \begin{equation*}
        \biggl\lVert \frac{1}{\sqrt{m}} \mQ_m(\mW + \mH) (\hat \mU_n - \gamma \hat \mV_n )^T \FSWH^T \biggr\rVert \leq 2K'.
    \end{equation*}
    From those results, we conclude the Lipschitz continuity of $f$ since
    \begin{align*}
        \left\lvert f(\mW+\mH) - f(\mW) \right\rvert &\leq 8 K_v K K' \biggl\lVert \frac{1}{\sqrt{m}} \Bigl[\FSWH-\FSW \Bigr] \biggr\rVert \\
        &\leq 8 K_v K K' \biggl\lVert \frac{1}{\sqrt{m}} \Bigl[\FSWH-\FSW \Bigr] \biggr\rVert_F \\
        &\leq \frac{8 K_v K K' K_\vsigma}{\sqrt{m}} \left\lVert \mH \mS \right\rVert_F \\
        &= \frac{8 K_v K K' K_\vsigma}{\sqrt{m}} \sqrt{ \Tr\left(\mH\mS\mS^T \mH^T\right) } \\
        &\leq \frac{8 K_v K K' K_\vsigma}{\sqrt{m}} \left\lVert \mS \right\rVert \lVert \mH \rVert_F.
    \end{align*}
    The last inequality is obtained because $\lvert \Tr(\mA\mB) \rvert \leq \lVert \mB \rVert \Tr(\mA)$ for some semi-positive-definite matrix $\mA$. We prove that $f$ is Lipschitz with parameter $\frac{8K_v K K' K_\vsigma}{\sqrt{m}} \left\lVert \mS \right\rVert$, and applying \lemref{lem:normal_concentration} gives
    \begin{align*}
            &\Pr\left(~\left\lvert \frac{1}{m}\vsigma^T \hat \mU_n \mQ_m(\mW) (\hat \mU_n - \gamma \hat \mV_n)^T \vsigma - \frac{1}{m}\vsigma^T \hat \mU_n\E[\mQ_m(\mW)](\hat \mU_n - \gamma \hat \mV_n)^T \vsigma  \right\rvert > t \right) \\
            &\quad \leq C e^{-\frac{cmt^2}{2^6K_v^2K^2K'^2 K_\vsigma^2 K_\varphi^2 \lVert \mS\rVert^2}},
    \end{align*}
    for some $C, c >0$ independent of other parameters.
\end{proof}
\begin{lemma}
     Under Assumptions~\ref{assumption:growth_rate}~and~\ref{assumption:regime_n}, let $\mQ_- \in \R^{n \times n}$ be the resolvent defined in \eqref{def:Q_-}, let $\vw_i \sim \N(\mathbf{0},\mI_d)$ be a Gaussian vector independent of $\mQ_-$, and let $\vsigma: \R \to \R$ be a real 1-Lipschitz function. Then
    \begin{align*}
        &\Pr\biggl(~\biggl\lvert \frac{1}{m} \sigma(\vw_i^T \hat \mS) \hat \mU_n \mQ_{-i} (\hat \mU_n - \gamma \hat \mV_n)^T \sigma(\hat \mS^T \vw_i) \\
        &\qquad - \frac{1}{m} \Tr \left(\hat \mU_n\E[\mQ_{-i}](\hat \mU_n - \gamma \hat \mV_n)^T \E[ \sigma(\hat \mS^T \vw_i) \sigma(\vw_i^T \hat \mS) ] \right) \biggr\lvert > t \biggr) \\
        &\leq C e^{-cm \max(t^2,t)},
    \end{align*}
    for some $C, c>0$ independent of $N, m$.
    \label{lem:concentration_coeff_D_i}
\end{lemma}
\begin{proof}
    We can observe that
    \begin{align}
        \begin{split}
            &\Pr\biggl(~\biggl\lvert \frac{1}{m} \sigma(\vw_i^T \hat \mS) \hat \mU_n \mQ_{-i} (\hat \mU_n - \gamma \hat \mV_n)^T \sigma(\hat \mS^T \vw_i) \\
            &\qquad - \frac{1}{m} \Tr \left(\hat \mU_n\E[\mQ_{-i}](\hat \mU_n - \gamma \hat \mV_n)^T \E[ \sigma(\hat \mS^T \vw_i) \sigma(\vw_i^T \hat \mS) ] \right) \biggr\lvert > t \biggr) \\
            &\leq \Pr\biggl(~\biggl\lvert \frac{1}{m} \sigma(\vw^T \hat \mS)^T \hat \mU_n \mQ_{-i} (\hat \mU_n - \gamma \hat \mV_n)^T \sigma(\vw^T \hat \mS) \\
            &\qquad - \frac{1}{m}\sigma(\vw^T \hat \mS)^T \hat \mU_n\E[\mQ_{-i}](\hat \mU_n - \gamma \hat \mV_n)^T \sigma(\vw^T \hat \mS)  \biggr\rvert > \frac{t}{2} \biggr)  \\
            &+\Pr\biggl(~\biggl\lvert \frac{1}{m} \sigma(\vw^T \hat \mS)^T \hat \mU_n \E[\mQ_{-i}] (\hat \mU_n - \gamma \hat \mV_n)^T \sigma(\vw^T \hat \mS) \\ 
            &\qquad - \frac{1}{m} \Tr \left(\hat \mU_n\E[\mQ_{-i}](\hat \mU_n - \gamma \hat \mV_n)^T \E[ \sigma(\hat \mS^T \vw_i) \sigma(\vw_i^T \hat \mS) ] \right) \biggr\lvert > \frac{t}{2} \biggr).
        \end{split}
        \label{eq:pb_divided_lem_concentration_Q_-}
    \end{align}
    From \lemref{lem:upper_bound_norm_resolvent}, there exists a real $K>0$ such that, for all $m$, we have
    \begin{equation*}
        \lVert \mQ_{-i} \rVert \leq K.
    \end{equation*}
    Besides, both $\lVert \hat \mU_n \rVert$ and $\lVert \hat \mV_n \rVert$ are upper bounded by $1$. We thus bound the probability of the right-hand part with \lemref{lem:normal_concentration_resolvent} as
    \begin{align}
        \begin{split}
            &\Pr\biggl(~\biggl\lvert \frac{1}{m} \sigma(\vw^T \hat \mS)^T \hat \mU_n \E[\mQ_{-i}] (\hat \mU_n - \gamma \hat \mV_n)^T \sigma(\vw^T \hat \mS) \\ 
            &\qquad - \frac{1}{m} \Tr \left(\hat \mU_n\E[\mQ_{-i}](\hat \mU_n - \gamma \hat \mV_n)^T \E[ \sigma(\hat \mS^T \vw_i) \sigma(\vw_i^T \hat \mS) ] \right) \biggr\lvert > t \biggr) \\
            &\leq C e^{-\frac{cm}{2 K_\vsigma^2 K_\varphi^2 \lVert \hat \mS \rVert^2} \min\left(\frac{ t^2}{2^8t_0^2K^2}, \frac{ t}{2K}\right)},
        \end{split}
        \label{eq:pb_divided_1_lem_concentration_Q_-}
    \end{align}
    for $t_0 = |\sigma(0)| + \sqrt{\frac{d}{m}} K_\vsigma K_\varphi \lVert \hat \mS \rVert$, and $c, C \in \R$ independent of all other parameters. Let define the real $K'>0$ and let $\mathcal{A}_{K'}$ be the probability space defined as
     %
     \begin{equation*}
         \mathcal{A}_{K'}=\{ \vw \in \R^m,~ \lVert \sigma(\vw^T \hat \mS) \rVert \leq K' \sqrt{m} \}.
     \end{equation*} 
    From \lemref{lem:sigma_bound}, we bound the second term $\Pr(\mathcal{A}_{K'}^c)$ as
    \begin{equation*}
        \Pr(\mathcal{A}_{K'}^c) = \Pr(\{\lVert \sigma(\vw^T \hat \mS) \rVert > K' \sqrt m \}) \leq C' e^{-\frac{c'mK'^2}{2K_\vsigma^2 K_\varphi^2 \lVert \mX \rVert^2}},
    \end{equation*}
    for some $c', C'>0$ independent of other parameters. Conditioning the random variable of interest with respect to $\mathcal{A}_{K'}$ and its complementary $\mathcal{A}_{K'}^c$ gives with \lemref{lem:normal_concentration_matrix} 
    \begin{align}
        \begin{split}
            &\Pr\left( \left\lvert \frac{1}{m} \sigma(\vw^T \hat \mS)^T \hat \mU_n \mQ_- (\hat \mU_n - \gamma \hat \mV_n)^T \sigma(\vw^T \hat \mS) - \frac{1}{m}\sigma(\vw^T \hat \mS)^T \hat \mU_n\E[\mQ_-](\hat \mU_n - \gamma \hat \mV_n)^T \sigma(\vw^T \hat \mS)  \right\rvert > t \right)  \\
            &\leq \Pr\Bigl( \Bigl\lvert \frac{1}{m} \sigma(\vw^T \hat \mS)^T \hat \mU_n \mQ_- (\hat \mU_n - \gamma \hat \mV_n)^T \sigma(\vw^T \hat \mS) \\
            &\qquad \qquad -\frac{1}{m}\sigma(\vw^T \hat \mS)^T \hat \mU_n\E[\mQ_-](\hat \mU_n - \gamma \hat \mV_n)^T \sigma(\vw^T \hat \mS)  \Bigr\rvert > t ~\cap~ \mathcal{A}_{K'} \Bigr) + \Pr(\mathcal{A}_{K'}^c) \\
            &\leq C'' e^{-c''mt^2}+C' e^{-\frac{c'mK'^2}{2K_\vsigma^2 K_\varphi^2 \lVert \hat \mS \rVert^2}},
        \end{split}
        \label{eq:pb_divided_2_lem_concentration_Q_-}
    \end{align}
    where $c'', C''>0$. Combing both \eqref{eq:pb_divided_1_lem_concentration_Q_-} and \eqref{eq:pb_divided_2_lem_concentration_Q_-} with \eqref{eq:pb_divided_lem_concentration_Q_-} gives
    \begin{align}
        \begin{split}
            &\Pr\biggl(\Bigl\lvert \frac{1}{m} \sigma(\vw_i^T \hat \mS) \hat \mU_n \mQ_{-i} (\hat \mU_n - \gamma \hat \mV_n)^T \sigma(\hat \mS^T \vw_i) \\
            &\qquad - \frac{1}{m} \Tr \left(\hat \mU_n\E[\mQ_{-i}](\hat \mU_n - \gamma \hat \mV_n)^T \E[ \sigma(\hat \mS^T \vw_i) \sigma(\vw_i^T \hat \mS) ] \right) \Bigr\lvert > t \biggr) \\
            &\leq C e^{-\frac{cm}{2 K_\vsigma^2 K_\varphi^2 \lVert \hat \mS \rVert^2} \min\left(\frac{ t^2}{2^10t_0^2K^2}, \frac{t}{4K}\right)} + C'' e^{-\frac{c''mt^2}{4}}+C' e^{-\frac{c'mK'^2}{2K_\vsigma^2 K_\varphi^2 \lVert \mX \rVert^2}}.
        \end{split}
    \end{align}
\end{proof}
\begin{lemma}
    Under Assumptions~\ref{assumption:growth_rate}~and~\ref{assumption:regime_n}, let $\lambda>0$, let $\mW \in \R^{N \times d}$, and let the resolvent
    \begin{equation*}
        \mQ_m(\mW)=\left[\frac{1}{m}(\hat \mU_n - \gamma \hat \mV_n)^T \FSW^T \FSW \hat \mU_n + \lambda \mI_{n} \right]^{-1}
    \end{equation*}
    defined as in \eqref{def:resolvent}. Let $\vu \in \R^{n}$ such that $\lVert \vu \rVert \leq K_\vu$ for $K_\vu>0$. Then
    \begin{equation*}
            \Pr\left(~\left\lvert \frac{\lambda^2}{n}\vu^T \mQ_m(\mW)^T \mQ_m(\mW) \vu - \frac{\lambda^2}{n}\vu^T \E[\mQ_m(\mW)^T \mQ_m(\mW)]\vu  \right\rvert > t \right)
            \leq C e^{-cn^2mt^2},
    \end{equation*}
    for some $C, c >0$ independent of $m, n$ and $N$.
    \label{lem:normal_concentration_msbe}
\end{lemma}
\begin{proof}
    Let the function $f: \mW \mapsto \frac{\lambda^2}{n} \vu^T \mQ_m(\mW)^T \mQ_m(\mW) \vu$. We want to show $f$ is Lipschitz in order to apply \lemref{lem:f_frobenius_lipschitz_concentration}. From \lemref{lem:upper_bound_norm_resolvent}, we  know there exists a real $K>0$ such that, for all $m$ and $\mW$, we have 
    \begin{equation*}
        \lVert \mQ_m(\mW) \rVert \leq K.
    \end{equation*}
    Furthermore, both $\lVert \hat \mU_n \rVert$ and $\lVert \hat \mV_n \rVert$ are upper bounded by $1$. Let $\mH \in \R^{N \times d}$, we have
    \begin{align*}
        \left\lvert f(\mW+\mH) - f(\mW) \right\rvert &=\left\lvert \frac{\lambda^2}{n} \vu^T \bigl[\mQ_m(\mW+\mH)^T \mQ_m(\mW+\mH) - \mQ_m(\mW)^T \mQ_m(\mW) \bigr] \vu \right\rvert \\
        &\leq\underbrace{\left\lvert \frac{\lambda^2}{n} \vu^T \mQ_m(\mW+\mH)^T \Bigl[\mQ_m(\mW+\mH)-\mQ_m(\mW) \Bigr] \vu^T \right\rvert}_{(1)} \\
        &\quad + \underbrace{\left\lvert \frac{\lambda^2}{n} \vu^T \Bigl[ \mQ_m(\mW+\mH)-\mQ_m(\mW)\Bigr]^T \mQ_m(\mW) \vu \right\rvert}_{(2)} \\
    \end{align*}
    For $(1)$, we have
    \begin{align*}
        &\left\lvert \frac{\lambda^2}{n} \vu^T \mQ_m(\mW+\mH)^T \Bigl[\mQ_m(\mW+\mH)-\mQ_m(\mW) \Bigr] \vu^T \right\rvert \\
        &=\biggl\lvert \frac{\lambda^2}{n}\frac{1}{m} \vu^T \mQ_m(\mW+\mH)^T \mQ_m(\mW+\mH) (\hat \mU_n-\gamma \hat \mV_n)^T\Bigl[\FSWH^T \FSWH \\
        &\quad -\FSW^T \FSW \Bigr] \hat \mU_n \mQ_m(\mW) \vu^T \biggr\rvert \\
        &=\biggl\lvert \frac{\lambda^2}{n}\frac{1}{m} \vu^T \mQ_m(\mW+\mH)^T \mQ_m(\mW+\mH) (\hat \mU_n-\gamma \hat \mV_n)^T\Bigl[\FSWH^T \bigl[\FSWH - \FSW \bigr] \\
        &\quad +\bigl[\FSWH - \FSW\bigr]^T \FSW \Bigr] \hat \mU_n \mQ_m(\mW) \vu^T \biggr\rvert \\
        &\leq \biggl\lvert \frac{\lambda^2}{n}\frac{1}{m} \vu^T \mQ_m(\mW+\mH)^T \mQ_m(\mW+\mH) (\hat \mU_n-\gamma \hat \mV_n)^T\FSWH^T \bigl[\FSWH \\
        &\qquad - \FSW \bigr] \hat \mU_n \mQ_m(\mW) \vu^T \biggr\rvert \\
        &\quad + \biggl\lvert \frac{\lambda^2}{n}\frac{1}{m} \vu^T \mQ_m(\mW+\mH)^T \mQ_m(\mW+\mH)(\hat \mU_n - \gamma \hat \mV_n)^T \bigl[\FSWH \\
        &\qquad - \FSW\bigr]^T \FSW \hat \mU_n \mQ_m(\mW) \vu^T \biggr\rvert \\
        &\leq \frac{\lambda^2}{n} K^2 K_\vu^2 \biggl\lVert \frac{1}{\sqrt{m}} \mQ_m(\mW+\mH) (\hat \mU_n-\gamma \hat \mV_n)^T\FSWH^T \biggr\rVert~\biggl\lVert \frac{1}{\sqrt{m}} \Bigl[\FSWH-\FSW \Bigr] \biggr\rVert \\
        &+ \frac{2\lambda^2}{n} K^2 K_\vu^2 \biggl\lVert \frac{1}{\sqrt{m}} \Bigl[\FSWH-\FSW \Bigr] \biggr\rVert~\biggl\lVert \frac{1}{\sqrt{m}} \FSW \hat \mU_n \mQ_m(\mW) \biggr\rVert
    \end{align*}
    From \lemref{lem:upper_bound_norm_resolvent_+_RF_matrix}, we know there exists a real $K'>0$ such that, for all $m$, we have 
    \begin{equation*}
        \biggl\lVert \frac{1}{\sqrt{m}} \FSW \hat \mU_n \mQ_m(\mW) \biggr\rVert \leq K'
    \end{equation*}
    and
    \begin{equation*}
        \biggl\lVert \frac{1}{\sqrt{m}} \mQ_m(\mW + \mH) (\hat \mU_n - \gamma \hat \mV_n )^T \FSWH^T \biggr\rVert \leq 2K'.
    \end{equation*}
    From those results, we conclude for $(1)$ that
    \begin{align*}
        &\left\lvert \frac{\lambda^2}{n} \vu^T \mQ_m(\mW+\mH)^T \Bigl[\mQ_m(\mW+\mH)-\mQ_m(\mW) \Bigr] \vu^T \right\rvert \\
        &\leq \frac{4\lambda^2 K^2 K_\vu^2 K'}{n} \biggl\lVert \frac{1}{\sqrt{m}} \Bigl[\FSWH-\FSW \Bigr] \biggr\rVert \\
        &\leq \frac{4\lambda^2 K^2 K_\vu^2 K'}{n} \biggl\lVert \frac{1}{\sqrt{m}} \Bigl[\FSWH-\FSW \Bigr] \biggr\rVert_F \\
        &\leq \frac{4\lambda^2 K^2 K_\vu^2 K' K_\vsigma }{n\sqrt{m}} \left\lVert \mH \mS \right\rVert_F \\
        &= \frac{4\lambda^2 K^2 K_\vu^2 K' K_\vsigma }{n\sqrt{m}} \sqrt{ \Tr\left(\mH\mS\mS^T \mH^T\right) } \\
        &\leq \frac{4\lambda^2 K^2 K_\vu^2 K' K_\vsigma }{n\sqrt{m}} \left\lVert \mS \right\rVert \lVert \mH \rVert_F.
    \end{align*}
    The last inequality is obtained because $\lvert \Tr(\mA\mB) \rvert \leq \lVert \mB \rVert \Tr(\mA)$ for some semi-positive-definite matrix $\mA$. With a similar reasoning, we can prove for $(2)$ that
    \begin{equation*}
        \left\lvert \frac{\lambda^2}{n} \vu^T \Bigl[ \mQ_m(\mW+\mH)-\mQ_m(\mW)\Bigr]^T \mQ_m(\mW) \vu \right\rvert \leq \frac{4\lambda^2 K^2 K_\vu^2 K' K_\vsigma }{n\sqrt{m}} \left\lVert \mS \right\rVert \lVert \mH \rVert_F.
    \end{equation*}
    We thus prove that $f$ is Lipschitz with parameter $\frac{8\lambda^2 K^2 K_\vu^2 K' K_\vsigma }{n\sqrt{m}} \left\lVert \mS \right\rVert$, and applying \lemref{lem:normal_concentration} gives
    \begin{align*}
            &\Pr\left(~\left\lvert \frac{\lambda^2}{n}\vu^T \mQ_m(\mW)^T \mQ_m(\mW) \vu - \frac{\lambda^2}{n}\vu^T \E[\mQ_m(\mW)^T \mQ_m(\mW)]\vu  \right\rvert > t \right) \\
            &\quad \leq C e^{-\frac{cn^2mt^2}{2^6 \lambda^2 K^4 K_\vu^4 K'^2 K_\vsigma^2 K_\varphi^2 \lVert \mS\rVert^2}},
    \end{align*}
    for some $C, c >0$ independent of other parameters.
\end{proof}
\section{Intermediary Lemmas}
\begin{lemma}[Normal Concentration]
    \label{lem:normal_concentration}
    (\citep[Corollary~2.6, Propositions~1.3,~1.8]{ledoux2001concentration} or \citep[Theorem~2.1.12]{tao2012topics}) For $d \in \mathbb{N}$, consider $\mu$ the canonical Gaussian probability on $\R^d$ defined through its density $d\mu(\vw)=(2 \pi)^{-\frac{d}{2}}e^{- \frac12 \lVert \vw\rVert^2}$ and $f : \R^d \to \R$ a $L_f$-Lipschitz function. Then
    \begin{align}
        \mu\left( \left\{ \left| f - \int f d\mu \right| \geq t \right\} \right) &\leq  Ce^{-c\frac{t^2}{L_f^2}},
    \end{align}
    where $C,c>0$ are independent of $d$ and $L_f$. 
\end{lemma}
\begin{lemma}[Resolvent Identity]
	\label{lem:resolvent_identity}
    For invertible matrices $\mA,\mB \in \R^{n \times n}$,
    \begin{equation*}
         \mA^{-1}-\mB^{-1}=\mA^{-1}(\mB-\mA)\mB^{-1}
    \end{equation*}
\end{lemma}
\begin{lemma}[Sherman–Morrison–Woodbury Matrix Identity]\label{lem:woodbury}
    \citep[Theorem~0.7.4]{horn2012matrix} Let $\mA \in \R^{n \times n}$ be a non-singular matrix with a known inverse $\mA^{-1}$; let $\mM=\mA + \mU \mC \mV$, in which $\mU \in \R^{k \times n}$, $\mV \in \R^{n \times k}$, and $\mC^{k \times k}$ is non-singular. If $\mM$ and $\mC^{-1}+\mV\mA^{-1}\mU$ are non-singular then
    \begin{equation}
        \left(\mA+\mU\mC\mV\right)^{-1}=\mA^{-1}-\mA^{-1}\mU\left(\mC^{-1}+\mV\mA^{-1}\mU\right)^{-1}\mV\mA^{-1},
    \end{equation}
    In particular $\left(\mA+\mU\mV\right)^{-1}\mU=\mA^{-1}\mU\left(\mI_n+\mV\mA^{-1}\mU\right)^{-1}$ and $\mV\left(\mA+\mU\mV\right)^{-1}=\left(\mI_n+\mV\mA^{-1}\mU\right)^{-1}\mV\mA^{-1}$.
\end{lemma}
\begin{lemma}[Sherman–Morrison Formula]\label{lem:sherman}
    Let $\mA \in \R^{n \times n}$ be a non-singular matrix with a known inverse $\mA^{-1}$; let $\mM=\mA + \vu\vv^T$, in which $\vu, \vv \in \R^{n}$. If $\mM$ is non-singular and $1+\vv^T\mA^{-1}\vu \neq 0$ then
    \begin{equation}
        \left(\mA+\vu\vv^T\right)^{-1}=\mA^{-1}-\frac{\mA^{-1}\vu\vv^T\mA^{-1}}{1+\vv^T\mA^{-1}\vu}.
    \end{equation}
    In particular, $\left(\mA+\vu\vv^T\right)^{-1}\vu=\frac{\mA^{-1}\vu}{1+\vv^T\mA^{-1}\vu}$ and $\vv^T\left(\mA+\vu\vv^T\right)^{-1}=\frac{\vv^T\mA^{-1}}{1+\vv^T\mA^{-1}\vu}$. This Lemma is an extension of Lemma~\ref{lem:woodbury}.
\end{lemma}
\begin{lemma}[Ostrowski’s Theorem]\label{lem:ostrowski}
    \citep[Theorem~4.5.9]{horn2012matrix} Let $\mA, \mS \in \R^{n \times n}$ with $\mA$ Hermitian and $\mS$ nonsingular. Let the eigenvalues of $\mA$, $\mS\mA\mS^T$, and $\mS\mS^T$ be arranged in nondecreasing order. Let $\sigma_1 \geq \ldots \geq \sigma_n > 0$ be the singular values of $\mS$. For each $k \in [n]$ there is a positive real number $\theta_k \in [\sigma_n^2, \sigma_1^2]$ such that
    \begin{equation*}
        \nu_k(\mS\mA\mS^T)=\theta_k \nu_k(\mA)
    \end{equation*}
\end{lemma}
\begin{lemma}[Weinstein–Aronszajn Identity]\label{lem:weinstein}
    For $\mA \in \R^{m \times n}$, $\mB \in \R^{n \times m}$ and $\lambda \in \R \setminus \{0\}$,
    \begin{equation*}
        \det(\mA\mB-\lambda \mI_{m})=(-\lambda )^{m-n}\det(\mB\mA-\lambda \mI_{n}).
    \end{equation*}
    It follows that the non-zero eigenvalues of $\mA\mB$ and $\mB\mA$ are the same.
\end{lemma}
\begin{lemma}
	Let $\mA \in \R^{n \times n}$ and $\lambda>0$. 
        \begin{equation*}
		\lVert \left( \mA + \lambda \mI_n \right)^{-1} \rVert \leq \frac{1}{\lambda} 
	\end{equation*}
        if and only if $\mA\mA^T + \lambda (\mA + \mA^T)$ is positive definite.
        In particular, for matrix $\mA \in \R^{n \times n}$ whose the Hermitian part $H(\mA)=\frac{\mA+\mA^T}{2}$ is semi-positive-definite we have
	\begin{equation*}
		\lVert \left( \mA + \lambda \mI_n \right)^{-1} \rVert \leq \frac{1}{\lambda} 
	\end{equation*}
        \label{lem:resolvent_bound_semi_positive_definite}
\end{lemma}
\begin{proof}
    \begin{align}
        \begin{split}
            \lVert \left( \mA + \lambda \mI_n \right)^{-1} \rVert^2 &= \nu_{max}\left(\left( \mA + \lambda \mI_n \right)^{-1 \, T}\left( \mA + \lambda \mI_n \right)^{-1}\right) \\
            &=\nu_{max}\left( \left[\left( \mA + \lambda \mI_n \right)\left( \mA^T + \lambda \mI_n \right)\right]^{-1}\right) \\
            &=\nu_{max}\left( \left ( \mA\mA^T + \lambda (\mA + \mA^T)  + \lambda^2 \mI_n \right)^{-1}\right) \\
            &=\nu_{min}\left( \left ( \mA\mA^T + \lambda (\mA + \mA^T)  + \lambda^2 \mI_n \right)\right)^{-1}
        \end{split} 
        \label{eq:eig_min_resolvent}
    \end{align}
    where $\nu_{max}(\mB)$ and $\nu_{min}(\mB)$ denotes the maximum eigenvalue and minimum eigenvalues of a matrix $\mB$. Since $\mA$ is positive-definite the matrix $\mA\mA^T + \lambda (\mA + \mA^T)$ is semi-positive-definite and has positive nonzeros eigenvalues. Therefore, $\nu_{min}\left( \left ( \mA\mA^T + \lambda (\mA + \mA^T)  + \lambda^2 \mI_n \right)\right) >\lambda^2$ and $\lVert \left( \mA + \lambda \mI_n \right)^{-1} \rVert \leq \frac{1}{\lambda}$

\end{proof}
\begin{lemma}
    \citep[Theorem~2]{yates1995framework} If a mapping $f:[0, \infty) \to [0, \infty)$  
    \begin{itemize}
        \item is monotonically increasing, i.e $x \geq x' \implies f(x) \geq f(x')$,
        \item is scalable, i.e $\forall \alpha > 1,~\alpha f(x) > f(\alpha x)$,
        \item admits $x_0 \in [0, \infty)$ such that $x_0 \geq f(x_0)$,
    \end{itemize}
    then $f$ has a unique fixed-point.
    \label{lem:fixed_point_standard_function}
\end{lemma}
\begin{lemma}
    Let $\mA \in \R^{m \times n}$ and $\mB \in \R^{n \times m}$. If $\mA \mB + \lambda \mI_m$ is invertible, then
    \begin{equation*}
        \bigl[\mA \mB + \lambda \mI_m \bigr]^{-1} \mA  =  \mA \bigl[\mB \mA + \lambda \mI_n \bigr]^{-1}.
    \end{equation*}
    \label{lem:push_through_identity}
\end{lemma}
\begin{proof}
    We have 
    \begin{equation*}
        \mA \bigl[\mB \mA + \lambda \mI_n \bigr] = \bigl[\mA \mB + \lambda \mI_m \bigr] \mA 
    \end{equation*}
    Since both $\mA \mB$ and $\mB\mA$ share the same non-zero eigenvalues from Lemma~\ref{lem:weinstein}, we deduce $\mB\mA + \lambda \mI_n$ is also invertible.
    By multiplying the equation above with both the inverse of $[\mB \mA + \lambda \mI_n]$ and $[\mA \mB + \lambda \mI_m \bigr]$, we get 
    \begin{equation*}
        \bigl[\mA \mB + \lambda \mI_m \bigr]^{-1} \mA  =  \mA \bigl[\mB \mA + \lambda \mI_n \bigr]^{-1}
    \end{equation*}
\end{proof}

\end{document}